\documentclass[a4paper,12pt,twoside,openright]{book}
\usepackage{custom_packages}

\begin{document}

\thispagestyle{empty}
%
%
%
%
%

\begin{center}
    {
    \setlength\intextsep{0pt}
    \begin{figure}[h!]
        \centering
        \includegraphics[width=0.45\textwidth]{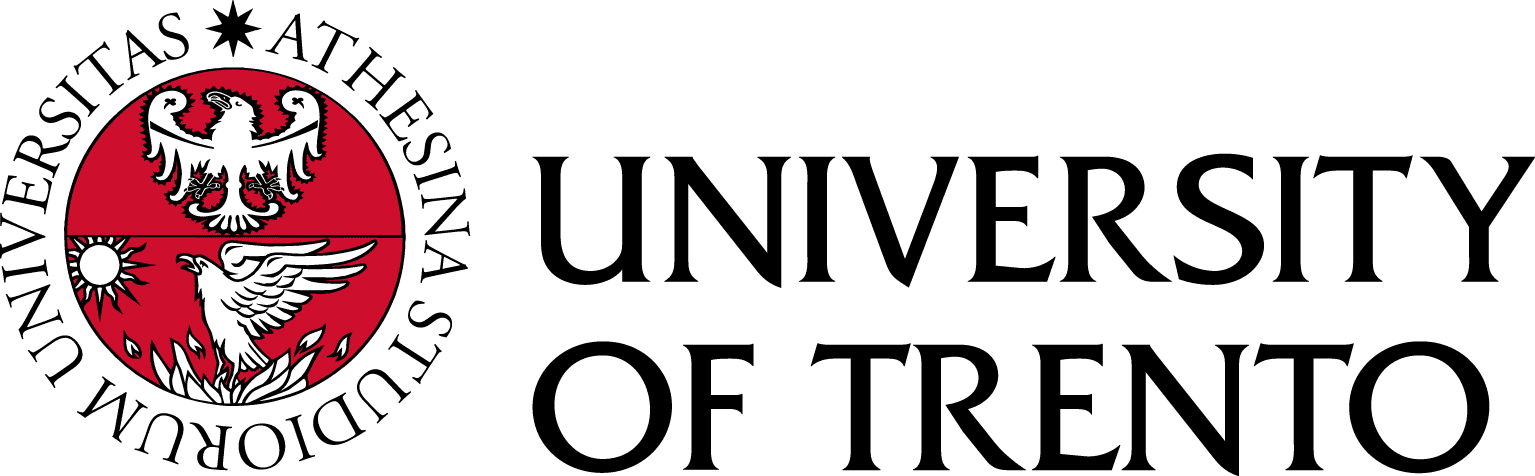}
        \label{fig:university-logo}
    \end{figure}
    }

    \par\noindent\rule{\textwidth}{0.1pt} 
    \small{DEPARTMENT OF INFORMATION ENGINEERING AND COMPUTER SCIENCE\\
    DEPARTMENT OF INDUSTRIAL ENGINEERING \\
    FONDAZIONE BRUNO KESSLER \\}

    \vspace{0.5 cm}  

    \large\textbf{Doctorate Program in Industrial Innovation}

    \vspace{2 cm}
    
    \Huge\textsc{Robotic Contextual Awareness for Human-Robot Collaboration and Environmental Understanding\\}

    \vspace{2 cm}

    \huge{Federico Rollo}

\end{center}

\vspace{2 cm}

\begin{tabular}{llll}
    \multicolumn{2}{l}{\large Accademic Advisor}   &  \multicolumn{2}{l}{\large Industrial Advisor}\\
     & \large Prof. Marco Roveri                &  & \large Dr. Arash Ajoudani\\
     & \large Universit\`a di Trento            &  & \large Istituto Italiano di Tecnologia\\
\end{tabular}

\vspace{1cm}


\begin{center}
    \vspace{1cm}
    \hrule
    \vspace{4pt}
    April 2026
\end{center}

\clearemptydoublepage

\frontmatter
\singlespacing
    \chapter*{}

\vspace*{\fill}

\begin{center}
    \hspace*{0.3\textwidth} 
    \parbox{0.6\textwidth}{ 
        \raggedleft 
        \emph{\large To my father,\\ who is always with me.}
    }
\end{center}

\vspace*{\fill}
    \chapter*{Disclaimer}

I, Federico Rollo ("the Author"), wrote this doctoral thesis in my personal capacity while working at Leonardo Spa ("the Company"), holding the position of Technical Coordinator of the Perceptive Navigation research area of the Robotics Laboratory in the Innovation Hubs \& Intellectual Property UO of the Corporate Division. 
The views and opinions expressed in this thesis are those of the Author only and do not necessarily reflect the official policy or position of the Company. 
This thesis has been written using open information sources only, \textit{i.e.}, public domain information gathered from unclassified knowledge sources, being the majority of them available on the internet, and not subject to any disclosure restriction of any kind. 
This includes some of the yet-unpublished results of various research activities carried out in Leonardo Spa with the thesis author's contribution or under his supervision. 
All cited sources (texts, patents, data, and images) have been duly referenced, where possible and where required. 
All cited trademarks and patents have been indicated as the property of their respective owners, where possible and where required. Examples given within this thesis must be considered only as the Author's assumptions and are not reflective of the position of the Company, or of any third party whatsoever. 
The Author has made reasonable efforts to ensure that all information and material included in this thesis were accurate at the time of writing, but the Author cannot guarantee their accuracy. 
The Author has not received any remuneration, consideration, or benefit for mentioning, reporting, including, or commenting about any paper, program, product, policy, law, theory, hypothesis, or intellectual property. 
For any errors or inadequacies that may remain in this work, of course, the responsibility is entirely on my own.
    \chapter*{Acknowledgment}

I want to express my sincere gratitude to everyone who has supported and guided me throughout my PhD journey. Without them, this path would not have been possible.

My deepest appreciation goes to my former senior seatmate, Andrea Zunino, whose invaluable teaching and insightful reviews were instrumental in initiating my research path. I am profoundly thankful to my supervisors, Prof. Marco Roveri and Dr. Arash Ajoudani, for their exceptional guidance and mentorship. Special thanks are also extended to Navvab Kashiri, my current principal investigator, and all the senior researchers who generously shared their expertise and provided invaluable assistance over the years.

I am immensely grateful to my current and former Robotics Lab teammates at Leonardo Spa, with whom I worked closely during my PhD: Edoardo Del Bianco, Valentina Pericu, Luigi Raiano, Enrico Mingo Hoffman, Fabio Amadio, Gennaro Raiola, Andrea Testa, Saber Mohammadi, and Marco Laghi, among others. Thanks for sharing this PhD journey with me over these years. The shared ideas, experiences, and team spirit have been crucial to my growth, and I cherish the friendships and the vibrant collaborative environment we've built. Thanks also go to my latest Robotics teammates: Simone Giampà, Andrea Monguzzi, Giuseppe Alfonso, and Marco Puliti, with whom I'll continue the journey in robotics after my graduation.

I also need to thank the researchers from the Istituto Italiano di Tecnologia (IIT) for the joyful times spent together. Especially Nikos Tsagarakis, who, together with Arash and their laboratories, Humanoids and Human Centered Mechatronics (HHCM) and Human-Robot Interfaces and Interaction (HRI²), followed part of my PhD, giving insightful and invaluable suggestions.

A huge thanks goes to Alessandro Garibbo, whose suggestions have been crucial for starting and concluding this path. I would like to thank all the Leonardo Spa staff whom I met during this time for their valuable support, with a special thanks to Alessandro Massa, Salvatore Scervo, and Pierpaolo Gambini for having guided the Leonardo Labs' evolution, and to CEO Roberto Cingolani and Co-General Manager Simone Ungaro, who are pushing and leading the innovation in our company.

Finally, I extend my heartfelt thanks to my family. To my mother, for her unwavering support from the very beginning. To my brother, for his wisdom and protection. To my nephews, who consistently fill my life with joy (and occasional delightful chaos). To my girlfriend, for her boundless love and steady support, especially during the most challenging times. And to my father, who I am certain would have been incredibly proud of this achievement and the path I've taken.
    \chapter*{Abstract}

The transition of autonomous mobile robots from controlled industrial settings to dynamic, human-centric environments, such as manufacturing, logistics, and healthcare, has made their safe and autonomous operation a critical area of research. These sophisticated machines must be capable of perceiving, understanding, and interacting with their surroundings to navigate freely and perform complex tasks. A significant obstacle to achieving this is the lack of comprehensive contextual awareness, which requires a robot to recognize its spatial environment and identify the objects and actors within it. Without this perceptual knowledge, robots struggle to plan adaptive behaviors or engage in meaningful interaction with humans.

This thesis presents novel solutions to this challenge by exploring two distinct but complementary research directions. The first direction involves human re-identification and tracking to improve \gls{hrc}. Our developed approach enables a mobile robot to recognize a specific person, facilitating targeted collaboration while ignoring other individuals. The second direction focuses on enhancing the robot's overall perceptual capabilities to understand its environment geometrically and semantically. Geometric information is vital for motion planning and collision avoidance, while semantic knowledge provides the robot with a richer understanding for more advanced interaction. Both solutions are driven by the improvement of the semantical understanding of robots that enhance their knowledge of their surroundings, allowing a smoother and more natural interaction between robots, humans, and the environment. The contributions of this work in human re-identification and environmental understanding represent a significant step toward a future where robots are more contextually aware, enabling safer coexistence and more effective collaboration.

\section*{Keywords}
\textit{Context Awareness, Environmental Understanding, Re-Identification,  SLAM, Loop Closure Detection, Semantic Mapping}

    \newacronym{fov}{FOV}{field-of-view}
\newacronym{slam}{SLAM}{Simultaneous Localization and Mapping}
\newacronym{lidar}{LiDAR}{Light Detection and Ranging}
\newacronym{3d}{3D}{three-dimensional}
\newacronym{reid}{Re-ID}{Re-identification}
\newacronym{mmt}{MMT}{Mutual Mean Teaching}
\newacronym{carpe}{CARPE-ID}{Continuous Adaptation for personalized Re-identification}
\newacronym{ot}{OT}{Object Tracking}
\newacronym{mot}{MOT}{Multi-Object Tracking}
\newacronym{sot}{SOT}{Single-Object Tracking}
\newacronym{yolo}{YOLO}{You Only Look Once}
\newacronym{hri}{HRI}{Human-Robot Interaction}
\newacronym{sc++}{SC++}{Scan Context++}
\newacronym[plural=sc,firstplural=Scan Contexts (SCs)]{sc}{SC}{Scan Context}
\newacronym{gsc}{GSC}{Gaussian Scan Context}
\newacronym{rgb}{RGB}{Red, Green and Blue}
\newacronym{rgbd}{RGB-D}{Red, Green, Blue and Depth}
\newacronym{rviz}{RViz}{ROS Visualization Tool}
\newacronym{ros}{ROS}{Robot Operating System}
\newacronym{ir}{IR}{Infra Red}
\newacronym[plural=led,firstplural=Light-Emitting Diodes (LEDs)]{led}{LED}{Light-Emitting Diode}
\newacronym{hog}{HOG}{Histograms of Oriented Gradients}
\newacronym[plural=cnn,firstplural=Convolutional Neural Networks (CNNs)]{cnn}{CNN}{Convolutional Neural Network}
\newacronym{ffm}{FFM}{Feature Funnel Model}
\newacronym{svm}{SVM}{Support Vector Machine}
\newacronym{cl}{CL}{Continual Learning}
\newacronym{si}{SI}{Synaptic Intelligence}
\newacronym{ewc}{EWC}{Elastic Weight Consolidation}
\newacronym{pn}{PN}{Progressive Networks}
\newacronym{aka}{AKA}{Adaptive Knowledge Accumulation}
\newacronym{vslam}{V-SLAM}{Visual SLAM}
\newacronym{loam}{LOAM}{LiDAR Odometry and Mapping}
\newacronym{imu}{IMU}{Inertial Measurement Unit}
\newacronym{lcd}{LCD}{Loop Closure Detection}
\newacronym{lreid}{L-REID}{Lifelong Re-Identification}
\newacronym{icp}{ICP}{Iterative Closest Point}
\newacronym{gicp}{GICP}{Generalized Iterative Closest Point}
\newacronym{2d}{2D}{two-dimentional}
\newacronym{lvi}{LVI}{LiDAR-Visual-Inertial}
\newacronym{isc}{ISC}{Intensity Scan Context}
\newacronym{bow}{BoW}{Bag-of-Words}
\newacronym{orb}{ORB}{Oriented FAST and rotated BRIEF}
\newacronym{sift}{SIFT}{Scale-Invariant Feature Transform}
\newacronym{vlad}{VLAD}{Vector of Locally Aggregated Descriptors}
\newacronym{dof}{DoF}{Degrees of Freedom}
\newacronym{nerf}{NeRF}{Neural Radiance Field}
\newacronym{3dgs}{3DGS}{3D Gaussian Splatting}
\newacronym{crf}{CRF}{Conditional Random Field}
\newacronym{tsdf}{TSDF}{Truncated Signed Distance Function}
\newacronym[plural=sdf,firstplural=Signed Distance Functions (SDFs)]{sdf}{SDF}{Signed Distance Function}
\newacronym{vio}{VIO}{Visual-Inertial Odometry}
\newacronym{uav}{UAV}{Unmanned Aerial Vehicle}
\newacronym{pnp}{PnP}{Perspective-n-Point}
\newacronym{gnss}{GNSS}{Global Navigation Satellite System}
\newacronym[plural=llm,firstplural=Large Language Models (LLMs)]{llm}{LLM}{Large Language Model}
\newacronym[plural=vlm,firstplural=Vision-Language Models (VLMs)]{vlm}{VLM}{Vision-Language Model}
\newacronym{clip}{CLIP}{Contrastive Language-Image Pre-training}
\newacronym{cp}{CP}{Correspondence Problem}
\newacronym{sfm}{SfM}{Structure from Motion}
\newacronym{da}{DA}{Data Association}
\newacronym{mtt}{MTT}{Multi-Target Tracking}
\newacronym{map}{MAP}{Maximum A Posteriori}
\newacronym{sota}{SoTA}{State-of-the-Art}
\newacronym{kf}{KF}{Kalman Filter}
\newacronym{ui}{UI}{User Interface}
\newacronym{ieee}{IEEE}{Institute of Electrical and Electronics Engineers}
\newacronym{ema}{EMA}{Exponential Moving Average}
\newacronym{dema}{DEMA}{Damped Exponential Moving Average}
\newacronym[plural=id,firstplural=Identifiers (IDs)]{id}{ID}{Identifier}
\newacronym[plural=gpu,firstplural=Graphic Processing Units (GPUs)]{gpu}{GPU}{Graphic Processing Unit}
\newacronym{dlo}{DLO}{Direct LiDAR Odometry}
\newacronym{isam2}{iSAM2}{Incremental Smoothing and Mapping}
\newacronym{pcl}{PCL}{Point Cloud Library}
\newacronym{bt}{BT}{Behaviour Tree}
\newacronym{ai}{AI}{Artificial Intelligence}
\newacronym{vbr}{VBR}{Vision Benchmark in Rome}
\newacronym{ate}{ATE}{Absolute Trajectory Error}
\newacronym{rpe}{RPE}{Relative Pose Error}
\newacronym{rmse}{RMSE}{Root Mean Square Error}
\newacronym{tve}{TVE}{Translation Vector Error}
\newacronym{te}{TE}{Translation Error}
\newacronym{oe}{OE}{Orientation Error}
\newacronym{p}{P}{Precision}
\newacronym{r}{R}{Recall}
\newacronym{pr}{PR}{Precision-Recall}
\newacronym{auc}{AUC}{Area Under the Curve}
\newacronym{tp}{TP}{True Positive}
\newacronym{fp}{FP}{False Positive}
\newacronym{tn}{TN}{True Negative}
\newacronym{fn}{FN}{False Negative}
\newacronym{surf}{SURF}{Speeded Up Robust Features}
\newacronym{fsm}{FSM}{Finite State Machine}
\newacronym{lior}{LIOR}{Low-Intensity Outlier Removal}
\newacronym{ddior}{DDIOR}{Dynamic Distance–Intensity Outlier Removal}
\newacronym{lidsor}{LIDSOR}{Low-Intensity Dynamic Statistical Outlier Removal}
\newacronym{hrc}{HRC}{Human-Robot Collaboration}
\newacronym{amcl}{AMCL}{Adaptive Monte Carlo Localization}
\newacronym{ukf}{UKF}{Unscented Kalman Filter}

\nomenclature{$\mathbb{N}$}{Set of natural numbers.}
\nomenclature{$\mathbb{R}$}{Set of relative numbers.}

\nomenclature{$x$}{Scalar.}
\nomenclature{$x_t \equiv x[t]$}{Scalar at time $t$.}
\nomenclature{$\vert x \vert$}{Modulus of scalar $x$, also known as the absolute value of $x$.}
\nomenclature{$\mathbf{x}$}{Vector.}
\nomenclature{$\vert \mathbf{x}\vert$}{Cardinality of vector $\mathbf{x}$, \textit{i.e.}, number of elements in the vector. The same holds for matrices and sets.}
\nomenclature{$\mathbf{x}_t \equiv\mathbf{x}[t]$}{Vector at time $t$.}
\nomenclature{$\mathrm{X}$}{Vectors set.}
\nomenclature{$\mathrm{X}_{1:t}$}{Sequence of vectors from $\mathbf{x}_1$ to $\mathbf{x}_t$. \textit{I.e.}, $\{\mathbf{x}_1, \dots, \mathbf{x}_t\}$.}

\nomenclature{$\mathbf{X}$}{Matrix.}
\nomenclature{$\mathbf{X}_{i,j}$}{Matrix entry on row $i$ and column $j$. When $i$ or $j$ is replaced by $:$, it means the entire row or column.}
\nomenclature{$\mathbb{X}$}{Matrices set.}
\nomenclature{$\mathbb{X}_{1:t}$}{Sequence of Matrices from $\mathbf{X}_1$ to $\mathbf{X}_t$. \textit{I.e.}, $\{\mathbf{X}_1, \dots, \mathbf{X}_t\}$.}
\nomenclature{$\mathbf{I}_k$}{Identity matrix of dimensions $k \times k$.}
\nomenclature{$x^*$}{Optimal value of ${x}$. Valid for scalar, vectors, and matrices.}
\nomenclature{$\hat{x}$}{Estimated value of ${x}$. Valid for scalar, vectors, and matrices.}

\nomenclature{$\in$}{Contained in set/sequence. \textit{E.g.}, $\mathbf{x} \in \mathrm{X}$, Vector $\mathbf{x}$ contained in Vector set $\mathrm{X}$.}
\nomenclature{$\notin$}{Not contained in set/sequence. \textit{E.g.}, $\mathbf{x} \notin \mathrm{X}$, Vector $\mathbf{x}$ not contained in Vector set $\mathrm{X}$.}
\nomenclature{$\equiv$}{Defined as. \textit{I.e.}, $x \equiv \dots$, mean $x$ is defined as or is equivalent to $\dots$}
\nomenclature{$\&$}{Conditional And.}
\nomenclature{$\vert$}{Conditional Or.}
\nomenclature{$log(x)$}{Natural logarithm.}
\nomenclature{$e^x \equiv exp(x)$}{Exponential.}
\nomenclature{$\mathrm{A} = \{x\vert y\}$}{Set $\mathrm{A}$ is equivalent to the set of all elements $x$, such that $\vert $ the condition $y$ is true (or holds).}

\nomenclature{$\mathbf{P(a)}$}{Probability distribution of $a$.}
\nomenclature{$\mathbf{P(a\vert b)}$}{Conditional probability distribution of $a$ given $b$.}

\nomenclature{$\sum_{i=1}^\mathrm{N}$}{Summation iterating from $i=1$ to $\mathrm{N}\in\mathbb{N}$.}
\nomenclature{$\sum_{\mathbf{x}\in\mathrm{X}}$}{Summation iterating over $\mathrm{X}$ set. The same is valid for matrices.}
\nomenclature{$\prod_{i=1}^\mathrm{N}$}{Product iterating from $i=1$ to $\mathrm{N}\in\mathbb{N}$.}
\nomenclature{$\prod_{\mathbf{x}\in\mathrm{X}}$}{Product iterating over $\mathrm{X}$ set. The same is valid for matrices.}

\nomenclature{$\mu$}{Mean value of a Gaussian distribution. Valid for scalar and vector.}
\nomenclature{$\sigma$}{standard deviation value of a Gaussian distribution. Valid for scalar and vector.}
\nomenclature{$\boldsymbol{\Sigma}$}{Covariance matrix of a Gaussian distribution.}
\nomenclature{$\mathrm{G} \equiv \mathcal{N}(\mu, \sigma)$}{Univariate Gaussian distribution.}
\nomenclature{$\boldsymbol{\mathcal{G}} \equiv \mathcal{N}(\boldsymbol{\mu}, \boldsymbol{\Sigma})$}{Multivariate Gaussian distribution.}

\nomenclature{$\mathbf{p} \in \mathbb{R}^3$}{Three-dimensional point in the Cartesian space.}
\nomenclature{$\mathbf{p} \in \mathbb{R}^4$}{Three-dimensional point in the Cartesian space plus intensity value.}
\nomenclature{$p^x,\ p^y,\ p^z,\ p^{\psi}$}{x, y, z and intensity values of point $\mathbf{p}$.}
\nomenclature{$^a\mathbf{p}\ \equiv\ ^a\mathbf{t}$}{Point $\mathbf{p}$ or vector $\mathbf{t}$ expressed in frame $a$.}

\nomenclature{$^a\mathbf{t}_{b} \equiv \{^a\mathbf{x}_{b},\ ^a\mathbf{y}_{b},\ ^a\mathbf{z}_{b}\}$}{Translation vector ($x,\ y,\ z$) from frame $b$ to frame $a$.}
\nomenclature{$^a\mathbf{R}_{b}$}{Rotation matrix from frame $b$ to frame $a$.}
\nomenclature{$^a\mathbf{H}_{b} $}{Homogeneous transformation from frame $b$ to frame $a$.}
\nomenclature{$^a\mathbf{T}_{b}$}{Homogeneous transformation from frame $b$ to frame $a$ containing only translational vector. \textit{I.e.}, $^a\mathbf{R}_{b} = \mathbf{I}_3$.}

\nomenclature{$\mathcal{L}$}{Loss function.}
\nomenclature{$s \equiv \mathcal{P}$}{Point cloud scan.}
\nomenclature{$\mathcal{S} \equiv \sum \mathcal{P}$}{Submap defined as sum of point clouds.}
\nomenclature{$\mathbb{S} \equiv \sum \mathcal{S}$}{Submap sum.}

\nomenclature{$\Vert\cdot\Vert_0$}{L0-Norm.}
\nomenclature{$\Vert\cdot\Vert_2$}{L2-Norm also known as Euclidean norm.}
\nomenclature{$\Vert\cdot\Vert_F$}{Frobenius norm.}
\nomenclature{$\langle \mathbf{a}, \mathbf{b} \rangle$}{Dot product between vectors $\mathbf{a}$ and $\mathbf{b}$.}
\nomenclature{$\lceil a\rceil$}{Ceiling function of a scalar $a$.  Uprounds the scalar $a$ up to the nearest integer.}

    \tableofcontents
    \listoftables
    \listoffigures
    \printglossary[type=\acronymtype, title=List of Acronyms]
    \printnomenclature

\mainmatter
\onehalfspacing

\chapter[Introduction]{Introduction} \label{chap:intro}
    In recent years, the primary focus of robotics has shifted toward enhancing the autonomy and motion capabilities of robots to operate within our complex, human-centric environments. This evolution is driven by the growing demand for mechatronic systems that assist with repetitive, dangerous, or physically demanding tasks. While the goal of a fully autonomous robot capable of adapting to any situation is still a significant research challenge, substantial progress has been made. A core requirement for achieving this level of autonomy is the robot's ability to effectively perceive, understand, and interact with its surroundings. Without a comprehensive grasp of its environment, a robot cannot act responsively or perform meaningful, real-world tasks.

For decades, the majority of industrial robots, particularly in sectors like the automotive and production industry, have operated as \textit{blind} machines. These robots are programmed to repeat the same high-accuracy motions within a highly structured environment. This approach, rooted in the Ford concept of the assembly line, works efficiently as long as the environment remains static. However, as soon as a variable changes, these robots become unable to complete their tasks without manual recalibration, leading to costly downtime. Modern solutions have begun to address this limitation by incorporating sensing capabilities, such as cameras and time-of-flight sensors, to adapt a robot's final pose to the object it is manipulating. This allows for smoother task execution and increases the robot's awareness of its immediate surroundings.

This early integration of sensors marked a crucial step forward, but it largely focused on a purely geometric understanding of the world. Advanced sensing modalities, such as \gls{lidar} and camera sensors, provide rich \gls{3d} data that enables robots to construct detailed maps for path planning and obstacle avoidance. While these geometric representations are excellent for describing the location and size of objects, they offer little insight into their semantic meaning, \textit{e.g.}, what they represent or why they are significant. This limitation is a significant barrier to operating in dynamic, unstructured environments.

\section{Problem statement}

Humans and many animals possess a far more profound perceptual capability. We not only perceive the geometric layout of our environment but also assign high-level semantic meaning to objects, places, and other entities. For instance, while geometric cues are essential for reactive behaviors, like catching a falling object, semantic understanding is indispensable for all high-level reasoning and decision-making. We instinctively know to grasp a knife by its handle, not its blade, and we avoid walking through mud, especially with new shoes. This ability to fuse geometric and semantic information is what enables us to make nuanced decisions and exhibit highly adaptive behaviors.

The progression from a purely geometric understanding to a more semantically informed perception is paramount for mobile robots to operate intelligently in complex, human-inhabited spaces. This interdisciplinary field, which merges \gls{ai} with spatial and geometrical concepts, is now known as Spatial \gls{ai}. While some tasks still benefit from standard computer vision and geometric algorithms, more complex problems, such as object detection, require data-driven solutions powered by neural networks.

This thesis revolves around the concept of robotic contextual awareness, which is the ability of a robot to perceive and comprehend its surroundings. The concept is wide and covers many aspects. The problems faced in this thesis are deeply discussed in Chapter~\ref{chap:problem} where the data association and the mapping problems are presented respectively in Section~\ref{sec:association-prob} and Section~\ref{sec:mapping-prob}.

\section{Presented solutions}

This thesis argues that to truly achieve advanced robotic autonomy and intuitive \gls{hri}, robots must develop a comprehensive contextual awareness. This means moving beyond simply knowing where things are to understanding what they are, who is present, and how these elements relate to each other within a dynamic scene. Only by integrating this deeper understanding can robots become truly robust, adaptable, and trustworthy collaborators.

The research is moving toward developing robots that can understand their context and act as an active part of the environment, rather than an outsider object. This includes enhanced capabilities for environmental understanding (Spatial \gls{ai}) and improved motion capabilities that can adapt to the environment. Current research on adaptive motion capabilities, often leveraging reinforcement learning and foundation models, shows great promise for specific situations due to their robustness. However, their generalization capabilities remain poor, as even small environmental changes can cause these controllers to fail because they have not encountered that exact situation during training. By combining deep contextual understanding with advanced motion capabilities, we can develop robots that are not just blind manipulators, but active, intelligent actors in our evolving world.

This PhD thesis addresses the critical challenge of achieving comprehensive contextual awareness for mobile robots operating in complex, semi-structured, and unstructured environments. For robots to become truly intelligent and autonomous, they must move beyond a purely geometric understanding of their surroundings and acquire a holistic situational awareness. This includes not only an understanding of the static environment but also the dynamic entities within it, particularly humans, and the intricate relationships between them.

A deep, contextual understanding is the cornerstone of effective robotic autonomy in human-centric spaces. Without it, a robot is limited to reactive, often brittle behaviors. For a robot to engage in socially compliant navigation, intuitive \gls{hrc}, and truly adaptive task execution, it must be able to understand what objects are, who is present, and how all these elements relate to each other. This thesis posits that true robotic context awareness emerges from the synergistic integration of two fundamental domains: environmental perception and dynamic entity understanding.

To achieve this comprehensive contextual awareness, this research is structured around two interconnected pillars: reliable object detection and \gls{reid} and robust environmental mapping and semantic understanding.

The first pillar of this research delves into the persistent perception of dynamic entities, primarily focusing on humans. Through the work on Person \gls{reid}, including contributions like FollowMe\,\cite{rollo2023followme}, continuous adaptation\,\cite{rollo2024carpe}, and personalized \gls{reid} through unsupervised continual learning\,\cite{rollo2025pesonalized}, the crucial need for robots to continuously track and recognize individuals across different viewpoints and over time is addressed. This capability is essential for enabling personalized interactions and human-aware navigation, allowing a robot to maintain a consistent understanding of a person's identity despite occlusions or changes in position.

The second pillar of this work focuses on endowing robots with the ability to build accurate and semantically rich representations of their operational space. This is explored through the work in \gls{slam} and Semantic \gls{slam}. This research, including contributions such as LEO-SLAM\,\cite{rollo2025leoslam}, Artifacts Mapping\,\cite{rollo2023artifacts}, Ground-Aware \gls{lidar} Filtering, and \gls{gsc}, enables a robot to not only localize itself and map its environment precisely but also to identify and localize salient objects and features. This capability is particularly critical in challenging scenarios, such as environments with significant reflections, where traditional methods often fail.

A central argument of this thesis is that neither of these pillars is sufficient on its own. While advanced Semantic SLAM provides the foundational environmental context, the \gls{reid} research enriches this understanding with dynamic human presence and identity. Conversely, the robust environmental understanding gained through the advanced mapping techniques offers vital contextual cues that enhance the reliability and adaptability of person \gls{reid} in real-world settings.

The combined approach forms the foundation for developing more sophisticated and trustworthy autonomous systems. By integrating environmental mapping with dynamic object \gls{reid}, the robots can move beyond simple, reactive behaviors to exhibit proactive, socially compliant, and intelligent actions. This research thus paves the way for a new generation of robots that are not just machines in our world, but active and understanding participants in human-centric environments.

\section{Innovative aspects}

In the following, the contributions are quickly introduced.

This thesis presents several key contributions in the robotic context awareness realm; the most significant are:
\begin{itemize}
    \item A robust \gls{reid} method (see Section~\ref{sec:re-id-method}) that accurately recognizes specific people or objects despite partial or total occlusions over extended periods. It enhances robustness to appearance changes through a statistical continual adaptation and a continual learning technique. This technique employs a parallel twin neural network for online training, which allows the \gls{reid} application to operate without interruption.
    \item A novel \gls{3d} \gls{lidar} \gls{slam} approach (see Section~\ref{subsec:leo-slam-method}) that uses a submap-based keyframe strategy. This approach significantly enhances odometry and loop closure for more accurate and reliable mapping.
    \item A statistical version of a \gls{sota} loop closure algorithm known as \gls{sc}, where, by employing Gaussian probabilities, place recognition capabilities are improved (see Section~\ref{subsec:gsc-method}).
    \item The introduction of multi-modal semantic mapping to substantially improve a robot's environmental understanding capabilities (see Section~\ref{sec:semantic-method}).
\end{itemize}

The following sections provide a brief overview of these contributions. 

\subsection{Visual re-identification for human-robot collaboration}

The first contribution on context awareness is human \gls{reid} for \gls{hri}. The first introduced approach is FollowMe\,\cite{rollo2023followme}, a robust framework for person-following based on visual human \gls{reid} and hand gesture detection. The proposed architecture consists of three integrated modules: Perception, Decision Making, and Navigation. These modules receive sensor data and issue command references to the robot.

FollowMe is adaptable to various floating-base systems because it is built upon established robotics tools like \gls{ros}\,\cite{quigley2009ros}, and \gls{sota} machine learning algorithms such as Yolact++\,\cite{bolya2020yolact++}, \gls{mmt} pre-training\,\cite{ge2020mutual}, and Mediapipe\,\cite{zhang2020mediapipe}. A key feature is that the user can move the robot without manual teleoperation. This approach uses visual \gls{reid}, which is light, robust, and personalized, meaning the target person doesn't need to wear special patterns or intrusive devices like beacon or \gls{ir} emitters. A hand gesture recognition module is also introduced, allowing the user to send intuitive commands, such as "stop", to the robot.

This work highlights that human detection and \gls{reid} using visual data are powerful tools for personalized \gls{hrc}. The system is user-friendly because the robot can autonomously collect and use calibration data to re-identify and follow the target person. While much research focuses on isolated topics like multi-object tracking, this work addresses the more complex, real-world challenge of integrating mobile robotics into a practical application environment.

The potential applications for this framework are broad, including carrying heavy items, assisting with manipulation tasks at different stations\,\cite{lamon2020visuo}, and providing support for people with care needs\,\cite{eisenbach2015user}. The \gls{reid} and hand gesture detection modules are qualitatively evaluated using dedicated experimental setups, demonstrating impressive accuracy. Additionally, the entire FollowMe framework is qualitatively evaluated in a simulated working area, where the robot successfully follows the target, responds to hand gestures, and navigates around static and dynamic obstacles.


To further minimize human intervention in robot recognition and \gls{reid}, a key limitation of the original FollowMe\,\cite{rollo2023followme} framework is addressed: its inability to track a person who changes their appearance, such as by changing their outfit, after the initial calibration. To overcome this, a novel \gls{reid} module that employs a deep learning approach based on feature extraction and continual adaptation is proposed. As the robot tracks the person, it continuously acquires new images, using this new appearance information to update an ideal target representation model. This allows the system to re-identify the person even if tracking is momentarily lost.

This new module, called \gls{carpe}, integrates a \gls{mot} algorithm with an adaptive \gls{reid} layer. The \gls{mot} algorithm, known as \textit{yolo\_tracking}, is built upon the popular \gls{yolo} framework\,\cite{redmon2016you} and the StrongSORT tracker\,\cite{du2023strongsort}. The additional \gls{reid} layer adapts to changes in the target's appearance, effectively managing the "\gls{id} jumps" that can occur due to occlusions or appearance differences.

In this work, key contributions are:
\begin{itemize}
    \item The development of \gls{carpe}, a framework that achieves personalized \gls{hri} tasks with a specific target through continual adaptation.
    \item A quantitative evaluation of the system through real-world tests in a Human-Robot collaborative scenario.
    \item An in-depth analysis of the system's limitations and a discussion of potential solutions.
\end{itemize}

While \gls{carpe} was a significant step toward a comprehensive tracking framework, it suffered from a major limitation: catastrophic forgetting. This means it could not re-identify previous appearances of the target person over time, but just for a confined amount of it.

An enhanced version of the \gls{carpe} framework to solve the catastrophic forgetting issue\,\cite{rollo2025pesonalized} is proposed. This approach uses the output from \gls{carpe} to train a parallel twin feature extractor network in an unsupervised continual learning manner. This method allows the network to autonomously embed and learn the appearances of the specific target, creating a highly personalized and robust \gls{reid} system. A key innovation of this final version is the development of a smart image pool acquisition method specifically designed to overcome the catastrophic forgetting problem.

The final contributions of this \gls{reid} work are:
\begin{itemize}
    \item A novel unsupervised continual learning approach that trains a parallel feature extractor network online to create a personalized \gls{reid} framework. This unique combination has not been utilized in other \gls{reid} or tracking approaches in the literature.
    \item The development of an autonomous smart image pool acquisition method to directly address the issue of catastrophic forgetting.
\end{itemize}

In its final iteration, the \gls{reid} framework addresses the major challenges of personalized person tracking. While the original FollowMe framework was robust for simple \gls{hri} tasks, the introduction of \gls{carpe}, along with its subsequent enhancements through continual learning and smart image pooling, has addressed its limitations. This demonstrates a robust \gls{reid} behavior and a superior level of human-robot integration, which are essential for a complete robotic context awareness.

\subsection{Geometrical robotic mapping and place recognition}

Geometrical mapping approaches are predominantly based on \gls{slam} techniques. This is essential because perfect localization is rarely known in most environments, and a robot must localize itself to accurately create a map.

For this reason, LEO\raisebox{0.1ex}{-}SLAM\,\cite{rollo2025leoslam}, a \gls{lidar}-based SLAM algorithm designed to generate an environmental map that accurately represents real-world surroundings and supports navigation and exploration tasks, is developed. This approach is built upon recent theoretical and algorithmic advancements. Using scan-matching algorithms, odometry is refined using \gls{lidar} scans, and \gls{sc++}\,\cite{kim2021scan} is used for robust \gls{lcd}. The key contributions lie in the unique alignment and loop closure strategies. Specifically:
\begin{itemize} 
    \item A multi-level alignment strategy that not only compares the current scan with the previous one (scan-to-scan, $\mathrm{s}2\mathrm{s}$) and with the last $\mathrm{N}_s$ submaps (scan-to-submaps, $\mathrm{s}2\mathbb{S}$), similar to \gls{sota} methods but, additionally introduce a submap-to-submaps alignment step ($\mathcal{S}2\mathbb{S}$), which is performed once a submap keyframe is finalized.
    \item A submap-based \gls{sc++} \gls{lcd} method that uses an adaptive search area. This provides richer features for comparison compared to analyzing a single, sparse scan.
\end{itemize}

This efficient submap representation produces denser and more distinctive point clouds, which facilitates robust scan matching and significantly improves \gls{lcd}.


During real-world robot experiments, the generated map was not completely accurate because \gls{lidar} sensors are susceptible to reflections from transparent or reflective surfaces. These reflected points often have low intensity and can be filtered out by setting a simple threshold. However, this method can mistakenly remove ground points, as a low incidence angle between the \gls{lidar} ray and a reflective ground surface can also result in a low-intensity return.

To address these issues, an efficient ground-aware intensity filter is introduced into LEO\raisebox{0.1ex}{-}SLAM. This filter enhances \gls{lidar}-based systems by preserving crucial ground information while effectively filtering out erroneous reflections. This solution specifically tackles two key challenges: consistent erroneous reflections and the loss of valid low-intensity ground points.

The main contributions for this part of the work are:
\begin{itemize}
    \item A simple yet effective intensity-based filter that accounts for the robot's height to retain low-reflectance ground points.
    \item A comprehensive evaluation and comparison of the filter's effectiveness using \gls{sota} odometry systems. This solution is tested on a quadruped robot using a custom indoor dataset that includes numerous glass doors, demonstrating its robust performance in challenging real-world environments.
\end{itemize}

Another challenge was the accuracy of the place recognition algorithm. The conventional approach of setting a loop closure threshold often leads to a trade-off between low accuracy and an increased risk of false positives.

To address this, an extension of \gls{sc++}\,\cite{kim2021scan} called \gls{gsc} is proposed. As its name implies, the method integrates a statistical analysis of the input point cloud to generate a more robust environmental representation within the context matrix. This statistical formulation allows the algorithm to capture the overall point distribution, thereby enhancing its resilience to the noise and outliers that are common in real-world scenarios.

This work represents a significant step forward in enhancing the robustness of \gls{slam} algorithms for real-world robotic applications. The key contributions are:
\begin{itemize}
    \item A more robust and computationally efficient \gls{lcd} method achieved by integrating a statistical analysis of the point cloud.
    \item An in-depth study and comparative evaluation of multiple statistical distance metrics, including a discussion of the factors that influence their relative performance.
\end{itemize}

\subsection{Introducing semantics for higher-level scene understanding and interaction}

As previously discussed, while geometrical mapping is crucial for robotic navigation and exploration, performing higher-level autonomous tasks requires semantic understanding. Many \gls{sota} robotic approaches, reported in Section~\ref{sec:semantic-works}, are based on a single camera, which limits mapping capabilities to small indoor environments. Conversely, solutions in autonomous driving often use both \gls{lidar} and \gls{rgb} cameras but neglect camera depth estimation, which is not essential in their context.

To bridge the strengths of both robotics and autonomous driving, a modular, multi-modal (\gls{rgbd} camera-\gls{lidar}) online semantic mapping framework is proposed. This architecture can fuse sensor information in real-time, adapting its approach based on object distance and sensor accuracy. Semantic information from images is used to enrich the filtered and stabilized positions of objects, leading to precise object localization. The objects' dimensions are simplified as spherical shapes. This framework relies on external geometric navigation systems, such as SLAM or other localization algorithms like \gls{amcl}\,\cite{xiaoyu2018adaptive}.

The proposed application demonstrates good accuracy for both near and distant objects, thanks to this novel camera-\gls{lidar} depth fusion technique. To the best of the author's knowledge, this specific fusion approach has not been explored in other robotic or autonomous driving semantic mapping works. This application also operates online on low-resource embedded systems, further highlighting its practical contributions. A custom \gls{rviz}\footnote{rviz:~\url{http://wiki.ros.org/rviz}} plugin is introduced to improve the user experience for visualizing and interacting with the semantic map.

To demonstrate the utility of these semantic maps, an easy-to-use framework for a mobile robotic collaborator to complete a "bring me" service is developed. This work leverages the knowledge from the semantic mapping framework to accomplish the task, representing a step towards a more comprehensive and adaptable application that can dynamically respond to workers' needs.

\section{Thesis structure}

The thesis will follow a structured approach that moves from a more human-centric view to an environmental one. Specifically, the context awareness of the mobile robot will be first considered in a \gls{hri} scenario, to a robot-environment interaction, concluding with a conceptual union between the two that brings the robot to a semantic environmental context-awareness and understanding. This is done because those two topics complement each other to ensure human-robot coexistence and robot autonomy. For these reasons, the following sections are ordered from human \gls{reid} and robot interaction to robot environmental understanding and mapping.
Specifically, in Chapter~\ref{chap:works} the \gls{sota} is analyzed and compared. The human \gls{reid} and tracking techniques for \gls{hri} are analyzed in Section~\ref{sec:reid-works}, while the robotic environmental mapping is analyzed in Section~\ref{sec:slam-works}, concluding with an in-depth analysis of the semantic SLAM world in Section~\ref{sec:semantic-works}.

Chapter~\ref{chap:problem} will highlight the problems present in those fields and will provide hints on how they can be faced and handled. In Section~\ref{sec:association-prob}, the correspondence problem is analyzed. As can be seen in the next sessions, this problem arises many times in robotics, as, also for humans, data association is one of the primary tasks performed to maintain consistency between the world perceived with our senses and the cognition and representation of those perceptions.  Section~\ref{sec:mapping-prob} instead will focus on the environmental mapping problem. The problem is analyzed from both geometrical and semantic perspectives, and their dependencies and differences are highlighted. The mapping problem is also dependent on the previous one, as data association and correspondences are a crucial part of the mapping process. 
The methodologies and algorithms developed to solve these problems are presented in Chapter~\ref{chap:method}. Here, the theoretical information about the proposed solutions is described along with an explanation of the ideas behind them. Specifically, in Section~\ref{sec:re-id-method}, the \gls{reid} application for \gls{hri} is handled along with some subsequent improvements for robustness and accuracy. Section~\ref{sec:mapping-method} contains the geometrical contribution proposed to improve the 3D SLAM problem in terms of new approaches to the \gls{lidar} odometry, place recognition, and \gls{lcd}. Finally, in Section~\ref{sec:semantic-method}, the semantic-aided mapping methods are presented. These methods make use of the expertise acquired with the previous two modules and put them together for an increased contextual awareness of the environment.
The presented methods are then thoroughly analyzed and proved in Chapter~\ref{chap:experiments} where each component is validated using different techniques and using benchmarks and real robot experiments. In Section~\ref{sec:re-id-exp}, the \gls{reid} capabilities of the robot are evaluated, and a specific \gls{hri} task has been considered to prove the usability of those capabilities in a real robotic scenario. Section~\ref{sec:mapping-exp} demonstrates the mapping capabilities in terms of accuracy and performance of the proposed SLAM and loop closure methods, and Section~\ref{sec:semantic-exp} presents the results of the semantic contextual awareness and an example of how it can be used for improving robotic capabilities. 
As a conclusion, Chapter~\ref{chap:conclusion} reports the conclusive remarks and the take-home messages of this document, along with some of the wrong choices made during this journey and the limitations that need to be handled for future works.

\chapter{Related works} \label{chap:works}
    Robotic systems operating in close proximity to humans demand a high degree of contextual awareness to ensure both safety and operational efficiency. This critical capability is built upon the robot's fundamental perceptive skills: reliably identifying collaborators, accurately mapping the workspace geometry, and semantically understanding the objects and areas within it. This chapter provides a comprehensive review of the current and historical \gls{sota} research that directly addresses the core perceptive challenges investigated in this thesis.

The discussion begins with the \gls{reid} in Section~\ref{sec:reid-works}, focusing on the essential task of maintaining a persistent identity for collaborators across various viewpoints and occlusions. Successively, the broader challenge of environmental perception is addressed, divided into two key areas: geometrical perception as reviewed through \gls{slam} and related works in Section~\ref{sec:slam-works}, and semantic understanding, the assignment of meaning to raw sensor data, in Section~\ref{sec:semantic-works}. Collectively, the works presented herein have significantly contributed to enhancing the robot's perceptive capabilities, forming the necessary foundation for the novel contributions presented in this thesis.

\section{Visual re-identification for human-robot collaboration} \label{sec:reid-works}

Visual tracking and \gls{reid} are critical capabilities for robots intended to operate alongside humans and dynamic obstacles in unstructured environments. They are essential not only for safe coexistence but also for recognizing specific individuals with whom the robot must collaborate in settings like manufacturing. A primary requirement for mobile collaborative robots is the ability to accompany a person between different workstations to execute specific tasks. While conventional methods involve human-operated controls (\textit{e.g.}, joysticks), a more autonomous and user-friendly solution involves a robot with visual recognition executing a person-following action without continuous human intervention.

\subsection{Person following} \label{subsubsec:person-follow}

Existing applications in both industry (\textit{e.g.}, Piaggio's Gita and Kilo transport robots\footnote{Piaggio Gita and Kilo: \href{https://www.piaggio.com/gb_EN/piaggio-world/about-us-piaggio/the-revolutionary-nature-of-gita-and-kilo/}{\nolinkurl{Gita-and-Kilo}}}) and research employ a variety of sensors to achieve person-following. Many of these traditional methods rely on extrinsic or intrusive localization devices. For instance, some use dedicated signal emitters, such as Wi-Fi transmitters\,\cite{geetha2021follow}, \gls{ir} \glspl{led}\,\cite{dang2011human, afghani2013follow}, or Bluetooth connections\,\cite{pradeep2017follow}. Other approaches utilize simple range-based sensors, such as sonar rings and rangefinders\,\cite{peng2016tracking}. A drawback of these frameworks is the dependency on the target wearing or carrying specific, easily recognizable devices. Furthermore, many of these systems lack integration with robust obstacle avoidance algorithms, although\,\cite{afghani2013follow} does use an ultrasonic sensor for this purpose.

More robust detection has been achieved using vision-based sensors, particularly \gls{rgbd} cameras. Object detection is often used for many tasks that span from object grasping\,\cite{del2024high} to semantic mapping\,\cite{rollo2023artifacts}. Early work involved detecting specific objects, such as a red T-shirt, and using a range sensor for distance estimation and obstacle avoidance\,\cite{pinrath2018simulation}. Other approaches use anthropometric features, such as shoulder length and head height, for distinction and \gls{reid}\,\cite{shimoyama2017human}. Fusion techniques are also common, such as combining camera-based person detection with laser range finders\,\cite{sonoura2008person}. Feature extraction has evolved from using basic color and contour information\,\cite{schlegel1998vision} to more sophisticated holistic features like \gls{hog}\,\cite{dalal2005histograms}.

More recently, sophisticated tracking and detection methods have been employed. Eisenbach et al.\,\cite{eisenbach2015user} presented a care robot that detects legs via a laser range finder and the upper body using an orientation-based decision tree of HOGs, improving the tracking with color and clothing texture matching. Other solutions have fused \gls{slam} with vision, as seen in Weber et al.\,\cite{weber2017follow}, which uses LSD-\gls{3d} and a \gls{cnn} for head detection. The emergence of affordable aerial robotics has also seen the development of person-following drones that use holistic information (skeleton points) and simple gestures for control, as in Naseer et al.\,\cite{naseer2013followme}. Collaborative robotics has leveraged tools like OpenPose\,\cite{cao2019openpose} to track a person's skeleton in \gls{3d} using stereo cameras for tasks like pick and place, often integrating haptic feedback\,\cite{lamon2020visuo}.

Crucially, most of these works lack the necessary robustness and personalization for long-term collaboration, as they primarily rely on transient or broadly shared features (\textit{e.g.}, location, pose, simple colors) that are hardly distinguishable across individuals or robust to changes.

\subsection{Visual object tracking}

To overcome the limitations of the works presented in the previous Section~\ref{subsubsec:person-follow}, one of the approaches that can be used to enhance collaborative tasks is \gls{ot}. The field of visual tracking is broadly divided into \gls{sot} and \gls{mot}. \gls{sot} algorithms track one selected object based on an initial bounding box, regardless of class, while \gls{mot} algorithms detect and track multiple objects simultaneously.

Recent surveys classify \gls{sot} techniques based on feature analysis, segmentation, estimation, and learning\,\cite{soleimanitaleb2022single}. Specifically, correlation and deep learning algorithms have received detailed attention\,\cite{zhang2021recent, javed2022visual}, particularly those employing Siamese networks and discriminative correlation filters. Conversely, other reviews focus on the recent developments in \gls{mot} algorithms\,\cite{luo2021multiple}.

While many robust \gls{mot} systems exist, a performance trade-off often emerges. For example, TMOT\,\cite{stadler2021improving} performs well in cluttered and occluded environments but with unsatisfactory time performance for real-time robotic applications. Faster \gls{mot} algorithms, such as those based on the LMOT framework\,\cite{mostafa2022lmot}, solve the timing issue, but these trackers typically cannot re-identify objects that leave and then re-enter the camera \gls{fov}. This limitation represents a key area of improvement and one of the strengths of the proposed works\,\cite{rollo2023followme, rollo2024carpe, rollo2025pesonalized}.

\subsection{Other proposals}

As established, general \gls{sot} and \gls{mot} algorithms often fail to meet the demands of personalized \gls{hri} due to their lack of robustness in re-identifying targets after partial or total occlusions. Consequently, the literature on robotics applications offers more specialized \gls{reid} methods.

The use of soft biometrics, such as skeleton points and face features, is widely employed for person tracking and \gls{reid} in robotic assistance. For instance, Patruno et al.\,\cite{patruno2019people} proposed creating a person descriptor using soft biometric features extracted from depth and color information. They utilized AlphaPose\,\cite{fang2022alphapose} to extract \gls{3d} skeleton points, normalizing the target person to a standard posture. A grid applied to this normalized posture, with the mean color of each cell, formed the \gls{reid} descriptor. However, this method is limited by its reliance on clothing color, making it prone to errors when people wear similar outfits. Similarly, Ye et al.\,\cite{ye2023robot} developed a person-following system that uses a predefined model and skeleton heuristics to achieve robustness against partial occlusion. Crucially, this system assumes the person always remains within the camera's \gls{fov}, which is not guaranteed in dynamic real-world environments.

Face recognition provides another specialized approach. Liu et al.\,\cite{liu2017online} proposed a method that trains a metric model offline and then uses online face information to match the target and update the model. They introduced the \gls{ffm} to merge appearance and skeleton information. Likewise, Wang et al.\,\cite{wang2019real} improved \gls{hri} by using an unsupervised face \gls{reid} approach with a pre-trained \gls{cnn} feature extractor. A major practical limitation of both face-based methods is the assumption that the human collaborator consistently faces the robot, an assumption often invalidated in real-world scenarios.

Other explored works use different sensor modalities and learning strategies. Koide et al.\,\cite{koide2020monocular} employed a \gls{rgb} monocular approach for tracking, extracting the skeleton using OpenPose\,\cite{cao2017realtime}, and tracking the target with an \gls{ukf}. For \gls{reid}, they combined convolutional channel features with boosting techniques. However, their method requires an initial deep feature appearance calibration, which is used statically for subsequent \gls{reid}, lacking dynamic adaptation.

In a different direction, Cocsar et al.\,\cite{cocsar2020human} utilized a thermal camera for tracking and \gls{reid}. They trained a network using a custom dataset. They sampled data through an entropy process to create discriminative descriptors. These descriptors were then used to train a \gls{svm} classifier. Unfortunately, this approach necessitates extensive training for thermal integration and, critically, cannot adjust to changes in the target's appearance, making it unsuitable for long-term personalized \gls{hri}.

The examination above reveals that existing methods suffer from significant limitations, particularly the inability to robustly handle target appearance shifts (\textit{e.g.}, clothing changes) and a reliance on intrusive assumptions (\textit{e.g.}, constant visibility, frontal face-to-robot orientation). These restrictions do not satisfy the established specifications for a reliable, personalized tracking system.

This thesis addresses these gaps by proposing a solution that tackles the problem of person \gls{reid} robustly, even in the presence of target appearance shifts and occlusions. A system that generalizes the person-following task using deep neural network features for \gls{reid} and visual data, similar to how a human would perceive, with additional depth data for localization, is presented. The key contribution lies in the development and seamless integration of the entire visual pipeline: visual person detection, \gls{reid}, localization modules, and hand gesture detection for commanding. This comprehensive pipeline, or its individual components, can be readily adapted for other \gls{hri} tasks. The specific methods used for this integrated approach are presented in detail in Section~\ref{sec:re-id-method}.

\subsection{Continual learning} \label{subsect:contlearnworks}

To further enhance the performance of the personalized tracking system, an unsupervised, online training technique that employs \gls{cl} and real-time dataset creation to improve \gls{reid} is developed\,\cite{rollo2025pesonalized}. Given the novelty of concurrently addressing both advanced \gls{reid} and \gls{cl} within a robotic framework, this section first concludes by summarizing the related \gls{reid} literature presented above before detailing the \gls{sota} in \gls{cl}.

As highlighted in the previous section and in the prior work \gls{carpe}\,\cite{rollo2024carpe}, most \gls{sot} and \gls{mot} algorithms do not meet the strict requirements of personalized robotics, particularly their limitations in re-identifying targets after prolonged occlusions. Alternative methods that utilize skeleton point extraction for color-based \gls{reid}\,\cite{patruno2019people} or structure-matching\,\cite{ye2023robot} have been explored. However, these techniques are susceptible to errors when clothing is similar and often rely on the restrictive assumption that the person remains within the camera's \gls{fov}. Similarly, face-based recognition\,\cite{liu2017online, wang2019real} is compromised by the impractical assumption that the target consistently faces the robot, an issue confirmed in the experiments.

In the FollowMe framework\,\cite{rollo2023followme}, a learning-based \gls{reid} approach that used a calibration phase to adapt to the target's initial appearance is presented. While this module significantly improved \gls{hrc}, it was inherently sensitive to subsequent target appearance changes.

These accumulated limitations across the \gls{sota} make most existing approaches unsuitable for robust, personalized human tracking. The approach here presented builds directly upon the preliminary work, the \gls{carpe} framework\,\cite{rollo2024carpe}, which successfully addressed challenges related to occlusion and appearance changes. However, as noted in the Introduction, \gls{carpe} is not immune to catastrophic forgetting, meaning it struggles to re-identify a person reverting to an older appearance. This latest work specifically proposes a novel solution, the use of \gls{cl}, to this critical challenge, which is not sufficiently studied in the \gls{hri} tracking field.

Continual Learning, also known as Lifelong Learning, is a critical paradigm focused on training neural networks to adapt to new incoming data streams sequentially without suffering from catastrophic forgetting, the abrupt and severe loss of previously acquired knowledge. This field attempts to solve the fundamental stability-plasticity dilemma, ensuring the model retains old knowledge (stability) while being flexible enough to integrate new information (plasticity)\,\cite{parisi2019continual, wang2023comprehensive}.

\gls{cl} methodologies can be broadly categorized into three main approaches:
\begin{itemize}
    \item \textit{Rehearsal/Memory-Based Methods}: These techniques maintain a limited memory buffer of representative data samples (exemplars) from previous tasks. By interleaving or "rehearsing" these old samples alongside the new data, the model is constantly reminded of its past knowledge. An early example of this concept was presented by Robins et al.\,\cite{robins1995catastrophic}. More recent exemplar-based approaches, such as iCARL\,\cite{rebuffi2017icarl} and GDumb\,\cite{prabhu2020gdumb}, focus on sophisticated selection and management of this memory buffer to maximize information retention. The work proposed in this thesis draws inspiration from this concept to develop the smart image pool acquisition method, reinforcing the online unsupervised continual learning in the framework.
    \item \textit{Regularization-Based Methods}: This category mitigates forgetting by imposing constraints on the network's parameters during the learning of a new task. The core idea is to identify the weights that are critical to the performance of previous tasks and penalize any drastic changes to them. Landmark techniques in this area include \gls{ewc}\,\cite{kirkpatrick2017overcoming} and \gls{si}\,\cite{zenke2017continual}, which compute importance based on network sensitivity to weight changes. While computationally efficient due to low memory overhead, these methods can suffer from a poor ability to generalize to highly dissimilar new tasks, a critical trade-off addressed in recent theoretical analyses\,\cite{li2023fixed}.
    \item \textit{Dynamic Architectural Methods}: These approaches address the stability-plasticity dilemma by either expanding or dynamically modifying the network architecture as new tasks arrive. Methods like \gls{pn} dynamically allocate new resources (\textit{e.g.}, adding new network branches) for each task while keeping parameters for old tasks frozen\,\cite{rusu2016progressive}. Other methods utilize task-specific and task-shared parameters or dynamically expand the network only when capacity is insufficient.
\end{itemize}

In the context of Person \gls{reid}, this concept is often termed \gls{lreid}, which is particularly challenging as it is an open-set, fine-grained problem where new identities constantly appear across various camera domains. Recent \gls{lreid} studies have tackled issues like re-indexing free learning, where the model must remain compatible with features calculated by older models\,\cite{cui2024learning}. Others have proposed systems like Adaptive Knowledge Accumulation\,\cite{pu2021lifelong} to continually accumulate knowledge and improve generalization across unseen domains, often drawing inspiration from human cognitive models. Furthermore, work on Continual Meta Metric Learning aims to incrementally learn a feature space that generalizes well to new tasks\,\cite{aljundi2022continual}.

In contrast, earlier works focusing on \gls{reid} using deep metric learning, such as those employing multi-stage training with incremental triplet margins\,\cite{zhang2019learning} or an Easy Positive strategy\,\cite{xuan2020improved}, are typically restricted to static image datasets. As highlighted by Lesort et al.\,\cite{lesort2020continual}, while the concept of \gls{cl} strongly aligns with the goals of autonomous agents and embodied \gls{ai}, the integration of these sophisticated \gls{cl} techniques into a real-time tracking and personalized robotic application remains a complex and sparsely explored area. The \gls{reid} works proposed in this thesis provide a key contribution to this novel intersection.

\section{Environmental geometric mapping} \label{sec:slam-works}

When everyone introduces geometric mapping, the focus falls on \gls{slam}. \gls{slam} is a cornerstone of autonomous robotics, defined as the computational problem of constructing or updating a map of an unknown environment while simultaneously keeping track of an agent's location within it. A core challenge in \gls{slam} is the inherent "chicken-and-egg" problem: a map is needed to accurately determine the robot's location (localization), but an accurate location is required to correctly build the map (mapping)\,\cite{durrant2006simultaneous, bailey2006simultaneous}. Without a consistent solution, small errors in either process quickly accumulate, leading to map inconsistencies and complete loss of localization, a phenomenon known as drift.

The field of \gls{slam} addresses this circular dependency through probabilistic estimation techniques, which formulate the problem as a single joint probability function over the robot's trajectory $\mathrm{X}_{1:t}$ and the map's features $\mathbf{M}$, effectively exploiting the correlation between localization and mapping estimates to achieve convergence:

\begin{equation} \label{eq:slam-prob}
    P(\mathrm{X}_{1:t}, \mathbf{M} \mid \mathrm{Z}_{1:t}, \mathrm{U}_{1:t})
\end{equation}

where $\mathrm{X}_{1:t} = \{\mathbf{x}_1, \dots, \mathbf{x}_t\}$ is the sequence of robot poses (or states) up to time $t$, $\mathbf{M}$ is the environment map, $\mathrm{U}_{1:t} = \{\mathbf{u}_1, \dots, \mathbf{u}_t\}$ is the sequence of control inputs (odometry or motion commands) up to time $t$ and $\mathrm{Z}_{1:t} = \{\mathbf{z}_1, \dots, \mathbf{z}_t\}$ is the sequence of observations (sensor readings) up to time $t$. This joint probability can be broken down using the Bayes filter and the Markov assumption as explicitly expressed in Section~\ref{sec:mapping-prob}.

While \gls{slam} can be implemented using various sensors, including cameras (\gls{vslam}) and 2D laser scanners, \gls{3d} \gls{lidar} has become the gold standard for high-precision, large-scale, and real-time mapping in complex environments, particularly in autonomous driving and advanced robotics.

\gls{3d} \gls{lidar} sensors emit laser pulses to measure distances, generating a dense and geometrically accurate point cloud of the environment. Unlike \gls{vslam}, \gls{lidar}-based systems are robust to variations in ambient light and texture, offering superior performance in poor lighting or feature-sparse scenes.

Modern \gls{3d} \gls{lidar} systems are typically structured into a front-end and a back-end.

The front-end takes care of the odometry and mapping. This module handles the high-frequency sensor data, primarily focusing on scan-to-scan or scan-to-map registration to estimate the robot's pose (position and orientation) between consecutive frames. Early but influential methods, such as \gls{loam}\,\cite{zhang2014loam}, achieved low drift and high efficiency by exploiting feature extraction from the point cloud data. Subsequent tightly coupled approaches, like Fast-LIO\,\cite{xu2022fast}, fuse \gls{lidar} measurements directly with data from an \gls{imu} at an early stage, significantly improving robustness and accuracy by compensating for sensor motion distortion.

The back-end instead focuses on the global optimization of the problem formulated in the front-end. This module periodically refines the overall consistency of the map and trajectory. It typically uses factor graphs, a graph-based structure where robot poses and landmark positions are nodes, and measurements are edges. A specialized factor graph structure composed only of pose nodes is called a pose graph. A critical function here is \gls{lcd}, which recognizes previously visited areas and adds constraints to the factor graph. This step is essential for correcting the cumulative drift that is unavoidable in the odometry stage, often leveraging non-linear optimization libraries like GTSAM\footnote{GTSAM: \url{https://gtsam.org/}} or g2o\footnote{G2O: \url{https://g2o.com/}}.

The \gls{sota} in \gls{3d} \gls{lidar} is now moving towards integrating these traditional geometric methods with deep learning to handle challenges such as sensor data degradation in high-dynamic or featureless environments, and to manage the compatibility of features in Lifelong \gls{slam} scenarios.

\subsection{3D LiDAR SLAM}

\gls{lidar}-based \gls{slam} has gained significant attention due to its robustness and suitability for mobile robotics, especially in large-scale and challenging environments. Several surveys have documented advancements in this field, including a recent study by Li et al.\,\cite{li2025review}, which categorizes modern \gls{slam} systems into three main approaches: \gls{lidar}-based, multi-sensor, and deep learning-based \gls{slam}. Their analysis examines key components, including front-end data association, \gls{lcd}, back-end optimization, and map construction.

While vision-based \gls{slam}\,\cite{zhang2015visual} offers advantages in feature extraction and \gls{lcd} due to the rich textures captured by \gls{rgb} cameras, it is highly susceptible to variations in illumination, occlusions, and viewpoint changes. As a result, \gls{lidar} \gls{slam} is often preferred in scenarios where these environmental factors pose significant challenges.

Alternatively, multi-sensor solutions leverage complementary modalities to enhance \gls{slam} performance. For example, the \gls{lvi} \gls{slam} system by Shan et al.\,\cite{shan2021lvi} tightly fuses multiple sensor streams, while other approaches\,\cite{chen2019suma++, rollo2023artifacts} integrate semantic information to enrich map representations. Despite these advancements, the \gls{2d} \gls{lidar} \gls{slam} system developed by Macenski et al.\,\cite{macenski2021slam} remains one of the most widely adopted solutions, stemming largely from its seamless integration with the ROS2 Navigation Stack for real-world deployment.

\gls{3d} \gls{lidar} \gls{slam} odometry primarily relies on point cloud matching using least-squares optimization techniques. This typically involves variants of the \gls{icp} algorithm\,\cite{besl1992method}, such as \gls{gicp}\,\cite{segal2009generalized}, to estimate the robot's motion between consecutive scans.

Methods like KISS-ICP\,\cite{vizzo2023kiss} focus on lightweight and efficient \gls{lidar} odometry, relying solely on geometric information and direct scan matching. Similarly, \gls{dlo}\,\cite{chen2022direct} proposes a fast and robust \gls{lidar}-based localization method that utilizes a custom \gls{icp} solver and combines both scan-to-scan and scan-to-map matching to maintain accuracy in challenging conditions. To counteract the effects of aggressive motions, approaches like POINT-LIO\,\cite{he2023point}, which is an improvement over FAST-LIO2\,\cite{xu2022fast}, integrate \gls{imu} data to perform high-frequency, point-wise \gls{lidar}-inertial odometry, employing an iterated extended \gls{kf} for real-time state estimation. However, a well-known limitation of all \gls{lidar} odometry systems is pose drift, \textit{i.e.}, the accumulation of small errors that results in map distortions, making them unreliable for long-term global consistency.

Several \gls{3d} \gls{lidar} \gls{slam} solutions take a feature-based approach, where raw point clouds are processed into distinct, robust features for odometry estimation. A prominent example is \gls{loam}\,\cite{zhang2014loam}, which extracts edge and plane features and sequentially matches them with two separate feature maps. This approach inspired time-efficient adaptations such as LeGO-\gls{loam}\,\cite{shan2018lego}, which incorporates ground plane constraints into a two-step optimization process, and F-\gls{loam}\,\cite{wang2021f}, which improves accuracy by introducing surface smoothness into feature computation while employing both scan-to-scan and scan-to-map strategies. A further refinement, LIO-SAM\,\cite{shan2020lio}, tightly integrates \gls{imu} data to enhance accuracy and performance compared to other \gls{loam}-based approaches. While feature-based approaches offer improved computational efficiency for low-resource devices, their accuracy is inherently limited because feature matching can provide less comprehensive information than direct geometric point cloud alignment. Furthermore, \gls{imu}-based methods introduce additional challenges, such as sensor calibration and synchronization, which can cause systematic errors if not handled properly.

To correct the cumulative drift from the odometry front-end, a complete \gls{slam} framework requires a back-end for global map optimization. For instance, ART-SLAM\,\cite{frosi2022art} combines scan-to-scan \gls{lidar} odometry, \gls{lcd}, and pose graph optimization. However, relying solely on scan-to-scan matching can lead to significant odometry drift, which pose-graph optimization alone may struggle to correct in complex environments.

To address these limitations, an approach that employs a multi-level scan matching strategy that avoids using the full map, instead relying on recent local submaps to maintain consistency between consecutive submaps, is proposed in Section~\ref{sec:mapping-method}. Additionally, submap-based \gls{sc++}\,\cite{kim2021scan} \gls{lcd} is used to identify revisited locations and correct the drift introduced by odometry errors. A pose graph is constructed during mapping and globally optimized using \gls{isam2}\,\cite{kaess2012isam2} when a loop closure is successfully detected.

\subsubsection{Point clouds wrong-reflections filtering through point intensity}

\gls{lidar} intensity data, which represents the returned energy level of the emitted laser beams, has emerged as a valuable resource for enhancing \gls{lidar} odometry and \gls{slam} systems. This measurement provides additional information about the reflectance properties of objects' materials\,\cite{kashani2015review}. For example, highly reflective metallic surfaces typically yield high-intensity returns, while materials like concrete exhibit medium- to low-intensity returns. By complementing geometric features, these intensity measurements add a form of semantic information to the point cloud, leading to more robust data interpretation.

Several studies have utilized intensity data to improve \gls{slam} components. Wang et al.\,\cite{wang2020intensity} proposed the \gls{isc}, extending the original \gls{sc} approach\,\cite{gkim-2018-iros} by integrating intensity information to enhance loop closure and place recognition capabilities.
He et al.\,\cite{he2023igicp} incorporated intensity into the \gls{icp} algorithm\,\cite{besl1992method}, leading to significant performance improvements in \gls{lidar} odometry systems.
Other works by Guadagnino et al.\,\cite{guadagnino2022fast} for sparse \gls{lidar} odometry and by Wang et al.\,\cite{wang2021intensity} for large-scale environment mapping also rely heavily on intensity. However, a common practice in these approaches is to filter out or threshold low-intensity points as outliers. This method can inadvertently remove genuine points of interest, such as those on the ground, which naturally exhibit low returns due to their low incidence angle $\theta$ (see the left image in Fig~\ref{fig:filter_compare} compared to the right one where the ground is maintained).

\begin{figure}[t]
    \centering
    \includegraphics[width=\linewidth]{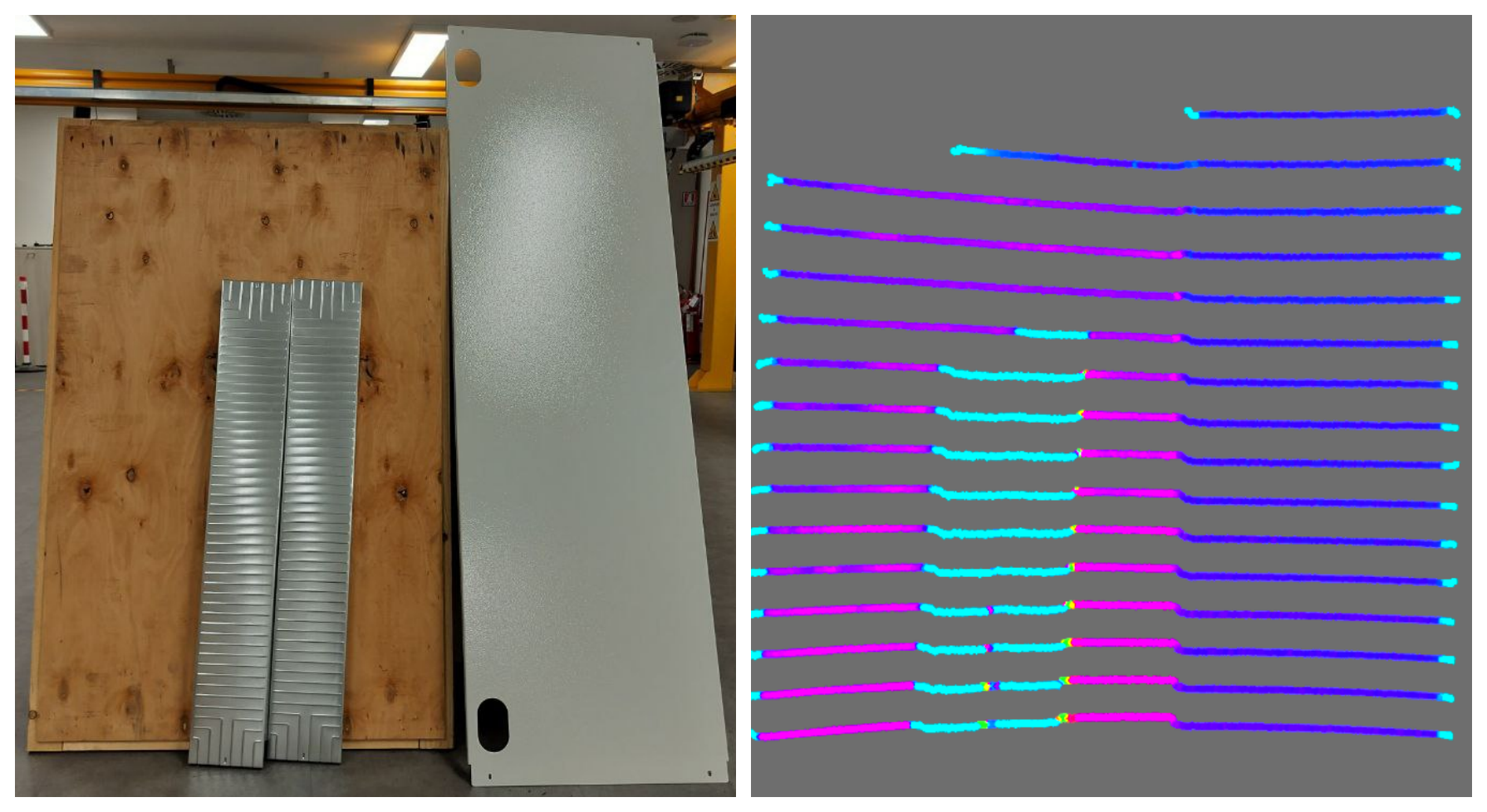}
    \caption{Comparison of \gls{lidar}-perceived intensities across diverse materials. The intensity ranges from the lower values of shiny metallic surfaces, represented in light blue, through the medium values of opaque metallic surfaces in violet, to the higher values of the wooden panel in pink.}
    \label{fig:lidar_intensity}
\end{figure}

The measured \gls{lidar} intensity is a complex function of the target surface's reflectance and roughness, as well as external factors like distance, incidence angle, and the sensor's transmitted energy\,\cite{kashani2015review}. For instance, a metallic surface may exhibit lower intensity values than a wooden panel (see Fig.~\ref{fig:lidar_intensity}), primarily due to variations in reflection, surface texture, and the surface's orientation relative to the \gls{lidar} scanner.

In indoor environments, highly reflective surfaces (\textit{e.g.}, mirrors) and transparent surfaces (\textit{e.g.}, windows) are common and often lead to erroneous point detections in the \gls{lidar} cloud. These artifacts pose significant challenges for odometry and \gls{slam} systems.

\begin{figure*}
    \begin{subfigure}[b]{\textwidth}
        \includegraphics[width=\linewidth]{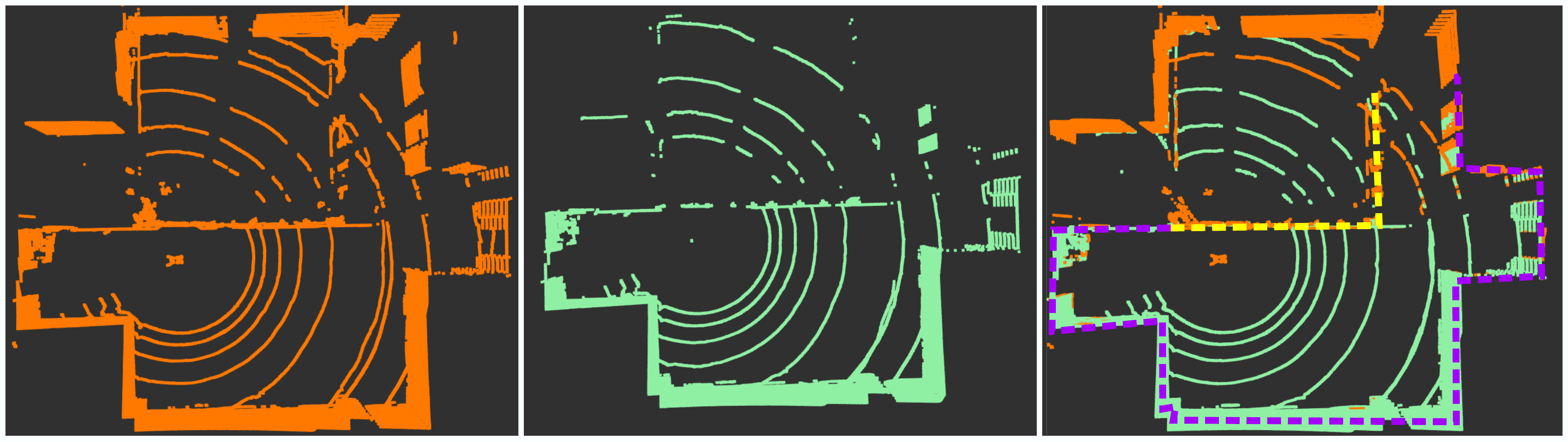}
        \caption{First filtering example with windows on both sides, in a narrow corridor.}
        \label{fig:filter1}
    \end{subfigure}
    \begin{subfigure}[b]{\textwidth}
        \includegraphics[width=\linewidth]{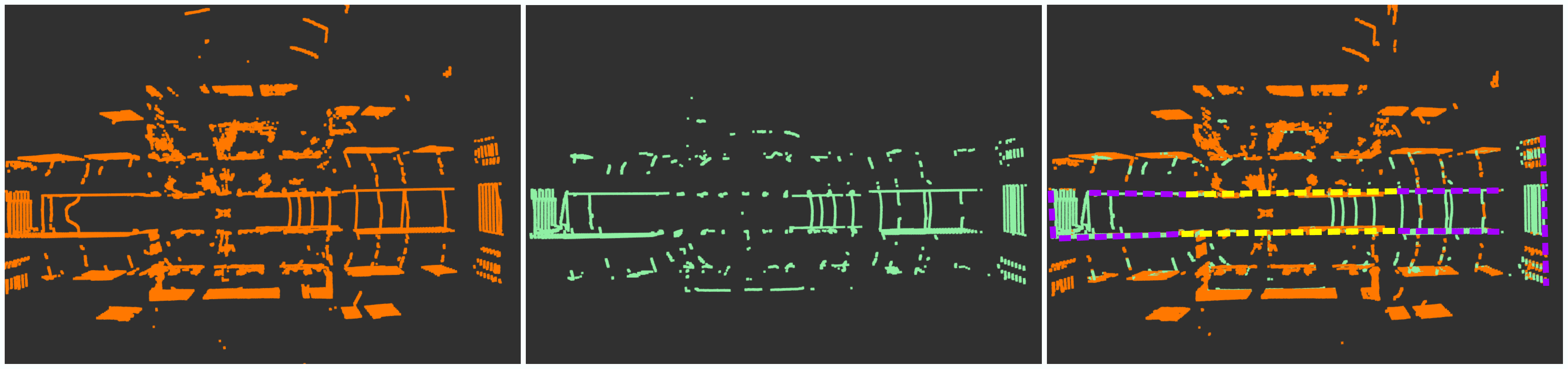}
        \caption{Second filtering example with windows only on one side, in a wide room.}
        \label{fig:filter2}
    \end{subfigure}
    \caption{Examples of the filtering effect are shown in two scenarios that show numerous erroneous reflections caused by the presence of windows. The original point cloud, collected from the robotic platform, is shown in orange, while the point cloud obtained using the ground-aware intensity filter appears in light green. The purple lines in the right images represent the structure of the environment/scenario, and the yellow lines indicate the positions of the windows. The overlap in the right image demonstrates that the proposed filter successfully removes almost all incorrect reflections while preserving ground points.}
    \label{fig:filter}
\end{figure*}

In indoor settings, reflections from objects like windows create significant complications. For example, in Fig.~\ref{fig:filter}, windows cause spurious point detections (orange cloud) due to reflections along the yellow-marked boundary. The filtered point cloud (light green) is compared against the ground truth (purple line).

This challenge is especially pronounced in scenarios like navigating in a room with windows on one side that create a mirroring effect, projecting reflections onto both sides of the robot (see Fig.~\ref{fig:filter1}) or in a narrow corridor with windows on both sides (see Fig.~\ref{fig:filter2}). In these complex scenarios, intensity information is beneficial not only for semantic data enrichment but also as a crucial tool for filtering erroneous measurements caused by reflections and refractions.

Intensity-based filtering has previously been investigated in autonomous driving applications to remove noise caused by weather phenomena like snow or rain. Early work by Hui et al.\,\cite{hui2008laser} used a simple intensity threshold to filter noisy points from raindrops and snowflakes. More recent and sophisticated methods include \gls{lior} by Park et al.\,\cite{park2020fast}, \gls{ddior} by Wang et al.\,\cite{wang2022scalable}, and \gls{lidsor} by Huang et al.\,\cite{huang2023lidsor}.

\begin{figure}[t]
    \includegraphics[width=\linewidth]{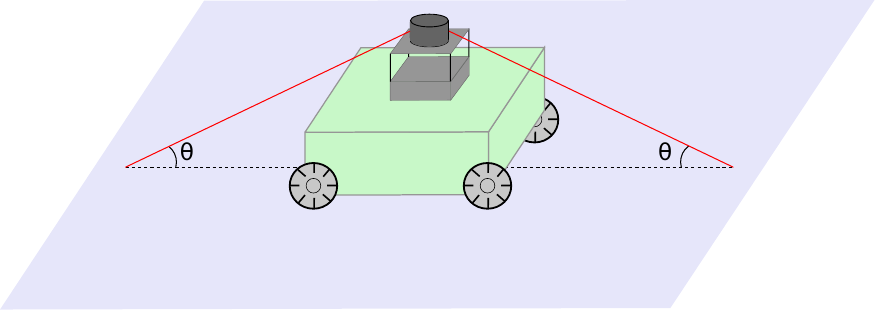}
    \caption{The incidence angle $\theta$ between the \gls{lidar} beam and the reflective surface significantly impacts the returned energy level, often reducing the point intensity. Points affected in this way would typically be removed by standard threshold filtering methods.}
    \label{fig:incidence_angle}
\end{figure}

These methods are specifically designed to address the sporadic and random noise characteristic of weather events. Consequently, they are often ineffective against consistent reflections (see left images in Fig.~\ref{fig:filter}), as they struggle to distinguish a steady, incorrect reflection from a real environmental point. Moreover, they do not intrinsically account for scenarios where erroneous low-intensity readings arise from non-weather factors, such as a low incidence angle between the \gls{lidar} ray and a reflective surface (see Fig.~\ref{fig:incidence_angle}). This is a critical point in indoor environments where surfaces, like the floor, may exhibit low-intensity values due to a combination of material reflectance and low incidence angles, a challenge that weather-based filters do not directly address.

In this thesis, a simple but effective approach for filtering out reflection points without removing the ones belonging to the ground is proposed. This solution enables the maintenance of ground information essential for navigation while obtaining a correct map of the environment (see Fig.~\ref{fig:filter}).

\subsection{Robust loop closure detection}

\gls{lcd} is a critical function in \gls{slam}, responsible for recognizing a previously visited location and thus providing a constraint that corrects the cumulative drift inherent in odometry. This section reviews the evolution of \gls{lcd}, with a specific focus on \gls{lidar}-based approaches. For a broader context, comprehensive surveys on \gls{lcd} techniques across various sensing modalities are available from Tsintotas et al.\,\cite{tsintotas2022revisiting} (focusing on visual \gls{lcd}) and Arshad et al.\,\cite{arshad2021role} (examining both visual and \gls{lidar} methods, with attention to deep learning).

\subsubsection{Visual LCD}
Visual \gls{lcd} has been extensively investigated, leveraging the rich texture and feature information captured by \gls{rgb} cameras.

Early influential methods include FAB-MAP\,\cite{cummins2008fab}, which used a probabilistic \gls{bow} model and Chow–Liu trees based on visual appearance. Later, SeqSLAM\,\cite{milford2012seqslam} improved robustness against environmental variations (\textit{e.g.}, lighting, seasons) by comparing sequences of images rather than individual frames. The widely adopted ORB-SLAM3\,\cite{campos2021orb} performs \gls{lcd} using a \gls{bow} model constructed from \gls{orb} features, refining loop candidates through geometric verification and back-end pose graph optimization to mitigate accumulated drift.

More recently, the field has seen a shift toward learning-based approaches. NetVLAD\,\cite{arandjelovic2016netvlad} introduced a \gls{cnn}-based layer that replicates \gls{vlad} encoding\,\cite{arandjelovic2013all}, enabling end-to-end training for visual place recognition. Furthermore, hybrid methods like SymbioLCD\,\cite{kim2021symbiolcd} integrate \gls{cnn}-extracted object semantics with traditional visual \gls{bow} features, exploiting the complementary strengths of deep representations and classical features to enhance robustness across diverse environments.

\subsubsection{LiDAR LCD}

\gls{lidar}-based \gls{slam} and \gls{lcd} methods are often preferred in industrial and outdoor robotics due to their high sensor precision and robustness against environmental variations, where visual techniques struggle.

Early influential \gls{lidar} \gls{slam} systems like \gls{loam}\,\cite{zhang2014loam} and its successors, such as LeGO-\gls{loam}\,\cite{shan2018lego}, extract edge and planar features for pose estimation and incorporate a basic loop closure mechanism within their global optimization framework. A key advancement in traditional \gls{lidar} \gls{lcd} is \gls{sc}\,\cite{gkim-2018-iros}, which forms the foundation of the proposed method in Section~\ref{subsec:gsc-method}. SC is an efficient and effective descriptor-based approach that generates a rotation-invariant representation of the point cloud. Subsequent improvements, such as \gls{sc++}\,\cite{kim2021scan}, incorporate rotation invariance and acceleration strategies, and \gls{isc}\,\cite{wang2020intensity} utilizes intensity values rather than height for feature representation.

Recent research has focused on enriching the \gls{lidar} input to enhance loop closure accuracy.
SA-\gls{loam}\,\cite{li2021sa} improves localization by integrating semantic features, while BoW3D\,\cite{cui2022bow3d} adapts the traditional BoW approach for real-time \gls{lcd} using \gls{3d} point cloud data.

Learning-based solutions have become increasingly prevalent\,\cite{kim2021symbiolcd, arce2023padloc, wang2024sglc}. PADLoC\,\cite{arce2023padloc} uses a transformer-based architecture that leverages panoptic segmentation during training to enhance both detection and registration. SGLC\,\cite{wang2024sglc} presents a semantic graph-guided framework that differentiates foreground from background regions for efficient detection and precise 6-\gls{dof} pose estimation. Similarly, Yang et al.\,\cite{yang2025lidar} utilize semantic graphs and graph-attention networks to improve the loop closure process. A more holistic strategy is seen in DLC-SLAM\,\cite{liu2023dlc}, which integrates a learning-based denoising module with a dedicated \gls{lcd} network to increase robustness in noisy environments.

While deep learning-based solutions offer cutting-edge performance, traditional methods remain a preferred choice within some robotics communities\,\cite{kim2021scan, gupta2025efficiently} due to their lower computational demands and independence from \gls{gpu} hardware, which is critical for energy-constrained platforms. For instance, Gupta et al.\,\cite{gupta2025efficiently} propose a robust \gls{lcd} pipeline based on point cloud density maps that efficiently detects loops across diverse \gls{lidar} configurations without deep learning. Nevertheless, ongoing advancements in hardware and software optimization are steadily shifting the trend toward more resource-intensive, learning-based approaches.

A notable limitation of \gls{sc}-based methods, despite their efficiency and effectiveness, is their susceptibility to outliers in \gls{lidar} point clouds, a common issue in real-world mapping due to sensor noise and reflective surfaces. To mitigate this, \gls{gsc} is introduced in Section~\ref{subsec:gsc-method}. It is a robust variant of \gls{sc++} that incorporates statistical analysis of the point cloud to improve the reliability and accuracy of \gls{sc++} in the presence of noise and outliers.

\section{Environmental semantic mapping and understanding} \label{sec:semantic-works}

The semantic mapping problem, \textit{e.g.}, integrating high-level semantic labels such as "chair", "wall", "door" into a geometric map has been a major focus in both robotics and autonomous driving. Several surveys offer valuable perspectives on this topic. Achour et al.\,\cite{achour2022collaborative} explore its application in \gls{hrc} for indoor environments and in a similar way as Pericu et al.\,\cite{pericu2025agnostic} employed semantic information to vocally commands robot actions; Xia et al.\,\cite{xia2020survey} provide a general analysis of semantic \gls{slam}, examining perception, robustness, and accuracy; and Kostavelis et al.\,\cite{kostavelis2015semantic} offer a useful historical reference by reviewing the early foundational developments before 2014.
One of the latest reviews is the one of Raychaudhuri et al.\,\cite{raychaudhuri2025semantic} which discusses semantic representations, map structures (spatial grids, topological, dense, hybrid), encodings (explicit/implicit), evaluation (task-level/extrinsic and map-level/intrinsic), and future directions. Also, Galagain et al.\,\cite{galagain2025semantic} review the semantic mapping under an embedded system perspective. They analyzed semantic-geometric \gls{slam}, \gls{nerf}, and \gls{3dgs} and tested some of them on a Nvidia Jetson Orin AGX to see if they are suitable for mobile robots with constrained performances. They demonstrated that although \gls{nerf} and \gls{3dgs} are good for realistic representation purposes, they are too slow for real-time applications. Latest trends try to adapt hardware structure to the new software, opposite to the common trend of adapting software to the available hardware.

The last decade in robotics has seen a rapid evolution of semantic mapping techniques, moving from feature-based object recognition to sophisticated deep learning and volumetric fusion approaches.

Initial successful examples demonstrated the performance benefits of incorporating semantic objects into the \gls{slam} loop. 
Civera et al.\,\cite{civera2011towards} presented a monocular \gls{slam} system that used a \gls{surf}\,\cite{bay2008speeded} feature extractor to check correspondences and reconstruct object geometry.
Salas et al.\,\cite{salas2013slam++} introduced an object-oriented \gls{3d} \gls{slam} that used \gls{icp}\,\cite{besl1992method} for object pose refinement, demonstrating that semantic objects can improve overall \gls{slam} performance.
Pillai et al.\,\cite{pillai2015monocular} developed a monocular \gls{slam}-aware object recognition system based on multi-view object proposals and efficient feature encoding, producing a semi-dense semantic map.

The advent of \gls{rgbd} cameras (like Microsoft Kinect) enabled approaches that tightly coupled geometry and semantics directly in \gls{3d}.
Tateno et al.\,\cite{tateno20162} proposed a framework that directly manages \gls{3d} objects, using a Kinect camera to reconstruct and classify objects while estimating their pose.
Xiang et al.\,\cite{xiang2017rnn} introduced the Data Associated Recurrent Neural Networks (DA-RNN) for semantic labeling of \gls{rgbd} videos, fusing its output with the KinectFusion algorithm\,\cite{newcombe2011kinectfusion} to merge semantic and geometric data.
McCormack et al.\,\cite{mccormac2017semanticfusion} leveraged a \gls{cnn} with the ElasticFusion \gls{slam} algorithm\,\cite{whelan2015elasticfusion} to achieve long-term dense correspondences and semantic labeling, even in loopy trajectories. Sunderhauf et al.\,\cite{sunderhauf2017meaningful} built on ORB-SLAM2\,\cite{mur2017orb} for geometric reconstruction, using Single-Shot multi-box Detector (SSD)\,\cite{liu2016ssd} and unsupervised \gls{3d} segmentation to place objects.

Recent works have focused on dynamic environments, real-time performance, and structured map representations.
Zeng et al.\,\cite{zeng2018semantic} introduced Contextual Temporal Mapping (CT-Map), modeling semantic inference as a \gls{crf} to account for contextual relations and temporal consistency.
MaskFusion\,\cite{runz2018maskfusion} provided a real-time, object-aware semantic and dynamic \gls{rgbd} \gls{slam} that continuously labels and handles dynamic objects, a significant improvement over predecessors.
Fusion++\,\cite{mccormac2018fusion} performed object-level \gls{slam} based on a \gls{3d} graph map of reconstructed objects, utilizing \gls{rgbd} cameras, Mask-RCNN\,\cite{he2017mask} instance segmentation, and the \gls{tsdf} for semantic reconstruction.
Grinvald et al.\,\cite{grinvald2019volumetric} incrementally built a volumetric object-centric map but suffered from limited time performance (1 Hz). Conversely, Pham et al.\,\cite{pham2019real} achieved real-time dense reconstruction and semantic segmentation by using an efficient super-voxel clustering method and \gls{crf} with higher-order constraints, running in parallel with a real-time \gls{3d} reconstructor.
Kimera\,\cite{rosinol2020kimera} is an open-source C++ library for real-time metric-semantic Visual-Inertial \gls{slam}, providing modular components for \gls{vio}, pose graph optimization, and dense \gls{3d} metric-semantic reconstruction.
Bultmann et al.\,\cite{bultmann2021real} used a \gls{uav} equipped with \gls{lidar}, an \gls{rgb} camera, and a thermal camera to augment \gls{3d} point clouds and image segmentation, generating an allocentric map.
Hughes et al.\,\cite{hughes2022hydra} presented a semantic mapping framework that uses only \gls{rgb} data and exploits \gls{3d} dynamic scene graphs\,\cite{rosinol20203d} to abstract different layers of inference (object, room, building), solving loop closure and mapping problems comprehensively.
Hau et al.\,\cite{hau2022object} used \gls{rgbd} cameras to reconstruct an allocentric semantic map, employing a \gls{cnn} keypoint extractor trained on synthetic data for pose estimation and a variant of the \gls{pnp} algorithm to recover object poses from multi-camera views.

Instead, in autonomous driving, the challenge often revolves around multi-sensor fusion in large, outdoor environments.
Liang et al.\,\cite{liang2019multi} addressed \gls{3d} object detection using \gls{lidar} and \gls{rgb} camera fusion to estimate object positions via ground estimation and depth completion, training their multi-task network with an end-to-end approach.
Chen et al.\,\cite{chen2019suma++} demonstrated building a semantic map through laser-based semantic segmentation of the point cloud, without requiring any camera data.
Li et al.\,\cite{li2020building} performed \gls{lidar}-based \gls{slam} for geometric mapping, then used a \gls{crf} to fuse and optimize camera semantic labels to obtain the final semantic map.
Berrio et al.\,\cite{berrio2021camera} used camera and \gls{lidar} data to construct a probabilistic semantic octree map, explicitly accounting for the uncertainties of all involved sensors.
Cheng et al.\,\cite{cheng2022vision} presented a recent work that uses an \gls{rgb} camera and \gls{lidar} for semantic segmentation, combined with direct sparse visual odometry and global optimization to include \gls{gnss} data in the mapping process.

This review indicates a key divergence in the \gls{sota}: robotics platforms often rely solely on camera measurements, with experiments typically confined to small indoor environments. Conversely, the autonomous driving scenario already embraces camera-\gls{lidar} fusion for semantic tasks, but often uses high-end, powerful \gls{lidar} sensors (\textit{e.g.}, 128-row vs. common 16-row in robotics) and focuses on outdoor driving challenges.

Hence, the proposed work in this thesis aims to stress the fact that \gls{rgbd} cameras and \gls{lidar} sensors are complementary sensors also in robotic semantic applications. For the semantic mapping application, the synergistic use of both sensors allows for correctly localizing objects across different distance ranges, thus significantly improving overall detection accuracy.

A particularly interesting and rapidly emerging area in both research and industry is Open-World Semantic Mapping. In contrast to the traditional semantic mapping approaches reviewed thus far, which rely on a fixed, predefined set of classes (\textit{e.g.}, "door", "chair", "fire extinguisher"), the open-world paradigm leverages the generalizability and grounding capabilities of \glspl{llm}, or, more accurately, \glspl{vlm}, to provide a mapping framework that does not require retraining for novel objects or concepts. This allows a robot to recognize and localize any object described in natural language, extending the map's semantic utility far beyond its initial training data. Although this field is quite new, a recent overview of existing works on open-world semantic mapping is provided by Miao et al.\,\cite{miao2025frontier}.

The field is evolving quickly, with a few foundational works setting the current \gls{sota}.
OpenScene\,\cite{peng2023openscene} is one of the first prominent open-world semantic mapping systems. OpenScene employs the powerful \gls{vlm} \gls{clip}\,\cite{radford2021learning} to bridge the visual and textual domains. The core idea is to associate \gls{clip} embeddings (high-dimensional vectors representing visual and semantic information) with each point in the \gls{3d} point cloud. Once the map is built, a user can query the map using a text input (\textit{e.g.}, "locating the fire extinguisher" or "show me all flat surfaces") and the system uses the text embedding to directly discriminate and segment the corresponding objects, affordances, or specifics within the map. This enables flexible and zero-shot semantic query capabilities.

Similar to OpenScene, OVO-SLAM (Open-Vocabulary Object \gls{slam})\,\cite{martins2024open} integrates \gls{clip} embeddings, but focuses on an online semantic mapping framework that maintains consistency as the robot moves. By structuring the map around individual objects identified through the \gls{vlm} and integrating these into the process, OVO-SLAM achieves real-time, open-vocabulary localization and mapping, allowing the map to be continuously queried and updated with new semantic concepts as they are encountered.

Taking a different approach to representation and visualization, Yang et al.\,\cite{yang2025opengs} integrated open-world semantic mapping into a \gls{3dgs} framework (OpenGS). \gls{3dgs} is a novel, highly efficient \gls{3d} rendering technique. OpenGS leverages this to provide not only open-vocabulary semantic recognition but also a smoother, high-fidelity visual experience when querying and visualizing the semantic map. This combination of open-world semantics with next-generation rendering addresses both the functional (recognition) and experiential (visualization) aspects of large-scale mapping.

The literature review on semantic mapping reveals a clear evolution from feature-based object recognition to sophisticated multimodal, deep learning-driven fusion systems. Initial works demonstrated the performance benefits of incorporating object geometry into the \gls{3d} loop\,\cite{civera2011towards, salas2013slam++}, leading to the wide adoption of \gls{rgbd} volumetric fusion methods that tightly coupled semantics and \gls{3d} structure\,\cite{mccormac2017semanticfusion, runz2018maskfusion}. More recent robotics research emphasizes building structured, allocentric maps using semantic scene graphs\,\cite{rosinol20203d, hughes2022hydra} and robust multi-sensor frameworks\,\cite{rosinol2020kimera}.

A fundamental divergence exists between domains: autonomous driving has successfully integrated high-power \gls{lidar}-camera fusion for outdoor semantic tasks\,\cite{liang2019multi, cheng2022vision}, whereas many robotics solutions historically favored camera-only approaches in small indoor spaces\,\cite{pillai2015monocular}. This disparity highlights a crucial opportunity: combining the geometric accuracy of \gls{lidar} with the rich semantic detail of \gls{rgbd} cameras is essential for creating highly accurate and semantically dense maps across the diverse operational range required by advanced mobile robots.

Finally, the emerging field of Open-World Semantic Mapping\,\cite{miao2025frontier} signals the next frontier. By leveraging \glspl{vlm} like \gls{clip}\,\cite{peng2023openscene, martins2024open}, these systems move beyond predefined classes to enable natural language querying of the environment, representing a significant step toward truly flexible and human-interpretable autonomous systems.

\chapter{Problem statement} \label{chap:problem}
    
Throughout this thesis, different problems were addressed. Some of them have similar foundations but evolve in various fields, while others are specific to their respective sectors. In the following, the main problems to enhance the context awareness of mobile robots are presented.

\section{Data association and corresponding problem} \label{sec:association-prob}

The fields of robotics, computer vision, and sensor fusion heavily rely on establishing accurate relationships between sets of measurements or observations, a challenge encapsulated by two related concepts: data association and the correspondence problem. While often used interchangeably, subtle differences in context and scope exist.

The \gls{cp} is fundamentally a geometric and often static challenge. It involves identifying pairs or sets of features, points, or observations that represent the same physical entity in the real world across different views, images, or datasets taken at the same or similar time.
This problem has been primarily faced in stereopsis (\gls{3d} reconstruction from two or more camera views), \gls{sfm}, and image mosaicking. In these cases, the goal is to find which point in one image A maps to which point in another image B.
Usually, it focused on spatial alignment and geometric consistency. The features are typically extracted from different sensors or viewpoints simultaneously.

\gls{da} is a broader and often dynamic problem rooted in state estimation and tracking. It concerns the task of assigning a set of new measurements (\textit{e.g.}, from a sensor sweep) to a set of existing tracks or entities that are being monitored over time.
It is essential in \gls{mtt}, \gls{slam}, and sensor fusion (\textit{e.g.}, radar, lidar, and camera fusion). Its goal is to correctly label a new measurement as belonging to an existing track, a new track, or as clutter (false alarm).
It focuses on temporal consistency and probabilistic modeling. It manages the uncertainty in measurement origins as targets move and sensors provide noisy data across sequential time steps.

The key differences are resumed as follows: \gls{cp} focuses on spatial/geometric alignment while \gls{da} consider also temporal or tracking consistency among frames, the input in \gls{cp} are generally features extracted from different views or contexts while DA has measurement extracted from sequential time steps and the goal is finding the correct geometric match in \gls{cp} while determining the origin or the belonging of the current measurement in \gls{da}. 

In essence, the correspondence problem is often a sub-problem that needs to be solved within a larger data association framework, especially in \gls{slam}, where associating features between two successive frames of a moving camera is a form of correspondence that is critical for the overall tracking process. However, data association extends this to the entire history of tracked objects and the probabilistic management of target identities over time.

\subsection{The data association problem in human re-identification} \label{subsec:reid-prob}

In the field of computer vision and surveillance, Human \gls{reid} is fundamentally cast as a specialized data association problem. Specifically, it is the challenge of establishing identity correspondence between a person's image, the query ($\mathbf{Q}$), and a collection of images from a database, the gallery ($\mathbf{J}$). Both sets are typically captured by non-overlapping camera views ($\mathbf{C}_i \neq \mathbf{C}_j$) at potentially different times ($t_1 \neq t_2$).

The core objective is to determine a binary correspondence function, $f: \mathbf{Q} \times \mathbf{J} \rightarrow \{0, 1\}$, where $f(\mathbf{Q_i}, \mathbf{J_i}) = 1$ if the query image $\mathbf{Q}_i \in \mathbf{Q}$ and the gallery image $\mathbf{J}_i \in \mathbf{J}$ depict the same individual, and $0$ otherwise.

Unlike classical geometric correspondence (\textit{e.g.}, matching a point in a stereo pair), the \gls{reid} data association problem is inherently ill-posed due to the extreme variability introduced by real-world surveillance environments. This variability is often categorized into two main challenges:
intra-class variation (appearance discrepancy of the same person) and inter-class similarity (ambiguity between different people).

In intra-class variation, the same person's appearance can change drastically between camera views, making the correspondence difficult to establish. Key factors include:
\begin{itemize}
    \item \textit{Viewpoint and pose change}: A person's body structure and visible features (\textit{e.g.}, a backpack) change significantly when viewed from a side-angle versus a front-angle. This causes major spatial misalignment of corresponding body parts across images.
    \item \textit{Illumination change and color distortion}: Different lighting conditions (shadows, indoor vs. outdoor, time of day) alter the color and texture of clothing, which are the primary visual cues for \gls{reid}.
    \item \textit{Occlusion and deformations}: Partial or complete obstruction by other people or objects, along with non-rigid human body movement, can hide critical features, reducing the reliable information available for correspondence matching.
    \item \textit{Appearance Changes}: Changes in a person's appearance, such as different clothes or hairstyle.
\end{itemize}

\textbf{inter-class similarity (ambiguity between different people)}
Instead, inter-class similarity means that different individuals can appear visually similar, leading to false correspondences.
\begin{itemize}
    \item \textit{Clothing homogeneity}: In many environments, different people wear identical or near-identical clothing (\textit{e.g.}, black coats, blue jeans), causing a high degree of visual ambiguity that confuses feature-based matching algorithms.
    \item \textit{Background clutter}: Similarities in non-person regions (background or objects carried) can dominate appearance-based feature vectors, leading to incorrect matches between different individuals observed in similar environments.
\end{itemize}

\subsubsection{The problem formulation for Re-ID}

To address this ill-posed correspondence, the \gls{reid} problem is typically reformulated as a metric learning or similarity ranking task. Instead of directly seeking a geometric correspondence, the goal shifts to learning a robust feature representation $\phi(\cdot)$ and a distance metric $D(\cdot, \cdot)$ such that:
\begin{equation}
    D(\phi(\mathbf{q}_i), \phi(\mathbf{j}_j)) \ll D(\phi(\mathbf{q}_i), \phi(\mathbf{j}_k))
\end{equation}
where $\mathbf{q}_i$, $\mathbf{j}_j$, and $\mathbf{j}_k$ are feature vectors representing a person (\textit{e.g.}, embedding vectors extracted with a specialized neural network), and the person \gls{id} associated with $\mathbf{q}_i$ is equal to the one of $\mathbf{j}_j$ and different from the one of $\mathbf{j}_k$.
The solution to the \gls{reid} correspondence problem, therefore, lies in developing a feature embedding and metric that is discriminative for identity while simultaneously being invariant to the environmental and appearance variations outlined above. This transformation from raw pixel space to an identity-invariant feature space is the central challenge addressed by modern \gls{reid} research.

\subsection{The corresponding problem in loop closure detection} \label{subsec:lcd-prob}

In \gls{slam} systems, the process of \gls{lcd} represents a critical instance of the corresponding problem or a simplified data association one, as already stated before. It addresses the challenge of recognizing that the robot or sensor has returned to a previously visited location after a long, potentially drift-accumulating trajectory.

The data association task in \gls{lcd} operates at a global, pose-graph level, focusing on establishing a connection between the current sensor measurement and a historical keyframe in the map.

The actors in this problem are
\begin{itemize}
    \item \textit{The measurement set}: The new measurement ($\mathbf{z}_k$) is the sensory data (\textit{e.g.}, an image, a \gls{lidar} scan) acquired at the current time step $k$.
    \item \textit{The track/map set}: The existing tracks/entities correspond to the set of historical keyframes or map nodes ($\mathbb{M} = \{\mathbf{M}_1, \mathbf{M}_2, \dots, \mathbf{M}_{k-L}\}$) stored by the \gls{slam} system, where $L$ is a large temporal separation.
\end{itemize}

The data association algorithm must decide if the current measurement $\mathbf{z}_k$ corresponds to any of the past map nodes $\mathbf{M}_i \in \mathbb{M}$. If an association is made ($\mathbf{z}_k \leftrightarrow \mathbf{M}_i$), it implies a spatial return to $\mathbf{M}_i$'s location, thereby creating a loop closure constraint.

The data association problem for \gls{lcd} is particularly challenging because the relationship must be established across significant spatial and temporal gaps.
Many difficulties arise in this problem, among which can be found:
\begin{itemize}
    \item \textit{Perceptual aliasing} (false positives): The most severe difficulty is perceptual aliasing, where distinct physical locations appear visually or structurally similar to the sensor (\textit{e.g.}, repeating corridors, symmetrical buildings, or similar foliage). The system must avoid falsely associating the current location with a similar-looking, but incorrect, past location. This failure introduces an erroneous constraint that can corrupt the entire map.
    \item \textit{Appearance change} (false negatives): Over long time spans, the environment's appearance can change significantly (\textit{e.g.}, lighting, weather, seasonal variations, presence of transient objects). These changes can prevent a correct association, leading the system to fail to recognize a true loop closure and thus continue to accumulate localization drift.
    \item \textit{Accumulated uncertainty}: Since the system's pose estimate has drifted significantly over the long trajectory, the expected transformation between the current pose and any past map node is highly uncertain. This makes probabilistic validation of the association much more difficult compared to local data association between sequential frames.
\end{itemize}

In summary, the data association in \gls{lcd} is a high-stakes, long-range identity validation task that must overcome the ambiguity of perceptual aliasing using techniques that are robust to drastic appearance changes. Once the data association is successfully established, the system leverages this constraint to perform global map optimization.

\subsection{Connection between loop closure detection and person re-identification}

The \gls{lcd} and Person \gls{reid} problems, while applied in different domains (\gls{slam} vs. surveillance/tracking), are fundamentally and structurally connected because they are both specialized variations of the \gls{da} problem.

The key connection is that \gls{lcd} can be formally defined as a place \gls{reid} problem.

The most direct connection lies in their shared objective function: finding a match across a significant gap in observation time and/or viewpoint, using only appearance-based features.

Both problems structure the correspondence as a retrieval task. While \gls{lcd} uses the current keyframe in input, \gls{reid} uses the current image. While \gls{lcd} uses a database of historical map nodes, \gls{reid} uses a gallery of person images captured in the past or a model as in the specific case considered. The goal of \gls{lcd} is to find a location correspondence between keyframes, while in \gls{reid} the correspondence is between person identities.

In both cases, the solution relies on learning a powerful feature representation $\phi(\cdot)$ and a distance metric $D(\cdot, \cdot)$ to solve the identity/location correspondence by ranking similarity scores.

Some algorithmic principles are the same. Among the most relevant are:
\begin{itemize}
    \item \textit{Feature embedding}: both \gls{lcd} and \gls{reid} use Deep \glspl{cnn} or geometric metrics to map the input to a compact, low-dimensional embedding vector or representation. The goal is to make this embedding highly discriminative yet robust to environmental variations.
    \item \textit{Metric learning}: Both tasks commonly employ metric comparison to ensure the learned metric space separates positive pairs (same identity/location) from negative pairs (different identity/location) by a large margin.
    \item \textit{Robustness to ill-posed challenges}: Both systems must overcome issues of high intra-class variation (where the appearance of the target, either a person or a place, changes) and Perceptual Aliasing (where different targets look visually similar).
\end{itemize}

\subsubsection{Fundamental differences and cross-pollination}

While the foundations are similar, the inherent constraints and resulting feature focus differ significantly. Place recognition often works with wide static environment structures, with difficulties stemming from dynamic objects and transient lighting changes, or reflections/refractions. In person \gls{reid}, the features are extracted from transient identity attributes, and the challenges arise from the background clutter, different camera viewpoints, and pose variation.

The structural similarity means that breakthroughs in one field often transfer to the other. For example, techniques developed for the highly complex Person \gls{reid} task (\textit{e.g.}, specialized attention mechanisms) can be adopted to create more robust, appearance-based descriptors for \gls{lcd}.
Also, the idea of dividing the image into local parts and establishing local correspondence before a global match (\textit{e.g.}, body parts in \gls{reid}) is sometimes applied in \gls{lcd} to mitigate the effect of dynamic objects, where a matching static image block can still confirm a loop closure.

\section{The localization and mapping problem} \label{sec:mapping-prob}

One core challenge faced by autonomous systems is the \gls{slam} problem. \gls{slam} is defined as the computational problem of incrementally building a map of an unknown environment while simultaneously keeping track of the agent's location within that map. The difficulty lies in the co-dependency: an accurate map is needed for localization, and accurate localization is needed to build a consistent map.

Formally, \gls{slam} aims to estimate the sequence of agent poses $\mathrm{X}_{1:t} = \{\mathbf{x}_1, \dots, \mathbf{x}_t\}$ and the map of environmental features $\mathbf{M}$ given a history of sensor observations $\mathrm{Z}_{1:t} = \{\mathbf{z}_1, \dots, \mathbf{z}_t\}$ and control inputs $\mathrm{U}_{1:t} = \{\mathbf{u}_1, \dots, \mathbf{u}_t\}$. The desired output is the joint posterior probability:

\begin{equation}
    P(\mathrm{X}_{1:t}, \mathbf{M} | \mathrm{Z}_{1:t}, \mathrm{U}_{1:t})
\end{equation}

which is exactly Eq.~\eqref{eq:slam-prob} presented in Section~\ref{sec:slam-works}. In that section, the \gls{slam} problem and its back-end and front-end formulation have been introduced for the sake of completeness in reviewing the current works. Here, an additional and more complete problem presentation is provided.
 
One of the primary goals is often to estimate only the current pose $\mathbf{x}_t$ recursively and the map $\mathbf{M}$ in an online fashion:
\begin{equation} \label{eq:online-slam-prob}
    P(\mathbf{x}_t, \mathbf{M} \mid \mathrm{Z}_{1:t}, \mathrm{U}_{1:t})
\end{equation}

This is typically computed in two steps: the prediction and the update step.
The prediction step is modeled using the posterior probability of state $\mathbf{x}$ at time $t$ and the map $\mathbf{M}$, knowing the control inputs $\mathrm{U}_{1:t}$ and the sensor measurements $\mathrm{Z}_{1:t-1}$: It is predicted from the state posterior probability and the previous joint posterior probability at time $t-1$ following Eq.~\eqref{eq:online-slam-prob}

\begin{equation}
    P(\mathbf{x}_t, \mathbf{M} \mid \mathrm{Z}_{1:t-1}, \mathrm{U}_{1:t}) = \int P(\mathbf{x}_t \mid \mathbf{x}_{t-1}, \mathbf{u}_t) \, P(\mathbf{x}_{t-1}, \mathbf{M} \mid \mathrm{Z}_{1:t-1}, \mathrm{U}_{1:t-1}) \, d\mathbf{x}_{t-1}
\end{equation}

where $P(\mathbf{x}_t \mid \mathbf{x}_{t-1}, \mathbf{u}_t)$ is the motion model used to estimate the robot motion in the last time step, a term often referred to as robot odometry, and $P(\mathbf{x}_{t-1}, \mathbf{M} \mid \mathrm{Z}_{1:t-1}, \mathrm{U}_{1:t-1})$ is the previous joint posterior probability (Eq.~\eqref{eq:online-slam-prob} at time $t-1$).

The prediction is then corrected in the update step using the latest observation $\mathbf{z}_t$:
\begin{equation}
    P(\mathbf{x}_t, \mathbf{M} \mid \mathrm{Z}_{1:t}, \mathrm{U}_{1:t}) = \frac{P(\mathbf{z}_t \mid \mathbf{x}_t, \mathbf{M}) \, P(\mathbf{x}_t, \mathbf{M} \mid \mathrm{Z}_{1:t-1}, \mathrm{U}_{1:t})}{P(\mathbf{z}_t \mid \mathrm{Z}_{1:t-1}, \mathrm{U}_{1:t})}
\end{equation}
 where $P(\mathbf{z}_t \mid \mathbf{x}_t, \mathbf{M})$ is the measurement model.
 
The previous formulation relies on two key assumptions:
\begin{itemize}
    \item \textit{Markov assumption for motion}: The current pose $\mathbf{x}_t$ depends only on the previous pose $\mathbf{x}_{t-1}$ and the current control $\mathbf{u}_t$:
    $$ P(\mathbf{x}_t \mid \mathrm{X}_{1:t-1}, \mathrm{U}_{1:t}, \mathbf{M}) = P(\mathbf{x}_t \mid \mathbf{x}_{t-1}, \mathbf{u}_t) $$
    \item \textit{Observation independence}: The current observation $\mathbf{z}_t$ is conditionally independent of all previous poses and observations given the current pose $\mathbf{x}_t$ and the map $\mathbf{M}$:
    $$ P(\mathbf{z}_t \mid \mathrm{X}_{1:t}, \mathrm{Z}_{1:t-1}, \mathrm{U}_{1:t}, \mathbf{M}) = P(\mathbf{z}_t \mid \mathbf{x}_t, \mathbf{M}) $$
\end{itemize}

The \gls{slam} problem can be factored using the previous assumption into a joint probability of all states, the map, the observations, and controls as: 

\begin{equation} \label{eq:joint_prob}
    P(\mathrm{X}_{1:t}, \mathbf{M}, \mathrm{Z}_{1:t}, \mathrm{U}_{1:t}) = P(\mathbf{x}_0) P(\mathbf{M}) \prod_{k=1}^t P(\mathbf{x}_k \mid \mathbf{x}_{k-1}, \mathbf{u}_k) \prod_{k=1}^t P(\mathbf{z}_k \mid \mathbf{x}_k, \mathbf{M})
\end{equation}

where $P(\mathbf{x}_0)$ is the initial pose distribution and $P(\mathbf{M})$ is the prior over the map, while the first product represents the motion model and the second the observation one. Modern \gls{slam} methods often focus on finding the most probable estimate as the \gls{map}, which maximizes the term $P(\mathrm{X}_{1:t}, \mathbf{M} \mid \mathrm{Z}_{1:t}, \mathrm{U}_{1:t})$, which is equivalent to maximizing the full joint distribution in Eq.~\eqref{eq:joint_prob} assuming a uniform denominator. Nowadays, the above problem is practically formulated using factor graphs\,\cite{dellaert2021factor}. A deep review of the \gls{slam} problem formulated using factor graphs can be found in the \gls{slam} Handbook\,\cite{slam_handbook}.

As already introduced, the \gls{slam} problem is often divided into back-end and front-end. The back-end solves the optimization problem formulated in the factor graph following the assumptions and joint probabilities presented above.
Instead, the front-end focuses on how the factor graph is created. In the following, the focus is on the front-end problem formulation, which is one of the main problems faced in this thesis.

In the following, the evolution of \gls{slam} solutions is distinguished by the type of information used for environmental representation and data association: Geometric (relying on raw spatial features) and Semantic (incorporating cognitive-level information).

\subsection{Geometric SLAM: foundational techniques} \label{subsec:geometric-mapping-prob}

Classical \gls{slam} focuses purely on geometric consistency and spatial fidelity within a local coordinate system.
The environment is modeled using low-level sensor data (\textit{e.g.}, points from \gls{lidar}) or abstract spatial primitives derived directly from them:
\begin{itemize}
        \item \textit{Sparse point features}: These are highly repeatable, unique corner points or keypoints (\textit{e.g.}, \gls{sift}, \gls{orb} features) used in \gls{vslam} for fast processing and robust tracking.
        \item \textit{Dense representations}: Voxel grids or \glspl{sdf} are employed to model the volume of the scene, typically in \gls{rgbd} or dense \gls{lidar} \gls{slam}, providing a high-fidelity \gls{3d} map.
        \item \textit{Structural primitives}: Line segments, planes, and cylinders, are utilized, particularly in \gls{lidar} and \gls{2d}-\gls{slam}, to capture the dominant structure of indoor environments.
\end{itemize}

The environment map $\mathbf{M}$ is typically represented as a factor graph, where nodes are keyframes (poses) and edges are the relative geometric transformations between them. Factor graphs are used to model and solve the probability in Eq.~\ref{eq:online-slam-prob}.

The previously presented prediction step, referred to here as the odometry step, suffers from drift accumulation. The primary weakness is that pose estimates from odometry are prone to accumulating small errors over time, causing the estimated trajectory to drift away from the true path.
To correct drift, the \gls{lcd} module is activated. \gls{lcd} is a geometric data association problem that seeks to identify if the current observation $\mathbf{z}_t$ matches a previously mapped location $\mathbf{M}_i$. Upon a successful match, a new, highly accurate constraint is added to the factor graph, which is then globally optimized to distribute the accumulated error across the entire map.

Nevertheless, geometric features are fragile. They are highly sensitive to viewpoint change, illumination variation, and the presence of dynamic objects, leading to frequent false negative loop closures (missed opportunities) or, critically, false positive loop closures (incorrect associations that corrupt the map).

\subsection{Semantic SLAM: cognitive integration} \label{subsec:semantic-mapping-prob}

Semantic \gls{slam} augments the geometric framework by integrating cognitive-level understanding of the environment, treating scenes not just as points and planes, but as collections of meaningful objects.

The \gls{slam} problem is enriched with more meaningful semantic features extracted via deep neural networks, including:
\begin{itemize}
        \item \textit{Object instances}: Bounding boxes and instance masks for specific objects (\textit{e.g.}, "chair 1", "table 2").
        \item \textit{Semantic labels}: Per-pixel classification (semantic segmentation) for surface types (\textit{e.g.}, "wall", "floor", "ceiling").
        \item \textit{Place categories}: Classification of the scene (\textit{e.g.}, "living room", "street intersection").
\end{itemize}

The map evolves into a semantic map that can be associated with a semantic factor graph. This graph retains the geometric structure but enriches it with semantic relationships and labels, enabling querying by object identity.

The introduction of semantics inherently enhances the robustness of data association by leveraging semantic invariance.
A chair remains a chair regardless of a shift in lighting or a change in viewing angle. By associating keyframes based on the consistent presence and geometric configuration of invariant semantic objects, the system achieves a more robust form of place recognition (or place \gls{reid}), mitigating the risks of perceptual aliasing that plague geometric methods.
Semantic relationships provide powerful constraints for global optimization. For instance, knowing that a "ceiling" is always parallel to a "floor" or that a "lamp" is typically above a "table" allows the system to enforce structural consistency, leading to a more accurate and globally plausible map than is possible with geometric errors alone.

The principal challenge of Semantic \gls{slam} lies in effectively fusing two distinct modalities:
\begin{itemize}
    \item \textit{Low-level geometry}: Precise, but fragile and abstract.
    \item \textit{High-level semantics}: Robust and meaningful, but inherently coarse and subject to object detection errors.
\end{itemize}
The problem transitions from a purely mathematical optimization problem to a challenge of robust fusion and data-driven reasoning to maximize the consistency and utility of the final map.

\chapter{Method} \label{chap:method}
    
Due to the multiplicity of methods presented, the proposed method is structured into three main sections: $(i)$ \gls{reid} for \Gls{hrc} in Section~\ref{sec:re-id-method}, which addresses the visual data association and tracking problem, $(ii)$ environmental perception through geometrical matching in Section~\ref{sec:mapping-method}, which tackles the \gls{slam} problem and the corresponding issue of loop closure, and $(iii)$ robot contextual understanding and interaction with semantics in Section~\ref{sec:semantic-method}, where a solution to enhance the robot's cognitive capabilities is introduced.

\section{Human-robot collaboration through re-Identification} \label{sec:re-id-method}

For successful interaction, robots must be able to detect and recognize their human collaborators. To emulate human abilities, visual \gls{reid} is employed to address this challenge, as mobile robots are nearly always equipped with camera sensors. Incremental solutions are presented that offer several improvements over the current \gls{sota} and address the limitations encountered in prior iterations.

\subsection{Baseline approach for visual Re-ID} \label{subsec:followme-method}

\begin{figure}
    \centering
    \includegraphics[width=1\linewidth]{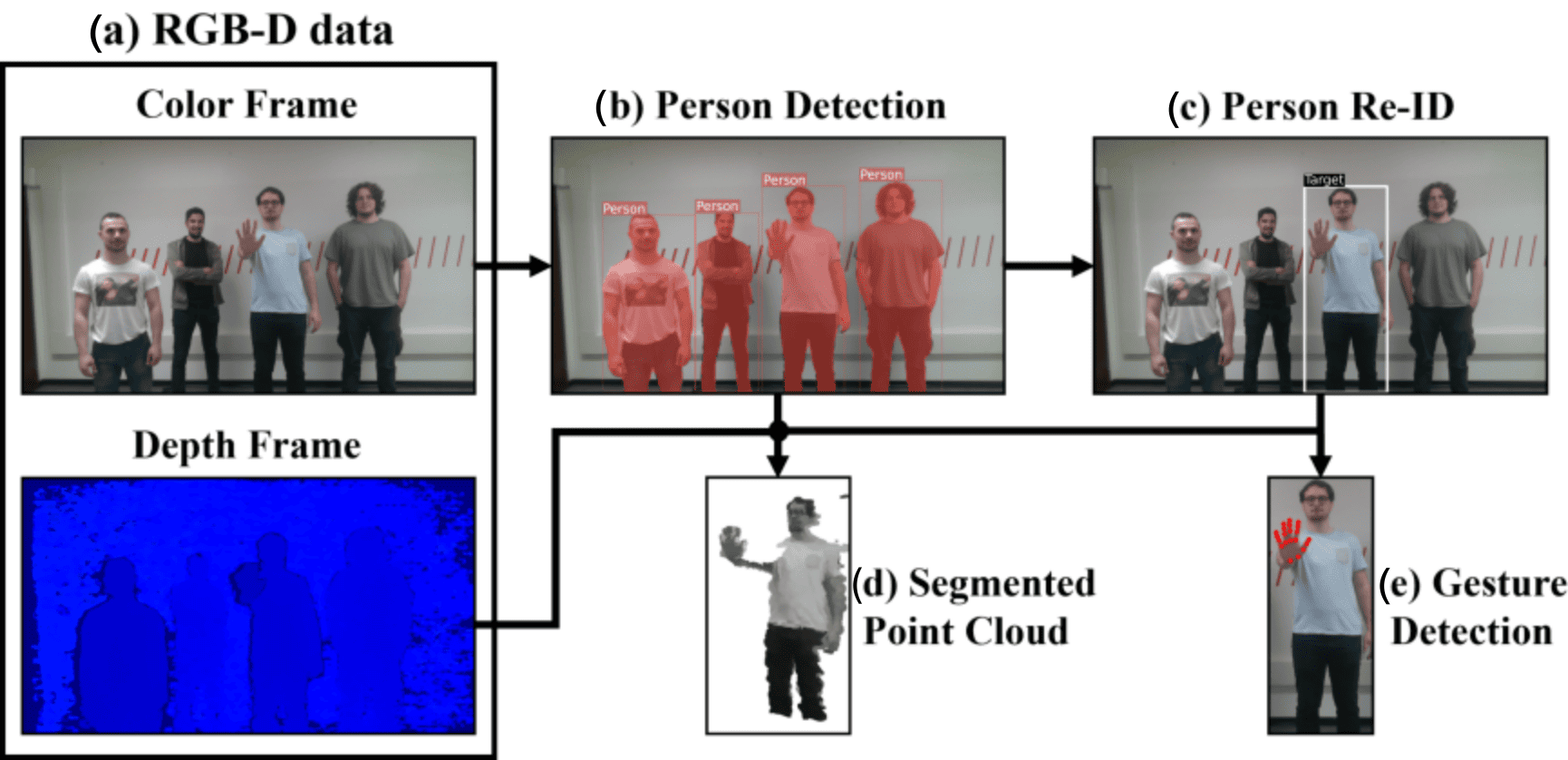}
    \caption{FollowMe perception pipeline: (a)  \gls{rgb} and Depth images acquisition; (b) person detection through Yolact++; (c) person \gls{reid} using a deep neural network and human features distance; (d) person localization using the point cloud and re-identified person mask; (e) gesture detection to send commands to the robot. Reprinted, with permission, from\,\cite{rollo2023followme}, \textsuperscript{\textcopyright} 2023 \gls{ieee}.}
    \label{fig:followme_pipeline}
\end{figure}

The first proposed approach, specifically FollowMe\,\cite{rollo2023followme}, utilizes visual and depth data obtained from an \gls{rgbd} camera to execute four consecutive steps: $(i)$ Detection, $(ii)$ \gls{reid}, $(iii)$ Localization and filtering, and $(iv)$ Gesture detection. The complete pipeline initially proposed for solving this problem is presented in Fig.~\ref{fig:followme_pipeline}.

\subsubsection{People detection} \label{subsec:follow_person_detection}

To detect individuals in the image (Fig.~\ref{fig:followme_pipeline}a), an instance segmentation algorithm is employed. Instance segmentation is a specific form of image segmentation that involves detecting object instances and delineating their boundaries. YOLACT++\,\cite{bolya2020yolact++} is a one-stage instance segmentation framework that offers several advantages over existing architectures, with prediction speed being one of the most significant. Although new solutions have been presented in the literature, YOLACT++ is among the existing frameworks capable of delivering "real-time" instance segmentation inference. In this work, a standard model pre-trained on the 80 classes of COCO\,\cite{lin2014microsoft} (which includes the ``person" class) is adopted (Fig.~\ref{fig:followme_pipeline}b). YOLACT++ extracts both bounding boxes and segmentation masks corresponding to the trained classes, where the mask values are $1$ for the pixels where the object is localized and $0$ elsewhere.
To precisely isolate the person's contours and completely remove background information, the original images are element-wise multiplied by their corresponding binary masks. This operation allows the \gls{reid} module to extract cleaner personal features based solely on the individual by eliminating background noise. The resulting images, each representing a detected person, are subsequently made available to the \gls{reid} module.

\subsubsection{Person re-Identification} \label{subsec:follow_reident}

To correctly re-identify an individual, an additional neural network is employed to extract distinctive features. This step is necessary because features derived from YOLACT++\,\cite{bolya2020yolact++} and \gls{yolo}\,\cite{redmon2016you} networks in general are highly generalized within the "person" class, meaning that different individuals possess very similar features in the \gls{yolo} embedding space.

Person \gls{reid} was originally developed for identifying a subject across non-overlapping camera views. Given a query image of a person, the objective is to recognize the same individual in images acquired by different cameras by analyzing and extracting appearance information alone (without relying on biometric cues). To mitigate the limitations of the \gls{yolo} embedding space, the popular \gls{mmt}\,\cite{ge2020mutual} framework is utilized for extracting features for the \gls{reid} module. Specifically, only the pre-training phase is considered. In this phase, a deep learning network is trained to recognize people in images. The segmented output images are passed into this network to extract the features that will subsequently be used to re-identify the target person. The features from the last layer of the trained network are specifically considered. The \gls{reid} process is divided into two distinct phases: calibration and identification.

The calibration phase is executed at the start of the application. The target person is instructed to move at various distances in front of the camera for a few seconds, simulating the movements and postures they will typically adopt while walking, in order to collect images. This calibration step is critical because the diversity in person representations captured during this stage directly influences the \gls{reid} system's robustness against illumination changes and occlusions. The collected images are divided into two groups: the person calibration set (with $\mathrm{N}_{c}$ elements) and the threshold set (with $\mathrm{N}_\lambda$ elements). The processed images from the person calibration set are input first into YOLACT++ for person masking and then into the \gls{reid} model for feature extraction.

Let $D$ denote the dimension of the feature vector $\mathbf{x}$ extracted by the \gls{reid} model; each person can be associated with a vector $\mathbf{x} = [x_1,\dots,x_D]^T$. For each image in the person calibration set, the corresponding features are extracted and then used to compute the associated mean $\boldsymbol{\mu}=[\mu_1,\dots,\mu_D]^T$ and standard deviation $\boldsymbol{\sigma}=[\sigma_1,\dots,\sigma_D]^T$, which are required for target identification in the second phase.

Thus, given a new image of a person, the corresponding feature vector $\mathbf{x'}$ is computed. This person can be identified as the target by measuring the following feature-space weighted distance between $\mathbf{x'}$ and the distribution defined by $\boldsymbol{\mu}$ and $\boldsymbol{\sigma}$:

\begin{equation}\label{eq:feat_dist}
    d_{\boldsymbol{\mu},\boldsymbol{\sigma}}(\mathbf{x'}) = \sqrt{\frac{1}{D}\sum_{i=1}^D \left(\frac{x'_i-\mu_i}{\sigma_i}\right)^2}\text{,}
\end{equation}
where the subscript $i$ represents the $i$-th index of the corresponding vector. 

The previously acquired threshold set is utilized to compute a distance threshold $\lambda_d$: if the measured distance $d_{\boldsymbol{\mu},\boldsymbol{\sigma}}(\mathbf{x'}) \leq \lambda_d$, the person associated with feature $\mathbf{x'}$ will be identified as the target observed during the calibration phase.

\begin{figure}
     \centering
     \includegraphics[width=\linewidth]{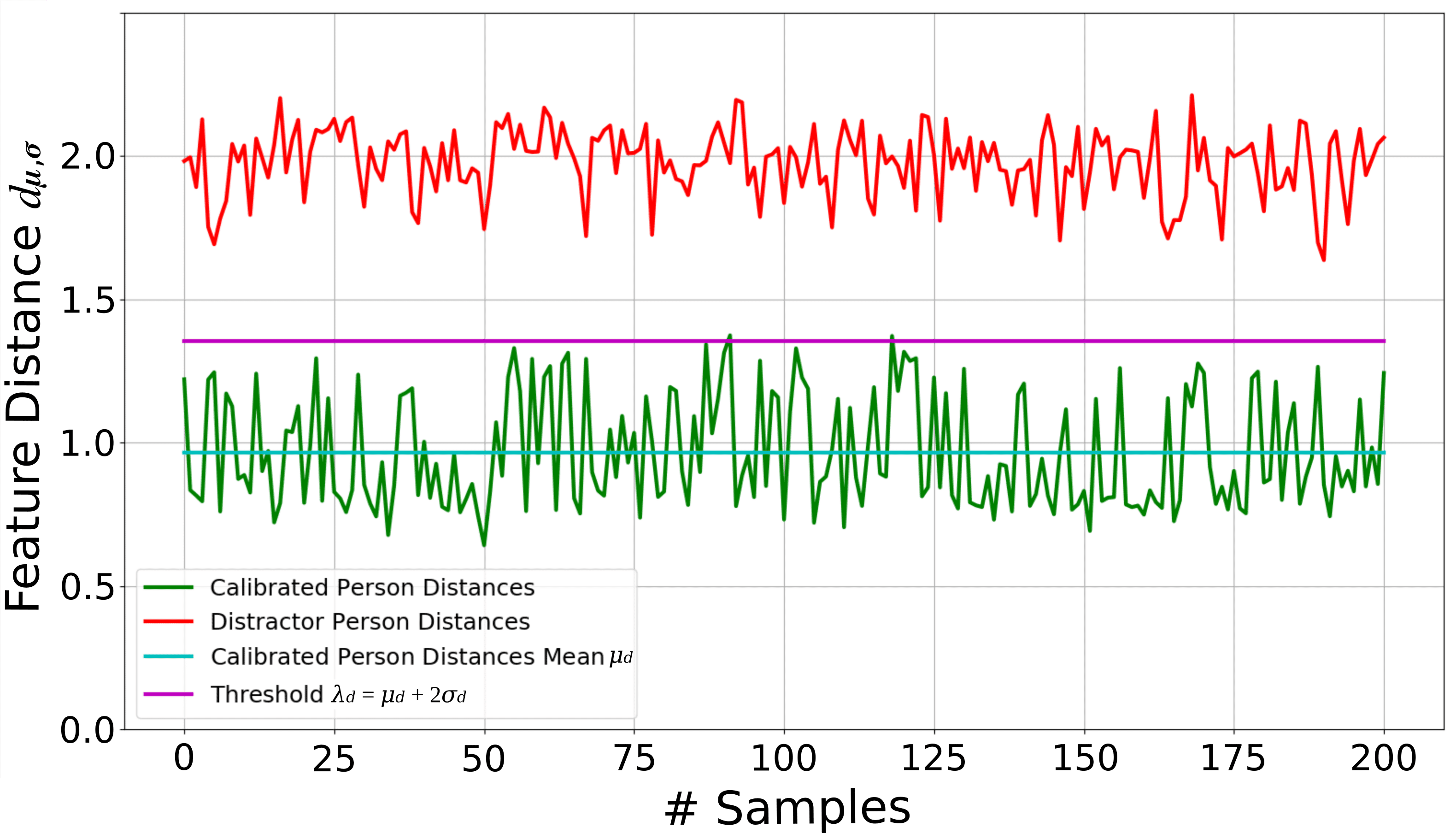}
     \caption{In green, the plot of the feature distances for a calibrated person used to compute the target threshold (magenta horizontal line). In red, the distances computed between the features of another person (the distractor) and the target template. The distances are computed over 200 sample images. Reprinted, with permission, from\,\cite{rollo2023followme}, \textsuperscript{\textcopyright} 2023 \gls{ieee}.}
     \label{fig:followme_threshold}
\end{figure}

Specifically, for each image in the threshold set, the associated feature distance, calculated using Eq.~\eqref{eq:feat_dist}, is computed, yielding the overall mean distance $\mu_d$ and standard deviation $\sigma_d$. The calibrated person threshold is then set as $\lambda_d=\mu_d\ +\ 2\sigma_d$. This setting is chosen to encompass the majority of target samples without being excessively high, which would risk failing to filter out distractor people (any person who is not the calibrated target). Fig.~\ref{fig:followme_threshold} illustrates an example set of distances for a calibrated person and a distractor, along with the calculated $\lambda_d$ value.
In this work, we prioritize minimizing the number of wrongly re-identified targets (false positives) because such errors are difficult to mitigate. Conversely, false negatives (the calibrated person is present but not re-identified) are managed. If the target is not recognized in certain frames, a \gls{kf} integrates the last known detection for a specified duration, thereby enhancing the application's robustness.
 
During the execution of the FollowMe application, new images are continuously acquired, and people are detected and filtered. In the identification step, similarly to the calibration phase, the background is removed from the processed images, which are then passed to the \gls{reid} model. The feature vectors obtained from the \gls{reid} inference for all detected people are collected. Finally, the feature distances (Eq.~\eqref{eq:feat_dist}) are calculated. The person exhibiting the smallest distance, provided it is less than the selected threshold $\lambda_d$, is chosen as the target (Fig.~\ref{fig:followme_pipeline}c). If the distances for all detected people exceed the threshold, no target is detected.

\subsubsection{Localization and filtering} \label{subsec:follow_localization}

In the localization and filtering step, the mask determined during the \gls{reid} process is used to compute the target's position, ${^\mathrm{c}}\mathbf{p}$, within the camera reference frame. Utilizing both \gls{rgb} and depth data, and leveraging the re-identified person's mask, the segmented target point cloud is constructed (Fig.~\ref{fig:followme_pipeline}d). The position ${^\mathrm{c}}\mathbf{p}$ is then calculated as the \gls{3d} centroid of this point cloud.

\begin{figure}
    \centering
    \includegraphics[width=\linewidth]{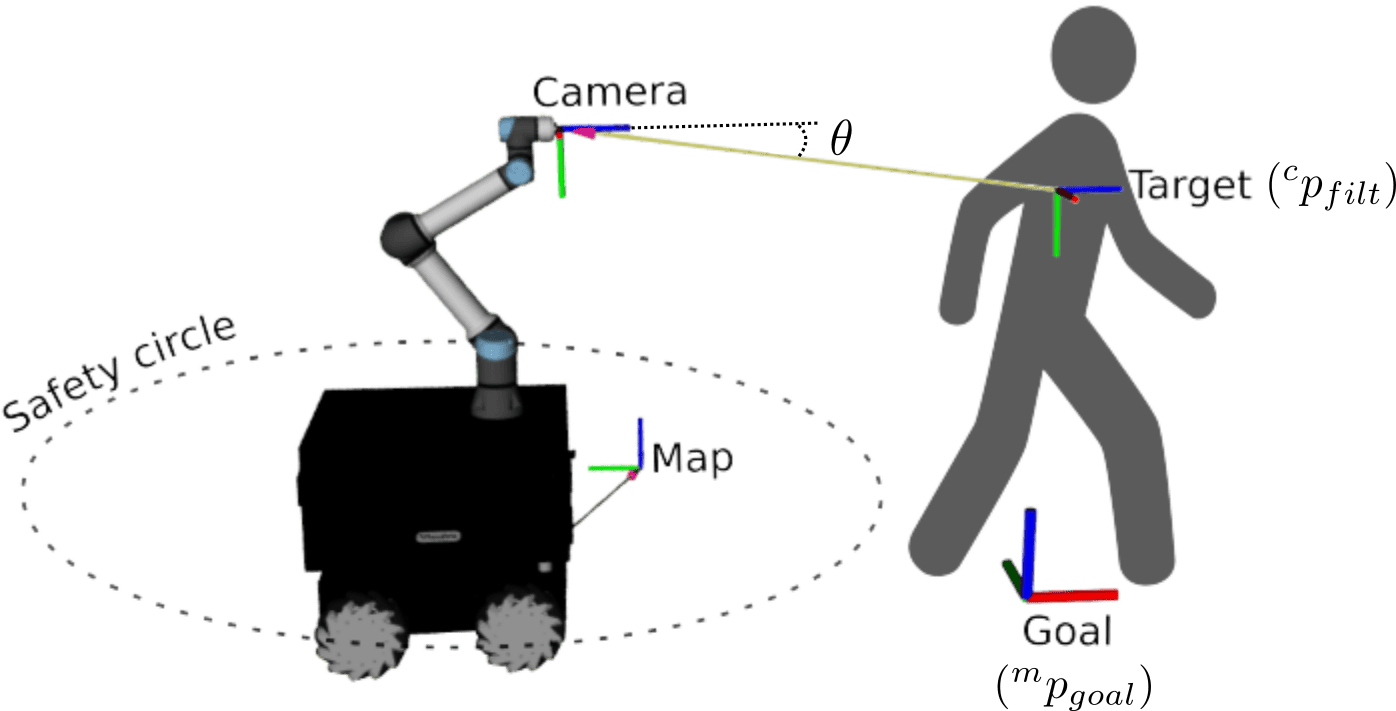}
    \caption{Overview of relevant transformation frames and representation of the safety circle used during the FollowMe application. Reprinted, with permission, from\,\cite{rollo2023followme}, \textsuperscript{\textcopyright} 2023 \gls{ieee}.}
    \label{fig:followme_setup}
\end{figure}

To filter out noise caused by the robot's movements, vibrations, and inherent mask detection errors, a \gls{kf} is applied to ${^\mathrm{c}}\mathbf{p}$. The \gls{kf} stops the integration when measurement updates are not received for a predefined duration, known as the expiration time, $t_{exp}$. Since the \gls{kf}-filtered position, ${^\mathrm{c}}\mathbf{p}_{filt}$, is defined in the moving camera frame, the following homogeneous transformation is applied:
\begin{equation} \label{eq:transformation}
    {^\mathrm{m}}\mathbf{p}_{filt} = {^\mathrm{m}}\mathbf{t}_c{^\mathrm{c}}\mathbf{p}_{filt}
\end{equation}
where ${^\mathrm{m}}\mathbf{p}_{filt}$ is the target position expressed in the map reference frame, and $^\mathrm{m}\mathbf{t}_\mathrm{c}$ is the translation from the camera reference frame $\mathrm{c}$ to the map frame $\mathrm{m}$, which is obtained from the navigation stack localization module.

Finally, the \gls{2d} goal position $^\mathrm{m}\mathbf{p}_{goal}$ is computed by projecting $^\mathrm{m}\mathbf{p}_{filt}$ onto the \gls{2d} map plane (XY plane). Concurrently, the heading angle $\theta$ between the camera reference frame and $^\mathrm{m}\mathbf{p}_{filt}$ is calculated. The goal data (i.e., $^\mathrm{m}\mathbf{p}_{goal}$ and $\theta$) are managed by the \gls{fsm}, which forwards them to the navigation module to drive the robot towards the target. An overview of the relevant transformation frames is presented in Fig.~\ref{fig:followme_setup}.

\subsubsection{Gesture detection} \label{subsec:follow_gesture}

To enable interaction between the robot and the human operator, hand gesture recognition capabilities have been developed for the robot. Hand gestures are a commonly used communication tool for exchanging non-verbal information, not just in robotics\,\cite{kaur2016review}. By utilizing gestures, it is possible to interact with the robot and convey commands that trigger specific robot functionalities and behaviors. In this work, gestures are used to stop and reactivate the FollowMe task, but other commands can be incorporated if necessary.

The gesture detection module was implemented using the Mediapipe hand tracking algorithm\,\cite{zhang2020mediapipe} combined with a \gls{svm} classifier. Mediapipe is a real-time framework that extracts 21 ordered hand key points in $2.5$ dimensions from a standard \gls{rgb} image, independent of the hand scale. These points provide the $x$, $y$, and $z$ positions relative to the hand's palm center. Mediapipe performs inference on the 21 key points with high prediction quality and real-time speed (Fig.~\ref{fig:followme_pipeline}e).

To train the \gls{svm} classifier to distinguish between gestures, the $21$ hand landmarks were flattened into a single feature vector of size $63 = 21 \times 3$.

\begin{figure}
    \centering
    \includegraphics[width=\linewidth]{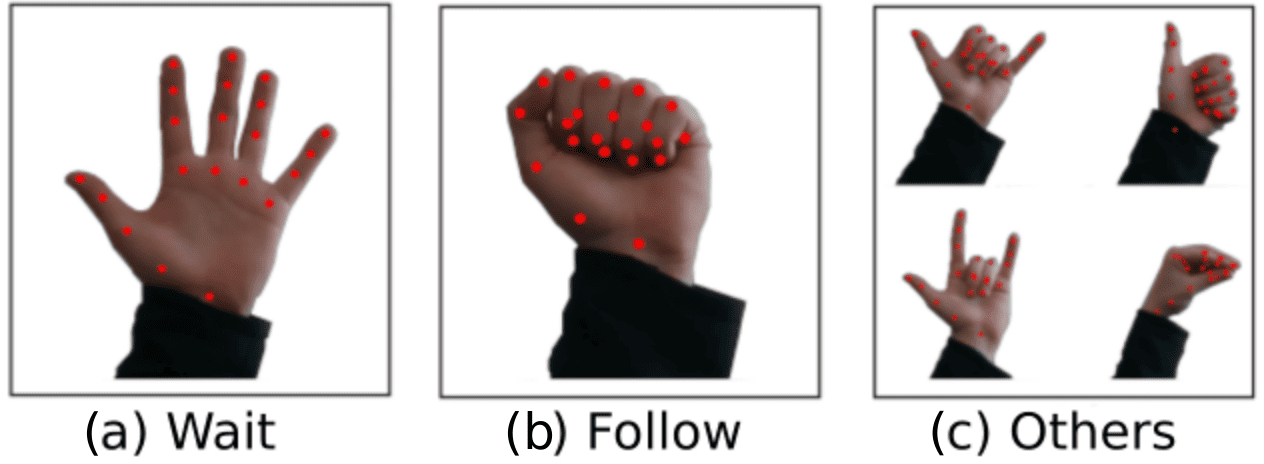}
    \caption{Classes considered to train the \gls{svm} for hand gesture detection. Mediapipe framework\,\cite{zhang2020mediapipe} is used to extract \gls{3d} key points (red points) relative to the centre of the hand. Reprinted, with permission, from\,\cite{rollo2023followme}, \textsuperscript{\textcopyright} 2023 \gls{ieee}.}
    \label{fig:followme_hand}
\end{figure}

The classes representing the gestures are three: \textit{(i)} an open hand for the \textit{wait} command (Fig.~\ref{fig:followme_hand}a), \textit{(ii)} a closed hand for the \textit{follow} command (Fig.~\ref{fig:followme_hand}b), and \textit{(iii)} all other configurations which are ignored (Fig.~\ref{fig:followme_hand}c).

A radial basis function kernel and a one-versus-one decision function were employed to train the multi-class \gls{svm} classifier. The classification output is filtered (\textit{i.e.}, $\xi$ equal consecutive classification outputs are required to send a command; the default choice is $\xi=5$) and sent to the \gls{fsm}, which processes it to command the robot's actions.

\subsubsection{Navigation} \label{subsec:follow_navigation}

The environments where humans and collaborative robots operate are typically noisy and populated. The robot must navigate freely while minimizing the probability of collision with obstacles to prevent damage. To accomplish this task, the \gls{ros} navigation stack\footnote{\gls{ros} Navigation Stack: \url{http://wiki.ros.org/navigation}} is utilized. Among the well-known advantages of using this navigation stack (\textit{e.g.}, customization, structured organization), there is its generalization into a standard navigation architecture, which allows for simple re-adaptability across different robotic platforms.

For safety reasons, a circular safety region is established around the robot (see Fig.~\ref{fig:followme_setup}). If the estimated position of the target falls within this safety circle, signifying a dangerous area, the robot immediately deletes the last navigation goal and stops. Non-target people are treated as obstacles and are avoided if possible; otherwise, the robot stops and attempts to find an alternative path to the goal. In the worst-case scenario, where the target is not re-identified, and the obstacle avoidance module fails to recognize the person as an obstacle (a rare occurrence), the robot still has time to stop, owing to the \gls{kf} expiration time, thereby nullifying the chances of collision.

The safety distance $d_{safe}$ can be computed as follows:
\begin{equation} \label{eq:d_safe}
    d_{safe} = 1.4\ v_{max}\ t_{exp}
\end{equation}
where $v_{max}$ is the maximum robot velocity and $t_{exp}$ is the \gls{kf} expiration time. In this way, the robot is designed to stop after traversing a distance of $v_{max} \times t_{exp}$ meters, leaving an additional $40\%$ margin of the required stopping distance from the person to ensure safety.

\begin{figure}[t]
    \centering
    \includegraphics[width=\linewidth]{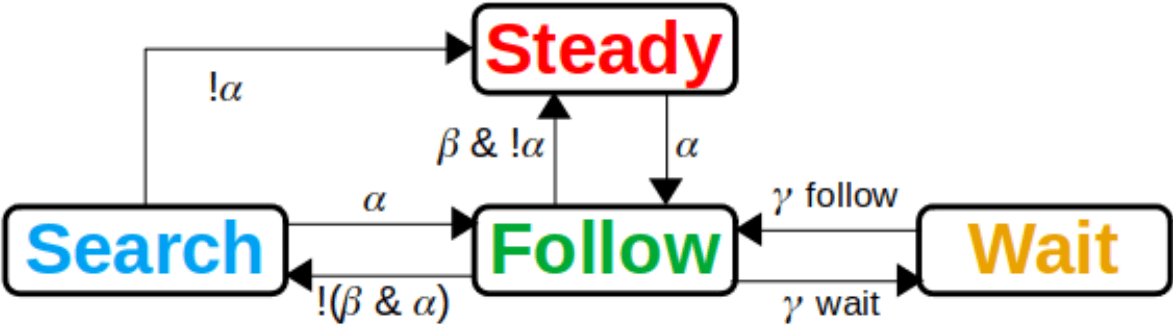}
    \caption{FollowMe \gls{fsm} diagram. The transitions between the states are highlighted with Greek letters and boolean operators \textit{and} (\&), \textit{not} (!). The transitions' meaning is: $(\alpha)$ the target is re-identified in the camera \gls{fov}, $(\beta)$ the last target position is inside the safety circle, and $(\gamma)$ the robot receives a command by the hand gestures module. Reprinted, with permission, from\,\cite{rollo2023followme}, \textsuperscript{\textcopyright} 2023 \gls{ieee}.}
    \label{fig:statemachine}
\end{figure}

\subsubsection{Decision making} \label{subsec:follow_decision}

The application workflow is managed using a \gls{fsm}, which collects all data from the perception and navigation modules and controls the robot's overall behavior.

The \gls{fsm} has four defined states:
\begin{itemize}
    \item \textit{Steady}: The robot is stopped and waits for the perception module to issue a \textit{follow} command.
    \item \textit{Follow}: The robot actively tracks and follows the target person.
    \item \textit{Search}: When the robot stops receiving target position updates, it initiates a search by rotating around its current position (for a maximum of one complete turn) toward the direction of the last known position.
    \item \textit{Wait}: The robot was paused by the target person using the \textit{wait} gesture command and remains stopped until the \textit{follow} gesture command is received.
\end{itemize}

The \gls{fsm}'s structure is visualized in Fig.~\ref{fig:statemachine}. The events that trigger the transitions (represented by the edges) can be summarized as follows: $(\alpha)$ the target is re-identified within the camera's \gls{fov}, $(\beta)$ the last target position is inside the safety circle, and $(\gamma)$ the robot receives a command from the hand gestures module.

\subsection{Introducing autonomous adaptation for smooth interaction in challenging situations} \label{subsec:carpe-method}

\begin{figure}
    \centering
    \includegraphics[trim={3cm 5cm 12.5cm 3.5cm},clip,width=\linewidth]{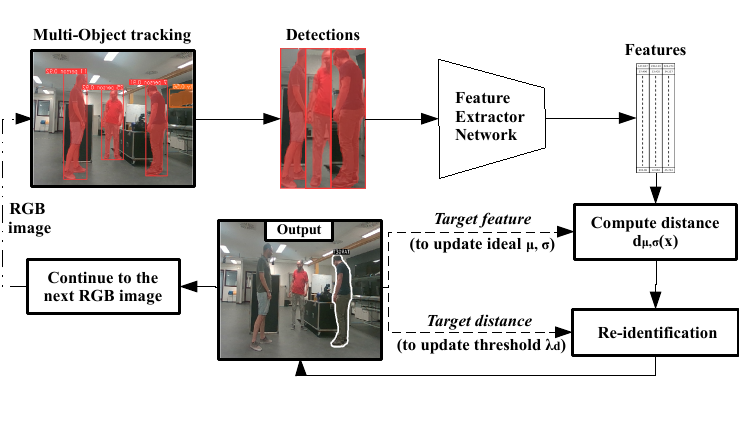}
    \caption{The pipeline framework begins with an image input and ends with a re-identified target output. The first module is the \gls{mot}, where a neural network provides a rough tracking of objects present in the image. 
    From the \gls{mot} module, the detections obtained are passed into a feature extractor, which generates the output feature vectors $\boldsymbol{x}$. The feature vectors $\boldsymbol{x}$ allow to compute the statistical distance $d_{\boldsymbol{\mu},\boldsymbol{\sigma}}(\mathbf{x})$ (as shown in Eq.~\eqref{eq:feat_dist}), which the \gls{reid} module will use. 
    If the target is successfully re-identified, the target statistical distance $d_{\boldsymbol{\mu},\boldsymbol{\sigma}}(\mathbf{x})$ and the target features $\boldsymbol{x}$ are used to update the re-identifier threshold $\lambda_d$ and the ideal target representation $\boldsymbol{\mu}$ and $\boldsymbol{\sigma}$ (represented by dashed lines). Reprinted, with permission, from\,\cite{rollo2024carpe}, \textsuperscript{\textcopyright} 2024 \gls{ieee}.
    }
    \label{fig:carpe_pipeline}
\end{figure}

Although the \gls{reid} framework presented in Section~\ref{subsec:followme-method} is sufficient for detecting and tracking a person in an image to perform a collaborative task, it requires initial calibration based on the current person's appearance. This presents a limitation when an individual is prone to changing their appearance frequently throughout the day or, certainly, over the course of a week. To overcome this issue, a new approach for tracking people, presented in Fig.~\ref{fig:carpe_pipeline}, is proposed. A pseudo-code implementation is also provided in Algorithm~\ref{alg:carpe-id}. The tracking process begins when a user selects an \gls{id}. Initially, the detected people's appearances, obtained using the \gls{mot} system, are collected in a database. This collection allows the user to choose the specific person they wish to follow.

Once the \gls{id} is selected, the \gls{rgb} image is fed into the \gls{mot} algorithm to obtain a preliminary detection and tracking, which assigns a unique \glspl{id} to each person. These detections are used to crop the people images, which are then input into a deep neural network trained for person recognition (more details are available in Section~\ref{subsec:carpe-eval}). This network extracts features, represented as a one-column vector, which are subsequently used for \gls{reid} and tracking of the target.

For each frame, the application executes two sequential steps: \gls{reid} of the target and updating the target's ideal representation.

\subsubsection{Re-identification} \label{subsec:reid}

In the \gls{reid} step, the same statistical distance presented in Eq.~\eqref{eq:feat_dist} used in FollowMe is computed from the features $\mathbf{x}$ extracted for the detected people.

As a reminder and to better specify the components of Eq.~\eqref{eq:feat_dist}, $i$ denotes the $i$-th index of a vector. The values of mean and standard deviation, denoted by $\boldsymbol{\mu}$ and $\boldsymbol{\sigma}$ respectively, represent the model of the ideal target to track. The mean $\boldsymbol{\mu}$ is initialized with the initial feature vector $\mathbf{x}$, while the standard deviation $\boldsymbol{\sigma}$ is initialized with a zero vector; both have dimension $D$.

In the next step, it is checked whether the tracking algorithm provides a detection with the same \gls{id} as the one selected by the user. If such a detection exists, it is output directly with its \gls{id}. If not, the algorithm proceeds to the \gls{reid} module. Here, the module compares the adaptive threshold $\lambda_d$ with the feature vector that has the smallest distance from the target model. If the module is unable to re-identify any person among the detected ones, the framework continues by analyzing the next \gls{rgb} frame. This step is similar to the one presented in Section~\ref{subsec:followme-method}, but includes a few improvements.

\begin{figure*}[ht]
    \centering
    \begin{subfigure}[b]{0.49\textwidth}
        \centering
        \includegraphics[width=\linewidth]{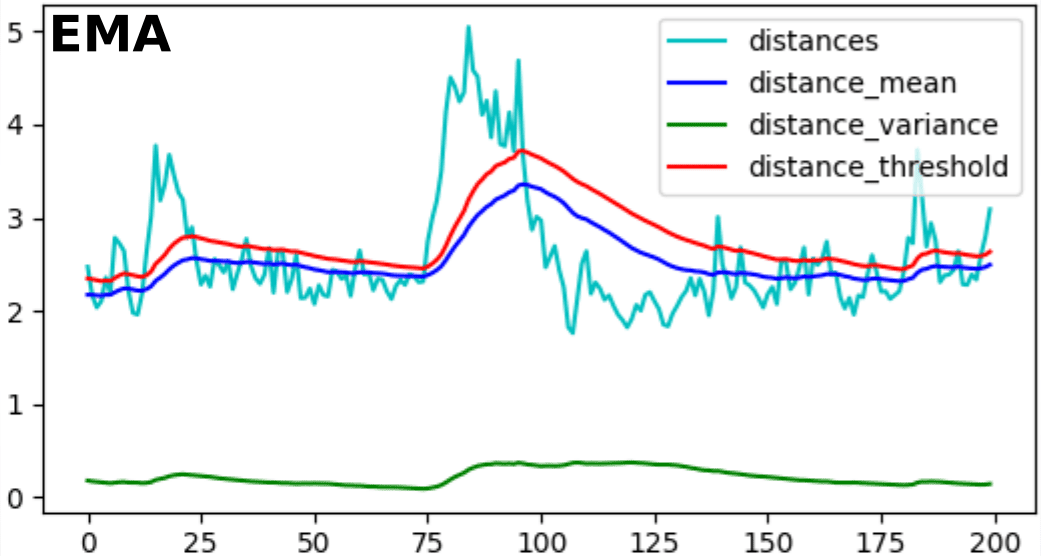}
        \caption{\gls{ema} function effects.}
        \label{fig:no_damp}
    \end{subfigure}
    \begin{subfigure}[b]{0.49\textwidth}
        \centering
        \includegraphics[width=\linewidth]{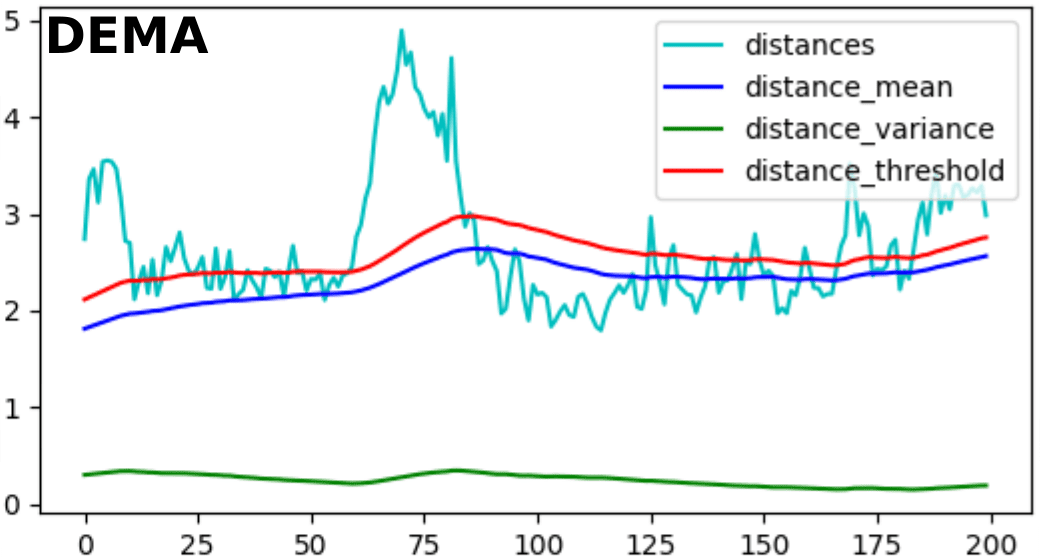}
        \caption{\gls{dema} function effects.}
        \label{fig:damp}
    \end{subfigure}
    \caption{This figure illustrates a comparison between \gls{dema} and \gls{ema} effects on threshold filtering and the ideal target representation. On the left, they follow the light blue line (\textit{i.e.}, the statistical distance) with a small delay. On the right, the damping guarantees that their values do not track the distance behaviour when peaks are present. Instead, they slow down, waiting for the adaptation of the target model to the newly acquired appearances. Reprinted, with permission, from\,\cite{rollo2024carpe}, \textsuperscript{\textcopyright} 2024 \gls{ieee}.}
    \label{fig:ema_comparison}
\end{figure*}

\subsubsection{Model update} \label{subsec:mod_update}

The primary contribution compared to the previous work lies in the model update step presented here. After the target is re-identified, its feature vector and corresponding statistical distance are used to continuously adjust the ideal target representation ($\boldsymbol{\mu}$ and $\boldsymbol{\sigma}$), as well as the threshold $\lambda_d$ used during the \gls{reid} process, according to the following equation:
\begin{equation}
	\lambda_d = \mu_d + 2\ \sigma_d
\end{equation}
where $\mu_d$ and $\sigma_d$ are the mean and the standard deviation of the target's statistical distances.

The variables $\boldsymbol{\mu}$, $\boldsymbol{\sigma}$, $\mu_d$, and $\sigma_d$ are updated online using a Damped version of the \gls{ema} ($\boldsymbol{\chi}^{dema}$), which is referred to as \gls{dema}. The formula for \gls{dema} is expressed as:
\begin{equation} \label{eq:dema}
	\boldsymbol{\chi}^{dema}_t =  \alpha_{damp}\ \boldsymbol{\psi} + (1 - \alpha_{damp})\ \boldsymbol{\chi}^{dema}_{t-1}\text{,}
\end{equation}

In this equation, $\boldsymbol{\psi}$ represents the new value at time $t$. The term $\alpha_{damp}$ is a weighting factor that defines the importance of the newly acquired value $\boldsymbol{\psi}$ relative to the \gls{dema} $\boldsymbol{\chi}^{dema}_{t-1}$ from the previous time step. The calculation of $\alpha_{damp}$ depends on the specific implementation, following the formula:
\begin{equation}
	\alpha_{damp} = \frac{2}{\mathrm{N}_{damp} + 1}\text{,}
\end{equation}

To calculate the damping factor for the \gls{dema}, the formula $\mathrm{N}_{damp} = \mathrm{N}\cdot \Delta$ is used, where $\mathrm{N}$ represents the number of \gls{dema} updates, and $\Delta$ depends on the type of information being updated. It is optimal to initialize $\mathrm{N}=0$ for faster initial convergence and to set a fixed upper-bound $\mathrm{N}_{max}$ to ensure that the \gls{dema} remains affected by new values over time, since $\alpha_{damp}$ approaches $0$ as $\mathrm{N}$ increases.

Two distinct damping factors are used for the \gls{dema}. One is for the ideal target features, $\boldsymbol{\mu}$ and $\boldsymbol{\sigma}$, which is designated $\Delta_{f}$ (see Eq.~\eqref{eq:deltaf}). The other is for the threshold $\lambda_d$, which is computed using $\mu_d$ and $\sigma_d$ and is called $\Delta_{\lambda_d}$ (see Eq.~\eqref{eq:deltal}).
\begin{align}
    \Delta_{f} &= min(1,\ \frac{d_{\boldsymbol{\mu},\boldsymbol{\sigma}}}{2})\text{,} \label{eq:deltaf}
    \\
    \Delta_{\lambda_d} &= max(1,\ 2\frac{d_{\boldsymbol{\mu},\boldsymbol{\sigma}}} {\lambda_d})\text{.} \label{eq:deltal}
\end{align}

Fig.~\ref{fig:ema_comparison} compares the \gls{ema} with and without these damping factors. It demonstrates that adding the damping factor $\Delta_{\lambda_d}$ attenuates the high-frequency noise of the threshold $\lambda_d$ (shown in red) and smoothes the entire line. This mechanism helps to prevent incorrect \gls{reid}s by reducing the $\lambda_d$ peaks that typically appear in the standard \gls{ema}.

When the damping factor $\Delta_{f}$ from Eq.~\eqref{eq:deltaf} is applied within \gls{dema} Eq.~\eqref{eq:dema} using $\mathrm{N}_{damp}$, it helps mitigate situations where the distance in Eq.~\eqref{eq:feat_dist} from the current feature is minimal. Such small distances could otherwise lead to overfitting the ideal representation, particularly when the target is stationary in the same pose. The damping factor $\Delta_{\lambda_d}$ in Eq.~\eqref{eq:deltal} is utilized to prevent the threshold from increasing excessively when the ratio between the distance $d_{\boldsymbol{\mu},\boldsymbol{\sigma}}$ and the threshold $\lambda_d$ is too large. This actively helps prevent incorrect \gls{reid}.

To enhance the algorithm's reliability, a \textit{blacklist manager} is introduced. This manager evaluates the \glspl{id} provided by the \gls{mot} and identifies which ones belong to distractors. By doing so, these distractor \glspl{id} are excluded during the \gls{reid} step. Whenever the target \gls{id} provided by the \gls{mot} remains consistent during tracking, the manager adds the \glspl{id} of the other individuals in the image (the distractors) to the blacklist. This approach minimizes \gls{reid} errors that might occur in challenging situations.

\subsection{Using unsupervised continual learning and parallel twin neural network to improve robustness online} \label{subsec:colp-method}

The \gls{carpe} framework\,\cite{rollo2024carpe} described in Section~\ref{subsec:carpe-method} is a robust approach for tracking people in images that can adapt to moderate changes in appearance. However, it is prone to a phenomenon known as catastrophic forgetting, meaning it tends to lose the memory of previously seen target appearances as it continually updates its model to accommodate new ones. This limitation is particularly challenging when the target person frequently changes their appearance.

To address this issue, an autonomous continual learning module is presented, aimed at fine-tuning the feature extraction network to extract features that are less reliant on the individuals' frequently changing appearances\,\cite{rollo2025pesonalized}. This continuously trained network specializes in recognizing the target by focusing on characteristics that are more consistent and unique to that person, making them distinguishable from the other actors present in the scene, known as distractors. The module autonomously acquires target and distractor images and performs parallel training once a sufficient number of samples is collected, thereby avoiding any delays in the tracking performance. A qualitative explanation of this concept is provided in Section~\ref{subsec:colp_eval}, where saliency maps, an explainability technique commonly used in computer vision, are employed to support the presented hypothesis.

This section is divided into several subsections to describe and validate the contributions of this work. It begins with the presentation of the idea of parallel training, along with considerations regarding its implementation. The explanation of the continual training setup and the smart image pool acquisition process is then presented. Finally, the features of post-training statistical model adaptation and early training are reported. These two modules play a central role in improving the overall accuracy of this framework.
\begin{figure}
    \centering
    \includegraphics[trim={9cm 4.3cm 7cm 5.7cm},clip,width=\linewidth]{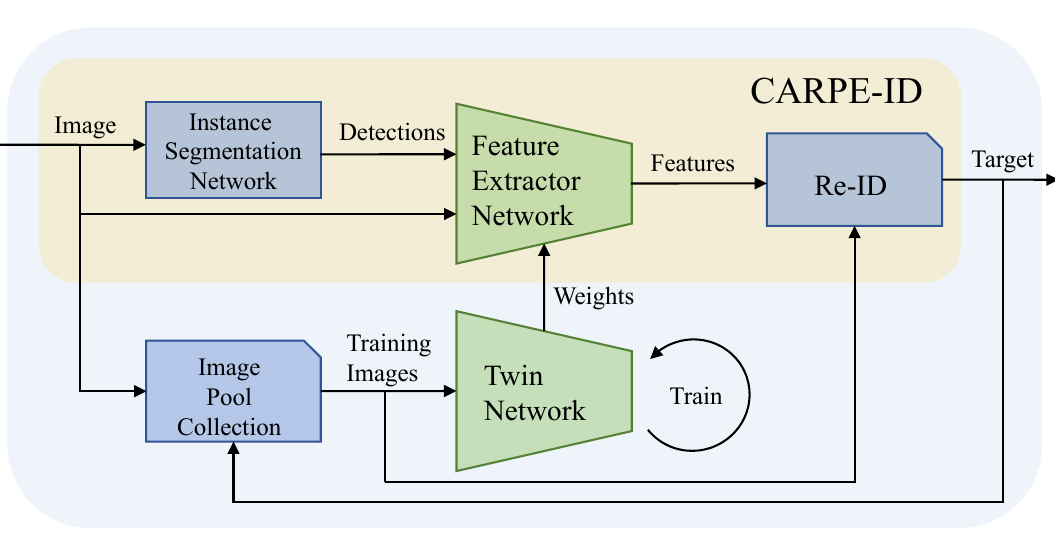}
    \caption{This work pipeline, integrated with the \gls{carpe} framework, outlines the algorithmic flow from the input image to the \gls{reid} output. The arrows symbolize the data exchange among the various components within the framework. Reprinted, with permission, from\,\cite{rollo2025pesonalized}, \textsuperscript{\textcopyright} 2025 \gls{ieee}.}
    \label{fig:colp-pipeline}
\end{figure}

\subsubsection{Parallel network} \label{subsect:parallel_train}

An essential requirement for online tracking is achieving optimal performance in terms of both tracking time and accuracy. It is crucial to maintain continuous detection at a high frame rate without losing the target. Conversely, implementing an online continual learning approach necessitates training the feature extractor neural network online. This training process can become impractical when the feature extractor network is rendered unavailable for inference while training\,\cite{parisi2019continual}. To address this challenge, a twin neural network structure is introduced, as illustrated in Fig.~\ref{fig:colp-pipeline}. This arrangement enables the concurrent training of the twin network while the principal network continues to execute the main tracking task. Once the twin network's training is completed, its new weights can be substituted into the principal network, thereby minimizing any delays in the tracking task.

Duplicating a neural network on \glspl{gpu} may impose a considerable computational burden. While feature extractor networks are generally lightweight, duplicating them can be difficult to handle in systems with standard computational capabilities, such as mobile robots, even though the advantages for the tracking task are clear. On the other hand, relying on a single network would result in losing track of the target during network training, especially if the target becomes occluded during this process, making subsequent re-recognition impossible.

\subsubsection{Training setup} \label{subsect:train_setup}

The people image buffer, obtained from the \gls{carpe} framework, consists of both target images and distractor images. In this context, a distractor refers to any person other than the target individual. This buffer is subdivided into $\mathrm{N}$ sets. Each set has a dimension that matches the batch size, $\mathrm{N}$, and contains $\frac{\mathrm{N}}{2}$ target and $\frac{\mathrm{N}}{2}$ distractor images. For each batch, a training iteration of the neural network is performed using the \textit{Soft Triplet loss} proposed by Qian et al.\,\cite{qian2019softtriple}, which is the standard Triplet loss function\,\cite{schroff2015facenet} in Eq.~\eqref{eq:triplet} embedded within the Log Softmax loss in Eq.~\ref{eq:softmax}.
\begin{align}
    \mathcal{L}_\mathcal{T} &= \sum_{i=1}^\mathrm{N}\left[\lVert\mathbf{x}_i^a - \mathbf{x}_i^p\rVert_2^2 - \lVert\mathbf{x}_i^a - \mathbf{x}_i^n\rVert_2^2 + \alpha\right], \label{eq:triplet} \\
    \mathcal{L}_\mathcal{S} &= -\log{\frac{\exp{\mathbf{x}_i}}{\sum_{j=1}^K\exp{\mathbf{x}_j}}} \label{eq:softmax}
\end{align}
where $\alpha$ is a bias weight. This combined loss function helps the model in discerning the similarities and differences between distinct people. It operates by considering three types of samples:
\begin{enumerate}
    \item \textit{Anchor} samples $\mathbf{x}_i^a$ representing target image features;
    \item \textit{Positive} samples $\mathbf{x}_i^p$ corresponding to other image features belonging to the same target;
    \item \textit{Negative} samples $\mathbf{x}_i^n$, denoting distractor image features.
\end{enumerate}

In practice, the method computes the Euclidean distances between all the embeddings within the batch, and for each positive and negative sample, the hard choices are maintained. Specifically, for each anchor, the farthest positive sample and the nearest negative sample are considered to obtain stronger feature embeddings. The Log Softmax is then computed for each resulting value, along with their mean weighted by a predefined margin $\alpha$. By using these samples, the network is trained so that the anchor sample is closer to the positive sample and further away from the negative one. This approach enables the network to learn how to distinguish between similar and dissimilar targets more effectively.

\subsubsection{Image pool collection} \label{subsect:image_collection}

\begin{figure}
    \centering
    \includegraphics[width=\linewidth]{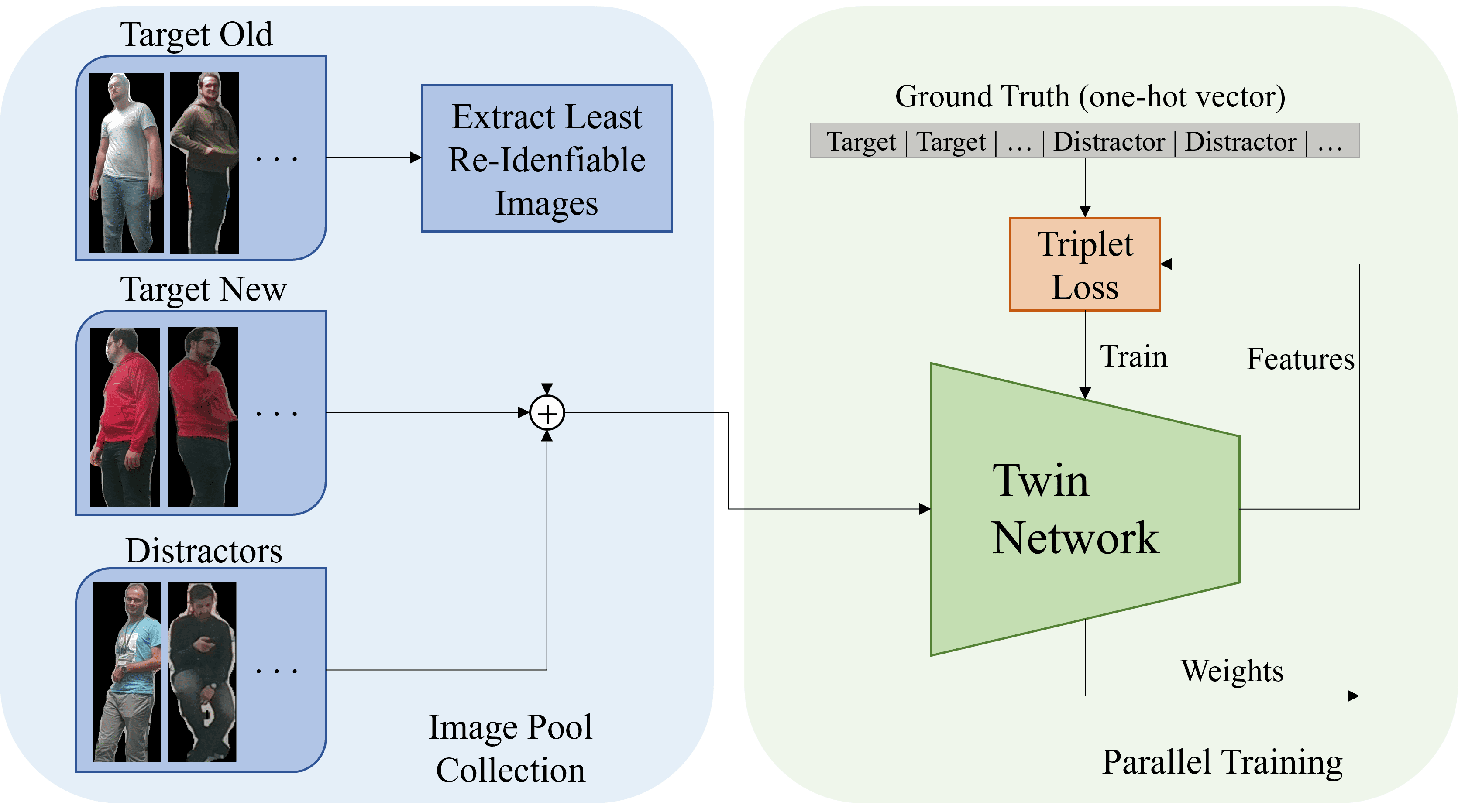}
    \caption{On the left, the pool collection of training images for adapting the twin network to target appearances, and on the right, the training process with Soft Triplet loss computation, is shown. Reprinted, with permission, from\,\cite{rollo2025pesonalized}, \textsuperscript{\textcopyright} 2025 \gls{ieee}.}
    \label{fig:colp_image_acquisition}
\end{figure}

In contemporary times, it is evident that datasets play a key role in the effectiveness of neural network training. If a dataset is poorly structured, it can adversely affect network performance. This holds not only in the context of continual learning but is further accentuated by the fact that datasets acquired in real-time have lower dimensions. Additionally, since such training data is collected online, there may not be substantial variation among the samples. Insufficient diversification among dataset samples can lead to overfitting in favor of the most recent inputs, thereby aggravating the problem of catastrophic forgetting\,\cite{parisi2019continual}.

To address this issue of improper network specialization, an intelligent image pool collection procedure is implemented that leverages the statistical model of \gls{carpe} to merge the latest acquired images with a set of older images to form the dataset. However, it is not feasible to retain all old images indefinitely, as this would lead to an unmanageable increase in dataset size, potentially causing costly training and \gls{gpu} overflow.

To overcome this challenge, an image selection mechanism is implemented that retains those images in the dataset that significantly differ in appearance from the current statistical model, as depicted in Fig.~\ref{fig:colp_image_acquisition}. To achieve this, the features extracted from the older target images and the statistical distance formula described in Eq.~\eqref{eq:feat_dist} are utilized to assess the similarity between each image embedding and the current model.

Next, the older images are arranged based on their calculated distances, sorting them in ascending order. The most recently acquired target images are then merged with those whose visual appearance differs the most from the model, provided the images still belong to the target of interest. Once this process is completed, the target image buffer is shuffled and added to the concurrently acquired distractor image buffer. The final combined image buffer is subsequently fed into the twin neural network for training. The size of the image buffer to be acquired is determined by the product of the selected batch size and the training iteration parameters set during initialization.

\subsubsection{Post-training statistical model adaptation} \label{subsect:model_adapt}

The feature vectors computed using the new network weights and the images used during the training described in the previous sections are subsequently used to refine the features and threshold statistical models. To ensure the compatibility of the new neural network weights with the statistical appearance model of \gls{carpe}, it is essential to update the model accordingly, as the introduction of these new weights can lead to changes in the neural network's output features that are no longer aligned with the previously acquired model. This adaptation approach allows the proposed framework to rapidly adjust to structural changes in the feature space, preventing misclassification and tracking losses (in this work, this variant is referred to as \textit{\gls{carpe} statistical model update}).

\subsubsection{Early training} \label{subsect:early_train}

While evaluating this framework, another challenge was encountered: cases where the tracked target moves out of the camera's \gls{fov} before the image buffer is sufficiently filled. This issue particularly occurs when the batch size and training iterations are set to high values. Under these circumstances, the framework may fail to start a single training iteration, consequently failing to incorporate the latest target appearances, which are essential for the \gls{reid} phase when the target re-enters the camera's \gls{fov}.

To address this issue, a solution that monitors the duration of target loss is introduced. If this period exceeds a predefined threshold, an early training session is initiated using the images currently available in the buffer. This approach ensures the framework can quickly overcome such challenging scenarios while maintaining tracking effectiveness.

\section{Environmental perception through geometrical matching} \label{sec:mapping-method}

As stated in Section~\ref{subsec:geometric-mapping-prob}, robots navigating in unknown environments require a robust and accurate \gls{slam} algorithm to build a map of the surroundings and localize themselves within it. Section~\ref{subsec:leo-slam-method} presents LEO\raisebox{0.1ex}{-}SLAM, a novel \gls{lidar}-based \gls{slam} algorithm designed to address the challenges of real-time mapping and localization in large-scale and dynamic environments. The proposed approach combines multi-level scan alignment with submap-based \gls{lcd} (Loop Closure Detection) to achieve accurate and efficient \gls{slam} performance. In Section~\ref{subsec:intensity-method}, a novel point cloud pre-processing technique is introduced that enhances the quality of the input data, thereby improving scan alignment and \gls{lcd}. Finally, in Section~\ref{subsec:gsc-method}, an improved \gls{lcd} method based on \gls{sota} algorithms is proposed, which enhances the robustness of the overall \gls{slam} system in noisy environments.

\subsection{Submap-based 3D LiDAR SLAM with multi-level scan matching} \label{subsec:leo-slam-method}

\begin{figure}
    \centering
    \includegraphics[width=\linewidth, trim={3cm 3cm 2.5cm 3cm}, clip]{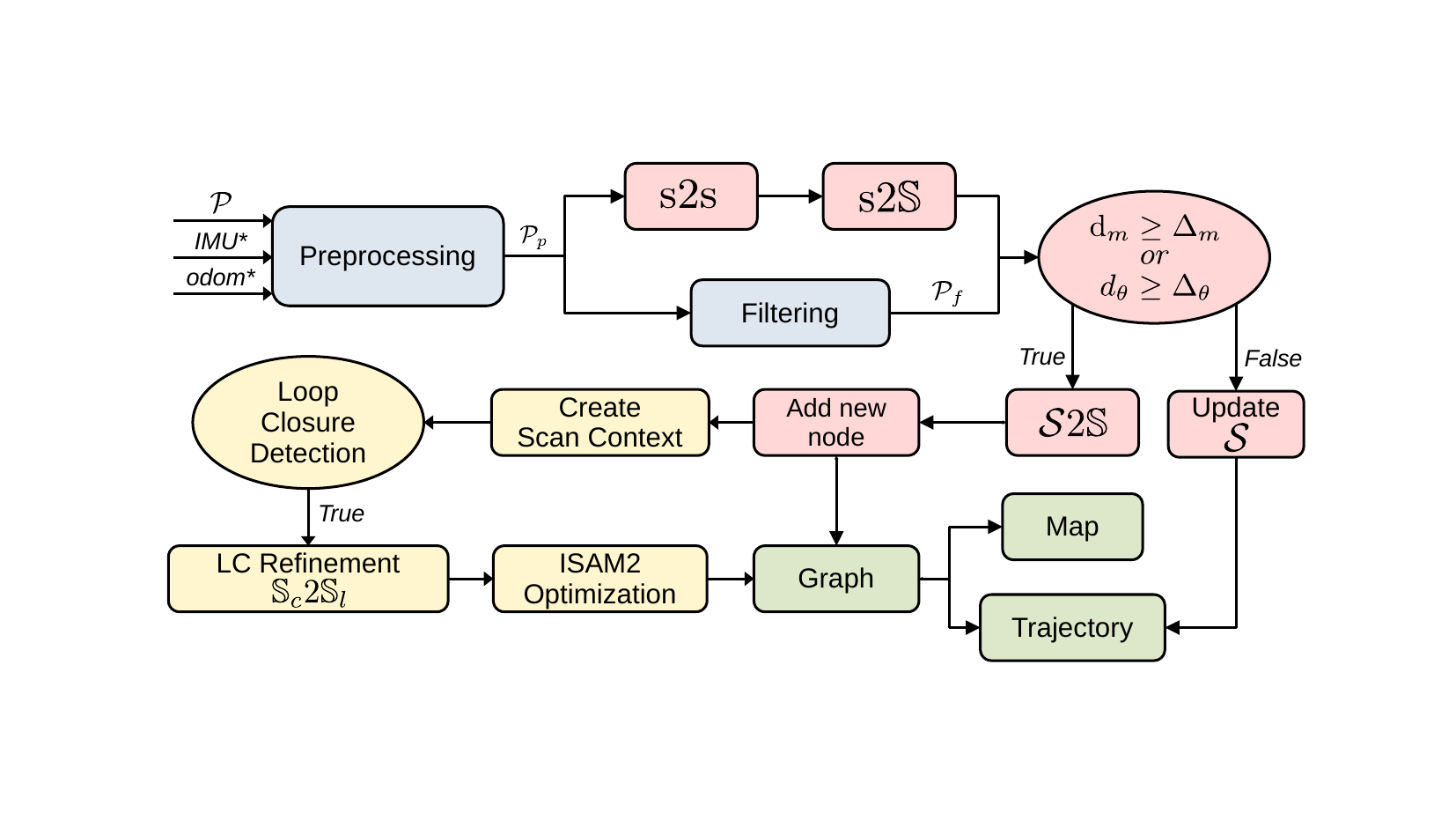}
    \caption{
     LEO\raisebox{0.1ex}{-}SLAM system architecture. The algorithm receives as input a point cloud $\mathcal{P}$ and optionally external odometry or \gls{imu} data. The point cloud is preprocessed, and the output $\mathcal{P}_p$ is passed in parallel into the scan-to-scan ($\mathrm{s}2\mathrm{s}$) and scan-to-submaps ($\mathrm{s}2\mathbb{S}$) alignment modules, as well as into a filtering module, which produces $\mathcal{P}_f$, later used for building the final map. If the robot has travelled a distance $d_m$ greater than $\Delta_m$, or rotated by an angle $d_\theta$ greater than or equal to $\Delta_\theta$, a submap-to-submaps ($\mathcal{S}2\mathbb{S}$) alignment is performed before adding a new node to the graph. If these distance and orientation conditions are not met, the current submap $\mathcal{S}$ and the trajectory are updated.
     When a new node is created, its submap $\mathcal{S}$ is used to compute a \gls{sc++}, and a \gls{lcd} step is performed. If a loop is detected, a submap-to-submaps ($\mathcal{S}2\mathbb{S}$) alignment is executed between the current node's submap $\mathcal{S}$ and the submaps around the detected loop-closure node $\mathbb{S}$. If the alignment converges, the loop is added to the pose graph, which is then optimized. The output of this optimization is used to correct both the trajectory and, consequently, the map. Reprinted, with permission, from\,\cite{rollo2025leoslam}, \textsuperscript{\textcopyright} 2025 \gls{ieee}.}
    \label{fig:leo-slam-architecture}
\end{figure}

This section presents the LEO\raisebox{0.1ex}{-}SLAM algorithm, beginning with the system overview illustrated in Fig.~\ref{fig:leo-slam-architecture}. This figure depicts the \gls{slam} workflow, with different colors representing the key conceptual blocks of the pipeline. Point cloud pre-processing and filtering are highlighted in light blue. The multi-level scan alignment strategy is shown in pink. Submap-based \gls{lcd} and pose graph optimization blocks are indicated in yellow. Finally, the algorithm outputs, namely the map and the reconstructed trajectory, are represented in green.

Algorithm\,\ref{alg:leo-slam} provides a simplified pseudocode representation of a single LEO\raisebox{0.1ex}{-}SLAM loop, which is the processing of an incoming point cloud $\mathcal{P}$ and, optionally, external odometry information $\mathbf{^\mathrm{o}H_\mathrm{b}}$. It outputs $^\mathrm{m}\mathbf{H}_{\mathrm{o}}$, the homogeneous transformation from the map frame $\mathrm{m}$ to the odometry frame $\mathrm{o}$, which corrects the robot’s pose. If no external odometry $\mathbf{^\mathrm{o}H_\mathrm{b}}$ is provided, the odometry frame $\mathrm{o}$ and the robot base frame $\mathrm{b}$ coincide, resulting in $^\mathrm{m}\mathbf{H}_{\mathrm{o}}\ =\ ^\mathrm{m}\mathbf{H}_{\mathrm{b}}$. The function output directly represents the robot’s pose relative to the initial map frame $\mathrm{m}$.

This \gls{slam} system is based on the concept of a node $\mathrm{n}$, which consists of a submap $\mathcal{S}$, an accumulation of point clouds, and a pose $^\mathrm{m}\mathbf{H}_{\mathrm{n}}$ relative to the fixed map frame $\mathrm{m}$.

\subsubsection{Point cloud preprocessing} \label{subsec:leoslam-preprocessing}

The input point cloud $\mathcal{P}$ undergoes two main steps: the \textit{preprocessing} and the \textit{filtering}, as illustrated in Fig.~\ref{fig:leo-slam-architecture}. In the \textit{preprocessing} stage, a set of \gls{3d} crop box filters is applied to remove points associated with the robot itself and those located too far from the \gls{lidar}, as distant points tend to be sparse and less useful for accurate mapping. Following this, a statistical outlier-rejection filter removes spurious points caused by motion, thereby improving the overall quality of the point cloud. These preprocessing operations ensure a balance between accuracy and computational efficiency, which is particularly crucial when dealing with sparse point clouds in the Multi-Level Alignment Strategy discussed in the next section. Additionally, the preprocessing block can incorporate external odometry or \gls{imu} data to provide an initial pose estimate for \gls{icp} alignment.

Once this initial step is completed, the preprocessed point cloud $\mathcal{P}_p$ is forwarded to both the Multi-Level Alignment layer and the subsequent \textit{filtering} stage. In this second stage, a ground-aware intensity filter is employed to discard points exceeding a specified intensity threshold, such as those from reflections on glass surfaces, while retaining low-intensity ground points. To further refine the data, a voxelization filter is applied to regulate the point cloud’s density, creating a more uniform distribution while preserving the environment's structure. While $\mathcal{P}_p$ serves as input to scan matching alignment, the final filtered point cloud $\mathcal{P}_f$ contributes to constructing a coherent map of the environment. This refined representation enhances visualization and facilitates navigation, ensuring that the generated map accurately reflects the real-world surroundings.

\subsubsection{Multi-level alignment strategy} \label{subsec:leoslam-alignment}

At each \gls{lidar} scan reception, the preprocessed point cloud $\mathcal{P}_p$ is first used to compute the robot's displacement. It is then transformed into the map frame $\mathrm{m}$ and accumulated within a submap $\mathcal{S}$. The scan collection continues until the robot's travelled distance $d_m$ from the last created node $\mathrm{n_{c-1}}$ exceeds a predefined Euclidean threshold $\Delta_m$, or its relative orientation change $d_\theta$ surpasses a set threshold $\Delta_{\theta}$. Once either of these conditions is met, a new node $\mathrm{n_c}$ is created, containing both the collected submap and the robot's current pose, which serves as the submap keyframe. This node is then added to a pose graph, which is structured as a factor graph that includes only pose factors.

To mitigate odometry drift, which commonly occurs in scan matching algorithms, a Multi-Level Alignment Strategy composed of three hierarchical alignment steps is employed:
\begin{enumerate}
    \item \textit{Scan-To-Scan} $\mathrm{s}2\mathrm{s}$;
    \item \textit{Scan-To-Submaps} $\mathrm{s}2\mathbb{S}$;
    \item \textit{Submap-To-Submaps} $\mathcal{S}2\mathbb{S}$.
\end{enumerate}
where $\mathrm{s}$ represents a single point cloud scan, $\mathcal{S}$ denotes a submap composed of multiple consecutive scans, and $\mathbb{S}$ is a subset of the $\mathrm{N}_s$ previously created submaps. Each level of alignment refines localization accuracy, ensuring both local and global consistency in the constructed map.

Alignment is performed at these three levels by exploiting the \gls{gicp} algorithm\,\cite{segal2009generalized}, specifically the small-GICP implementation\,\cite{koide2024small_gicp}. The $\mathrm{s}2\mathrm{s}$ alignment corrects small drifts by aligning the current \gls{lidar} scan $\mathcal{P}_t$ with the previous one $\mathcal{P}_{t-1}$. The resulting transformation serves as an initial guess for the $\mathrm{s}2\mathbb{S}$ alignment, where the scan is aligned with a collection of the $\mathrm{N}_s$ previous submaps to ensure local consistency.

Approaches such as \gls{dlo}\,\cite{chen2022direct} use keyframes as their nodes, \textit{i.e.}, a structure composed of a single point cloud and its associated pose. They utilize the $\mathrm{N}_s$ nearest keyframes in Euclidean space, summed with those extracted on the convex and concave hulls of the stored trajectory, to build the submap. In this way, they exploit keyframes from the entire trajectory rather than just the previous consecutive ones, as in the presented case. While this approach benefits robot pose refinement in an odometry system, it is less suitable for long trajectories and final map construction, as it may introduce discontinuities when the robot revisits a previously seen area, thereby skewing the final output map and trajectory.

Once a new node is ready, a final $\mathcal{S}2\mathbb{S}$ alignment is performed between the node submap $\mathcal{S}$ and $\mathbb{S}$, which is the subset of the $\mathrm{N}_s$ previously created submaps. This step improves long-term accuracy and reduces drift accumulation. Finally, after the multi-level alignment concludes, the submap keyframe is fixed, the node is finalized, and it is added to the pose graph. The node is then analyzed using \gls{sc++} for submap-based \gls{lcd}, as presented in the next section.

\subsubsection{Loop closure with submap-based Scan Context} \label{subsec:leoslam-loop_closure}

When a new node $\mathrm{n_c}$ is finalized, its associated submap $\mathcal{S}$ point cloud is used to compute a \gls{sc}, as presented by Kim et al.\,\cite{kim2021scan}. The \gls{sc} is a \gls{2d} descriptor that partitions the point cloud into rings and sectors, recording the maximal point height within each cell. This \gls{lcd} technique performs a global search against all previously generated \glspl{sc}. It utilizes a Ring Key (a 1D vector encoding ring values) to execute a preliminary, optimized check to determine if two \glspl{sc} are sufficiently similar to warrant a full comparison. Furthermore, a Sector Key embeds sector values into a vector, similar to the Ring Key, enabling the generated \gls{sc++} image to be circularly shifted to achieve orientation invariance.

Although this technique has proven effective in outdoor environments, it exhibits several limitations, particularly in indoor settings. The first limitation stems from the low channel count typical of mobile robot \gls{lidar} sensors (typically 16 channels) compared to the 64 or more channels commonly used in autonomous vehicles. To address this, a submap-based \gls{sc++} is introduced. Since each submap is an aggregate of multiple point clouds, it contains more environmental information, thereby yielding descriptors that resemble scans obtained from high-ring-count \gls{lidar} sensors.

Another limitation is that \glspl{sc} records the maximum point height for each entry, which is suboptimal indoors due to the presence of ceilings. To mitigate this, the ceiling is explicitly removed from the submap point cloud in indoor environments. This prevents the \gls{sc++} from becoming overly uniform and ignoring crucial environmental details. For example, in a corridor with obstacles, the maximum height would be uniformly defined by the ceiling, thus eliminating valuable height variations caused by the obstacles or the corridor structure itself.

Finally, structured environments often contain visually similar locations, leading to potential incorrect matches, even when spatially distant, as \gls{sc++} performs a global search across all previously acquired contexts. To address this, a search area ellipsoid $\mathcal{\mathbf{a_s}}$ is introduced. This area expands as the robot moves, growing proportionally to the motion uncertainty, which can be estimated from the robot's motion or \gls{lidar} odometry.

For each node $\mathrm{n_c}$ added to the graph, the ellipsoid is centered at the node's pose $^\mathrm{m}\mathbf{H}_{\mathrm{n_c}}$ and defines which \glspl{sc} fall within its area $\mathcal{\mathbf{a_s}}$. \gls{sc++} \gls{lcd} is then performed only on the variable-dimension node set $\mathbb{K}$, which consists of $k$ nodes whose distance vector $\mathrm{\mathbf{d_j}}$, with $j=1,\dots,k$, in Eq.~\eqref{eq:node_distance}, falls within the ellipsoid search area $\mathcal{\mathbf{a_s}}$, \textit{i.e.}, satisfying Eq.~\eqref{eq:ellipsoid_distance}.
\begin{equation} \label{eq:node_distance}
    \mathrm{\mathbf{d_j}}\ =\ ^\mathrm{n_c}\mathbf{t}_\mathrm{n_j}\ =\ ^\mathrm{m}\mathbf{t}_\mathrm{n_c}\ -\ ^\mathrm{m}\mathbf{t}_\mathrm{n_j},
\end{equation}
\begin{equation} \label{eq:ellipsoid_distance}
    \left(\frac{\mathrm{\mathbf{d_j^x}}}{\mathcal{\mathbf{a_s^x}}}\right)^2 + \left(\frac{\mathrm{\mathbf{d_j^y}}}{\mathcal{\mathbf{a_s^y}}}\right)^2 + \left(\frac{\mathrm{\mathbf{d_j^z}}}{\mathcal{\mathbf{a_s^z}}}\right)^2 \leq 1,
\end{equation}
where $x$, $y$, and $z$ are the components of vectors $\mathrm{\mathbf{d_j}}$ and $\mathcal{\mathbf{a_s}}$.

If a loop is detected between $\mathrm{n_c}$ and $\mathrm{n_l}$, the translation vector $\mathrm{\mathbf{d_j}}$ with $j = l$ in Eq.~\eqref{eq:node_distance} is used as the initial guess for the \gls{gicp} algorithm. Two point cloud sets, $\mathbb{S}_c$ and $\mathbb{S}_l$, are created for the current node $\mathrm{n_c}$ and the loop node $\mathrm{n_l}$, respectively. Each set contains the nodes in the neighborhood of the associated node. Specifically, $\mathbb{S}_c$ contains the $v$ nodes preceding the current node, while $\mathbb{S}_l$ contains the $v$ nodes preceding and the $v$ nodes succeeding the loop node $\mathrm{n_l}$, where $v \in \mathbb{N}$ is a predefined positive integer. The \gls{gicp} algorithm is then computed to align the $\mathbb{S}_c$ point clouds with the $\mathbb{S}_l$ point clouds, using $\mathrm{\mathbf{d_l}}\ =\ ^\mathrm{n_c}\mathbf{t}_\mathrm{n_l}$ as the initial guess. This $\mathbb{S}_c2\mathbb{S}_l$ alignment is employed to improve loop closure accuracy and mitigate false alignments in low-texture environments.

If the alignment converges, the \gls{gicp} homogeneous transformation output $\mathbf{H}^{ICP}_{lc}$ is used to correct the current node pose $^\mathrm{m}\mathbf{H}_{\mathrm{n_c}}$, as expressed in Eq.~\eqref{eq:node_correction}:
\begin{equation} \label{eq:node_correction}
    ^\mathrm{m}\mathbf{H}_{\mathrm{n_c}}^{'} = \mathbf{H}^{ICP}_{lc}\ *\ ^\mathrm{m}\mathbf{H}_{\mathrm{n_c}},
\end{equation}
where $^\mathrm{m}\mathbf{H}_\mathrm{n_c}^{'}$ is the corrected pose of node $\mathrm{n_c}$. The estimated final pose between the two nodes in Eq.~\eqref{eq:loop_h_current} is then used to close the loop, adding the corrected pose factor $^\mathrm{n_c}\mathbf{H}_{\mathrm{n_j}}^{'}$ between node $\mathrm{n_c}$ and node $\mathrm{n_j}$:
\begin{equation} \label{eq:loop_h_current}
    ^\mathrm{n_c}\mathbf{H}_{\mathrm{n_j}}^{'}\ =\ ^\mathrm{m}\mathbf{H}_{\mathrm{n_c}}^{'-1}\ *\ ^\mathrm{m}\mathbf{H}_{\mathrm{n_j}}.
\end{equation}

When the loop is correctly closed between two non-consecutive nodes, the pose graph is optimized using \gls{isam2}\,\cite{kaess2012isam2}, and the resulting new node poses are used to refine the nodes' keyframes.

\subsection{Point cloud ground-aware intensity filtering} \label{subsec:intensity-method}

As stated in Section~\ref{subsec:leoslam-preprocessing}, point cloud filtering is a crucial step in the LEO\raisebox{0.1ex}{-}SLAM pipeline, as it enhances the quality of the input data, leading to improved scan alignment and \gls{lcd}. In this section, a novel ground-aware intensity filter is presented that effectively removes points with intensities outside a specified range while preserving low-intensity ground points. Two lightweight implementations of this filter are proposed: the first, named the \textit{naive intensity filter}, is simpler and faster; while the second implementation, referred to as the \textit{normal intensity filter}, is more complex.

Both filters iterate through all the points $^\mathrm{l}\mathbf{p}_i$ of the point cloud $^\mathrm{l}\mathcal{P} = \{^\mathrm{l}\mathbf{p}_1, ^\mathrm{l}\mathbf{p}_2, \dots, ^\mathrm{l}\mathbf{p}_n\}$, expressed in the \gls{lidar} frame $\mathrm{l}$, and check whether their intensity $\psi_i$ falls outside the specified boundary $\psi^* = [\psi^*_{min}, \psi^*_{max}]$. This condition defines the intensity range to be filtered out, as expressed in Eq.~\eqref{eq:intensity_filter}:
\begin{equation} \label{eq:intensity_filter}
    \psi_i \notin \psi^* \equiv (\psi_i < \psi^*_{min})\ \&\ (\psi_i > \psi^*_{max})
\end{equation}

\begin{figure}
    \centering
    \includegraphics[width=\linewidth]{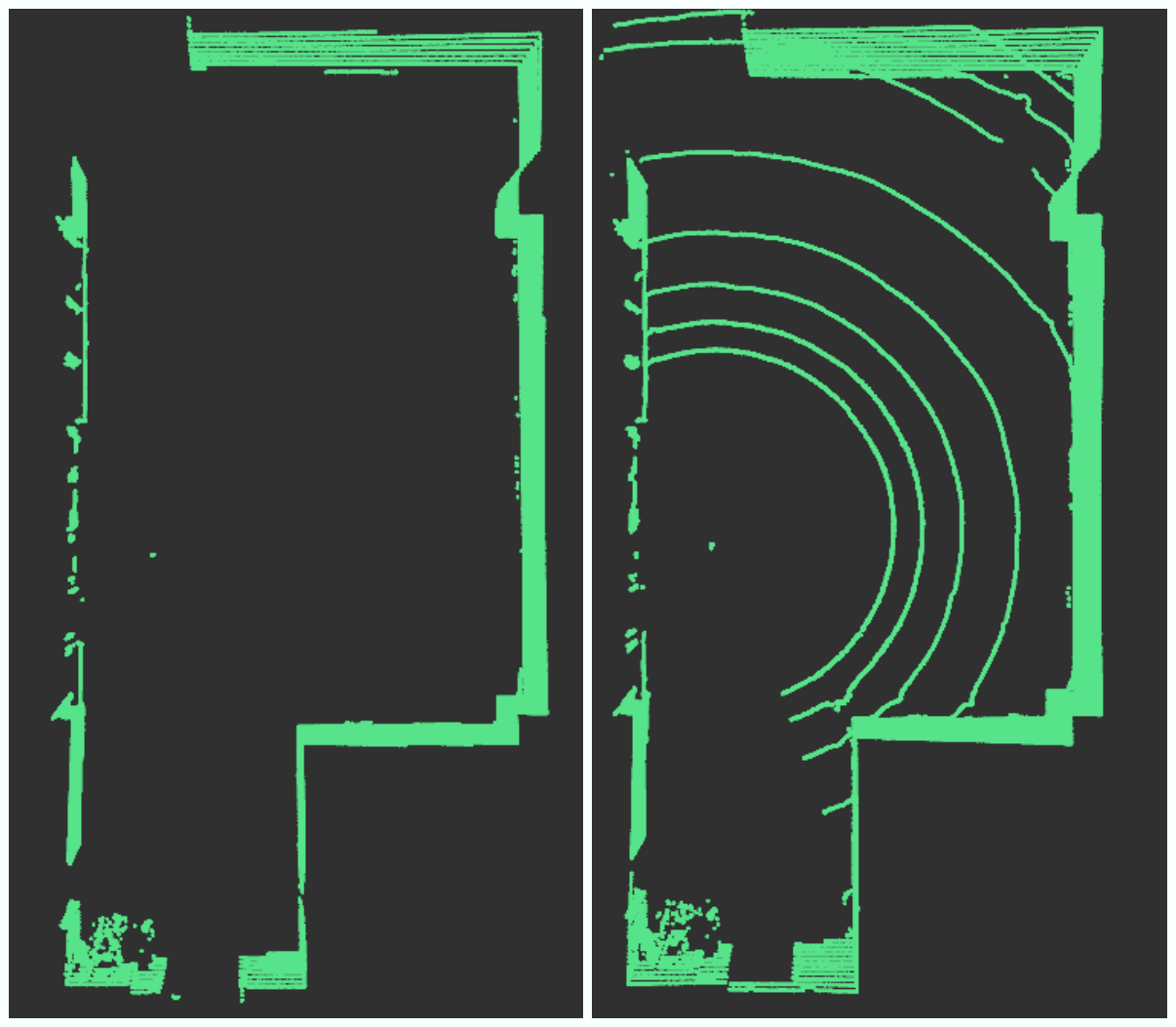}
    \caption{Comparison of intensity filtering: on the left, using only intensity thresholding; on the right, incorporating the ground awareness into the filtering process.}
    \label{fig:filter_compare}
\end{figure}

However, points belonging to the ground often return low reflected energy due to the low incidence angle $\theta$ between the \gls{lidar} ray and the ground surface. These ground points are typically filtered out by a standard thresholding filter, as seen in the left image of Fig.~\ref{fig:filter_compare}, which can cause problems for \gls{lidar} odometry and \gls{slam} systems. Fig.~\ref{fig:filter_compare} demonstrates the benefits of incorporating ground awareness through a comparison between the standard intensity thresholding filter and one enhanced with ground awareness.

\subsubsection{Naive intensity filter} \label{subsec:naive_filter}

As the name suggests, the first implementation, also known as the \textit{naive intensity filter}, is a straightforward application of a bound filter with basic ground awareness.

To address the issue of ground filtering, all points located below the robot's footprint, \textit{i.e.}, the projection of the robot's base link onto the ground, are retained. This approach ensures that ground points, which provide valuable environmental information, are preserved. However, it also retains incorrect reflections detected below the robot’s height. While such false detections could be considered minimal and insignificant compared to the other points being filtered out, it is important to acknowledge their presence.

Considering $^\mathrm{l}\mathbf{t}_\mathrm{f}$ as the translation vector between the \gls{lidar} frame $\mathrm{l}$ and the footprint frame $\mathrm{f}$, the third component, $^\mathrm{l}z_\mathrm{f}$, translates from the \gls{lidar} frame $\mathrm{l}$ to the footprint frame $\mathrm{f}$ along the $z$-axis. This component can be used to check whether the points $^\mathrm{l}\mathbf{p}_i$ in the \gls{lidar} frame $\mathrm{l}$ lie below the footprint frame $\mathrm{f}$.

Note that for fixed-wheeled robots, $^\mathrm{l}\mathbf{t}_\mathrm{f}$ is static and does not change during navigation. In this experimental setup, however, a quadruped robot is used, where $^\mathrm{l}\mathbf{t}_\mathrm{f}$ can be computed using the positions of the feet, $^\mathrm{l}\mathbf{t}_\mathrm{f_i}$, with $\mathrm{f_i}$ representing the pose of the feet in contact with the ground. Another important consideration is that the \gls{lidar} $xy$-plane is assumed to be parallel to the base footprint $xy$-plane, meaning that only \gls{lidar}-to-footprint translations are necessary. If this assumption is not valid, such as when the quadruped robot is tilted, or the \gls{lidar} is positioned with pitch or roll angles, fthe full homogeneous transformation $^\mathrm{l}\mathbf{H}_\mathrm{f}$ should be used to roto-translate the \gls{lidar} points into the footprint frame before checking their height. The details of these platform-dependent factors are beyond the scope of this paper and are left to the reader.

Taking into account the above considerations and assumptions, the final \textit{naive intensity filter} equation is:
\begin{equation}
    ^\mathrm{l}\mathcal{P}^* = \{^\mathrm{l}\mathbf{p}_i\ \in\ ^\mathrm{l}\mathcal{P}\ |\ (\psi_i \notin \psi^*)\ \&\ (^\mathrm{l}p_i^z <\ ^\mathrm{l}z_\mathrm{f}) \}
\end{equation}
where $^\mathrm{l}\mathcal{P}^*$ is the ground-aware intensity filtered point cloud, $^\mathrm{l}p_i^z$ is the $z$ component of point $^\mathrm{l}\mathbf{p}_i$, "$|$" and "$\&$" denote "\textit{such that}" and "\textit{conditional and}" respectively, and $\psi_i \notin \psi^*$ is defined in Eq.~\eqref{eq:intensity_filter}.

\subsubsection{Normal intensity filter} \label{subsec:normal_filter}

The second proposed solution is more sophisticated than the previous one; however, its benefits do not necessarily outweigh the issues it introduces. Similar to the naive approach, the normal filter checks whether the point intensity $^\mathrm{l}\mathbf{p}^\psi_i$ lies outside the boundaries $\psi^* = [\psi^*_{min}, \psi^*_{max}]$, and whether $^\mathrm{l}\mathbf{p}_i$ belongs to the ground. However, the ground check is now more advanced, involving the computation of the normal vector $^\mathrm{l}\boldsymbol{\eta}_i$ for each cloud point $^\mathrm{l}\mathbf{p}_i$.

The \textit{Normal Intensity Filter} consists of two primary steps:
\begin{enumerate}
    \item Compute the normal vectors $^\mathrm{l}\boldsymbol{\eta}_i$ for the points $^\mathrm{l}\mathbf{p}_i$.
    \item Check if the point's normal vector $^\mathrm{l}\boldsymbol{\eta}_i$ is parallel to the $z$-axis of the robot's footprint frame $\mathrm{f}$.
\end{enumerate}

In the norm computation step, the normal vectors of the points are calculated using the technique presented by Rusu et al.\,\cite{rusu2010semantic}, with the implementation provided by the \gls{pcl}\footnote{\gls{pcl} library normal estimation: \url{https://pointclouds.org/documentation/tutorials/normal_estimation.html}}.

To check whether a normal vector $^\mathrm{l}\boldsymbol{\eta}_i$ associated with a point $^\mathrm{l}\mathbf{p}_i$ is parallel to the $z$-axis of the robot's footprint frame $\mathrm{f}$, the vector must first be roto-translated into the footprint frame using the following transformation:
\begin{equation}
    ^\mathrm{f}\boldsymbol{\eta}_i =\ ^\mathrm{f}\mathbf{H}_l\ ^\mathrm{l}\boldsymbol{\eta}_i
\end{equation}
where $^\mathrm{f}\mathbf{H}_\mathrm{l}$ is the homogeneous transformation between the \gls{lidar} and the footprint frame, and $^\mathrm{f}\boldsymbol{\eta}_i$ is the normal vector expressed in the footprint frame.

Once the normal vector is expressed in the footprint frame, the angle between the vector $^\mathrm{f}\boldsymbol{\eta}_i$ and the $z$-axis of the $\mathrm{f}$ frame, defined as $^\mathrm{f}\mathbf{z} = \mathbf{[0,\ 0,\ 1]^T}$, can be calculated. To compute this angle, the vector cosine similarity measure, $\theta_{\eta_i}$, is used, which is given by:
\begin{equation} \label{eq:cosine}
    \theta_{\eta_i} = \frac{\langle ^\mathrm{f}\boldsymbol{\eta}_i,^\mathrm{f}\mathbf{z} \rangle}{\|^\mathrm{f}\boldsymbol{\eta}_i\|_2 \|^\mathrm{f}\mathbf{z}\|_2}
\end{equation}
where $\langle \cdot, \cdot \rangle$ denotes the dot product , and $\|\boldsymbol{x}\|_2$ represents the vector $L_2$-norm.

Using the cosine similarity $\theta_{\eta_i}$ from Eq.~\eqref{eq:cosine}, it can be estimated whether the point $^\mathrm{l}\mathbf{p}_i$, associated with its normal vector $^\mathrm{l}\boldsymbol{\eta}_i$, lies on a surface parallel to the robot's base $xy$-plane (\textit{i.e.}, the ground). When $^\mathrm{f}\boldsymbol{\eta}_i$ is parallel to $^\mathrm{f}\mathbf{z}$, the cosine value $\theta_{\eta_i}$ equals $1$ (since the angle is $0$ or $\pi$, but typically normalized to $0$ for parallel surfaces). However, due to noisy measurements and small delays in sensor data collection, it is good practice to apply a minimal threshold $\epsilon_\eta$ and check whether $\theta_{\eta_i}$ falls above a value close to $1$.

Given the above explanations, the final filter equation can be expressed as:
\begin{equation}
    ^\mathrm{l}\mathcal{P}^* = \{^\mathrm{l}\mathbf{p}_i\ \in\ ^\mathrm{l}\mathcal{P}\ |\ \theta_{\eta_i} > 1-\epsilon_\eta\  \}
\end{equation}

\subsection{Statistical loop closure detection with Gaussian Scan Context} \label{subsec:gsc-method}

\begin{figure}
    \centering
    \includegraphics[width=\linewidth]{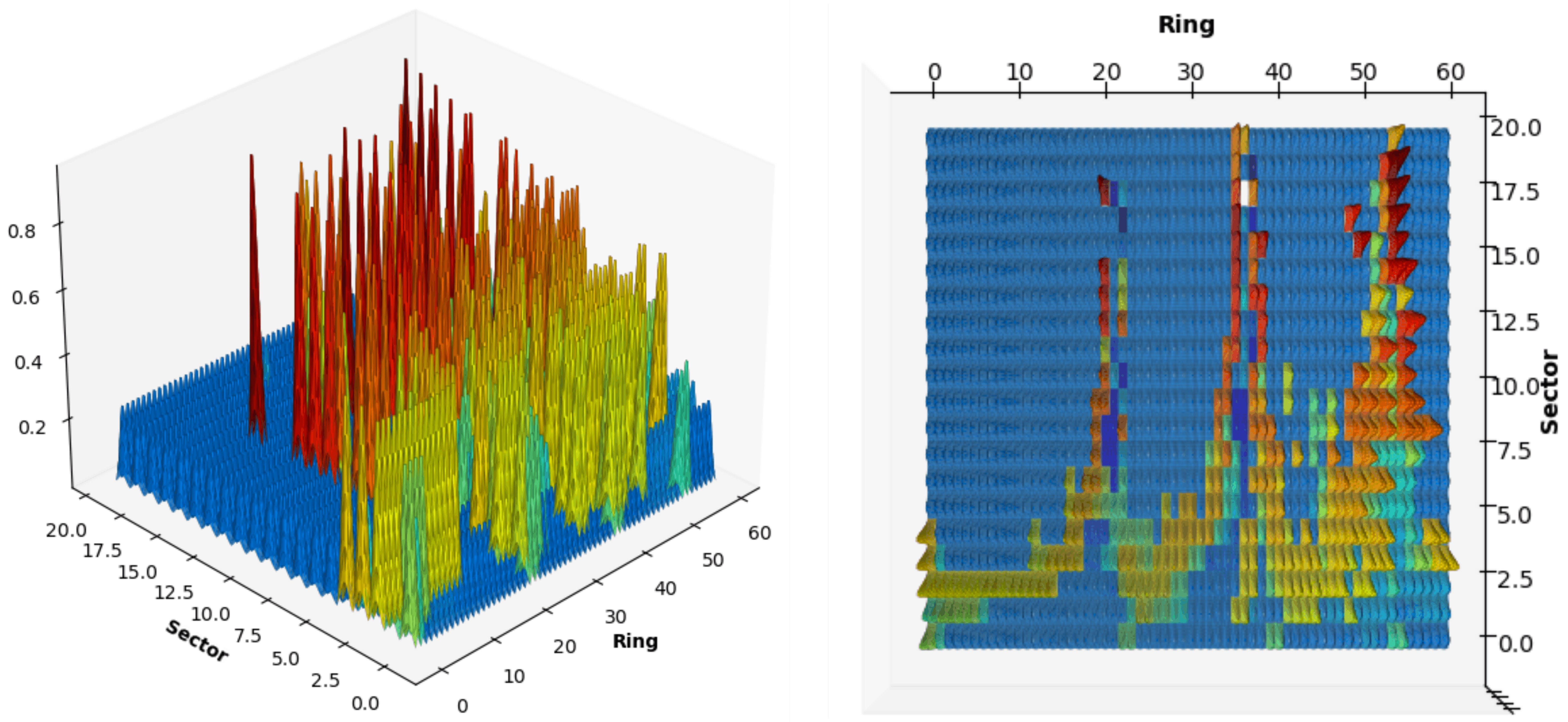}
    \caption{The diagonal and top views of a \gls{gsc}, where each entry represents a multivariate normal distribution, are shown on the left and right, respectively. Each Gaussian distribution contributes to the computation of the \gls{gsc} matrix.}
    \label{fig:gsc}
\end{figure}

In Section~\ref{subsec:leo-slam-method}, the use of \gls{sc++} for the \gls{lcd} step of LEO\raisebox{0.1ex}{-}SLAM is reported. Although this approach is one of the most widely used among the \gls{sota} algorithms due to its simplicity and effectiveness, it relies on certain assumptions regarding the noise present in the point clouds. This technique performs well when the input point cloud contains few outliers or minimal noise, a condition not always met in real-world scenarios. To address this problem, the \gls{gsc}, illustrated in Fig.~\ref{fig:gsc}, is introduced. The \gls{gsc} is a statistical enhancement designed to improve the robustness and performance of \gls{lcd} by explicitly incorporating Gaussian-based analysis of the point cloud data. Before proposing the specific methodology, a deeper overview of \gls{sc++} is provided in the next section to thoroughly establish its foundations.

\subsubsection{Scan Context++: methodology overview} \label{subsec:sc}

\begin{figure}
    \centering
    \includegraphics[width=\linewidth]{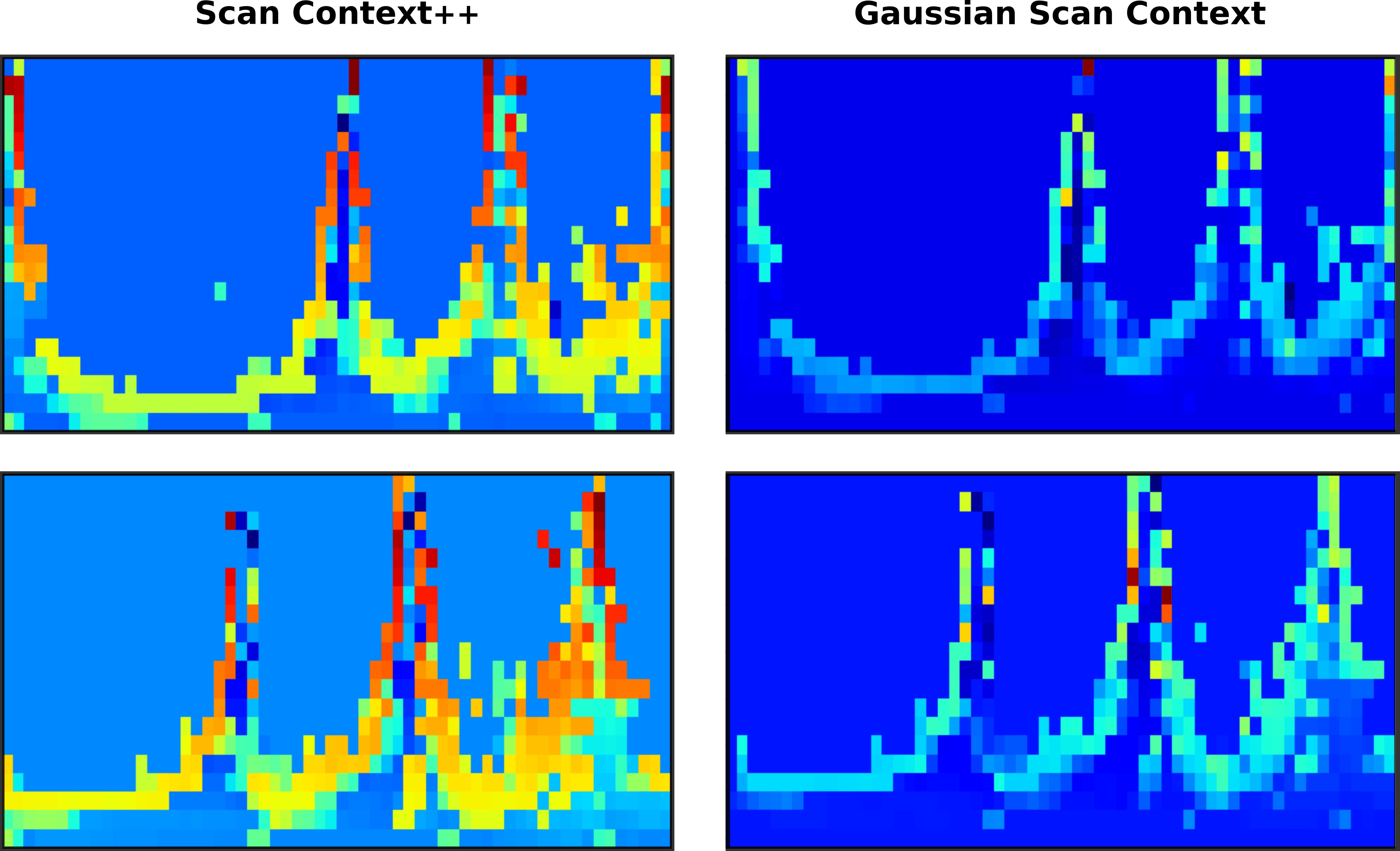}
    \caption{The plots compare \gls{sc++} (left) and \gls{gsc} (right) during a loop closure in the KITTI sequence $\mathrm{00}$. The \gls{gsc} matrix, displayed at the lower right, is derived from the Gaussian distributions illustrated in Fig.~\ref{fig:gsc}.}
    \label{fig:sc}
\end{figure}

\gls{sc++} is an enhanced \gls{lidar}-based place recognition framework designed to address rotational and lateral variations prevalent in urban driving environments. 

The process extends the original \gls{sc} descriptor by introducing mechanisms that improve efficiency and robustness without relying on learning-based components. The procedure begins by transforming the \gls{3d} \gls{lidar} point cloud data into a compact \gls{2d} matrix $\mathbf{C} \in \mathbb{R}^{\mathrm{N}_r \times \mathrm{N}_s}$, referred to as the \gls{sc++} (see Fig.~\ref{fig:sc}). Here, $\mathrm{N}_r$ represents the number of radial bins (rings) and $\mathrm{N}_s$ the number of angular bins (sectors).

The point cloud is first projected onto a polar grid $\mathcal{P}$, defined by $\mathrm{N}_r$ concentric rings and $\mathrm{N}_s$ angular sectors. Each cell (bin) in the polar grid is denoted by $\mathbf{b}_{i,j}$ and contains $D_{i,j}$ points $\boldsymbol{p} \in \mathbb{R}^4$, each representing a single point $(p^x, p^y, p^z, p^\psi)$ expressed in the \gls{lidar} frame, falling within the spatial extent of that bin, such that $\mathbf{b}_{i,j} \in \mathbb{R}^{D_{i,j} \times 4}$, where $ i = 1, \dots, \mathrm{N}_r$ and $j = 1, \dots, \mathrm{N}_s$. The ring index $i$ and the sector index $j$ are computed as:
\begin{align}
    i =& \max(\min(\mathrm{N}_r, \lceil\frac{p_r}{\max_r}\rceil \cdot \mathrm{N}_r),1), \\
    j =& \max(\min(\mathrm{N}_s, \lceil\frac{p_\theta}{\max_\theta}\rceil \cdot \mathrm{N}_s),1),
\end{align}
where $p_r$ and $p_\theta$ denote the radial distance and azimuthal angle of the point $\boldsymbol{p}$ in polar coordinates; $\max_r$ and $\max_\theta = 360^\circ$ represent their respective maximum values; and $\lceil\cdot\rceil$ denotes the ceiling function.

Each bin $\mathbf{b}_{i,j}$ is subsequently used to compute the \gls{sc++} cell $\mathbf{C}_{i,j}$ by storing the maximum height value of the points contained within the corresponding bin $\mathbf{b}_{i,j}$, i.e., $p^z$, as defined by the following equation:
\begin{equation} \label{eq:scentry}
    \mathbf{C}_{i,j} = \max_{\boldsymbol{p}\in \mathbf{b}_{i,j} }\ p^z.
\end{equation}

This representation captures the egocentric spatial structure of the environment but remains inherently rotation-variant.

To achieve \textit{rotation invariance}, \gls{sc++} computes a summary vector, referred to as the \textit{ring key}, which is obtained by taking the $L_0$-norm (i.e., counting the non-zero entries) across each row of $\mathbf{C}$. Formally, the ring key $\mathbf{k} \in \mathbb{R}^{\mathrm{N}_r}$ is defined as:
\begin{equation}
    k_i = \|\mathbf{C}_{i,:}\|_0 \quad \text{for } i = 1, 2, \dots, \mathrm{N}_r.
\end{equation}

These ring keys are stored in a k-d tree to enable efficient nearest-neighbor search during the place-recognition query stage.

Once potential matches are retrieved using the ring keys, the system performs \textit{descriptor alignment} to compensate for rotational discrepancies. This is achieved by cyclically shifting the columns of the candidate matrix $\mathbf{C}_c$ and computing the cosine similarity with the query matrix $\mathbf{C}_q$ at each shift. The optimal alignment is determined as:
\begin{equation}
    \text{sim}(\mathbf{C}_q, \mathbf{C}_c^{(\theta)}) = \frac{\langle \mathbf{C}_q, \mathbf{C}_c^{(\theta)} \rangle}{\|\mathbf{C}_q\|_F \|\mathbf{C}_c^{(\theta)}\|_F},
\end{equation}
where $\theta$ denotes the circular shift index and $\|\cdot\|_F$ represents the Frobenius norm.

To enhance robustness against \textit{lateral variations}, which arise when the same location is revisited from different lateral offsets (\textit{e.g.}, distinct driving lanes), \gls{sc++} augments the original point cloud by synthetically shifting it laterally before generating new \gls{sc++} descriptors. These \textit{virtual \gls{sc}s} emulate the effect of observing the same scene from slightly displaced viewpoints, enabling the system to maintain reliable matching across such offsets.

Subsequently, a \textit{false positive rejection} mechanism is employed. This procedure re-evaluates the similarity of the full \gls{sc++} descriptors (beyond the ring keys) and applies a predefined threshold $\tau$ to determine whether the match is accepted or rejected according to the following criterion:
\begin{equation}
    \text{sim}(\mathbf{C}_q, \mathbf{C}_c^{(\theta^*)}) > \tau,
\end{equation}
where $\theta^*$ is the optimal rotation alignment identified earlier.

Through these combined steps, \gls{sc++} attains efficient and accurate place recognition, making it suitable for real-world autonomous navigation scenarios.

\subsubsection{Gaussian extension} \label{subsec:gscext}

Although the latest version of \gls{sc++} offers substantial improvements and is sufficiently robust for integration into \gls{3d} \gls{lidar} \gls{slam} systems, it employs a relatively simplistic strategy to populate the context matrix $\mathbf{C}$ (see Eq.~\eqref{eq:scentry}). Specifically, each matrix entry is determined by selecting the maximum height value among the points within the corresponding bin $\boldsymbol{b}_{i,j}$. While computationally efficient, this maximum-value approach exhibits limited robustness to outliers, such as spurious reflections or sensor noise, which are commonly encountered in both indoor and outdoor \gls{lidar} mapping environments.

To overcome this limitation, a statistical enhancement that accounts for the full distribution of points within each bin is introduced. For each bin $\boldsymbol{b}_{i,j}$, an associated multivariate normal distribution ${\boldsymbol{\mathcal{G}} }_{i,j} \equiv \mathcal{N}(\boldsymbol{\mu}_{i,j}, \mathbf{\Sigma}_{i,j})$ is defined, where the mean vector $\boldsymbol{\mu}_{i,j}$ and covariance matrix $\mathbf{\Sigma}_{i,j}$ are computed as follows:
\begin{align}
    \boldsymbol{\mu}_{i,j} &= \frac{1}{\mathrm{N}} \sum_{\boldsymbol{p}\in \mathbf{b}_{i,j}} \mathbf{p} , \\
    \mathbf{\Sigma}_{i,j} &= \frac{1}{\mathrm{N}-1} \sum_{\boldsymbol{p}\in \mathbf{b}_{i,j}} (\mathbf{p} - \boldsymbol{\mu})(\mathbf{p} - \boldsymbol{\mu})^\top,
\end{align}
where $\mathrm{N} = |\mathbf{b}_{i,j}|$ is the number of points within bin $\mathbf{b}_{i,j}$. The covariance matrix is calculated using a denominator of $\mathrm{N} - 1$ to yield an unbiased estimator.

As demonstrated in Section~\ref{subsec:gsc-exp}, this straightforward statistical model yields significant improvements in both the accuracy and robustness of \gls{lcd}. While a univariate Gaussian $\mathcal{N}(\mu, \sigma^2)$ based solely on height values could have been employed with the same results, the multivariate formulation is used because it captures the spatial structure of the data more effectively, which can support future extensions and more sophisticated implementations.

After computing all Gaussian distributions $\boldsymbol{\mathcal{G}}_{i,j}$, the \gls{gsc} matrix $\mathbf{C}$ can be populated as follows:
\begin{equation} \label{eq:gscentry}
    \mathbf{C}_{i,j} = \boldsymbol{\mu}^z_{i,j} + \alpha\ \Sigma^{zz}_{i,j},
\end{equation}
where $\boldsymbol{\mu}^z_{i,j}$ and $\Sigma^{zz}_{i,j}$ represent the mean and variance of the height ($z$) component of the bin $\boldsymbol{b}_{i,j}$, and $\alpha \in \mathbb{R}^+$ is a tunable hyperparameter that controls the contribution of the variance term. This formulation preserves the core concept of estimating the maximum height, as originally applied in Eq.~\eqref{eq:scentry}, but achieves this by statistically modeling the complete height distribution of the points within each bin. Consequently, the resulting context descriptor becomes more resilient to outliers and noise while retaining meaningful geometric discrimination between locations.

To further improve robustness against outliers and heavy-tailed noise, the weighted Gaussian matrix $\boldsymbol{\mathcal{G}}_h \equiv \mathcal{N}_h(\boldsymbol{\mu}_h, \mathbf{\Sigma}_h)$ is introduced, which employs Huber weighting to produce a more compact and representative Gaussian distribution for each bin. This method further reduces the influence of outlier points that might otherwise distort the statistical characterization of the point cloud segment.

The central principle of the Huber weight is to down-weight points that deviate substantially from the estimated mean $\boldsymbol{\mu}$, thereby mitigating their effect on the resulting distribution $\boldsymbol{\mathcal{G}}_h$. The residual $r$ is defined as the Euclidean distance between a point $\boldsymbol{p}$ and the bin mean:
\begin{equation}
    r = \|\boldsymbol{p} - \boldsymbol{\mu} \|_2
\end{equation}

The corresponding weight is derived from the Huber loss function, parameterized by a user-defined threshold $\delta$:
\begin{equation}
    \rho_{\delta}(r) =
    \begin{cases}
        \frac{1}{2}r^2 & \text{if } |r| \leq \delta \\
        \delta(|r| - \frac{1}{2}\delta) & \text{if } |r| > \delta
    \end{cases}
\end{equation}

This piecewise function exhibits quadratic behavior for small residuals, enhancing sensitivity to inliers, and linear behavior for large residuals, thereby reducing the influence of outliers. By incorporating this weighting scheme into the computation of the Gaussian parameters, a more robust and representative characterization of the spatial structure within each bin is achieved.

To further suppress the effect of outliers, a simplified Huber weight function derived from the original Huber loss is employed, defined as:
\begin{equation}
    \omega_k =
    \begin{cases}
        1 & \text{if } |r_k| \leq \delta \\
        \frac{\delta}{|r_k|} & \text{if } |r_k| > \delta
    \end{cases}
\end{equation}
where $r_k$ denotes the residual for point $\mathbf{p}_k$, and $\delta$ is a user-defined threshold that controls the transition between full weighting and reduced weighting.

Using these weights, the Huber-weighted mean $\boldsymbol{\mu}_h$ and covariance matrix $\mathbf{\Sigma}_h$ for each bin is computed as follows:
\begin{align}
    \boldsymbol{\mu}_{h} &= \frac{\sum_{k=1}^\mathrm{N} \mathbf{\omega}_k\ \mathbf{p}_k}{ \sum_{k=1}^\mathrm{N} \mathbf{\omega}_k} \label{eq:huber_mean}\\
    \mathbf{\Sigma}_{h}  &=  \frac{\sum_{k=1}^\mathrm{N} \mathbf{\omega}_k\ (\mathbf{p}_k - \boldsymbol{\mu}_{h})(\mathbf{p}_k - \boldsymbol{\mu}_{h})^T}{ \sum_{k=1}^\mathrm{N} \mathbf{\omega}_k} \label{eq:huber_cov}
\end{align}
where $\mathrm{N}$ denotes the number of points in the bin, and $\omega_k$ represents the Huber weight associated with point $\mathbf{p}_k$. 

Applying Huber weighting to estimate Gaussian statistics offers several notable advantages:
\begin{itemize}
    \item The different regime for large residuals mitigates the influence of outliers, preventing them from disproportionately affecting the statistics.
    \item The parameter $\delta$ offers a tunable trade-off between sensitivity to inliers and robustness against outliers.
    \item Huber weighting preserves more information than hard outlier rejection or trimming approaches.
    \item The resulting Gaussian estimates (mean and covariance) more accurately capture the underlying structure of noisy or imperfect data.
\end{itemize}

After computing all Huber-weighted Gaussian distributions, the final Huber-enhanced \gls{gsc} matrix $\mathbf{C}_h$ is built using Eq.~\eqref{eq:gscentry}, where $\boldsymbol{\mu}^z_{i,j}$ and $\Sigma^{zz}_{i,j}$ are replaced by their Huber-weighted counterparts $\boldsymbol{\mu}^z_{h_{i,j}}$ and $\Sigma^{zz}_{h_{i,j}}$ respectively. As shown in Fig.~\ref{fig:sc}, this process yields a denoised matrix that highlights the scan characteristics, thereby improving \gls{lcd} performance.

\section{Robot environmental contextual understanding and interaction with semantics} \label{sec:semantic-method}

In this section, a solution is proposed to augment the geometric mapping capabilities of a mobile robot with semantic high-level understanding. A mobile robot would benefit enormously from this information, as it would expand the knowledge of its surroundings. This additional context can be utilized by the robot to make informed decisions and to interact with and manipulate the environment.

\subsection{Semantic mapping of objects of interest} \label{subsec:artifact-method}

\begin{figure}
    \centering
    \includegraphics[width=\linewidth]{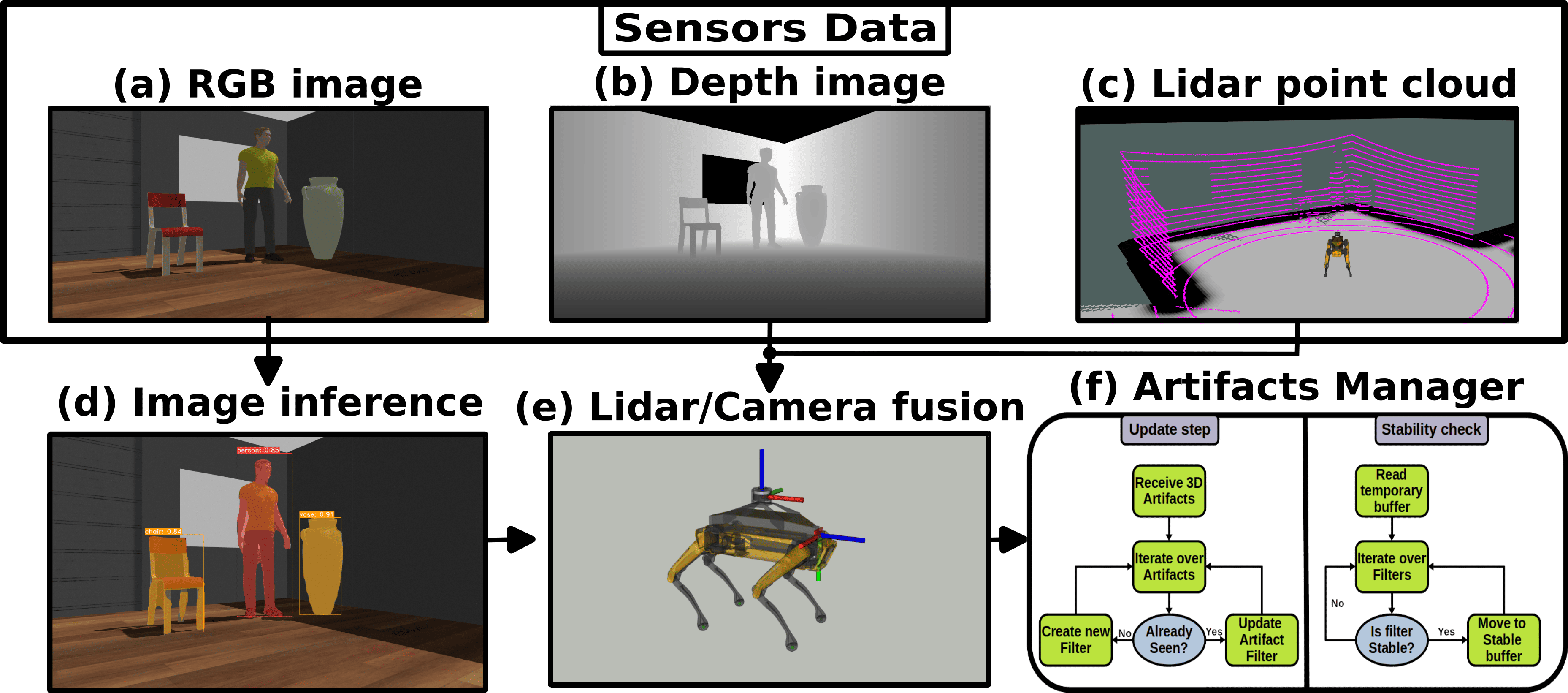}
    \caption{This figure represents the whole Artifacts Mapping pipeline. The top block groups the sensors' data readings: (a) camera \gls{rgb} image, (b) camera depth image, and (c) \gls{lidar} point cloud. At the bottom, there are (d) the \gls{rgb} image inference performed with a Deep Neural Network for instance segmentation, (e) the multi-modal sensor fusion for detection and localization which uses as input the camera depth, the \gls{lidar} point cloud and the Neural Network inference, and (f) a representation of the artifacts manager state-machines used to handle the sensor fusion detections and stabilize them. Reprinted, with permission, from\,\cite{rollo2023artifacts}, \textsuperscript{\textcopyright} 2023 \gls{ieee}.}
    \label{fig:artifacts-pipeline}
\end{figure}

This section presents Artifacts Mapping, a semantic mapping framework designed to map objects of interest in the \gls{3d} space. As shown in the pipeline in Fig.~\ref{fig:artifacts-pipeline}, this framework comprises two main functional blocks: object perception and object management.

\subsubsection{Artifacts detection and position estimation} \label{subsect:perception}

The perception part can be further divided into two components: \textit{(i)} \gls{2d} object segmentation, and \textit{(ii)} \gls{3d} object position estimation using camera-\gls{lidar} filtering.

\paragraph{\textit{\gls{2d} object segmentation}}
In this phase, a deep neural network\,\cite{bolya2020yolact++} is used to infer predefined object classes and their masks from \gls{rgb} images (see Fig.~\ref{fig:artifacts-pipeline}a). During navigation, the robot captures images of the environment using its mounted camera. These images are then processed by an instance segmentation deep neural network, which outputs the classification labels and masks (i.e., a binary image with 1 where the object is found and 0 elsewhere) for each object recognized on the image (see Fig.~\ref{fig:artifacts-pipeline}d).

The outputs are grouped and passed to the next module, which converts these \gls{2d} data into \gls{3d} data. An optional feature provided in this module is the possibility to filter out specific classes in real-time upon request, allowing the robot to map different objects online depending on the current requirements. Other implementation aspects will be further explained in Section~\ref{subsec:artifacts-experiments}.

\paragraph{\textit{\gls{3d} object position estimation using camera-\gls{lidar} filtering}}
This module fuses \gls{rgbd} camera and \gls{lidar} measurements to provide a precise estimate of the objects' positions in the environment. The input is composed of the classification labels and masks found in the previous module, along with depth information extracted from the camera (see Fig.~\ref{fig:artifacts-pipeline}b) and the \gls{lidar} (see Fig.~\ref{fig:artifacts-pipeline}c). Sensors' depth measurements are first analyzed separately in the following discussion.

The depth image obtained from the camera (see Fig.~\ref{fig:artifacts-pipeline}b) is filtered using the recognized object masks through element-wise matrix multiplication. The resulting data, which contains only the depth information of the object plus sensor noise and environment outliers, is used to construct a \gls{3d} point cloud by projecting the \gls{2d} image points into \gls{3d} space using the following camera model equation:
\begin{equation} \label{eq:projection3d}
    \begin{bmatrix}
        ^\mathrm{c}x \\ ^\mathrm{c}y \\ ^\mathrm{c}z 
    \end{bmatrix}
    =
    \begin{bmatrix}
        \frac{1}{f_x} & 0  & -\frac{p_x}{f_x} \\
        0  & \frac{1}{f_y} & -\frac{p_y}{f_y} \\
        0  & 0  & 1
    \end{bmatrix}
    \begin{bmatrix}
        u \\ v \\ 1
    \end{bmatrix}
    \ ^\mathrm{c}z
\end{equation}
where $^\mathrm{c}x$, $^\mathrm{c}y$, $^\mathrm{c}z$ are the \gls{3d} point coordinates with respect to the camera frame; $u$, $v$ are the pixel coordinates on the image plane; and $f_x$, $f_y$, $p_x$, and $p_y$ are the camera intrinsic parameters (focal distances and sensor principal point coordinates). Note that $^\mathrm{c}z$ is the depth value measured by the camera depth sensor.

The obtained point cloud $^\mathrm{c}\mathcal{P}$is first processed using a voxel grid downsampling filter\footnote{Downsampling filter:~\url{https://pointclouds.org/documentation/tutorials/voxel_grid.html}} to reduce the point count. Subsequently, a radius outlier filter\footnote{radius outlier removal:~\url{https://pointclouds.org/documentation/tutorials/remove_outliers.html}} is applied to remove outliers induced by sensor noise and segmentation imperfections. The final filtered point cloud is then used to compute the camera artifact centroid $^\mathrm{c}\boldsymbol{\zeta}$ as the mean of its constituent points.

The \gls{3d} \gls{lidar} centroid estimation is computed as follows. The \gls{3d} \gls{lidar} points $^\mathrm{l}p\in\ ^\mathrm{l}\mathcal{P}$ (see Fig.~\ref{fig:artifacts-pipeline}c) are projected onto the \gls{2d} detected mask images using the extrinsic and intrinsic calibration parameters, as described by Eq.~\eqref{eq:lidar_proj}. The object points of interest are then extracted from the \gls{lidar} point cloud (\textit{i.e.}, the points whose \gls{2d} projections fall within the mask).
\begin{equation} \label{eq:lidar_proj}
    ^\mathrm{l}z
    \begin{bmatrix}
        u \\ v \\ 1
    \end{bmatrix}
    =
    \begin{bmatrix}
        f_x & 0  & p_x \\
        0  & f_y & p_y \\
        0  & 0  & 1
    \end{bmatrix}
    \begin{bmatrix}
        ^\mathrm{c}\mathbf{R}_\mathrm{l} & ^\mathrm{c}\mathbf{t}_\mathrm{l}
    \end{bmatrix}
    \begin{bmatrix}
        ^\mathrm{l}x \\ ^\mathrm{l}y \\ ^\mathrm{l}z \\ 1
    \end{bmatrix}\text{ ,}
\end{equation}
where $^\mathrm{c}\mathbf{R}_\mathrm{l}\in\mathbb{R}^{3\times 3}$ and $^\mathrm{c}\mathbf{t}_\mathrm{l}\in\mathbb{R}^{3\times 1}$ are the rotation matrix and the translation vector between the \gls{lidar} and the camera frames, and $^\mathrm{l}x$, $^\mathrm{l}y$, $^\mathrm{l}z$ are the \gls{3d} coordinates of the \gls{lidar} point $^\mathrm{l}p$. The remaining parameters are the same as those defined in Eq.~\eqref{eq:projection3d}.

The extracted point cloud representing the noisy artifact will then be filtered using a radius-based outlier filter similar to the one used for the camera. Both radius filter parameters are directly dependent on the number of point cloud points, because different distances and object sizes affect point cloud density and, consequently, the filtering efficacy. Finally, the point cloud mean is computed to obtain the \gls{lidar} artifact centroid $^\mathrm{l}\boldsymbol{\zeta}$.

\begin{figure}
    \centering
    \includegraphics[width=\linewidth]{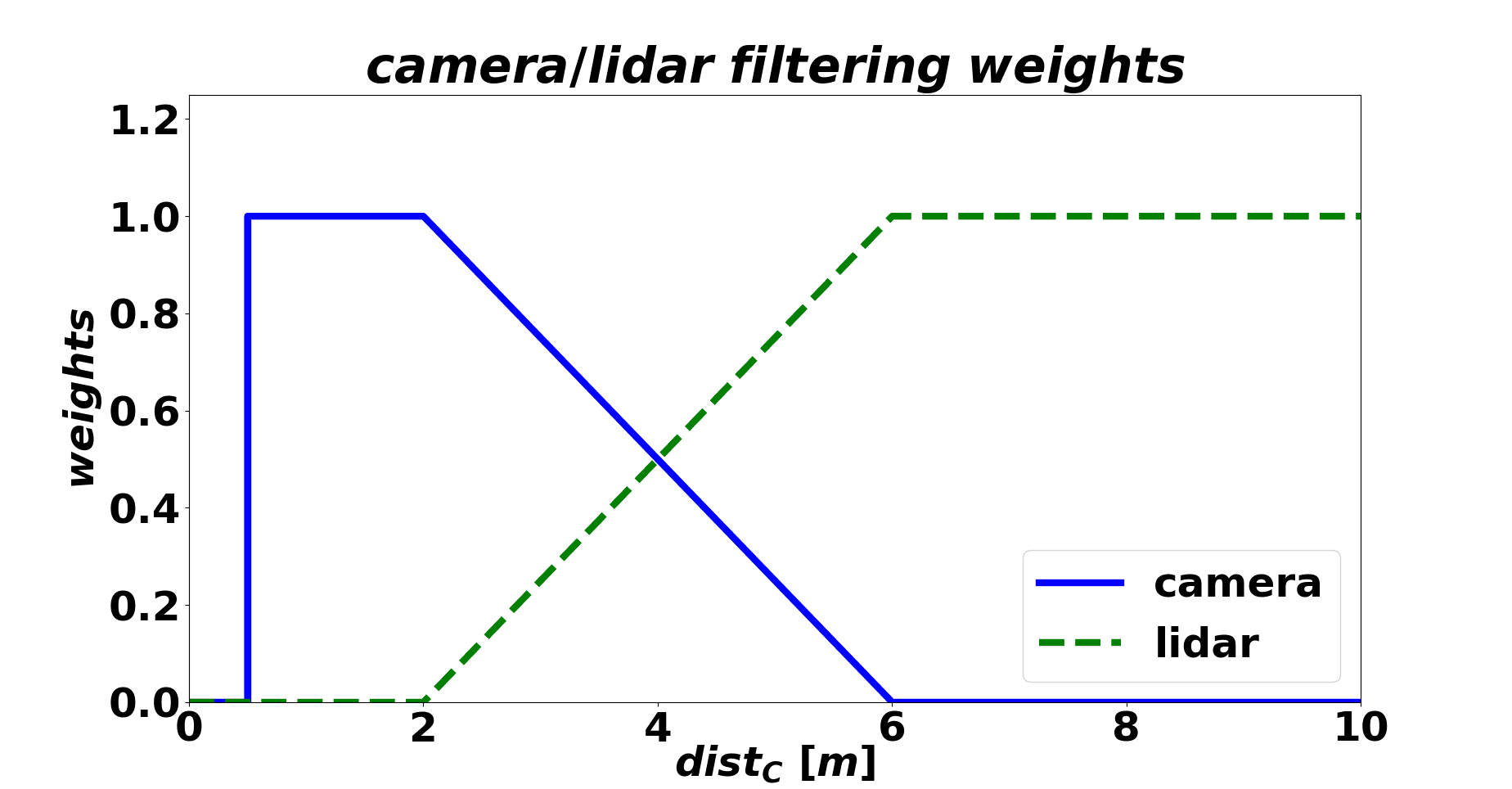}
    \caption{An example of the contribution weights of camera and \gls{lidar} for sensor fusion. The camera weight ($\xi$) is shown in blue, while the \gls{lidar} weight ($1-\xi$) is shown in dashed green. In this specific example, the specifications of a generic \gls{rgbd} camera are considered: $\text{min}_{c}=0.5$, $\text{acc}_{c}=2.0$ and $\text{max}_{c}=6.0$. Reprinted, with permission, from\,\cite{rollo2023artifacts}, \textsuperscript{\textcopyright} 2023 \gls{ieee}.}
    \label{fig:artifacts-weight}
\end{figure}

Once both centroid measurements are available, they are fused into the final artifact centroid $\boldsymbol{\zeta}$ following the piecewise rules defined in the equation:
\begin{equation} \label{eq:filtering}
    \boldsymbol{\zeta}=
    \begin{cases}
        \begin{tabular}{ll}
            $0$ & \text{If $ \text{dist}_{c} < \text{min}_{c} $ } \\
            $^\mathrm{c}\boldsymbol{\zeta}$ & \text{If $ \text{min}_{c} \leq \text{dist}_{c} \leq \text{acc}_{c} $ } \\
            $\xi \cdot\ ^\mathrm{c}\boldsymbol{\zeta} + (1 - \xi) \cdot\ ^\mathrm{l}\boldsymbol{\zeta}$ & \text{If $ \text{acc}_{c} \leq \text{dist}_{c} \leq \text{max}_{c} $ } \\
            $^\mathrm{l}\boldsymbol{\zeta}$ & \text{If $\text{dist}_{c} > \text{max}_{c}$}\text{ ,}
        \end{tabular}
    \end{cases}
\end{equation}
where $\text{dist}_{c}$ is the Euclidean distance between the \gls{3d} point estimates and the camera; $\text{min}_{c}$ and $\text{max}_{c}$ are the minimum and maximum perception distances of the depth camera; $\text{acc}_{c}$ is the distance within which the camera provides sufficiently accurate measurements to be used alone for object localization (a value typically provided by sensor vendors); $^\mathrm{l}\boldsymbol{\zeta}\in \mathbb{R}^3$ and $^\mathrm{c}\boldsymbol{\zeta}\in\mathbb{R}^3$ are the \gls{lidar} and camera \gls{3d} centroid estimates, respectively; and $\xi \in [0, 1] \in \mathbb{R}$ is the fusion weight represented by the blue slope of the segments between $\text{acc}_c$ and $\text{max}_c$ in Fig.~\ref{fig:artifacts-weight}, computed as follows:
\begin{equation} \label{eq:xi}
    \xi = 1-\frac{1}{\text{max}_{c} - \text{acc}_{c}}(\text{dist}_{c}-\text{acc}_{c})
\end{equation}

Using the filtered camera and \gls{lidar} point clouds, a rough \gls{3d} radius estimate $\rho$ for the objects is obtained. The camera radius $^\mathrm{c}\rho$ and the \gls{lidar} radius $^\mathrm{l}\rho$ are computed as the mean of the two largest dimensions along the $x$, $y$, and $z$ point cloud axes. The final radius $\rho$ is computed following the same centroid fusion rules of Eq.~\eqref{eq:filtering}, substituting $\boldsymbol{\zeta}$ with $\rho$, $^\mathrm{c}\boldsymbol{\zeta}$ with $^\mathrm{c}\rho$ and $^\mathrm{l}\boldsymbol{\zeta}$ with $^\mathrm{l}\rho$.

Additionally, the artifact's view angle $\phi$ with respect to the robot is computed. This angle is rotated with respect to the map reference frame for implementation reasons, using the equation:
\begin{equation} \label{eq:artifact_angle}
    \phi = \arctan2(r_{21}, r_{11}) + \arctan2(^ry,\ ^rx)
\end{equation}
where $r_{ij}$ is the entry at row $i$ and column $j$ of the rotation matrix $^\mathrm{m}\mathbf{R}_\mathrm{r}\in\mathbb{R}^{3\times 3}$ between the map frame $\mathrm{m}$ and the robot frame $\mathrm{r}$, and $^\mathrm{r}y$ and $^\mathrm{r}x$ are the $y$ and $x$ positions of the artifact centroid with respect to the robot base. The two additive terms of Eq.~\eqref{eq:artifact_angle} represent, respectively, the heading angle between the robot and the map and the angle between the robot and the \gls{3d} centroid.

\subsubsection{Artifacts manager for data association} \label{subsect:manager}

The manager (see Fig.~\ref{fig:artifacts-pipeline}f) is necessary to filter outliers and stabilize the artifact position estimations provided by the sensor fusion module. This entire procedure is commonly known as data association\,\cite{achour2022collaborative, xiang2017rnn}. The manager is composed of two asynchronously running, parallel modules: \textit{(i)} object position filtering and \textit{(ii)} object position stabilization.

\paragraph{\textit{Position filtering}}

The perceived artifacts are temporarily stored in a dedicated data structure called the \textit{temporary buffer}. Once the manager receives the \gls{3d} artifacts' position estimations from the perception module, it checks if the artifacts were previously observed, \textit{i.e.}, whether the distance between one of the stored artifacts and the current measurement is less than the stored artifact's \gls{3d} radius. If an artifact has been seen before, the corresponding entry in the temporary buffer is updated. Otherwise, for each newly observed artifact, the manager creates a new artifact instance in the temporary buffer. These instances have their own moving-average filter, which estimates the average of the artifact centroid position and its radius using Eq.~\eqref{eq:movingavg} and computes a variance based on the distances between the position measurements and the moving average within the filter horizon using Eq.~\eqref{eq:variance}.
\begin{align}
    \boldsymbol{\mu} &= \frac{1}{\mathrm{N}} \sum_{\boldsymbol{\kappa}\in\mathrm{K}_N} \boldsymbol{\kappa}\label{eq:movingavg}\\
    \sigma &= \frac{1}{\mathrm{N}} \sum_{\boldsymbol{\kappa}\in\mathrm{K}_N} ||\boldsymbol{\kappa} - \boldsymbol{\mu}||^2 \label{eq:variance}\text{ ,}
\end{align}
where $\mathrm{N}\in\mathbb{N}$ is the number of measurements in the moving average set $\mathrm{K}_N$ of \gls{3d} points, $\boldsymbol{\kappa}\in\mathbb{R}^3$ represents the current \gls{3d} position measurement, $\boldsymbol{\mu}\in\mathbb{R}^3$ is the \gls{3d} mean position, and $\sigma\in\mathbb{R}$ represents the variance of the filter.

\paragraph{\textit{Position stabilization}}

This module evaluates the stability of the artifacts in the temporary buffer and stores stable artifacts in a secondary, similar structure, the \textit{stable buffer}. If an artifact in the temporary buffer is determined to be stable, the stabilizer transfers the artifact from the temporary buffer to the stable one. An artifact is considered stable when its moving average filter variance $\sigma$ is less than half its \gls{3d} artifact radius $\rho$, and at least half the average filter set $\mathrm{K}_N$ is filled. This condition ensures there are enough consistent object position estimates, allowing the average object position to be used to fix the object position on the map.

At the end of the Artifacts Mapping application, an additional final data association step is performed. Artifacts of the same class that exhibit overlap on the $XY$ plane are merged into a single artifact. This step reduces the number of duplicated objects that can appear on the map due to different \gls{fov} and occlusions. Following this process, the stable artifacts buffer is saved in a YAML file, which can then be loaded into the user interface application presented in the next section.

\paragraph{\textit{User Interface for goal sending}} \label{subsect:application}

\begin{figure}[t]
    \centering
    \includegraphics[width=\linewidth]{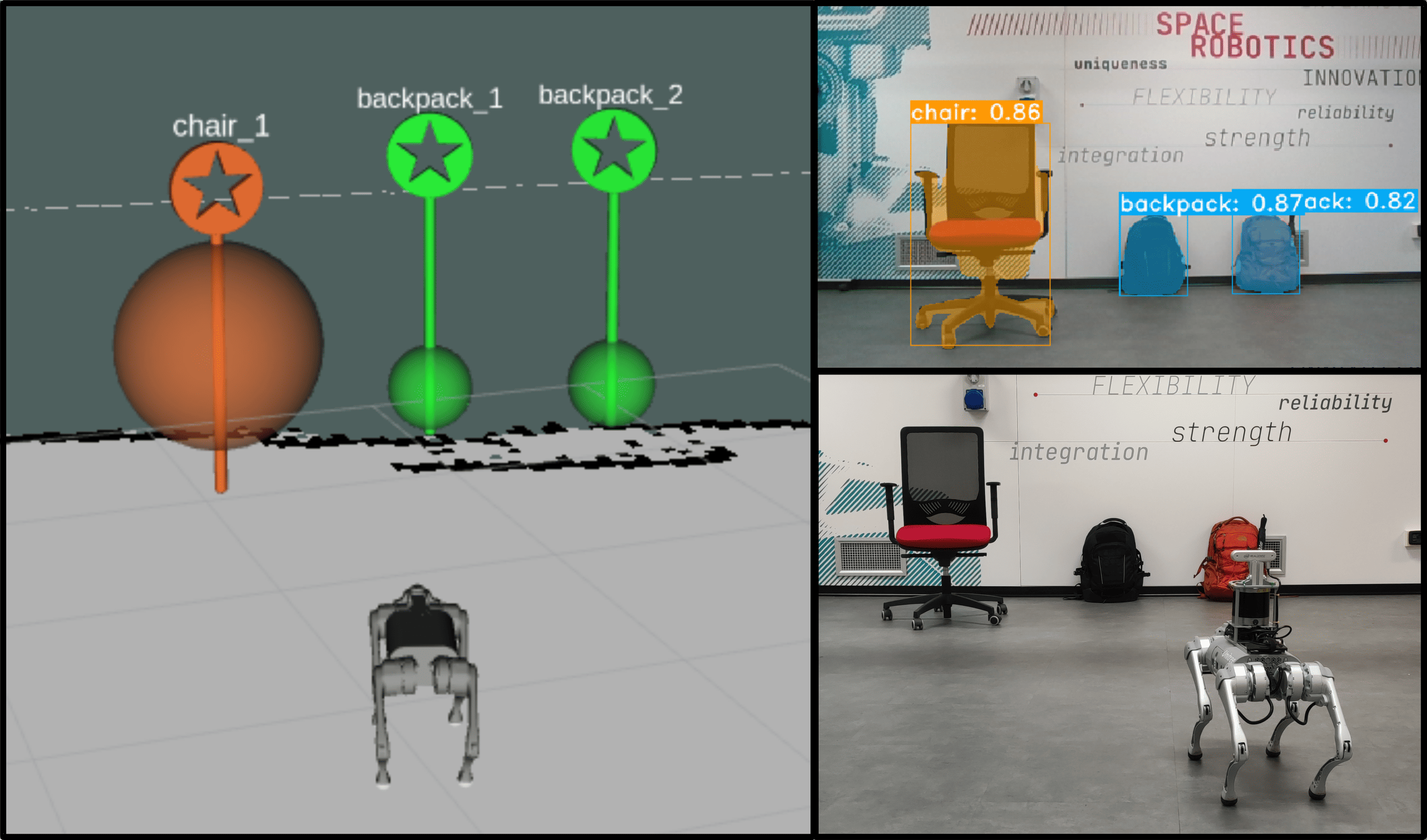}
    \caption{An example of the framework during an experiment. On the left is the visual application where objects are shown with a landmark and a spherical region of interest for the location in the \gls{rviz}. On the top right is the instance segmentation inference of the image taken from the robot camera, while on the bottom right is the external representation of the experimental scene. Reprinted, with permission, from\,\cite{rollo2023artifacts}, \textsuperscript{\textcopyright} 2023 \gls{ieee}.}
    \label{fig:artifact_example}
\end{figure}

A \gls{ui} application based on a \gls{rviz} plugin (see Fig.~\ref{fig:artifact_example}) was developed to provide intuitive visualization of the artifacts on the map, to send commands to the robot for moving near an artifact of interest, and to delete unnecessary or incorrectly identified artifacts. These artifacts can be loaded from the YAML file obtained using the Artifacts Mapping application. Through the \gls{ui} application, the user can send \emph{nav\_msgs/goal} \gls{ros} messages, which the robot can utilize to navigate towards the object (\textit{e.g.}, using the \gls{ros} navigation stack, see Section~\ref{sec:semantic-exp}). The user can interact with the artifacts in \gls{rviz} by right-clicking them and selecting the \textit{Go To} or \textit{Delete} action. Since the artifact's centroid lies within its bounding shape, the navigation goal is automatically offset to a position in front of the artifact, ensuring the robot stops before colliding with the object.

The other available option is artifact deletion. Suppose the user notices that an artifact is wrongly identified (in terms of classification or position). The user can delete it, and once the \gls{ui} application is closed, the loaded YAML file is updated with only the remaining artifacts.

\subsubsection{Example of semantic mapping usage in loco-manipulation applications} \label{subsec:locoman-method}

\begin{figure}
    \centering
    \includegraphics[width=\linewidth]{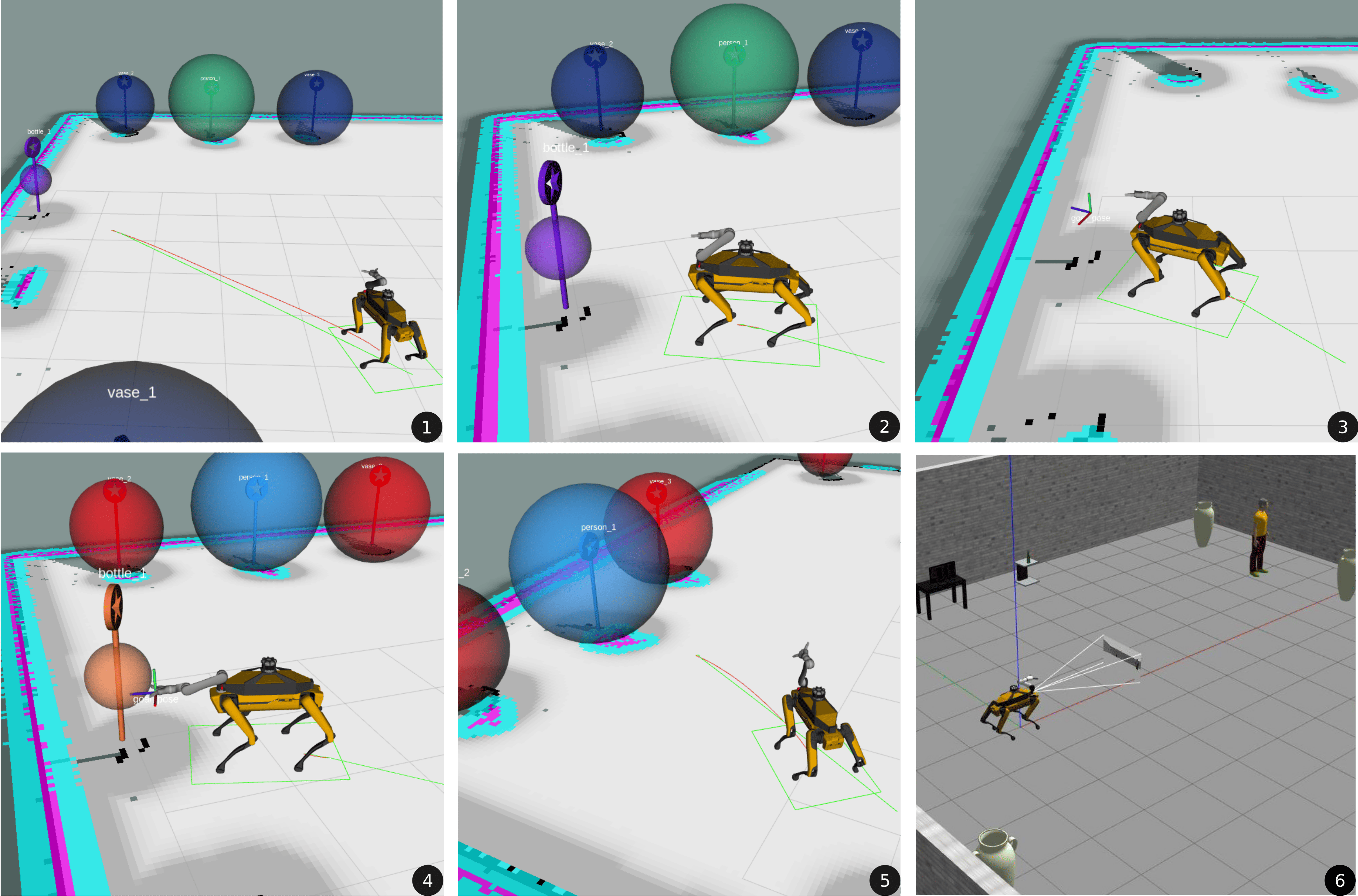}
    \caption{Autonomous actions held during the execution of the application. Image 1 represents the navigation towards the requested object. In image 2, the object is approached, while in image 3, the object pose is estimated. In images 4 and 5, the object is picked and brought to the person. Image 6 shows the whole simulated world. Reprinted, with permission, from\,\cite{rollo2023semantic}, \textsuperscript{\textcopyright} 2023 Springer Nature.}
    \label{fig:pipeline}
\end{figure}

In this section, an overview is provided of how Artifacts Mapping can be employed in a simple pick-and-place loco-manipulation task.

To manage the application workflow, a \gls{bt}\footnote{\gls{bt} library: \url{https://py-trees.readthedocs.io/en/devel/}} is utilized.
The \gls{bt} is composed of a sequence of six actions: \textit{navigate\_to\_object}, \textit{approach\_object}, \textit{pose\_estimation}, \textit{pick\_object}, \textit{bring\_to\_user}, and \textit{release\_object}, along with a safety action \textit{abort} which is activated when any of the other actions fails. Most of these actions are represented in Fig.~\ref{fig:pipeline}. The \gls{bt} waits until the user initiates a request, which is defined by the string associated with the artifact, e.g., "bottle\_1". After the request is received, the \gls{bt} starts ticking and executes the behaviors explained hereafter.

\paragraph{\textit{navigate\_to\_object}}
The rough positions of both the requested object and the target person are already known, having been previously documented during the mapping process\,\cite{rollo2023artifacts}. This action involves retrieving the object's stored position and sending a navigation goal to the robot. The goal is offset to an obstacle-free region located immediately in front of the required object, prompting the robot to calculate its path and navigate toward the location.

\paragraph{\textit{approach\_object}}
The object approaching phase is necessary to facilitate a smooth picking action by refining the robot's position relative to the artifact. This refinement is required because the goal position received in the previous step is imprecise. During this phase, the robot autonomously approaches the artifact, striving to center it in the camera's field of view and reduce the distance such that the object falls within the robot's manipulability space.

To accomplish this, the robot employs an instance segmentation network to segment the object and uses the resulting mask to crop both the \gls{rgbd} images. Subsequently, a point cloud is built using the masked depth image and the camera's intrinsic parameters. For each non-zero pixel of the masked depth, the point $^\mathrm{b}\mathbf{p}$ in the robot base frame $\mathrm{b}$ is computed as $^\mathrm{b}\mathbf{p} =\ ^\mathrm{b}\mathbf{t}_c + ^\mathrm{c}\mathbf{p}$, where $^\mathrm{b}\mathbf{t}_c$ is the translation vector from the base frame $\mathrm{b}$ to the camera frame $\mathrm{c}$. The point $^\mathrm{c}\mathbf{p}$ in the camera frame $\mathrm{c}$ is calculated like in Eq~\ref{eq:projection3d}:
\begin{equation} \label{eq:camera_pcd}
    ^\mathrm{c}\mathbf{p}
    =    
    \begin{bmatrix}
        \frac{1}{f_x} & 0  & -\frac{p_x}{f_x} \\
        0  & \frac{1}{f_y} & -\frac{p_y}{f_y} \\
        0  & 0  & 1 
    \end{bmatrix}
    \begin{bmatrix}
        u \\ v \\ 1 
    \end{bmatrix}
    \ ^\mathrm{c}d
\end{equation}
where $f_x$, $f_y$, $p_x$, and $p_y$ are the camera intrinsic parameters (\textit{i.e.}, the focal lengths and principal points along the $x$ and $y$ image axes), $u$ and $v$ are the depth pixel position along the $x$ and $y$ axes, and $^\mathrm{c}d$ is the depth value of the considered pixel.

Using this point cloud, the object's centroid $^\mathrm{b}\boldsymbol{\zeta}$ expressed in the base frame $\mathrm{b}$, is computed. Two key metrics are derived from this centroid: the \gls{3d} Euclidean distance to the robot base, $d_{obj}$, and the heading angle between the robot and the object, $\theta_{obj}$, calculated by:
\begin{equation} \label{eq:h_angle}
    \theta_{obj} = \arctan2(y, x)
\end{equation}
where $x$ and $y$ are the coordinates of the centroid $^\mathrm{b}\boldsymbol{\zeta}$ (with the $xy$ plane assumed parallel to the ground floor), and $\arctan2$ is the two-argument arc tangent function that accounts for the quadrant.

Two linear proportional controllers (\textit{e.g.}, Eq.~\eqref{eq:controller}) are then used to regulate the robot base position by sending velocity commands (linear velocity along the $x$-axis, $v_x$, and angular velocity around the $z$-axis, $\omega_z$). The controllers use optimal distance $d^*$ and optimal angle $\theta^*$ as references:
\begin{equation} \label{eq:controller}
    u = P (p - p^*)\text{ ,}
\end{equation}
where $u$ is the linear $v_x$ or angular $\omega_z$ velocity output, $P$ is the proportional gain, $p$ is the current distance $d_{obj}$ or the current heading angle $\theta_{obj}$, and $p^*$ is the optimal distance $d^*$ or angle reference $\theta^*$.

\paragraph{\textit{pose\_estimation}}
Once the robot is well positioned, the homogeneous transformation $^\mathrm{b}\mathbf{H}_\mathrm{g}$ of the grasping pose $\mathrm{g}$ expressed in the base frame $\mathrm{b}$ is computed. Using the instance segmentation neural network and the \gls{rgbd} camera, the translation vector $^\mathrm{b}\mathbf{\mathbf{t}}_\mathrm{g}$ from base frame $\mathrm{b}$ to the grasping pose $\mathrm{g}$ is computed in the same way as $^\mathrm{b}\boldsymbol{\zeta}$ in the \textit{approach\_object} behavior. The rotation matrix $^\mathrm{b}\mathbf{R}_\mathrm{g}$ between the robot base $\mathrm{b}$ and the grasping pose $\mathrm{g}$ is simply set equal to an identity matrix $\mathbf{I_{3}}$ ($^\mathrm{b}\mathbf{R}_\mathrm{g} = \mathbf{I_{3}}$). This yields the homogeneous transformation:
\begin{equation}
    ^\mathrm{b}\mathbf{H}_\mathrm{g}=
    \begin{bmatrix}
        ^\mathrm{b}\mathbf{R}_\mathrm{g} & ^\mathrm{b}\mathbf{t}_\mathrm{g} \\
        0\ 0\ 0 & 1
    \end{bmatrix}
\end{equation}

\paragraph{\textit{pick\_object}}
In this phase, the robot receives the desired grasping homogeneous transformation $^\mathrm{b}\mathbf{H}_\mathrm{g}$ and roto-translates it into the end-effector frame $e$ to determine the final end-effector pose for the action:
\begin{equation}
    ^\mathrm{e}\mathbf{H}_\mathrm{g} =\ ^\mathrm{e}\mathbf{H}_\mathrm{b}\ ^\mathrm{b}\mathbf{H}_\mathrm{g}
\end{equation}
In this experiment, the simulation uses an optimization-based whole-body inverse dynamics controller\,\cite{raiola2022wolf} to move the arm towards the target pose. However, the specific method for achieving this target pose depends entirely on the robot configuration used. When the robotic end-effector reaches the picking position, the gripper closes to grasp the artifact, and the arm subsequently moves to a folded position for safe transportation.

\paragraph{\textit{bring\_to\_user}}
Once the artifact is successfully grasped, the robot navigates through the environment to deliver it to the user.

\paragraph{\textit{release\_object}}
Finally, when the robot reaches the user's position, it releases the requested object to the user. After this step, the pipeline is completed, and the robot becomes available to receive other tasks.

\paragraph{\textit{abort}}
The abort status is essential for managing unexpected behaviors or failures. In this state, all currently running goals or requests are immediately halted, the mission is deleted, and the robot enters an idle state, promptly reporting the error to the user while awaiting further commands.

\chapter{Experiments} \label{chap:experiments}
    \section{Re-identification validations} \label{sec:re-id-exp}

In this section, the proposed evaluation methodology is presented to formally validate the solutions detailed in Section~\ref{sec:re-id-method} for the \gls{reid} problem stated in Section~\ref{subsec:reid-prob}.

\subsection{Re-ID baseline evaluation} \label{subsec:followme_eval}

The evaluation begins with the first \gls{reid} pipeline presented in Section~\ref{subsec:followme-method}, FollowMe\,\cite{rollo2023followme}. Three different experiments are conducted to validate the system: individual modules (\textit{i.e.}, hand gesture classification, visual \gls{reid} and the integrated system framework.

\subsubsection{Experimental setup} \label{subsec:reid_setup}

The experimental setup utilizes an \textit{Intel\textregistered\,RealSense™ D415}\footnote{\textit{Intel\textregistered\,RealSense™ D415}: \url{https://www.intelrealsense.com/depth-camera-d415/}} camera, a notebook equipped with an \textit{Intel\textregistered\,Core™ i9-11950H} processor and an \textit{NVIDIA Geforce RTX 3080 Laptop} \gls{gpu}, and a Robotnik RB-Kairos+ 5e\footnote{RB-Kairos+ Mobile Manipulator: \url{https://robotnik.eu/products/mobile-manipulators/rb-kairos/\#more-versions}} serving as the assistant robot. 

The camera was fixed onto the UR5e robotic arm wrist using a \gls{3d} printed adapter, and the robotic arm was positioned vertically to point the camera forward (see Fig.~\ref{fig:followme_setup}). The safety circle radius is set to $1.25\,\text{m}$, determined by applying Eq.~\eqref{eq:d_safe} with a maximum velocity $v_{\max} = 0.3\,\text{m/s}$ and an expected response time $t_{\exp} = 3\,\text{s}$. The navigation relies on an Adaptive Monte Carlo Localization algorithm\,\cite{xiaoyu2018adaptive} to estimate the robot's position using odometry and laser scan data.

Regarding the \gls{reid} experiments, a pre-trained IBN-ResNet-50 network\,\cite{pan2018two} on the popular MSMT17 dataset\,\cite{wei2018person} is employed. The feature dimension $D$ of the network is set to $256$. This dimension is a popular choice in \gls{reid} settings as it represents a good trade-off between performance and feature vector length.

\begin{center}
    \begin{table}[h]
        \centering
        \caption{Classification metrics over a test set of images. on the left, hand gestures, and on the right, \gls{reid}. Reprinted, with permission, from\,\cite{rollo2023followme}, \textsuperscript{\textcopyright} 2023 \gls{ieee}.}
        \begin{tabular}{|c|ccc||cc|}
            \hline
                    & \multicolumn{3}{c||}{Hand Gestures}  & \multicolumn{2}{c|}{\gls{reid}} \\
            \hline
                    & \textit{Wait} & \textit{Follow}    & \textit{Others} & \textit{Target} & \textit{No Target} \\
            \hline
            \textbf{Precision}    & 0.98      & 0.99      & 0.96  & 0.96       & 0.91   \\
            \textbf{Recall}   & 0.99      & 0.96      & 0.97   & 0.91      & 0.96 \\
            \textbf{F1-score}  & 0.98      & 0.97      & 0.96 &  0.93      & 0.93  \\
           \hline\hline
             \textbf{Accuracy} & \multicolumn{3}{c||}{0.97}  & \multicolumn{2}{c|}{0.94}\\
             \hline
        \end{tabular}
        \label{tab:follow_gesture_metrics}
    \end{table}
\end{center}

\subsubsection{Hand gesture classification experiments} \label{subsec:follow_gestureexperiment}

\begin{figure}[ht]
    \centering
     \begin{subfigure}{0.49\linewidth}
         \centering
         \includegraphics[width=\linewidth, trim={0cm 0cm 0cm 0cm}]{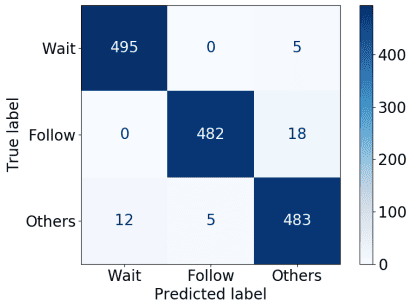}
         \caption{Average confusion matrix for hand gesture over a test set of images.}
         \label{fig:handconfusionmatrix}
     \end{subfigure}
     \begin{subfigure}{0.49\linewidth}
         \centering
         \includegraphics[width=\linewidth, trim={0cm 0cm 0cm 0cm}]{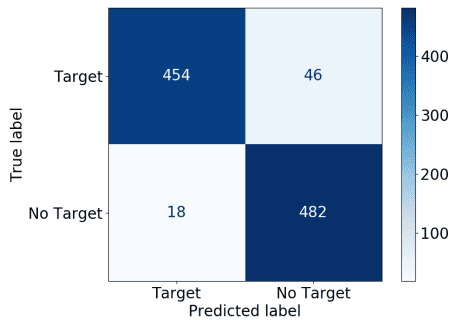}
         \caption{Confusion matrix for \gls{reid} over a test set of images.}
         \label{fig:reident_matrix}
     \end{subfigure}
    \caption{Comparison between panoptic and instance segmentation inferences. Reprinted, with permission, from\,\cite{rollo2023followme}, \textsuperscript{\textcopyright} 2023 \gls{ieee}.}
    \label{fig:panopticsegm}
\end{figure}

To train the \gls{svm} classifier, an initial dataset of $400$ hand images in different positions was collected. For the experimental validation of this module, a standard classification validation was used involving a group of eight subjects, for each subject and each class (i.e., \textit{Wait}, \textit{Follow}, \textit{Others}), hand key-points were extracted from $500$ images, resulting in a total of $4000$ images per class ($12000$ images total). These data were classified using the trained \gls{svm}. Using the results and the ground truth labels, the mean confusion matrix (averaged over the subjects) and various classification metrics were computed and are presented in Fig.~\ref{fig:handconfusionmatrix} and the left part of Table \ref{tab:follow_gesture_metrics}, respectively. The resulting metrics confirm the algorithm's strong ability to distinguish between the intended gestures across a diverse group of actors.

\subsubsection{Re-identification experiments} \label{subsec:follow_reidentexperiment}

To validate the \gls{reid} module, a dataset comprising $8500$ images was collected. The calibration phase required $500$ images with a single subject for each person ($500 \times 8 = 4000$ images). The remaining $4500$ images were collected according to the following scheme: $500$ images with all subjects present, and $500$ images for each person, with all subjects except the person of interest included.

The person \gls{reid} module was calibrated using the $500$ calibration images, of which two-thirds were used for feature extraction and one-third for threshold computation. For the testing phase, $1000$ images per subject were used, divided into two sets: the set where all people were present (Set A) and the set where all people except the calibrated one were present (Set B). The classification was evaluated as follows:
\begin{itemize}
    \item \textit{Set A} (Target Present): Classification is correct if and only if the module correctly re-identifies the target, \textit{i.e.}, it does not re-identify another person or no one.
    \item \textit{Set B} (Target Absent): Classification is considered correct if and only if the module does not re-identify anyone, \textit{i.e.}, all feature distances are above the pre-computed threshold.
\end{itemize}
The results of this process are represented in the confusion matrix in Fig.~\ref{fig:reident_matrix} and the right part of Table \ref{tab:follow_gesture_metrics}. The numbers are averaged per subject: one subject is used for calibration, and the others are used as distractors, alternating between them and then averaging the final results. The numerical results confirm that the chosen threshold computation favors False Negatives (top-right value in Fig.~\ref{fig:reident_matrix}) while significantly penalizing False Positives (bottom-left value in Fig.~\ref{fig:reident_matrix}), which is the desired conservative behavior for reliable \gls{reid}. This is also evident from the \textit{Target} column of Table \ref{tab:follow_gesture_metrics}, where Precision exceeds Recall.

\subsubsection{FollowMe whole experiments} \label{subsec:follow_wholeexperiment}

\begin{figure}[h]
    \centering
    \includegraphics[width=0.8\linewidth, trim={0cm 0cm 0cm 0cm}]{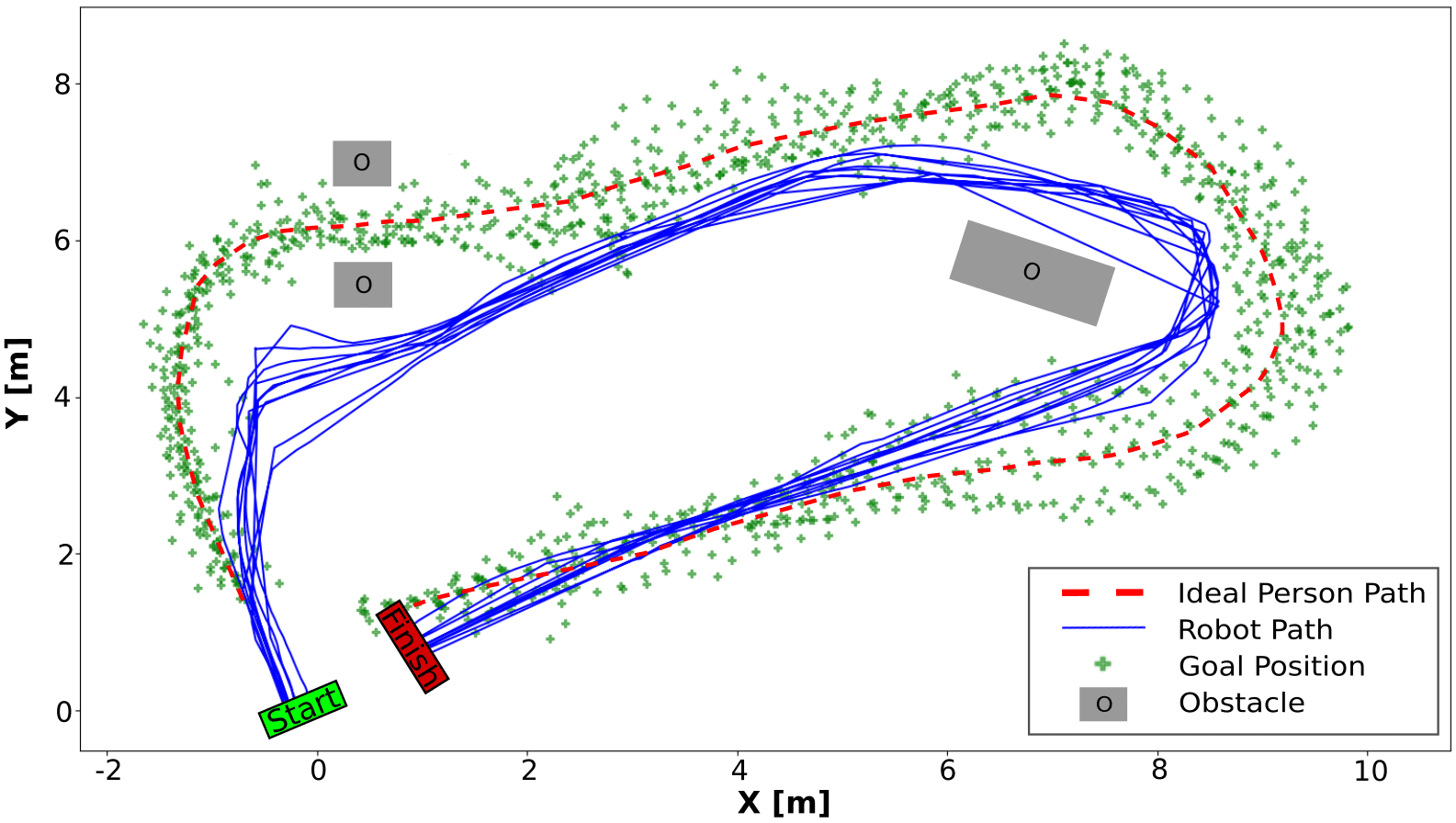}
    \caption{Results of the FollowMe whole experiment: each person has to follow an ideal path (red dashed line) while the robot (blue line) has to follow him/her. The goal positions computed from the perception module data are represented with green plus signs. The robot is placed in the green start position and has to follow the target until the red finish position is reached. Reprinted, with permission, from\,\cite{rollo2023followme}, \textsuperscript{\textcopyright} 2023 \gls{ieee}.}
    \vspace{-0.5cm}
    \label{fig:path_chart}
\end{figure} 

The complete FollowMe framework was validated with a group of ten subjects. The robot starts from a fixed position in a laboratory area (approximately $100\,\text{m}^2$) where its empty map is known and some random unknown obstacles are placed. Each person was instructed to follow a predefined path while other subjects moved around (generally, no more than four people were present along the route). At the start, the robot waits for a \textit{Follow} gesture command. When the target reaches the final station, the robot must be stopped using the \textit{Wait} hand gesture. The experiment was performed once for each person. The qualitative results are shown in Fig.~\ref{fig:path_chart}. The robot (blue path) successfully detects the person's position (green $+$). It follows the subject, avoiding unknown obstacles and maintaining a path pattern similar to the target's ideal path (dashed red line).

The time performance of the overall proposed framework was computed to estimate its time-frequency. The processing frequency, from camera acquisition to robot goal sending, ranges from $10\,\text{Hz}$ to $7\,\text{Hz}$ depending on the number of people in the image (1 to 10). This system runs online, with noticeable speed reductions as the number of detected people increases. However, in a typical real-world application scenario, it is rare to detect more than 10 non-occluded people.

\subsection{Autonomous adaptation evaluation} \label{subsec:carpe-eval}

In this section, a demonstration is provided of how the \gls{carpe} method presented in Section~\ref{subsec:carpe-method} improves upon the FollowMe baseline approach described in Section~\ref{subsec:followme-method}.

To test the \gls{carpe} framework, two experiments were conducted. Following a similar approach to the baseline evaluation in Section~\ref{subsec:followme_eval}, the framework is first validated by capturing videos from different angles using a fixed camera in a laboratory setup. Secondly, to further validate the proposed method, it is tested in a \gls{hri} scenario, specifically a robot person-following scenario.

Similarly to the experimental setup in Section~\ref{subsec:reid_setup}, the system was run on a notebook with an \textit{Intel\textregistered\,Core™ i9-11950H} processor and an \textit{NVIDIA Geforce RTX 3080 Laptop} \gls{gpu}. For image acquisition, the \textit{Intel\textregistered\,RealSense™ D455} camera and a Robotnik RB-Kairos+ 5e are used as the assistant robot. To extract features (a one-column vector of dimension $D=256$), the same IBN-ResNet-50\,\cite{pan2018two} network trained on the MSMT17 dataset\,\cite{wei2018person} is utilized.

\subsubsection{Individual experiments} \label{subsec:carpe_individual_exp}

The PersonPath22 dataset\,\cite{personpath22}, a visual person-tracking dataset, was initially considered for evaluating this framework. However, this dataset was found to be unsuitable for the following reasons:
\begin{itemize}
    \item The videos do not depict \gls{hrc} scenarios. Instead, they mainly comprise security camera footage or videos of crowded environments where people are far from the camera and appear in the image for only short periods.
    \item The videos are too brief to accurately validate the \gls{reid} module's ability to handle identity changes over time.
    \item In most cases, the \gls{mot} algorithm used with this dataset already performed the correct tracking. People do not frequently enter and exit the camera's \gls{fov}, so the \gls{reid} module is unnecessary.
\end{itemize}
To address these limitations in validating the framework, a custom dataset was collected, comprising $18$ videos totaling $53$ minutes of footage. Additionally, for videos featuring multiple actors, the framework was evaluated separately for each person, bringing the total analyzed duration to $113$ minutes. The videos were shot in two laboratory setups, featuring both single-person and group scenarios. Participants were asked to exit and re-enter the camera's \gls{fov} and to change their appearance (e.g., by varying their clothes) to rigorously validate the framework's ability to re-identify individuals who are totally or partially occluded and to adapt the model based on newly acquired appearances.

\begin{figure}[h]
    \begin{subfigure}[b]{0.49\textwidth}
        \includegraphics[width=\linewidth]{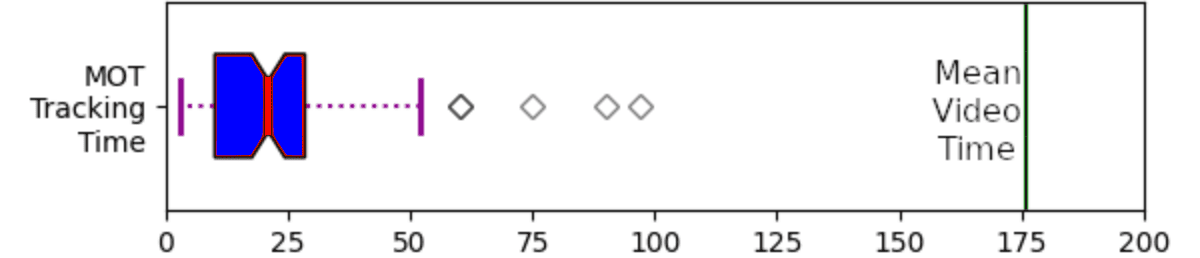}
        \caption{\gls{mot} tracking time (in seconds).}
        \label{fig:time_plot_mot}
    \end{subfigure}
    \begin{subfigure}[b]{0.49\textwidth}
        \includegraphics[width=\linewidth]{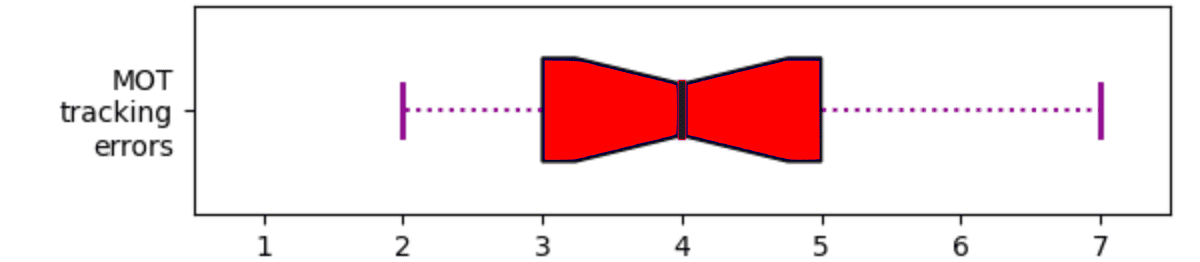}
        \caption{\gls{mot} errors.}
        \label{fig:error_plot_mot}
    \end{subfigure}\\
    \vfill
    \begin{subfigure}[b]{0.49\textwidth}
        \includegraphics[width=\linewidth]{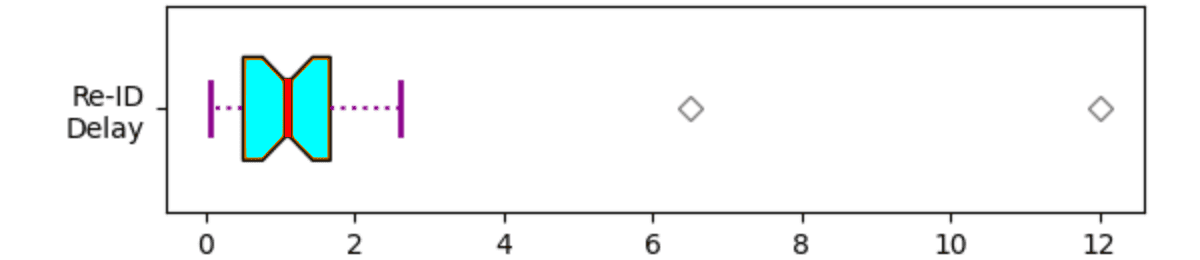}
        \caption{The delay (in seconds) for \gls{reid}.}
        \label{fig:time_plot_reid}
    \end{subfigure}
    \begin{subfigure}[b]{0.49\textwidth}
        \includegraphics[width=\linewidth]{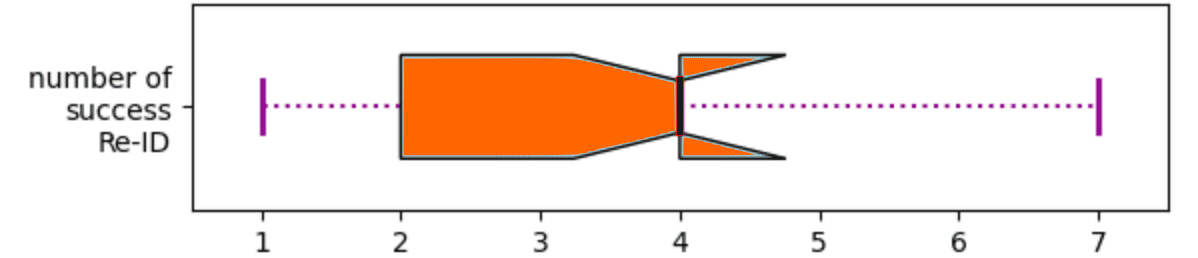}
        \caption{The number of \gls{reid}.}
        \label{fig:error_plot_reid}
    \end{subfigure}
    \caption{Individual experiments statistics. Reprinted, with permission, from\,\cite{rollo2024carpe}, \textsuperscript{\textcopyright} 2024 \gls{ieee}.}
    \label{fig:plots}
\end{figure}

The experiments were analyzed, and the following performance statistics were computed:
\begin{itemize}
    \item The \gls{sota} \gls{mot} tracker exhibited minimum, mean, and maximum tracking lengths of $3.02$, $21.2$, and $52.2$ seconds, respectively, with some outliers extending the maximum to $97.17$ seconds (see Fig.~\ref{fig:time_plot_mot}).
    \item The minimum, mean, and maximum \gls{reid} delay (the time taken by the framework to re-identify the target) were found to be $0.06$, $1.1$, and $2.6$ seconds, respectively (as shown in Fig.~\ref{fig:time_plot_reid}). However, there were two rare instances where the tracker took significantly longer, $6.5$ and $12.1$ seconds, for the \gls{reid} process.
    \item The \gls{mot} algorithm had a minimum and maximum failure rate of $2$ and $7$ times, respectively, with an average failure rate of $4$ times for each video, as shown in Fig.~\ref{fig:error_plot_mot}. 
    \item The overall presented framework, across all videos, made only $2$ errors in re-identifying the target. This low error count demonstrates the framework's reliability for tracking targets that are totally or partially occluded.
    \item The minimum, mean, and maximum number of successful \gls{reid}s made by the proposed framework were $1$, $4$, and $7$, respectively (See Fig.~\ref{fig:error_plot_reid}).
\end{itemize}

\textbf{Following task experiments} \label{subsec:carpe-whole-exp}

\begin{figure}[h]
    \centering
    \includegraphics[width=\linewidth]{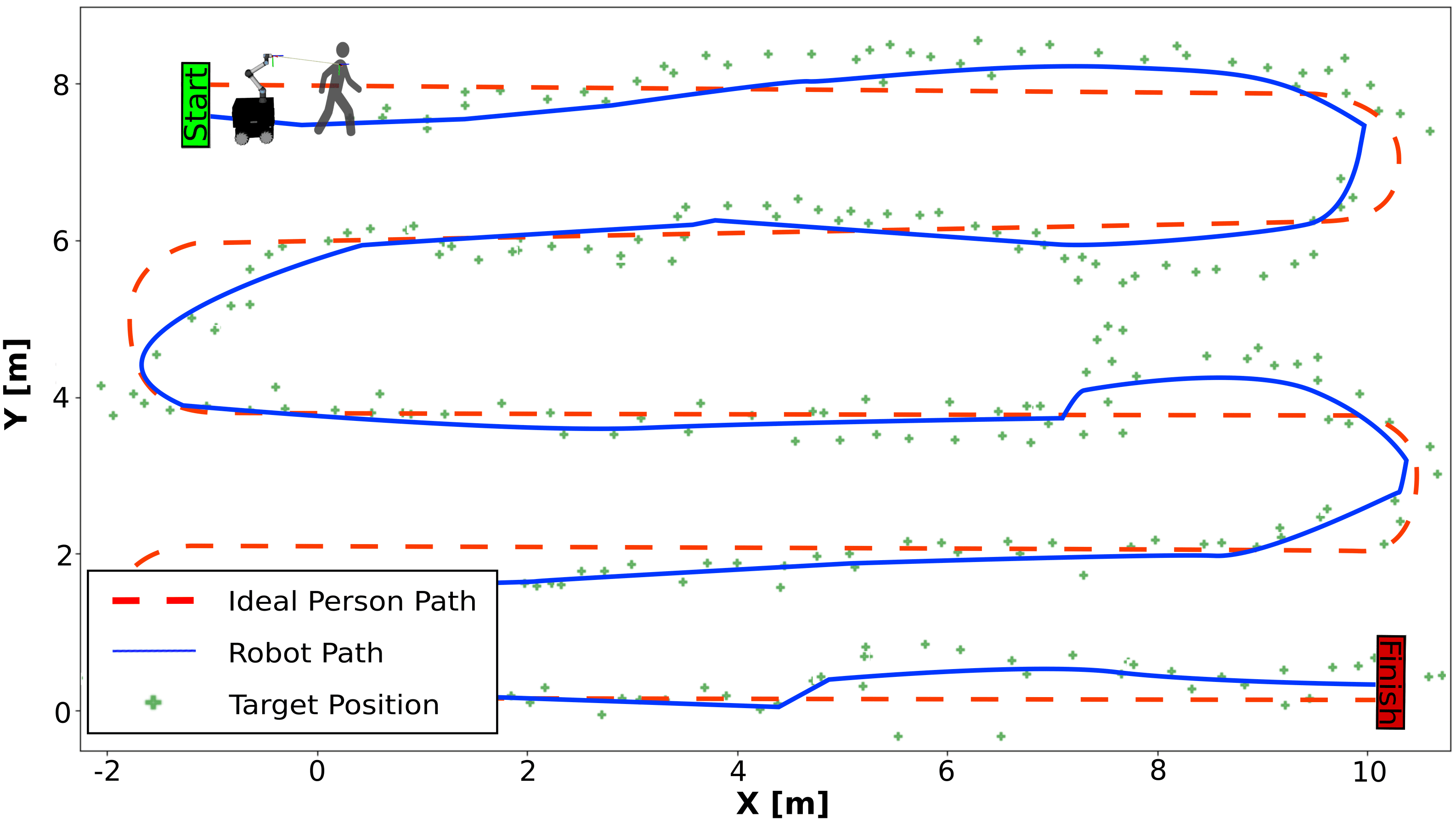}
    \caption{This is an example of a person-following experiment. The person being followed must roughly follow a particular path (indicated by the red dashed line), while the robot (indicated by the blue line) follows them. The green plus signs represent the target positions, which are calculated using the \gls{carpe} framework tracking output. The robot starts at the green position and needs to follow the person until they reach the red finish position. Reprinted, with permission, from\,\cite{rollo2024carpe}, \textsuperscript{\textcopyright} 2024 \gls{ieee}.}
    \label{fig:carpe_follow}
\end{figure}

Real-world experiments were conducted to validate the proposed framework using a robot-person-following scenario similar to the one described in Section~\ref{subsec:follow_wholeexperiment}. In this setup, the target person was asked to move around while the robot followed them, avoiding obstacles. However, in this experiment, the targets actively changed appearance over time and navigated through barriers to test the framework's robustness to partial or complete occlusions.

The framework successfully tracked, re-identified, and followed the target person in all experiments, even in a laboratory setting where other people were working. An example of the experimental setup and results is represented in Fig.~\ref{fig:carpe_follow}. The robot (blue line) successfully follows the target, who navigated along the ideal path (red dashed line). Ten experiments were performed with five different people as targets, covering a total distance of $837$ meters.

\subsubsection{Discussion} \label{subsect:res_discussion}

The evaluation results confirm the need for personalized approaches for \gls{reid} in \gls{hri} tasks. The required recognition capacities are challenging to achieve because the target appears in the image at different times and spaces and with varying appearances. The results achieved in Section~\ref{subsec:carpe_individual_exp} show that personalized approaches are necessary for re-identifying the target person in \gls{hri} tasks.

A more reliable application framework is required because \gls{sota} and \gls{mot} methods proved limited in these experiments. This statement is supported by the results presented in Fig.~\ref{fig:time_plot_mot}, where the \gls{mot} algorithm exhibits a mean tracking time of $21.2$ seconds, significantly shorter than the mean video time of $176$ seconds (minimum video time $94$ seconds), indicating frequent track losses.

In Fig.~\ref{fig:time_plot_reid}, the time delay between when the target person appears in the camera's \gls{fov} and the time they are re-identified is shown. The mean delay obtained ($1.1\,\text{s}$) is acceptable for \gls{hri} applications. However, there were two cases in which the re-identifier failed to re-identify the target within $6$ seconds. This occurred because the target person changed their appearance outside the camera's \gls{fov} (e.g., by changing clothing), which momentarily confounded the \gls{reid} module and led to a longer delay before successful \gls{reid}.

Fig.~\ref{fig:error_plot_mot} and Fig.~\ref{fig:error_plot_reid} compare the number of times the \gls{mot} algorithm lost track of the target with the number of times the \gls{reid} framework successfully re-identified the target. Both metrics have a mean value of $4$, indicating that the \gls{mot} errors are reliably recovered by the proposed framework. However, the figures also reveal that \gls{mot} mistakes occur more frequently than successful re-identifications. This disparity arises because the \gls{mot} algorithm may quickly switch between track \glspl{id} over a short period, while the \gls{reid} delay takes longer to confirm the true target identity. For example, in one video, the \gls{mot} changed the target's \gls{id} three times in $1.4$ seconds, but the re-identifier could only successfully re-identify the last correct track in real-time.

In the person-following scenario, Fig.~\ref{fig:carpe_follow} demonstrates a practical application of this framework. Despite minor \gls{reid} delays, the robot completed the task of following the target without any human intervention in all experiments.

\subsection{Contrasting catastrofic forgetting evaluation} \label{subsec:colp_eval}

The evaluation procedures for the proposed framework are outlined in the following. The obtained results are presented, followed by the results from the ablation study. Finally, the findings are elucidated using saliency analysis.

\subsubsection{Experimental setup} \label{subsect:colp_exp_set}

Similar to the previous \gls{reid} frameworks presented in Section~\ref{subsec:followme_eval} and Section~\ref{subsec:carpe-eval}, and for the same reasons, the proposed framework is validated on a custom dataset comprising $10$ videos. Its performance is compared with the results obtained using the \gls{carpe} method and a plain \gls{mot} approach. This evaluation focuses on two key aspects: tracking time (see Fig.~\ref{fig:tracking_time}) and the number of successful re-identifications (see Fig.~\ref{fig:reid}). For each video, the experiments are repeated five times for each framework to ensure stable results.

The batch size and training iteration parameters for the network adaptation process are set to $16$ and $18$, respectively, as these values represent the optimal choice determined in the ablation study presented later. These parameters are used for both the \textit{\gls{carpe} statistical model update} and the \textit{early training} process presented in Section~\ref{subsec:colp-method}.

In the video sequences considered, the target person changes their appearance (\textit{e.g.}, wearing different clothing), and crucially, initial appearances are occasionally reused upon returning to the camera's \gls{fov}. This specific condition was intentionally introduced to highlight the challenge of catastrophic forgetting. Videos where such challenging situations do not occur have been intentionally avoided, as the frameworks yield identical results in simpler scenarios.

\paragraph{\textit{Custom Dataset Requirements}}

A custom dataset was created since limitations similar to those described in Section~\ref{subsec:carpe_individual_exp} were encountered with publicly available datasets. Publicly available datasets were not suitable for evaluating the proposed framework due to the following typical characteristics:
\begin{itemize}
    \item They mainly consist of security camera footage recorded in crowded environments where individuals briefly appear in the frame without significant changes in appearance.
    \item The videos are typically short.
    \item Once individuals exit the camera's \gls{fov}, they rarely return to it.
\end{itemize}

Given these characteristics, validating this framework using such datasets would not be beneficial, as \gls{sot} and \gls{mot} algorithms can already handle these simpler scenarios effectively. To demonstrate the true effectiveness of the framework, the videos must require people to repeatedly exit and re-enter the camera's \gls{fov} from different positions. Furthermore, the target's appearance is required to change during tracking to demonstrate two critical aspects: first, the proposed framework's ability to adapt to appearance changes during tracking (robustness), and second, its capacity to autonomously learn and distinguish the target's unique characteristics. In other words, the framework must learn to recognize the target by its distinctive features, thereby actively mitigating catastrophic forgetting. 

\paragraph{\textit{System Configuration}}

Exactly as in the previous two validations (Section~\ref{subsec:followme_eval} and Section~\ref{subsec:carpe-eval}), the framework is tested on a notebook equipped with an \textit{Intel\textregistered\,Core™ i9-11950H} processor and an \textit{NVIDIA GeForce RTX 3080 Laptop} \gls{gpu}. An \textit{Intel\textregistered\,RealSense™ D455} camera is used for image acquisition. To handle feature extraction, an IBN-ResNet-50 network\,\cite{pan2018two} pre-trained on the widely adopted MSMT17 dataset\,\cite{wei2018person}, following the method outlined in Ge et al.\,\cite{ge2020mutual}, is used. This pre-training process establishes the initial network's weights and configures the output feature dimension $D$ to be $256$.

\begin{table}[ht!]
    \centering
    \caption{Mean tracking times obtained with the ablation study. The bold numbers represent the best performance achieved among the four experiments for each batch size and iteration number trial, taking into account variations in statistical model updates and/or early training. Reprinted, with permission, from\,\cite{rollo2025pesonalized}, \textsuperscript{\textcopyright} 2025 \gls{ieee}.}
    \resizebox{\textwidth}{!}{\begin{tabular}{|c|c||c|c||c|c||c|c||c|c||c|c||c|c||c|c||c|c||c|c|}
        \hline
        & model update & False & True & False & True & False & True & False & True & False & True & False & True & False & True & False & True & False & True \\
        \hline
        \makecell{Early \\ train} & \makecell{train iter $\rightarrow$ \\ batch size $\downarrow$}  & \multicolumn{2}{c||}{8} & \multicolumn{2}{c||}{10} & \multicolumn{2}{c||}{12} & \multicolumn{2}{c||}{14} & \multicolumn{2}{c||}{16} & \multicolumn{2}{c||}{18} & \multicolumn{2}{c||}{20} & \multicolumn{2}{c||}{22} & \multicolumn{2}{c|}{24}  \\
        \hline\hline
        False & \multirow{2}{*}{8}  & 94.4	& 94 & 95.5 & 95.5 & 96.2 &	98.4 &	99.9 & 101.1 & 96.3 & 97.7 & 100.8 & 102.1 & 99 & 100.6 & 101.3 & 103.5 & 101.8 & 102.5 
        \\
        \cline{1-0} \cline{3-20}
        True  &                     & 91.7 & \textbf{99.4} & 96.1 & \textbf{102.6} & 95.6 & \textbf{113.3} & 91.8 & \textbf{108.9} & 99.2 & \textbf{114.2} & 93.4 & \textbf{109.9} & 102.5 & \textbf{108.5} & 92 & \textbf{117.3} & 92 & \textbf{117.3} 
        \\
        \hline\hline
        False & \multirow{2}{*}{12}  & 95.2 &	96 &	91.5 &	94.6 &	\textbf{105.6} & 98.4 & 99.9	& 98.6 &	106.6 &	102.9 & 101.5 & \textbf{109.3} & 108.6 & 105.4 & 109.5 & 111.4	& 104.1 & 107.1 
        \\
        \cline{1-0} \cline{3-20}
        True  &                     & 91.4 & \textbf{99.3 }& 91.7& \textbf{113.4} & 100.8 & 98.1	& \textbf{104.4 }& 	100.7 &	104.4 &	\textbf{108.8} & 96.7 & 106.7 & \textbf{111.1} & 109.2 & 100 & \textbf{111.7} & 103.2 & \textbf{119.3}
        \\
        \hline\hline
        False & \multirow{2}{*}{16}  & 100.2 & 98.3 & 105.1	& 100 & 96.3 & 101.6 & 101.5 & 98.2	& 111 & 111.6 & 103.1 & 105 & 100.5 & 110.3 & 90.6 & 104.8 & 96.4 & 96.8 
        \\
        \cline{1-0} \cline{3-20}
        True  &                     & 93.8 & \textbf{106.4} & 94.8 & \textbf{116.6} & 79.1 & \textbf{115.2} & 99 & \textbf{118.8} & 	96.9 & \textbf{115.2} & 81.3 & \textbf{121.8} & 82.0 & \textbf{118.9} & 79.4 & \textbf{119} & 79.6 & \textbf{119.8}
        \\
        \hline\hline
        False & \multirow{2}{*}{24}  & 101.4 &	99.7 & 103.8 & 108.6 & 101.6 & 106.5 & 103.9 & 	107.8 & 96.4 & 95.4 & 98.8 & 99.5 & 106.3 &	105.6 & 96.8 & 103.5 & 94.2 & 96.6 
        \\
        \cline{1-0} \cline{3-20}
        True  &                     & 83.2 & \textbf{118.4} & 96.7 & \textbf{116} & 82.3 & \textbf{119.3} & 90.9 & \textbf{119.6} & 79.6 & \textbf{120.84} & 79.5 & \textbf{116.1} &  83.4 &	\textbf{119}	& 96.8 & \textbf{116.9} & 79.6 &	\textbf{120.1}
        \\
        \hline\hline
        False & \multirow{2}{*}{32}  & 108.1 & 110.8 & 93.7	& 111.1 & 96.4 & 97.1 & 102.3 & 102 & 104 & 103.3 & 92 & 96.2 & 99.9 & 96.2 & 98.1 & 99 & 96.9 & 96 
        \\
        \cline{1-0} \cline{3-20}
        True  &                     & 110.1 & \textbf{118.1}	& 82.3 & \textbf{118.7} & 79.5 & \textbf{119.5} & 79.7 & \textbf{115.4} & 84.6 & \textbf{118.9} & 79.4 & \textbf{118.6} & 79.4 & \textbf{119.6} & 79.4 & \textbf{119.2} & 79.5 & \textbf{118.6} 
        \\
        \hline
    \end{tabular}}
    \label{tab:ablation}
\end{table}

\subsubsection{Obtained results} \label{subsect:colpe_exp_results}

\begin{figure}[h]
    \begin{subfigure}[b]{0.5\textwidth}
        \includegraphics[width=\linewidth]{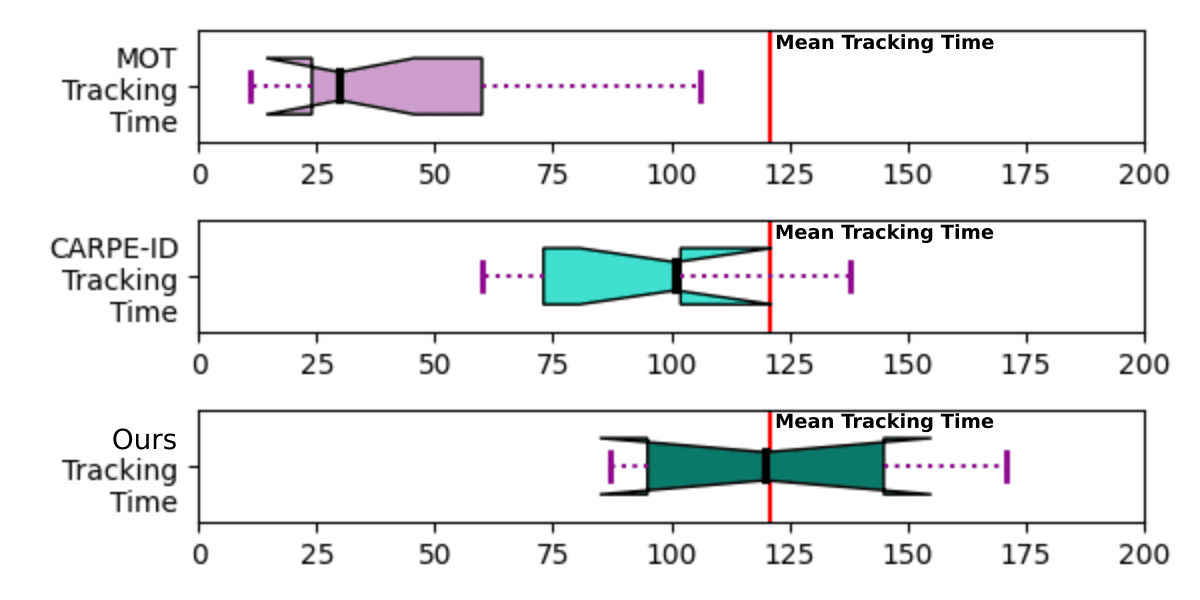}
        \caption{Tracking time comparison.}
        \label{fig:tracking_time}
    \end{subfigure}
    \begin{subfigure}[b]{0.5\textwidth}
        \includegraphics[width=\linewidth]{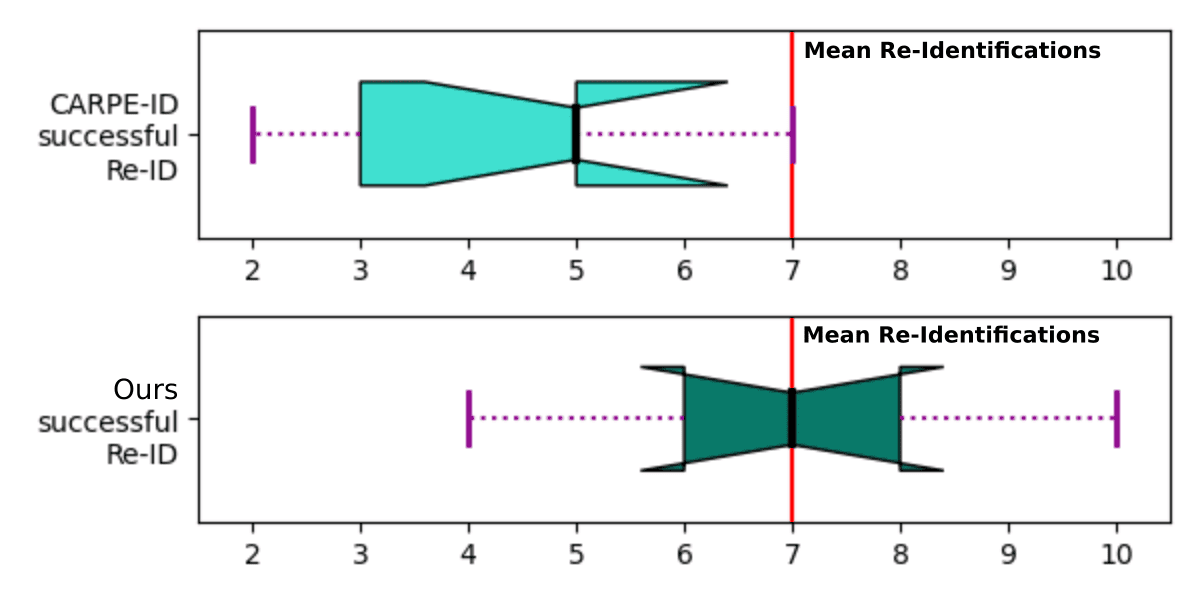}
        \caption{Successful \gls{reid} comparison.}
        \label{fig:reid}
    \end{subfigure}
    \caption{Comparison between the experimental results of \gls{sota} \gls{mot} algorithms: (in light purple), \gls{carpe} (in light blue), and the proposed framework (in dark green); including tracking time (a) and successful \gls{reid} (b). The ground truth for the mean tracking time and the \gls{reid}s are displayed in red. The failures of the \gls{mot} algorithm are used as a reference to evaluate the successful \gls{reid} of \gls{carpe} and the proposed framework; hence, the \gls{mot} algorithm has been excluded from comparison in (b). Reprinted, with permission, from\,\cite{rollo2025pesonalized}, \textsuperscript{\textcopyright} 2025 \gls{ieee}.}
    \label{fig:colp_experiments}
\end{figure}

Fig.~\ref{fig:colp_experiments} presents a comparison among a \gls{sota} \gls{mot} system\footnote{Yolov8 Tracking: \url{https://github.com/mikel-brostrom/yolo_tracking}}, the \gls{carpe} framework, and the proposed framework. The box plots in Fig.~\ref{fig:tracking_time} and Fig.~\ref{fig:reid} clearly show that the proposed framework consistently outperforms both the original \gls{carpe} approach and the baseline \gls{mot} system across the performance metrics. The \gls{mot} system is not directly included in Fig.~\ref{fig:reid} because it does not incorporate a \gls{reid} step; however, its tracking failures serve as the reference for determining whether \gls{carpe} or the proposed framework successfully re-identifies the target.

The proposed framework achieves a remarkable mean tracking time of $121.8$ seconds, which is nearly identical to the ground-truth mean appearance time of $122.7$ seconds (the total duration during which the target remains in the camera's \gls{fov}). This results in only a negligible $0.9$ seconds mean deviation in tracking accuracy. In contrast, \gls{carpe} achieved a mean tracking time of $102.3$ seconds, with a substantially larger mean tracking accuracy deviation of $20.4$ seconds.

Furthermore, the proposed framework exhibits high reliability in handling ID switches within the \gls{mot} system, successfully re-identifying the tracked target in every case. Indeed, the proposed approach achieves a mean of $7$ successful \gls{reid}s, matching the ground truth exactly. In contrast, \gls{carpe} only achieves a mean of $5$ because it fails to track the target until the end of several videos due to catastrophic forgetting.

It is important to note that the proposed framework operates in real time, as it does not rely on offline data or require the analysis of an entire dataset, which is beneficial for applications such as real-time camera streaming. This highlights its robustness, as it processes images at an online rate of $7\,\text{Hz}$. While a high presence of distractors in an image can hinder algorithm performance, this framework is specifically designed for middle-to-near distances between the robot and the human collaborator. This proximity typically keeps the number of distractors stable, preventing a significant drop in performance.

\subsubsection{Ablation study} \label{subsect:colp-abl-study}

\begin{figure}[h]
    \centering
    \includegraphics[width=\linewidth]{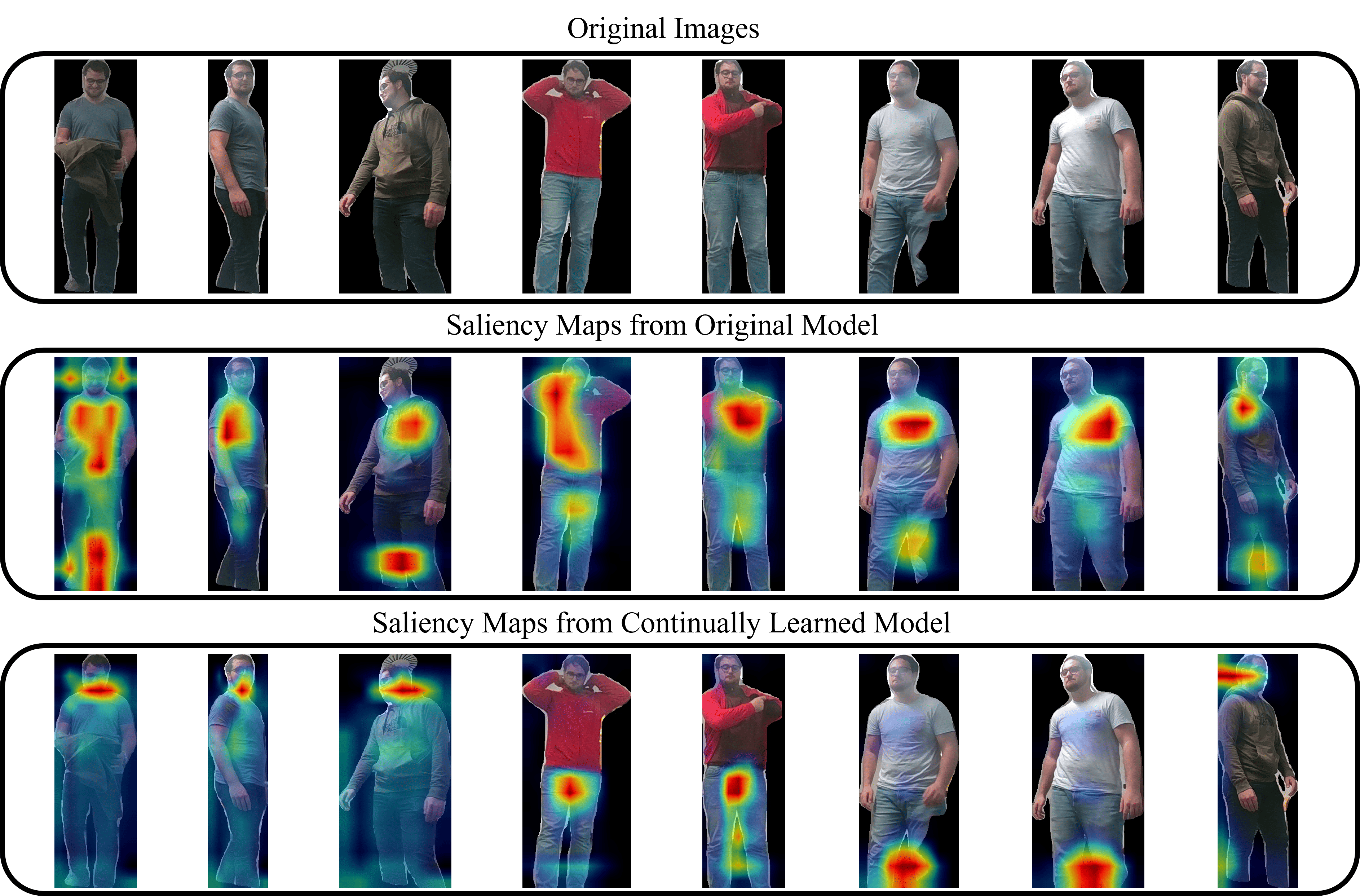}
    \caption{Saliency maps comparison obtained with Grad-Cam\,\cite{selvaraju2017grad}. The saliencies of the original pre-trained network model (second row) are compared with the ones obtained using the weights of the continually learned model (last row). The image background is intentionally removed to reduce its influence on the extracted features, allowing the network to focus just on the people's characteristics. Reprinted, with permission, from\,\cite{rollo2025pesonalized}, \textsuperscript{\textcopyright} 2025 \gls{ieee}.}
    \label{fig:saliency}
\end{figure}

An ablation study was conducted to determine the optimal batch size and number of training iterations for continual twin network training. This study is paramount for this application because it addresses the inherent trade-off between maximizing tracking performance and minimizing the risk of network overfitting during on-the-fly training.

The ablation study was organized using four distinct setup configurations for each parameter combination (see Section~\ref{subsec:colp-method} for parameter descriptions):
\begin{itemize}
    \item With the \textit{\gls{carpe} statistical model update} and with the \textit{early training} process.
    \item Without the \textit{statistical model update} but with the \textit{early training} process.
    \item With the \textit{statistical model update} but without the \textit{early training} process.
    \item Without the \textit{statistical model update} and without the \textit{early training} process.
\end{itemize}

Consequently, for each combination of batch size and training iterations, the study involved these four configurations to compare the outcomes and understand the precise influence of each proposed feature on the overall framework.

The study systematically tested five batch sizes ($8, 12, 16, 24, 32$) and ten training iterations ($8, 10, 12, 14, 16, 18, 20, 22, 24$) to identify the optimal settings. This process was applied to each video in the experiments dataset, and the results were subsequently averaged and presented in Table\,\ref{tab:ablation} to eliminate redundancy and provide a consolidated view. 

While several parameter combinations produced favorable results, a batch size of $16$ and an iteration count of $18$ were identified as the optimal parameters for this framework. Additionally, the study confirmed that both the \textit{\gls{carpe} statistical model update} and \textit{early training} features are essential for achieving robust results.

The \textit{\gls{carpe} statistical model update} effectively improves \gls{reid}, particularly when the network is trained just before the target disappears from the image. In such scenarios, the network's feature output deviates from the previously established statistical model, necessitating post-training adaptation to prevent the loss of the target. This feature thus proves essential in the selected scenarios, fortifying the overall framework.

Furthermore, \textit{early training} has proven crucial for achieving high accuracy. The value of this feature lies in enabling the feature network to train on all acquired images up to the point of target loss, and then fine-tune it with a larger number of valuable target appearances. This aligns perfectly with the \textit{statistical model update}, as training occurs precisely when the target is lost, which can cause misalignment between the feature network's output and the previously acquired statistical model.

This dual dependency is evident in Table\,\ref{tab:ablation}, where the impact of early training is heavily influenced by the presence or absence of the \textit{statistical model update}, with most of the best and worst results observed accordingly within the early training row. These experiments provided clear guidance on the best parameters and configurations to maximize the performance of this framework.

\subsubsection{Visualization with saliences} \label{subsect:colp_saliences}

To visually demonstrate why the proposed approach enhances \gls{reid} performance, a comparison of saliency maps is conducted. These maps are obtained using two sets of weights: those from the standard feature extractor network pre-trained on the MSMT17 dataset\,\cite{wei2018person}, and those obtained using the continual learning approach, as illustrated in Fig.~\ref{fig:saliency}.

The well-established explainable \gls{ai} framework Grad-CAM\,\cite{selvaraju2017grad} is used to generate this visual output. 

The comparison of saliency maps presented in the figure reinforces the thesis of network specialization. Saliency maps are generated for random images extracted from the custom experiments dataset. The figure clearly shows that the network's saliency maps with its original weights are relatively sparse and often focus on disparate image regions. In sharp contrast, the saliency maps generated by the network trained through continual learning are more specialized and tend to focus on visual cues that remain consistent across video sequences. For instance, in these experiments, as the targets frequently changed their upper-body clothing, the networks continually adapted and specialized in recognizing either their faces or their trousers/shoes.

Furthermore, incorporating the Soft Triplet loss during training serves a dual purpose: it not only compels the network to learn to recognize the most distinctive features of the target but also equips it to differentiate these features from those of the distractors effectively. This visual evidence explains why the proposed algorithm substantially outperforms the baseline model: the features learned by the continually trained network are finely tuned to focus on the most reliable and consistent visual cues for robust \gls{reid}.

\section{Geometrical SLAM validations} \label{sec:mapping-exp}

In this section, the results obtained to address the geometrical \gls{slam} problem are presented. The results for the \gls{3d} \gls{lidar} \gls{slam} algorithm are reported in Section~\ref{sec:slam-exp}, while the results of the ground-aware filtering technique are presented in Section~\ref{sec:intensity-exp}. Section~\ref{subsec:gsc-exp} contains the comparative results of the statistical \gls{lcd} proposal.

\subsection{Submap-based 3D LiDAR SLAM with multi-level scan matching experiments} \label{sec:slam-exp}

\begin{table*}[t]
    \centering
    \caption{Comparison of the proposed \gls{lidar} \gls{slam} method with \gls{sota} approaches provided by the \gls{vbr} dataset benchmark. Lower \gls{ate} and \gls{rpe} values indicate more accurate trajectory estimation and, consequently, more precise map reconstruction. Reprinted, with permission, from\,\cite{rollo2025leoslam}, \textsuperscript{\textcopyright} 2025 \gls{ieee}.}
    \resizebox{\textwidth}{!}{\begin{tabular}{|c|c|cccc|}
        \hline
        \textbf{Method} & \textbf{Metric}  & \textbf{Colosseum} & \textbf{Diag} & \textbf{Pincio} & \textbf{Spagna} \\
        \hline
        \multirow{2}{*}{\textbf{LEO\raisebox{0.1ex}{-}SLAM}} & \gls{ate} (m) ↓ & \textbf{ 0.40748} & 0.63319 & 1.019423 &  \textbf{0.775329} \\
                                           & \gls{rpe} (m) ↓ & \textbf{0.33596} & \textbf{0.30345} & \textbf{0.328775} & 1.027941 \\
        \hline
        \multirow{2}{*}{\textbf{KISS-ICP}} & \gls{ate} (m) ↓ & 1.51690 & 1.39727 & 0.78472 & 1.04755 \\
                                           & \gls{rpe} (m) ↓ & 0.45299 & 1.7908 & 0.48580 & 0.39860 \\
        \hline
        \multirow{2}{*}{\textbf{PIN-SLAM}} & \gls{ate} (m) ↓ & 0.50647 & \textbf{0.36240} & \textbf{0.64738} & 5.68224 \\
                                           & \gls{rpe} (m) ↓ & 0.49052 & 0.46836 & 0.45373 & \textbf{0.35910} \\
        \hline
    \end{tabular}}
    \label{tab:slam_comparison}
\end{table*}

\begin{figure}[h]
    \centering
    \begin{subfigure}[b]{0.24\textwidth}
        \includegraphics[width=\linewidth]{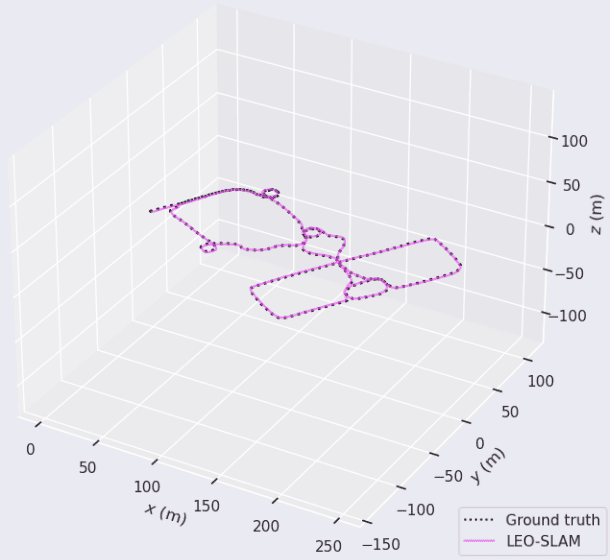}
        \caption{Pincio session.}
        \label{fig:pincio_traj}
    \end{subfigure}
    \begin{subfigure}[b]{0.24\textwidth}
        \includegraphics[width=\linewidth]{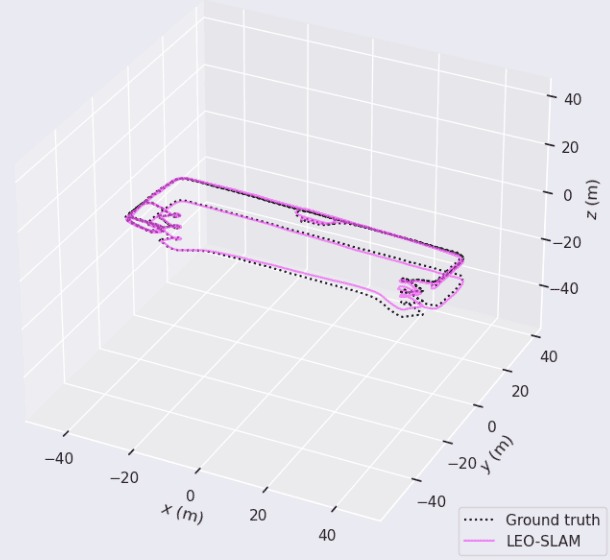}
        \caption{Diag session.}
        \label{fig:daig_traj}
    \end{subfigure}
        \begin{subfigure}[b]{0.24\textwidth}
        \includegraphics[width=\linewidth]{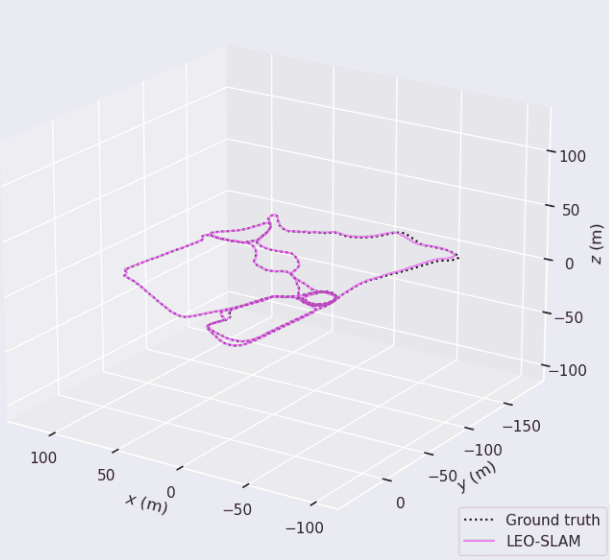}
        \caption{Spagna session.}
        \label{fig:spagna_traj}
    \end{subfigure}
        \begin{subfigure}[b]{0.24\textwidth}
        \includegraphics[width=\linewidth]{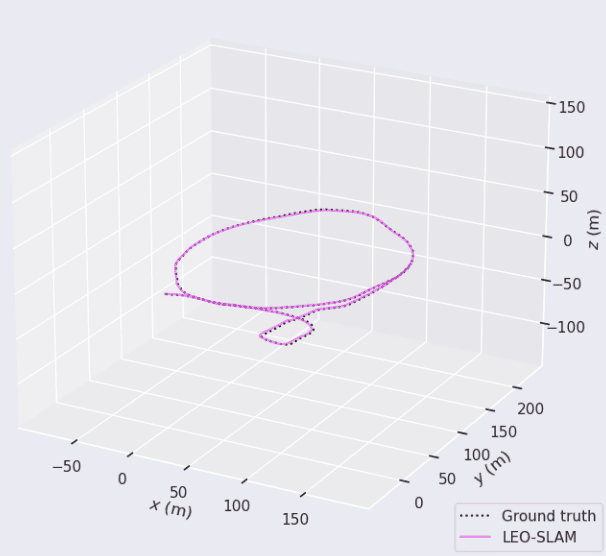}
        \caption{Colosseum session.}
        \label{fig:colosseum_traj}
    \end{subfigure}
    \caption{Comparison between the trajectories generated by LEO\raisebox{0.1ex}{-}SLAM and the \gls{vbr} ground truth across four different sessions. Reprinted, with permission, from\,\cite{rollo2025leoslam}, \textsuperscript{\textcopyright} 2025 \gls{ieee}.}
    \label{fig:trajectories}
\end{figure}

\begin{figure}[h]
    \centering
    \begin{subfigure}[b]{0.24\textwidth}
        \includegraphics[width=\linewidth]{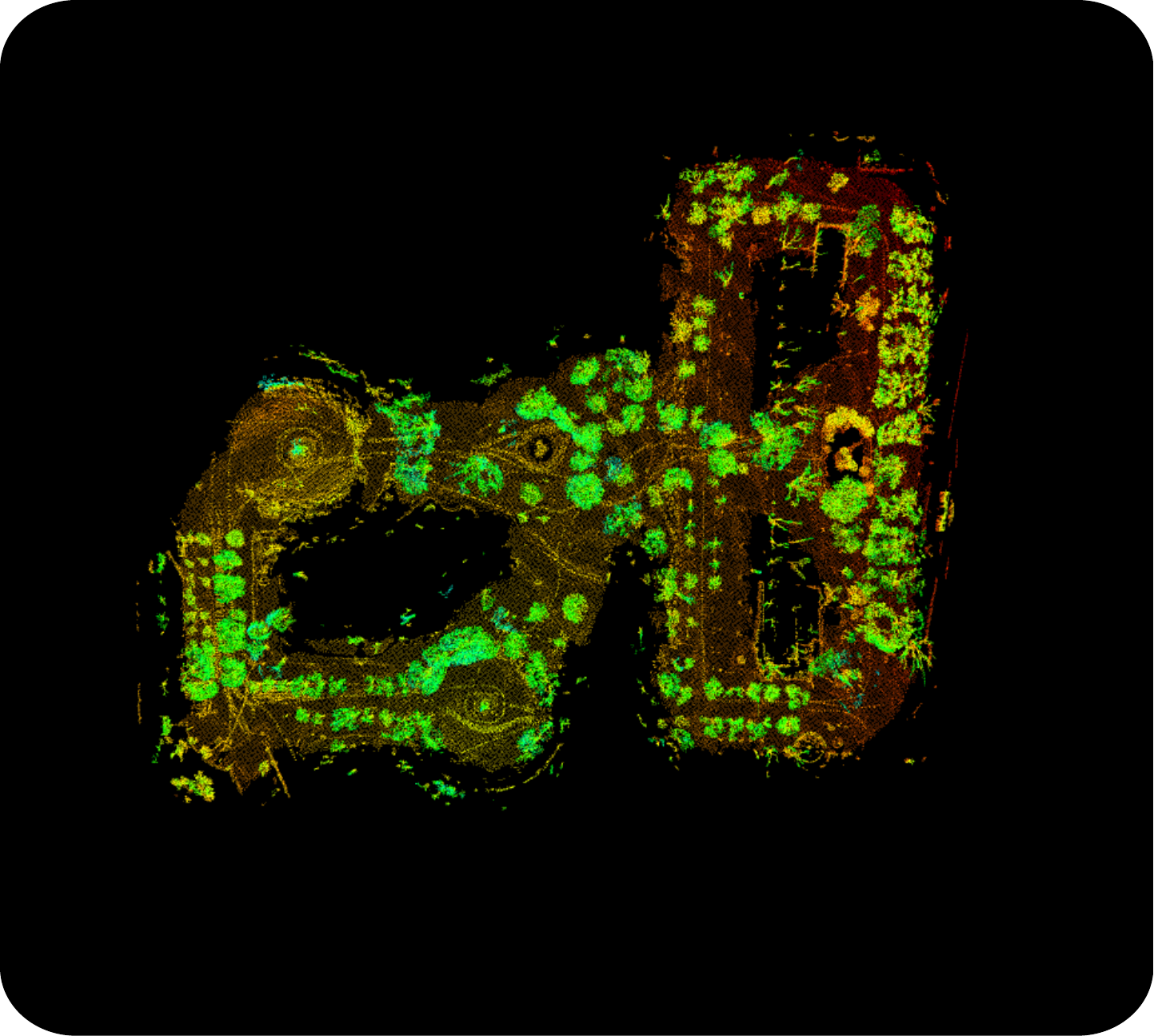}
        \caption{Pincio session.}
        \label{fig:pincio_map}
    \end{subfigure}
    \begin{subfigure}[b]{0.24\textwidth}
        \includegraphics[width=\linewidth]{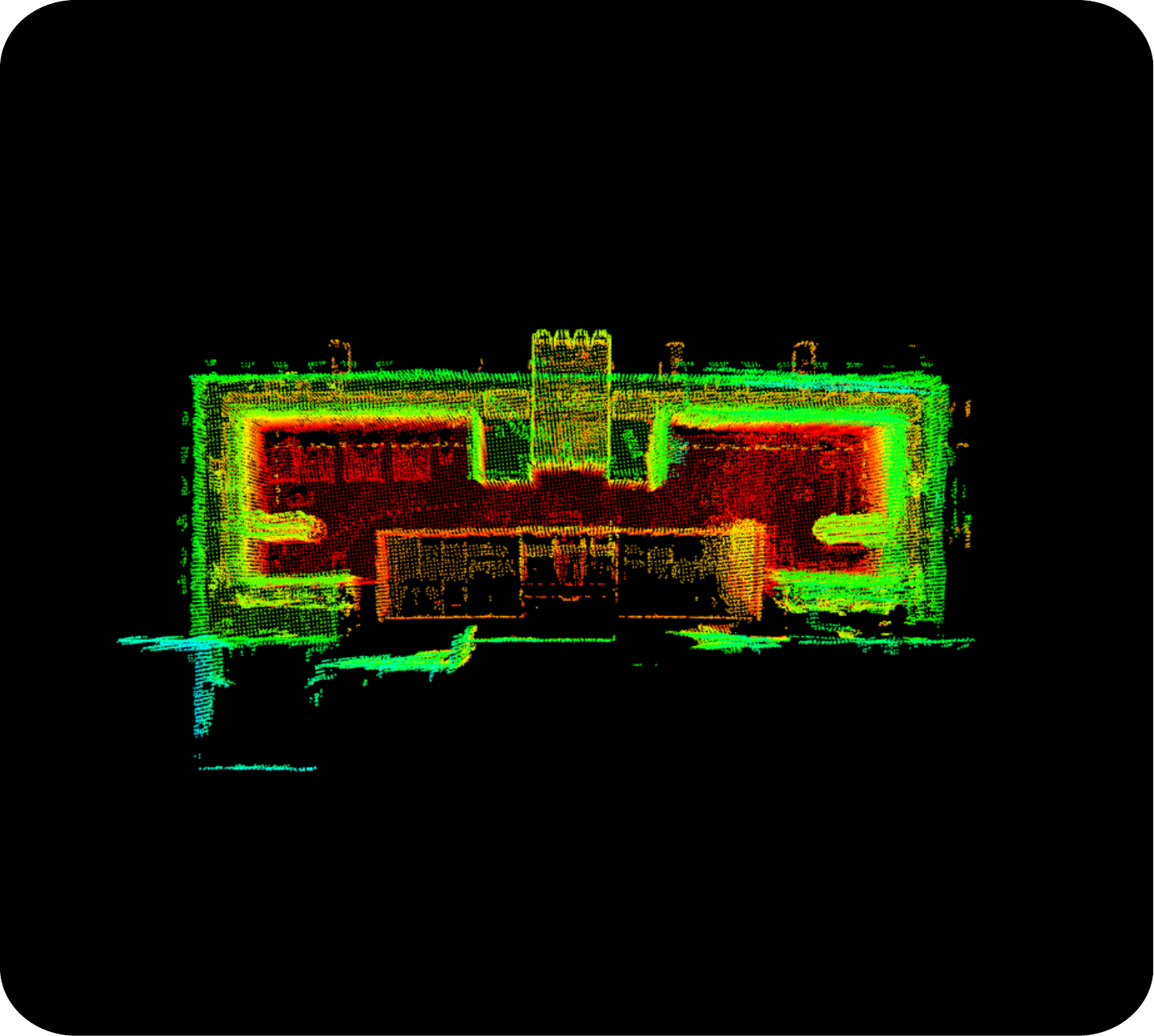}
        \caption{Diag session.}
        \label{fig:diag_map}
    \end{subfigure}
    \begin{subfigure}[b]{0.24\textwidth}
        \includegraphics[width=\linewidth]{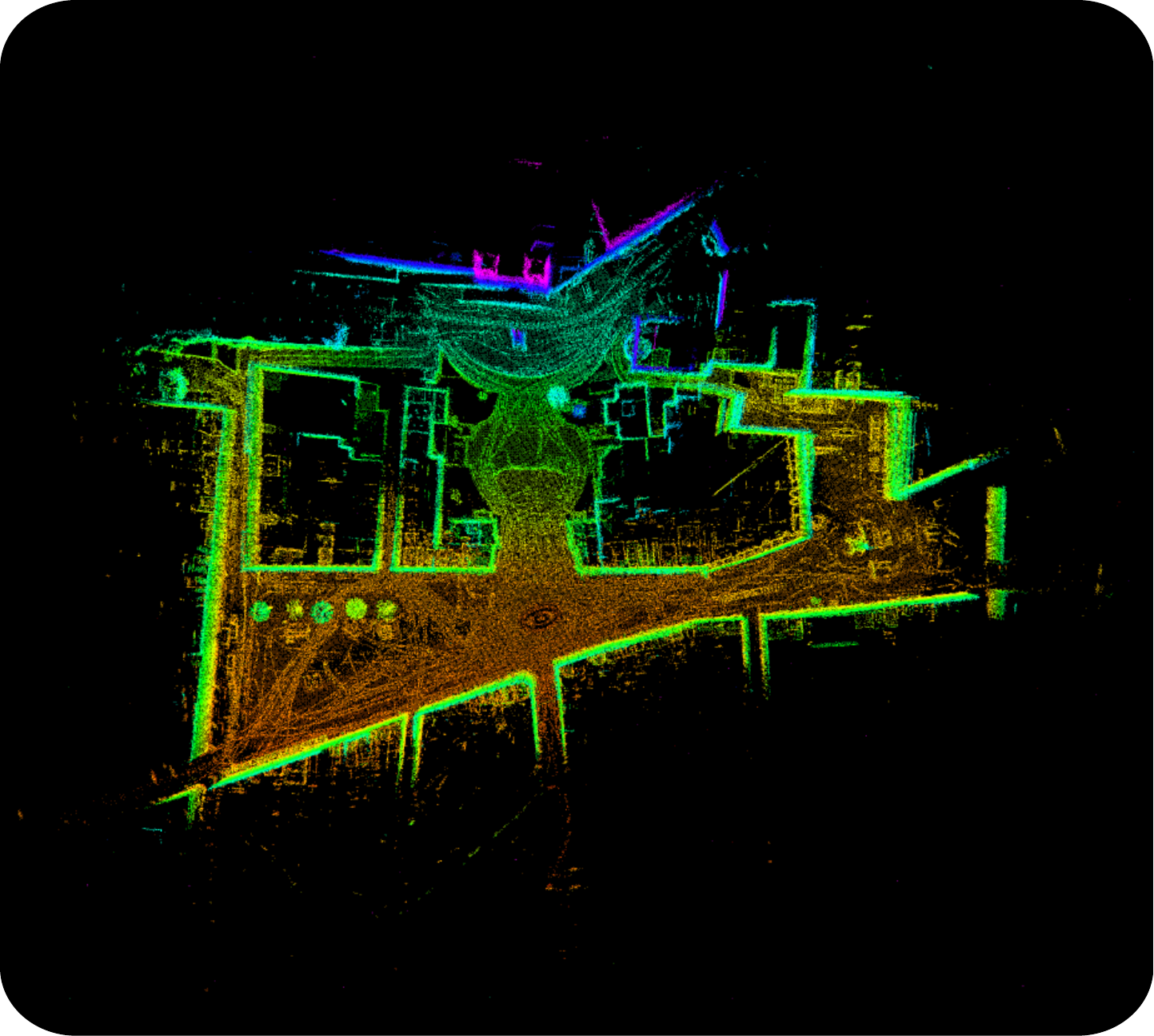}
        \caption{Spagna session.}
        \label{fig:spagna_map}
    \end{subfigure}
    \begin{subfigure}[b]{0.24\textwidth}
        \includegraphics[width=\linewidth]{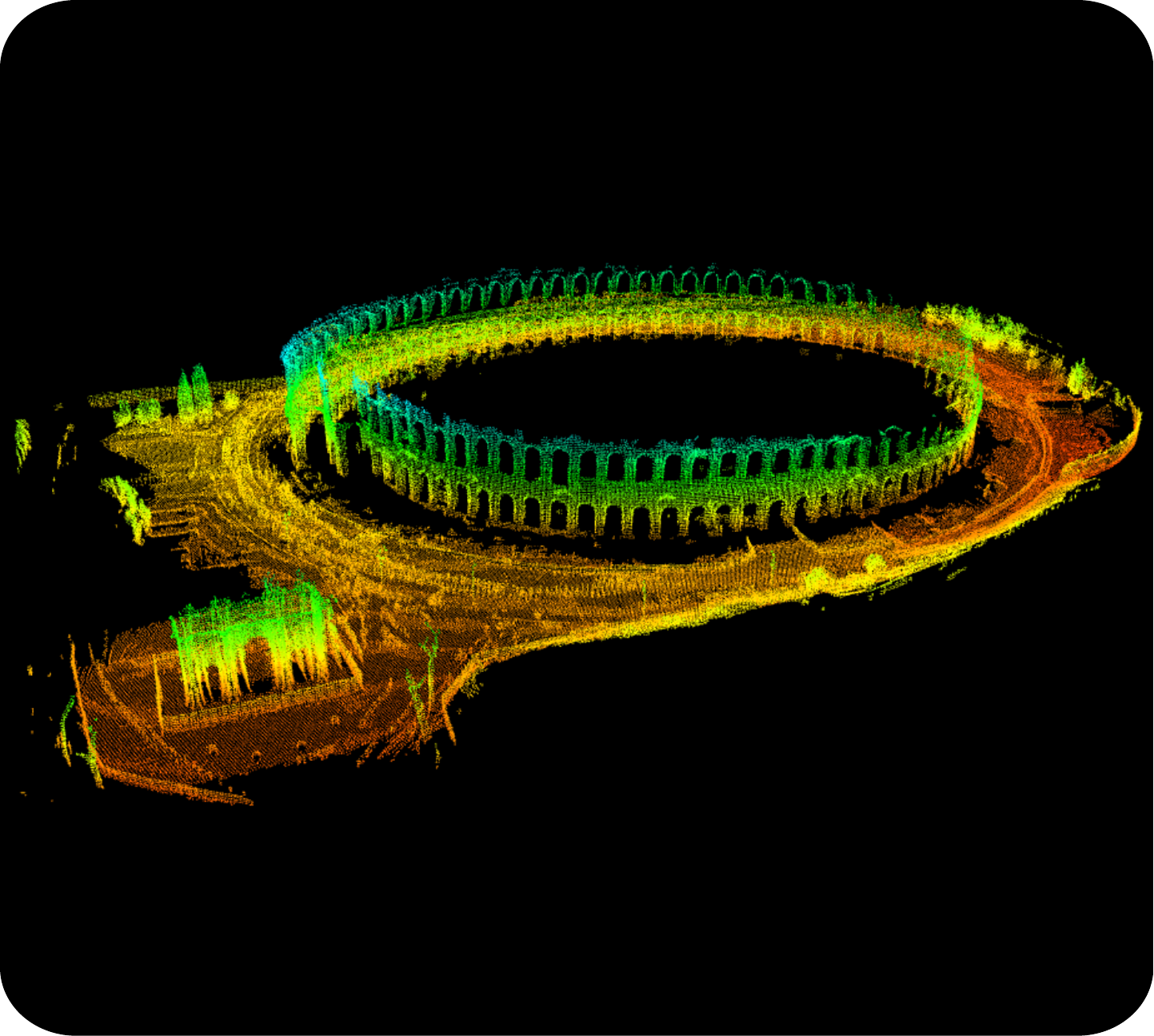}
        \caption{Colosseum session.}
        \label{fig:colosseum_map}
    \end{subfigure}
    \caption{Final maps obtained using LEO\raisebox{0.1ex}{-}SLAM from the four analyzed \gls{vbr} sessions. Reprinted, with permission, from\,\cite{rollo2025leoslam}, \textsuperscript{\textcopyright} 2025 \gls{ieee}.}
    \label{fig:maps}
\end{figure}

To validate the LEO\raisebox{0.1ex}{-}SLAM algorithm, the \gls{vbr} dataset\,\cite{brizi2024vbr} is used. The proposed \gls{slam} method is compared with two \gls{sota} algorithms: KISS-ICP\,\cite{vizzo2023kiss} and PIN-SLAM\,\cite{pan2024pin}, as reported in Table\,\ref{tab:slam_comparison}. The proposed algorithm and KISS-ICP require only a single CPU to run, whereas PIN-SLAM requires a \gls{gpu} to operate in real time. The \gls{vbr} dataset is chosen because it provides a comprehensive set of sensor data and reliable ground-truth estimations. The dataset includes a sensor session containing \gls{lidar}, \gls{imu}, and camera data. The \gls{lidar} scan and \gls{imu} data are utilized in these experiments to provide an initial guess for the scan-to-scan (s2s) alignment. Four out of the eight sessions provided in the \gls{vbr} training dataset are considered: Colosseum, Diag, Pincio, and Spagna.

The qualitative results of these experiments are shown in Fig.~\ref{fig:trajectories}, which compares the estimated trajectories with the ground-truth trajectories, and in Fig.~\ref{fig:maps}, which presents the final point cloud maps. 

The accuracy metrics used are the \gls{rmse} of the \gls{ate} and the \gls{rpe}, as expressed in Eq.~\eqref{eq:ate} and Eq.~\eqref{eq:rpe}, calculated on the translational component of the trajectory compared to the \gls{vbr} ground truth. The EVO framework\,\cite{grupp2017evo} is used to compute these metrics, which superimposes the estimated trajectory with the ground truth using the $SE(3)$ Umeyama alignment method\,\cite{umeyama1991least}.

\begin{align} 
    ATE = & \sqrt{\frac{1}{\mathrm{N}}\sum_{i=1}^\mathrm{N}||\hat{\textbf{p}}_i - \textbf{p}_i||^2}, \label{eq:ate} \\
    RPE = & \sqrt{\frac{1}{\mathrm{N}}\sum_{i=1}^\mathrm{N}||(\hat{\textbf{p}}_{i+1} - \hat{\textbf{p}}_{i}) - (\textbf{p}_{i+1} - \textbf{p}_{i})||^2}, \label{eq:rpe}
\end{align}
where $\hat{\textbf{p}}_i$ denotes the estimated trajectory position at time $i$, $\textbf{p}_i$ is the corresponding ground-truth position at time $i$, and $\mathrm{N}$ is the total number of estimated trajectory positions.

In the experiments, the voxelization leaf size of the small-GICP algorithm was set to $0.25$ meters to improve processing efficiency while preserving relevant environmental features. The validation results are presented in Table\,\ref{tab:slam_comparison}, where the LEO\raisebox{0.1ex}{-}SLAM  algorithm is compared against several algorithmic results provided by the \gls{vbr} benchmark.

\subsubsection{Discussion}

In the experiments, outdoor environments are primarily considered, except for the \textit{Diag} session, which transitions from outdoor to indoor environments, thereby increasing mapping difficulty.

As shown in Table\,\ref{tab:slam_comparison}, LEO\raisebox{0.1ex}{-}SLAM achieves improved mapping accuracy, enabling better map usability for navigation. The proposed algorithm outperforms current \gls{sota} methods in \gls{rpe} estimation across almost all sequences and achieves better \gls{ate} results in $2$ out of $4$ sequences. \gls{ate} measures the global accuracy of the estimated trajectory with respect to the entire ground truth trajectory, whereas \gls{rpe} evaluates the between-node estimation accuracy, \textit{i.e.}, the displacement error between consecutive nodes with respect to the ground truth displacements. As shown in Table\,\ref{tab:slam_comparison}, \gls{ate} and \gls{rpe} are not strictly correlated; thus, a better \gls{ate} does not necessarily imply a better \gls{rpe}. This is because a trajectory with smaller between-node error (\gls{rpe}) may still accumulate significant drift, leading to a higher overall \gls{ate}.

LEO\raisebox{0.1ex}{-}SLAM runs in real time on CPU-based platforms, with an average computation time of $185\,\text{ms}$ per full processing loop (see Algorithm\,\ref{alg:leo-slam} for reference).

The introduction of submap keyframes and submap-to-submap alignment improves accuracy by yielding denser and richer point clouds. This is particularly beneficial for \gls{lidar} sensors with fewer channels, a common scenario in robotics. In such cases, a keyframe consisting of a single point cloud may lack sufficient information to robustly handle potential alignment errors.
The reconstructed maps generated by the proposed algorithm are sufficiently smooth for subsequent planning and navigation tasks or for visualization.

One limitation of LEO\raisebox{0.1ex}{-}SLAM is the potential drift introduced in featureless environments, such as long corridors. This is a common issue in pure \gls{lidar} odometry \gls{slam} algorithms and is difficult to correct solely through loop closure. However, using external odometry or \gls{imu} data as an initial guess helps mitigate this problem.

Planned improvements to address this limitation include the integration of multi-sensory solutions, such as the deployment of visual cues to refine odometry using cameras, and the autonomous adaptation of parameters based on the environmental context, aiming to reduce drift in low-feature environments.
This work also encounters challenges on highly dynamic vehicles, where point clouds become distorted due to rapid motions, particularly during high-speed rotations. A potential solution is the application of point cloud deskewing, as proposed in Vizzo et al.\,\cite{vizzo2023kiss}.

\subsection{Point cloud ground-aware intensity filtering experiments} \label{sec:intensity-exp}

\begin{table*}[h]
    \centering
    \caption{The \gls{rmse} results from the experiments on the custom dataset using the selected frameworks are presented, considering the following three scenarios: (i) no point cloud filter applied, (ii) applying only the intensity thresholding filter, and (iii) applying the intensity thresholding filter with additional ground awareness. }
    \resizebox{\textwidth}{!}{\begin{tabular}{|c|c|ccc|}
    \hline
        Framework & Filtering Mode & \gls{te} [m] & \gls{oe} [rad] & \gls{tve} [m] (x y z) \\
        \hline
        \hline
        & No Filter & 22.529 & 0.270 & 18.848 12.129  2.298  \\
         Kiss-ICP\,\cite{vizzo2023ral} & Intensity & 2.585 & 0.100 & \textbf{0.278} 1.152 2.297 \\
         & Ground + Intensity & \textbf{1.112} & \textbf{0.075} &  0.478 \textbf{0.986} \textbf{0.187} \\
         \hline
         &  No Filter  & \textbf{0.410} & 0.068 & 0.164 \textbf{0.109} 0.359 \\
         \gls{dlo}\,\cite{chen2022direct} & Intensity & 2.395 & 0.098 & 0.179 0.408 1.979 \\
         & Ground + Intensity  & 0.428 & \textbf{0.0637} & \textbf{ 0.156} 0.397  \textbf{0.031}  \\
         \hline
    \end{tabular}}
    \label{tab:intensity-results}
\end{table*}

To validate the ground-aware intensity filter implementation, \gls{sota} \gls{lidar} odometry algorithms such as KISS-ICP\,\cite{vizzo2023ral} and \gls{dlo}\,\cite{chen2022direct} are used. Due to the sparsity of the point cloud in the proposed scenario, it was decided to remove one of the two downsampling filters used in KISS-ICP before scan matching, as the authors suggested, to obtain better results.

A custom dataset was created in an indoor scenario with a high number of windows, leading to numerous erroneous reflections detected by the \gls{lidar}. The experimental setup consists of a quadruped robot, the Unitree B1, equipped with a Velodyne VLP-16 \gls{lidar}. To filter out low-intensity points, a filtered range, $\psi^* = [0, 15]$, is selected from the full intensity range $\psi = [0, 255]$. The terrain is mostly flat, with some ramps along the paths. A custom dataset is created because, to the best of the author's knowledge, commonly used public datasets, such as the KITTI dataset\,\cite{geiger2013vision}, do not feature a significant number of erroneous reflections caused by intensity variations. As a result, these datasets are unsuitable for evaluating the proposed filtering methods.

Additionally, this dataset does not include a ground-truth trajectory to calculate the \gls{ate} or \gls{rpe}, unlike the previous validation in Section~\ref{sec:slam-exp}. However, for each path, the starting and finishing poses are aligned, enabling the final odometry error to be computed with minimal uncertainty. Using the initial and final robot poses, the following three types of mistakes are evaluated:

\begin{itemize}
    \item \textit{\gls{te}} $[\text{m}]$: The Euclidean distance between the origins of the final and initial frames.
    \item \textit{\gls{oe}} $[\text{rad}]$: The rotation angle between the orientations of the final and initial frames, expressed in axis-angle form.
    \item \textit{\gls{tve}} $[\text{m}]$: The translational error along the $x$, $y$, and $z$ axes.
\end{itemize}

To validate the performance, the \gls{rmse} of these errors is computed across the following three scenarios:
\begin{enumerate}
    \item Original \gls{lidar} output, containing incorrect reflections.
    \item Filtered \gls{lidar} output, obtained by removing points with intensity values outside the threshold interval $\psi^*$.
    \item Filtered \gls{lidar} output with intensity filtering within $\psi^*$, augmented with the additional ground awareness as described in Section~\ref{subsec:intensity-method}.
\end{enumerate}

The \gls{rmse} results obtained in these experimental scenarios are reported in Table\,\ref{tab:intensity-results}.

The \gls{tve} is computed by subtracting the initial position $^B\mathbf{t}_i$ from the final position $^B\mathbf{t}_f$. The \gls{te} is the Euclidean norm of \gls{tve}. In contrast, the \gls{oe} was calculated by determining the relative quaternion orientation $^B\mathbf{q}_r = \, ^B\mathbf{q}_i^{-1} * \, ^B\mathbf{q}_f$ and converting the resulting angle into axis-angle form. Finally, the \gls{rmse} is computed for each scenario using these error values.

It is important to note that only the \textit{naive intensity filter} presented in Section~\ref{subsec:naive_filter} is used, since the alternative solution proved unusable due to the high point cloud sparsity.

The proposed intensity filter solution has proven to be robust and stable, offering improvements to \gls{sota} odometry systems. The experimental results are evaluated in the following, and the two filter implementations are subsequently compared to provide insights into which solution is more suitable.

\subsubsection{Ground-aware filter evaluation} \label{subsubsec:filter_eval}

The results reported in Table\,\ref{tab:intensity-results} demonstrate that this filtering method is beneficial and essential for the first framework evaluated, \textit{i.e.}, KISS-ICP\,\cite{vizzo2023ral}. In contrast, the second framework, \textit{i.e.}, \gls{dlo}\,\cite{chen2022direct}, proved to be robust enough to handle erroneous reflections, yielding comparable results in both unfiltered and ground-aware filtered scenarios.

It has been observed that KISS-ICP is sensitive to erroneous reflections, which consistently impact its performance. While introducing a standard intensity filter significantly improved the results, it still produced significant errors along the $z$-axis. This is because, as previously stated, the standard intensity filter removes ground points, which are crucial for the \gls{icp} algorithm to recover errors along the $z$-axis accurately.

An interesting observation is that in KISS-ICP, while the ground-aware filter performed better in almost all metrics, the $x$-axis component of the \gls{tve} showed slightly higher errors compared to the solution using only intensity filtering. This is due to \gls{icp} matching errors between ground-level range points in the current scan and those in the previous point cloud. \gls{lidar} sensors emit infrared rays in rows (or channels), with each channel projecting a ring onto the ground, as shown in the right image of Fig.~\ref{fig:filter_compare}.
Since the point cloud is \gls{lidar}-centric, these projected rings move along with the \gls{lidar}, making them difficult to distinguish in consecutive scans, as the robot remains centered within them.
As the robot typically moves along the $x$-axis of the \gls{lidar}, the \gls{icp} algorithm tends to match the rings of two consecutive point clouds, introducing small errors along the $x$-axis. These errors are partially compensated for by matching points on walls and obstacles, but the remaining errors accumulate over time, leading to slightly worse results when the ground points are preserved.

\gls{dlo}, on the other hand, proved to be robust to erroneous reflections, yielding similar results in both the unfiltered and ground-aware scenarios. However, an interesting consideration arises: if odometry is used to create a point cloud map, the presence of erroneous reflections will affect the final results, resulting in an erroneous map. This is clearly demonstrated in Fig.~\ref{fig:filter}, where the comparison between the filtered and unfiltered point clouds in high-reflection environments shows that the unfiltered point cloud becomes essentially unusable. Additionally, when a standard intensity filter is applied to refine the map, the absence of ground points in the filtered point cloud negatively impacts the accuracy of \gls{icp} matching along the \gls{lidar} $z$-axis, as evidenced in Table\,\ref{tab:intensity-results}, resulting in degraded performance in the final application. This highlights that, in mapping scenarios, this ground-aware filter is essential for producing an accurate final map without compromising odometry performance.

Lastly, an interesting observation made during the experiments was that \gls{dlo} experienced transient odometry errors in the unfiltered scenario, whereas its performance was smoother in the ground-aware filtering scenario. \gls{dlo} seems to be able to recover these accumulated errors, at least partially, by performing scan matching between the most recent point cloud scan and a piece of the entire accumulated map when places are already seen and revisited. This hypothesis could be further supported by using more advanced metrics, such as the \gls{ate} and \gls{rpe}, throughout the entire trajectory. However, due to the absence of ground truth in the experiments (caused by the lack of motion capture systems), the authors are unable to fully validate this hypothesis and leave it open for future studies and deeper analysis.

\subsubsection{Solutions comparison} \label{subsubsec:filter_impl_compare}

Although only one solution was experimentally tested, a conceptual comparison of both proposed ground-aware intensity filter implementations is provided, highlighting their respective pros and cons.

The first solution, referred to hereafter as the \textit{naive solution}, is straightforward to implement. However, its main issue is that it retains all the erroneously reflected points below the robot's height. While many of these retained points may still correspond to the ground, facilitating their merging with actual ground points, this correspondence is not guaranteed.

The second solution, known as the \textit{normal solution}, is more advanced and can distinguish and retain additional surfaces (like low tables) by comparing the \gls{lidar}'s homogeneous transformation with the point cloud normals. However, this comes at the cost of increased computational time and complexity. Furthermore, it still faces challenges with misreflected points that are parallel to the \gls{lidar}'s base. Additionally, the normal computation is not robust in cases of \gls{lidar} sparsity, which is common in robots using \gls{lidar} sensors with fewer channels (\textit{e.g.}, 16 or 32). This limitation makes the \textit{normal solution} less effective in such situations, a frequent scenario in robotic applications.

The \textit{naive solution} has a mathematical complexity of $\mathcal{O}(n)$, as it processes each point only once (a constant time operation per point). In contrast, the \textit{normal solution} has a complexity of $\mathcal{O}(n^2)$. This is because it first computes the normal for each point (which takes $\mathcal{O}(n)$ time due to neighborhood searches), and then checks both the intensity and the cosine of the angle between the \gls{lidar}'s $z$-axis vector and the point's normal (which adds $\mathcal{O}(n)$ complexity).

In summary, the \textit{naive solution} is faster, works well when the point cloud is sparse, and is easier to implement.  On the other hand, the \textit{normal solution} is computationally more expensive due to the norm computation. While it offers the advantage of maintaining $xy$-plane-parallel surfaces that are above the ground, this feature is less common in typical scenarios due to the generally low height of robots. However, such situations can still occur, making the \textit{normal solution} beneficial in those specific cases.

The \textit{naive solution} is generally preferred because delays are harmful in \gls{lidar} odometry and \gls{slam} systems. This choice becomes even more beneficial in outdoor environments, where the ground may not be perfectly parallel to the \gls{lidar}, creating difficulties for the normal-based solution.
Additionally, in scenarios where the robot is in a pit, the surrounding ground may be higher than the robot's height, which compromises the effectiveness of both solutions. However, there are two important considerations to note for outdoor use. First, when the ground is not parallel to the robot's base, the incidence angle of the \gls{lidar} should be large enough to ensure the points fall outside the intensity threshold $\psi^*$. Secondly, in outdoor environments, the ground material is often opaque, which helps mitigate issues with incorrect reflections, especially when the incidence angle allows for proper point detection.

To address the limitations of these single-sensor solutions, a multi-sensor setup, such as integrating cameras and \gls{lidar}, could be employed. Alternatively, intensity and semantic information could be used to more accurately segment true ground points. However, in robots with constrained resources, using these methods would lead to higher power and computational demands, which could degrade the performance of odometry and \gls{slam}. Therefore, simpler yet effective solutions, like the one presented in this paper, remain the preferred choice.

\subsection{Gaussian Scan Context experiments} \label{subsec:gsc-exp}

\begin{figure}[h]
    \centering
    \includegraphics[width=0.9\linewidth]{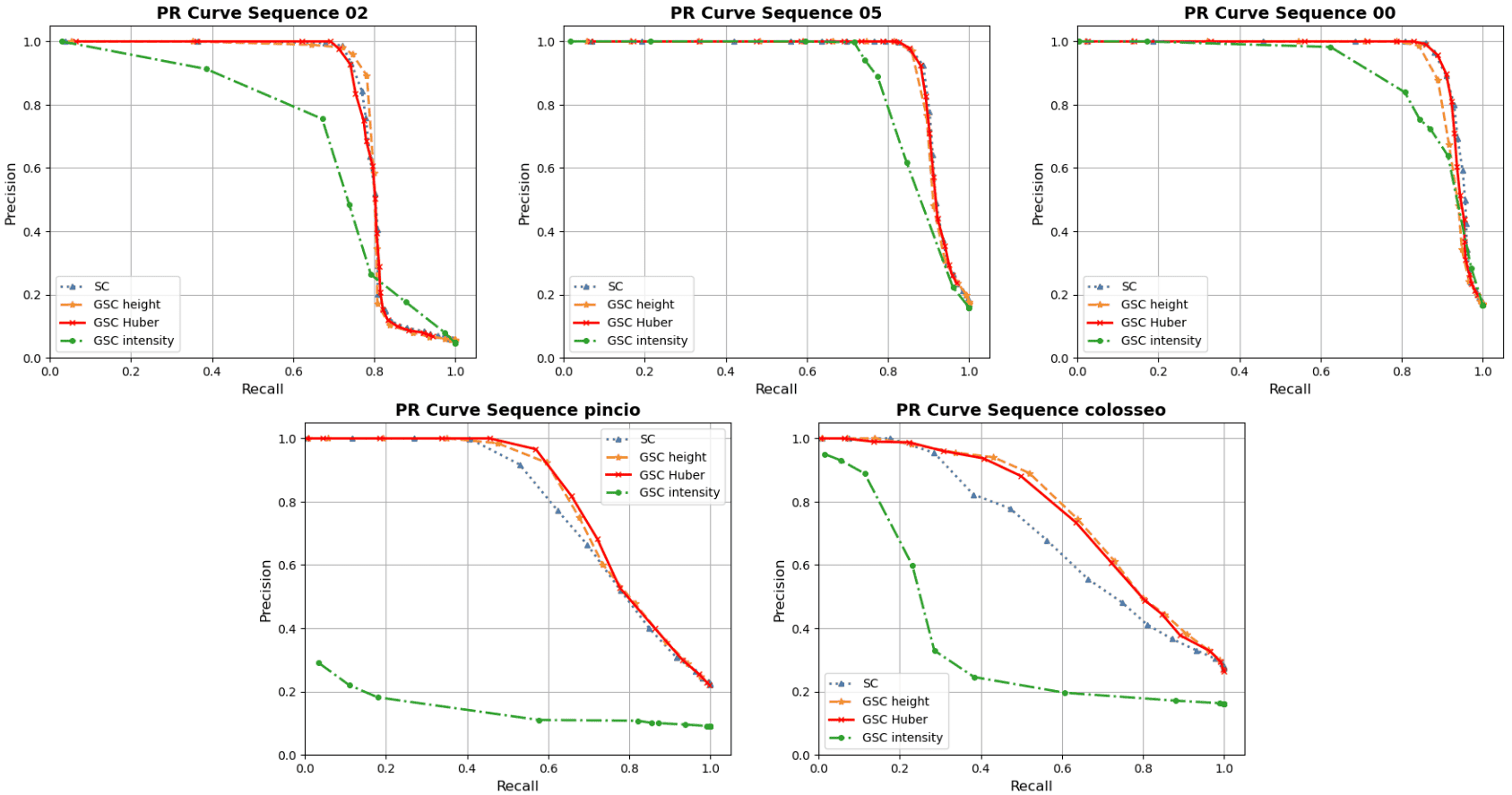}
    \caption{Precision–Recall curves obtained from the experimental results are presented. These graphs compare the \gls{sc++} baseline with three \gls{gsc} variants. Each \gls{gsc} variant computes the matrix entries using Eq.~\eqref{eq:gscentry}, but with different inputs: \gls{gsc} height (orange) uses the $z$ values, \gls{gsc} Huber (red) applies Huber-weighted $z$ values, and \gls{gsc} intensity uses the $i$ values.}
    \label{fig:prcurves}
\end{figure}

To validate the performance of the \gls{gsc} across diverse environments, two datasets are used for evaluation: the well-known KITTI dataset\,\cite{geiger2013vision} and the \gls{vbr} dataset\,\cite{brizi2024vbr}, collected in Rome. Specifically, sequences \texttt{00}, \texttt{02}, and \texttt{05} from KITTI are selected, as well as \texttt{Pincio} and \texttt{Colosseo} from \gls{vbr}, due to their high occurrence of loop closures.

The parameters for the Gaussian context are fixed at $\alpha = 2$ and $\delta = 1$. Both \gls{sc++} and \gls{gsc} are evaluated on the keyframes, where each keyframe corresponds to a point cloud with a ground-truth pose expressed relative to a common map frame.

\begin{table}[t]
    \centering
    \caption{Comparison of maximum recall values, under the constraint of perfect precision (100\%), between \gls{sc++} and \gls{gsc} for each selected sequence.}
    \begin{tabular}{|c|c|c|c|c|}
         \hline
         Sequence & \gls{sc++}\,\cite{kim2021scan} & \gls{gsc} Height & \gls{gsc} Huber    \\
         \hline\hline
         KITTI $\mathrm{02}$ & 57.8\%     & 59.4\%  & \textbf{69.7\%} \\
         \hline
         KITTI $\mathrm{05}$ & 81.2\%     & 82.6\%  & \textbf{83.1\%} \\
         \hline
         KITTI $\mathrm{00}$ & 80.8\%     & 81.2\%  & \textbf{83.7\%} \\
         \hline
         \gls{vbr} $\mathrm{Pincio}$ & 36.1\%   & 42.3\%  & \textbf{49.3\%} \\
         \hline
         \gls{vbr} $\mathrm{Colosseo}$ & \textbf{18.4\%} & 16.2\%  & 14.5\%          \\
         \hline
    \end{tabular}
    \label{tab:best}
\end{table}

\subsubsection{Evaluation metrics and conditions}

To assess performance, \Gls{p} and \Gls{r} are computed for both algorithms across each sequence. Ground-truth loop closures were identified based on the Euclidean distance $d_{ij}$ between the poses of two keyframes $i$ and $j$. The distance threshold is set as $d_t = 5\,\mathrm{m}$, and the following conditions are defined to determine the correctness of a detected loop closure:
\begin{itemize}
    \item A loop closure is detected between the current keyframe $\mathbf{k}_n$ and a previously observed keyframe $\mathbf{k}_j$, where $j \in \{1, \dots, \mathrm{N} - \xi\}$. The parameter $\xi \in \mathbb{N}$ excludes the most recent $\xi$ keyframes to prevent trivial and near-trivial matches.
    \item The Euclidean distance between the current keyframe $\mathbf{k}_n$ and the matched keyframe $\mathbf{k}_j$ is below the threshold, \textit{i.e.}, $d_{nj} \leq d_t$, indicating that they correspond to the same physical location. This condition is meaningful only if condition (1) holds.
    \item There exists a ground-truth match $\mathbf{k}_p$ among the previous keyframes such that $d_{np} \leq d_t$. This ensures that a valid loop closure is possible for the current keyframe.
\end{itemize}

Using these conditions, each keyframe match is classified as: (i) \gls{tp} (True Positive) if both conditions (1) and (2) are satisfied; (ii) \gls{fp} (False Positive) when Condition (1) is satisfied, but condition (2) is not; (iii) \gls{fn} (False Negative) if Condition (1) is not satisfied, but condition (3) is; and (iv) \gls{tn} (True Negative) in all remaining cases.

This evaluation framework enables a rigorous and interpretable comparison between \gls{sc++} and \gls{gsc} across diverse environmental conditions and sensor data characteristics.

Based on the classification results, \gls{p} and \gls{r} are computed as follows:
\begin{align}
    \text{P} = \frac{\text{TP}}{\text{TP}+\text{FP}}\ \ \ \ \
    \text{R} = \frac{\text{TP}}{\text{TP}+\text{FN}}
\end{align}

These metrics were evaluated for each validation sequence by varying the \gls{reid} threshold $\tau$ over a range from $0$ to $1$. This ablation study enables the analysis of how the algorithms' performance evolves with respect to the threshold, offering insights into their sensitivity to parameter tuning.

The resulting \gls{pr} curves are presented in Fig.~\ref{fig:prcurves}. These curves illustrate the robustness of each algorithm to variations in the threshold $\tau$. In particular, the \gls{auc} of the \gls{pr} curve serves as a measure of overall performance: the closer the \gls{auc} is to $1$, the more consistently high-performing the algorithm remains across threshold values. In the ideal case, where \gls{auc} = $1$, the algorithm achieves perfect performance regardless of the selected threshold. 

\subsubsection{GSC variants and results}

The performance of \gls{sc++} is compared with three variants of the proposed method:
\begin{itemize}
    \item \textit{\gls{gsc} Height}: computes the standard mean and standard deviation of point heights, then uses Eq.~\eqref{eq:gscentry} to populate the Gaussian context entries.
    \item \textit{\gls{gsc} Huber Height}: computes the mean and standard deviation of point heights using Huber weighting (as defined in Eq.~\eqref{eq:huber_mean} and Eq.~\eqref{eq:huber_cov}) and applies Eq.~\eqref{eq:gscentry} to obtain the context entries.
    \item \textit{\gls{gsc} Intensity}: computes the mean and standard deviation of point intensity values and applies a modified version of Eq.~\eqref{eq:gscentry} adapted for intensity instead of height.
\end{itemize}

In the \gls{lcd} problem, the primary objective is to maximize both precision and recall. However, achieving $100$\% precision (\textit{i.e.}, $\text{FP} = 0$) is particularly critical, as false positives can significantly degrade the accuracy of downstream \gls{slam} optimization. Therefore, for each sequence, the maximum recall achieved is reported under the constraint of perfect precision. These results are summarized in Table~\ref{tab:best}. The \gls{gsc} Intensity variant is excluded from this table, as it does not consistently achieve $100$\% precision and performs substantially worse than the other methods.

For computational performance, the methods are benchmarked on a machine equipped with an \textit{Intel\textregistered\,Core™ i9-11950H} processor. On average, \gls{sc++} requires $54\,\text{ms}$ per keyframe to generate the \gls{sc++} and execute the loop search, while the proposed \gls{gsc} Huber variant requires $67\,\text{ms}$. This additional computational cost is justified by its improved robustness, and the overhead can be mitigated through optimization strategies, such as parallelized computation of means and covariances, which substantially reduce computation time for large point clouds.

\subsubsection{Discussion}

The results presented in Tab.~\ref{tab:best} and Fig.~\ref{fig:prcurves} demonstrate that the proposed \gls{gsc} variants consistently outperform the original \gls{sc++}\,\cite{kim2021scan} in most evaluated scenarios. 

For the KITTI dataset, the \gls{pr} curves of \gls{sc++} and \gls{gsc} appear qualitatively similar across sequences, suggesting that both methods maintain robustness to variations in the \gls{reid} threshold. However, a closer inspection of the maximum recall achieved at $100$\% precision (Tab.~\ref{tab:best}) reveals a more nuanced outcome: \gls{gsc}, particularly the Huber-weighted height variant, significantly surpasses \gls{sc++}. In sequence $\texttt{02}$, \gls{gsc} Huber achieves nearly a $12\%$ improvement in recall. This indicates that incorporating all points within each context bin and regularizing their statistical representation using the Huber weight produces more reliable and discriminative context descriptors. The Huber function's capacity to suppress the influence of outliers likely contributes to this substantial gain.

The advantages of the proposed approach are even more evident in the \gls{vbr}, which contains more challenging and noisier real-world urban scenes. As shown in Fig.~\ref{fig:prcurves}, the \gls{gsc} variants exhibit markedly greater robustness to threshold variation than \gls{sc++}. This suggests that statistical modeling, particularly with robust estimators such as the Huber weight, is highly effective in environments where \gls{lidar} noise and dynamic elements are more prevalent. In the $\texttt{Pincio}$ sequence, the proposed method delivers an improvement of over $13\%$ in recall at perfect precision, underlining its strength in high-ambiguity contexts.

In contrast, in the $\texttt{Colosseo}$ sequence, \gls{gsc} performs slightly worse than \gls{sc++} ($2.2\%$ standard \gls{gsc} and $3.9\%$ less recall for Huber \gls{gsc}). This is attributed to the scene's repetitive geometric and structural layout, which may hinder statistical estimation or reduce the distinctiveness of Gaussian descriptors. Nonetheless, the overall trend still favors \gls{gsc}, particularly in terms of robustness and recall consistency across diverse environments.

Although the \gls{gsc} Intensity variant does not match the performance of its height-based counterparts, it is included in the evaluation to explore concepts introduced in \gls{isc}\,\cite{wang2020intensity}. \gls{isc} previously demonstrated gains over the original \gls{sc++}\,\cite{gkim-2018-iros} by leveraging point intensity information. However, the results suggest that directly modeling intensity values within a Gaussian framework results in poor performance. This may be due to the inherently high variance and unreliability of intensity readings, which are influenced by factors such as surface reflectivity, angle of incidence, and lighting conditions. These factors cause significant fluctuations in intensity values within a single bin, undermining the effectiveness of statistical modeling.

Finally, in terms of computational efficiency, the proposed method incurs only a modest overhead compared to \gls{sc++}. On the testing platform (\textit{Intel\textregistered\,Core™ i9-11950H}), \gls{gsc} with Huber weighting requires approximately $67\,\text{ms}$ per keyframe, compared to $54\,\text{ms}$ for \gls{sc++}. The performance improvements justify the additional processing time and can be mitigated by efficiently parallelizing the mean and covariance computations. Consequently, \gls{gsc} can serve as a drop-in replacement for \gls{sc++} in \gls{slam} systems requiring robust \gls{lcd}, without compromising real-time performance.

\section{Robot environmental contextual understanding  and interaction experiments} \label{sec:semantic-exp}

This section reports the results obtained with the methods presented in Section~\ref{sec:semantic-method}. While Section~\ref{sec:re-id-exp} focuses on testing contextual awareness in Human-Robot collaborative scenarios, this section focuses on the robot's environmental understanding capabilities for recognizing which objects populate the surroundings and how it can interact with them.

\subsection{Semantic mapping experiments} \label{subsec:artifacts-experiments}

The experiments are performed both in simulation and using a real robot in a laboratory environment. The experimental setup is the same for both: some chosen objects are randomly positioned in the experiment area, and the robot, following a predefined path, maps the objects of interest it encounters. This strategy is chosen because the objective is to validate the artifacts mapping during an application, for example, during a patrol. In other application scenarios, \textit{e.g.}, search and rescue, the proposed framework could run in parallel with an exploration algorithm, and the robot could trigger the exploration module whenever an object of interest is encountered to obtain a precise localization.

In the experiments, the data fusion approach is compared with mono-sensor applications (\textit{i.e.}, using only an \gls{rgbd} camera or only the \gls{lidar}) to demonstrate that data fusion significantly improves detection accuracy and decreases errors. For each environment setup, the experiments are repeated three times, once for each sensor configuration: only camera, only \gls{lidar}, and both.

This work focuses only on semantic mapping and does not account for the robot localization, which is assumed to be given. Additional mapping errors resulting from localization are not considered in the final evaluation, even if they negatively affect the application. Moreover, it is important to note that quadrupedal robots' movements are jerky, which can affect the sensors.

The parameters $min_{c}$, $acc_{c}$, and $max_{c}$ of Eq.~\eqref{eq:filtering} are set to $0.3$, $4$, and $6$, respectively, based on the camera hardware information provided by the camera vendor (Intel Realsense).

The final validation performance is measured as the number of correctly identified objects divided by the total number of objects. Also, the number of correctly detected objects relative to the total number of detections is evaluated.

The object is considered \textit{found} if the difference between the estimated and true positions is less than the true object radius and the associated class label is correct. The errors are categorized into duplicated objects, incorrect localization, and classification. Duplications occur when multiple artifacts are mapped to a single physical object. They could be caused by wrong artifact radius computation due to occlusions or distinct points of view (\textit{i.e.}, detected from different perspectives, such as front and behind). Localization is considered incorrect if the artifact's estimated position is outside the real object's shape, and classification is erroneous if the artifact's class label is wrong. 

For the simulation, the Whole-body Locomotion Framework\,\cite{raiola2022wolf} is used on a notebook with an \textit{Intel\textregistered\,Core™ i9-11950H} processor and an \textit{NVIDIA Geforce RTX 3080 Laptop} \gls{gpu}. In the real scenario, a Unitree Go1\footnote{Unitree Go1:~\url{https://www.unitree.com/en/go1/}} quadrupedal robot equipped with a RoboSense RS-Helios16 \gls{lidar}\footnote{RoboSense RS-Helios16:~\url{https://www.robosense.ai/en/rslidar/RS-Helios}}, an \textit{Intel\textregistered\,RealSense™ D455}\footnote{\textit{Intel\textregistered\,RealSense™ D455}:~\url{https://www.intelrealsense.com/depth-camera-d455/}} and three Nvidia Jetson\footnote{Nvidia Jetson:~\url{https://www.nvidia.com/it-it/autonomous-machines/embedded-systems/}} (two Jetson Nano 4GB and one Nvidia Xavier NX) are used for the evaluation. The experiments are performed using the instance segmentation algorithms Yolact++\,\cite{bolya2020yolact++} and YolactEdge\,\cite{liu2021yolactedge} trained on the COCO\,\cite{lin2014microsoft} dataset.

\begin{figure}[h]
    \centering
    \includegraphics[width=\linewidth]{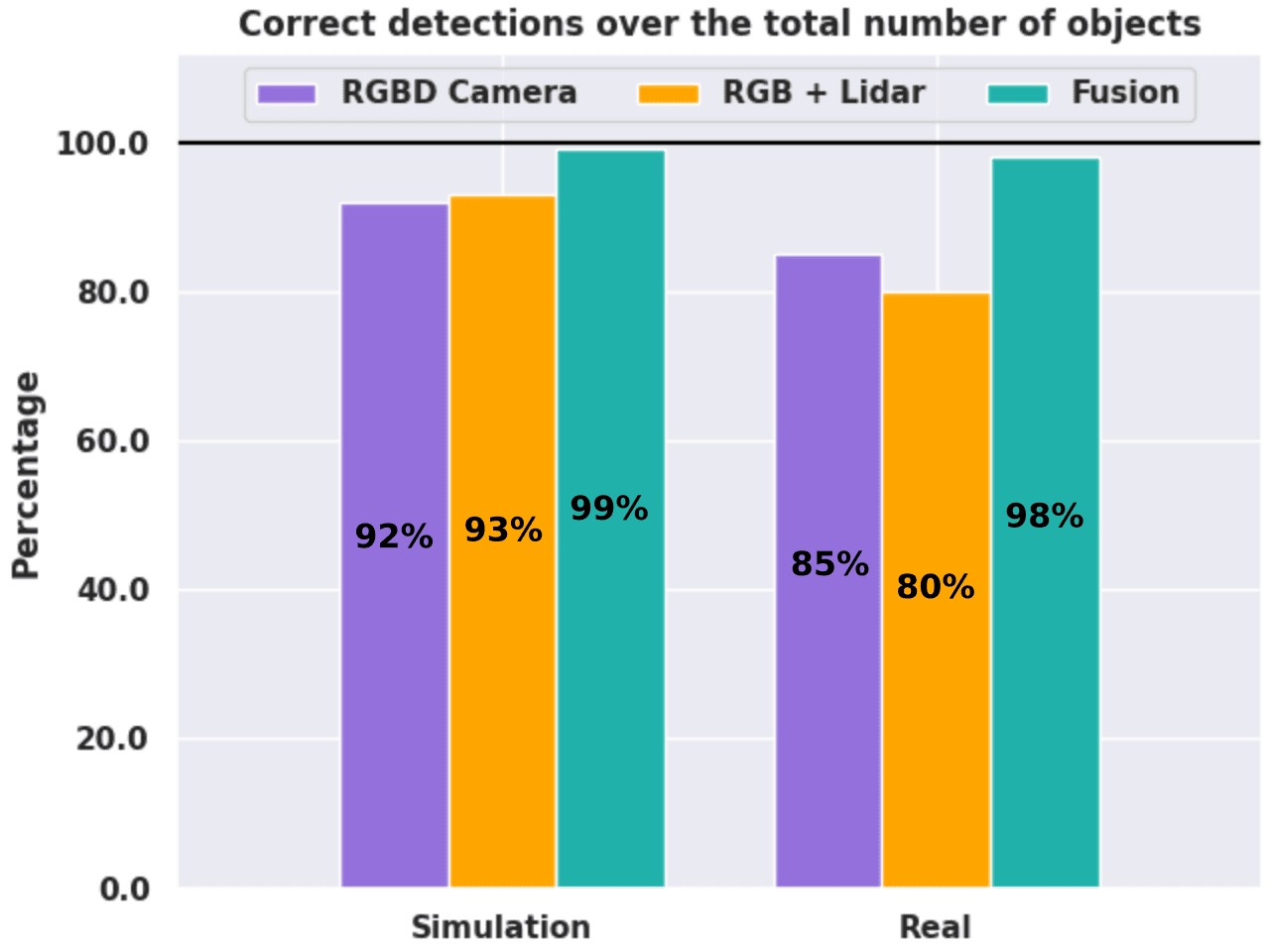}
    \caption{Percentage of the correctly mapped and labelled objects concerning the total number of objects on the scene. On the left are the simulation results; on the right are the real experiments. Each block contains three histograms representing the three \textit{sensor configurations} used during the experiments: only \gls{rgbd} camera, \gls{rgb} plus \gls{lidar}, and \gls{rgbd} and \gls{lidar}. Reprinted, with permission, from\,\cite{rollo2023artifacts}, \textsuperscript{\textcopyright} 2023 \gls{ieee}.}
    \label{fig:correct_detections}
\end{figure}

\subsubsection{Simulation experiments} \label{subsec:simulation}

\begin{center}
    \begin{table*} [h!]
        \centering
        \caption{Detection results of the simulation and real experiments. Reprinted, with permission, from\,\cite{rollo2023artifacts}, \textsuperscript{\textcopyright} 2023 \gls{ieee}.}
        \resizebox{\textwidth}{!}{\begin{tabular}{|c|ccc||ccc|}
            \hline
                & \multicolumn{3}{c||}{Simulation}  & \multicolumn{3}{c|}{Real} \\
            \hline
                & \textit{Camera} & \textit{\gls{lidar}}    & \textit{Fusion} & \textit{Camera} & \textit{\gls{lidar}} & \textit{Fusion} \\
            \hline
            \textbf{Correct detection}     & 386 & 391 & \bf{416 }    & 86   & 81   & \bf{99 } \\
            \textbf{Wrong localization}    & 12  & 13  & \bf{7 }       & 10   & 14   &  \bf{2 }  \\
            \textbf{Duplication}           & 19  & 24  & \bf{15 }       & 10   & 14   & \bf{6 }   \\
            \textbf{Wrong classification}  & \bf{0 }  & \bf{0}   & \bf{0  }     & 7    & 11   & \bf{6 } \\
            \hline\hline
            \textbf{Total detections}      & 417 & 428 & 433     & 113 & 120 & 113 \\
            \hline\hline
            \textbf{Total objects} & \multicolumn{3}{c||}{422}  & \multicolumn{3}{c|}{101}\\
            \hline
        \end{tabular}}
        \label{tab:metrics}
    \end{table*}
\end{center}

The Gazebo simulator\footnote{Gazebo simulator: \url{https://gazebosim.org/home}} is utilized to simulate the robot's behavior in two distinct environments: the Office world\footnote{Clearpath robotics worlds: \url{https://github.com/clearpathrobotics/cpr_gazebo/tree/noetic-devel/cpr_office_gazebo}} and the Maze world. In each environment, a predefined number of objects are randomly positioned at every iteration. The objects chosen for the simulation evaluation are \textit{vase}, \textit{couch}, \textit{plant}, and \textit{person}. Specifically, the Office world contains $5$ vases, $12$ couches, $6$ plants, and $11$ people, while the Maze world contains $15$ vases, $13$ couches, $12$ plants, and $12$ people. The robot's path is predetermined by a set of randomly chosen waypoints on the map. In total, $10$ experiments were conducted for each \textit{sensor configuration} ($5$ per environment, with different setups), yielding $30$ experiments.

The results of the simulation experiments are shown in the left panel of Fig.~\ref{fig:correct_detections} as the number of correctly detected objects. Specifically, across the three ordered \textit{sensor configurations} (\textit{i.e.}, only camera, only \gls{lidar}, and both), the percentages of correctly localized and classified objects obtained were $92\%$, $93\%$, and $99\%$, respectively.

\begin{figure}[h]
    \centering
    \includegraphics[width=\linewidth]{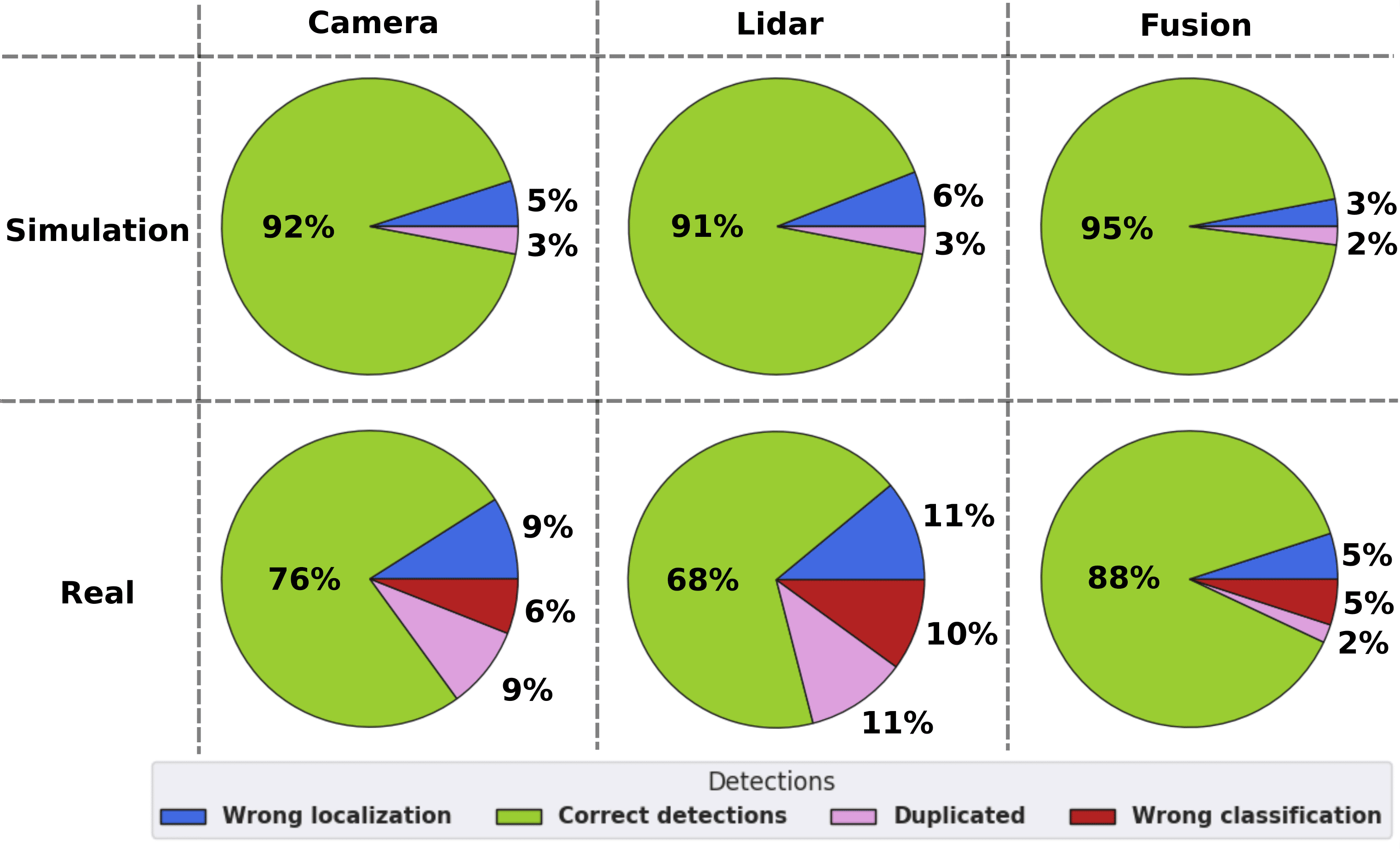}
    \caption{Distribution of correctly and wrongly detected artifacts among the total generated detections. The pie charts show the distribution of correctly detected artifacts (green), doubled objects (blue), wrongly localized (pink), and wrongly classified (red). The top row shows the simulation results, while the bottom row shows the real ones. For each row, experiments are divided into three columns depending on the \textit{sensors configuration} used: only camera, only \gls{lidar}, or both. Reprinted, with permission, from\,\cite{rollo2023artifacts}, \textsuperscript{\textcopyright} 2023 \gls{ieee}.}
    \label{fig:total_detections}
\end{figure}

Moreover, the distribution of total detections is shown in the left column of Table~\ref{tab:metrics} and in the top row of Fig.~\ref{fig:total_detections} for the simulation experiment. Among all the detections produced, considering the three \textit{sensor configurations} in order, $92\%$, $91\%$, and $95\%$ were correct, while the remaining $8\%$, $9\%$, and $5\%$ were wrong.

The farthest object correctly detected during the camera-\gls{lidar} sensor fusion experiments in simulation was at $15.47\,\mathrm{m}$ from the robot, while the nearest was at $1.23\,\mathrm{m}$.

\subsubsection{Laboratory experiments} \label{subec:realexp}

The real-world experiments were conducted in a laboratory setting, with two scenarios: a one-room environment and a full-floor environment in which the robot could move through corridors. In these environments, umbrellas, chairs, cabinets, backpacks, and TVs were arranged in varying numbers. For each \textit{sensor configuration}, $6$ experiments were conducted ($3$ per environment), for a total of $18$. For each trial, the objects were randomly moved, and the illumination was changed, \textit{i.e.}, switching off lights or closing shutters, to introduce environmental variability.

The results of the laboratory experiments are shown in the right panel of Fig.~\ref{fig:correct_detections} as correctly detected objects. Specifically, considering the three \textit{sensor configurations} in order (\textit{i.e.}, only \gls{rgbd} camera, only \gls{rgb} plus \gls{lidar}, and both), the percentage of correctly localized and classified objects obtained was $85\%$, $80\%$, and $98\%$, respectively.

Moreover, the distribution of detections is shown in the right column of Table~\ref{tab:metrics} and in the bottom part of Fig.~\ref{fig:total_detections} for the real experiment. Among all the detections produced, $76\%$, $68\%$, and $88\%$ were correct, while the remaining $24\%$, $32\%$, and $12\%$ of them were wrong.

The farthest object correctly detected during the camera-\gls{lidar} sensor fusion experiments was at a distance of $10.37\,\mathrm{m}$ from the robot, while the nearest was at $0.98\,\mathrm{m}$.

\begin{figure}[h]
    \centering
\includegraphics[width=\linewidth]{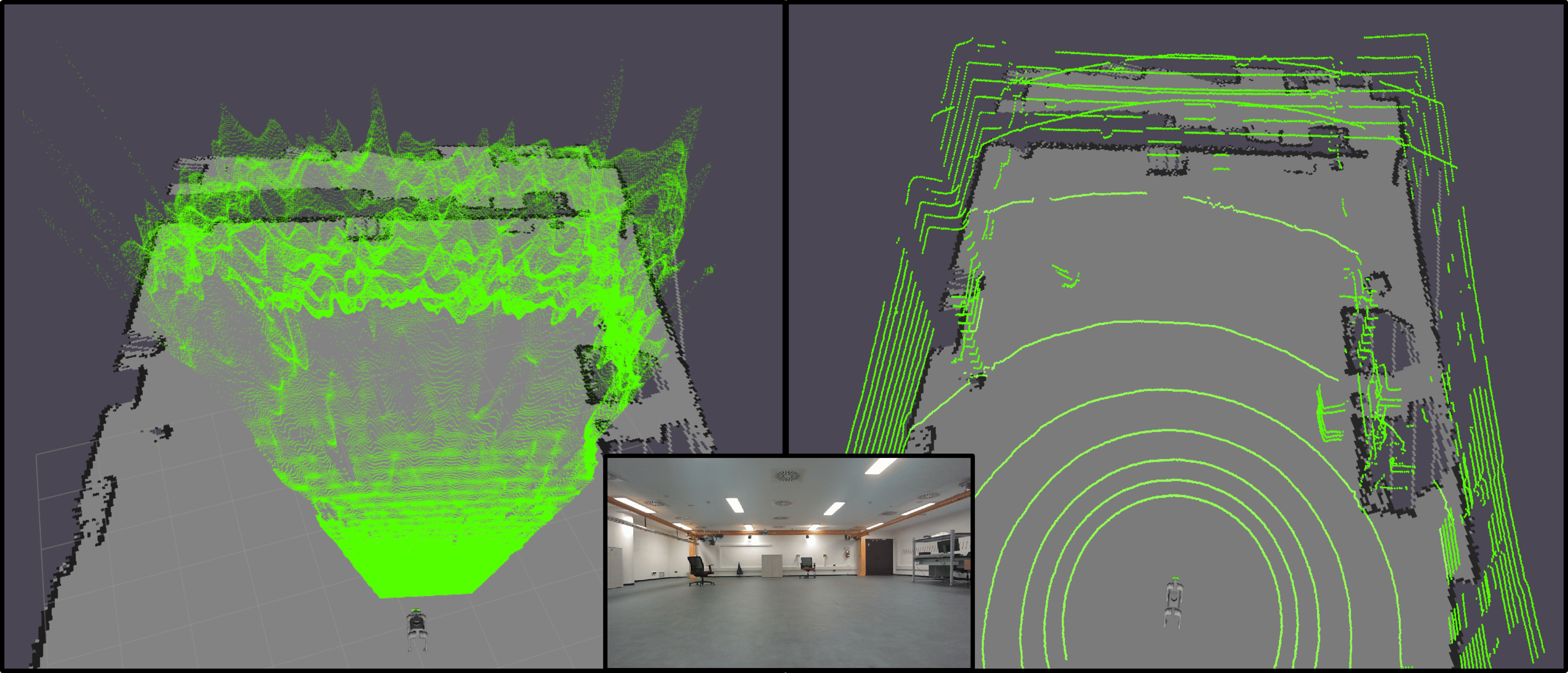}
    \caption{Qualitative comparison between \gls{rgbd} camera (left image) and \gls{lidar} (right image) point cloud detections at an approximate distance of $10\,\mathrm{m}$ from the wall. At the bottom center is a representation of the scene taken with the robot camera at that instant. At large distances, camera data are noisier and less accurate than the \gls{lidar} one. Still, at short distances, cameras provide a denser, more accurate point cloud, whereas \gls{lidar} data are sparser. From this comparison, it can be deduced that a visual-\gls{lidar} sensor fusion can enhance semantic mapping. Reprinted, with permission, from\,\cite{rollo2023artifacts}, \textsuperscript{\textcopyright} 2023 \gls{ieee}.}
    \label{fig:comparison}
\end{figure}

\subsubsection{Using semantic mapping for autonomous loco-manipulation}

A sample experiment demonstrating how to use such semantic maps has been performed in simulation using the setup described above. In the experiments, the robot is asked to bring an object to one of the people in the simulation (\textit{e.g.}, bring a bottle to the person with id 1). The experiment is repeated for different world configurations in which people and the objects are randomly moved (see an example in the $6^{th}$ block of Fig.~\ref{fig:pipeline}). The proposed pipeline always completed the task following the steps presented in Section~\ref{subsec:locoman-method}. 

\subsubsection{Discussion} \label{subec:expdiscussion}

The first thing to point out is that the farthest distances of the detected objects were greater than $10\,\mathrm{m}$ both in simulation and in real experiments. This distance is chosen to show a qualitative comparison between the \gls{lidar} and \gls{rgbd} measurements in Fig.~\ref{fig:comparison}. The figure qualitatively upholds the thesis that a \gls{lidar} sensor, along with the camera, is necessary to improve semantic mapping and, in general, other detection algorithms in wide areas.

Moreover, the results obtained from the experiments clearly show that in this framework, the use of both sensors improves the robustness of the application and decreases the detection errors. These improvements are less evident in a simulation environment, where almost ideal sensors are used (\textit{i.e.}, the noise representation is not as realistic as in Fig.~\ref{fig:comparison}), but they significantly impact real scenarios where sensor noise is more prevalent.

The \gls{lidar} can map far obstacles precisely, while the camera introduces many errors at high distances. If only the camera is adopted, one solution to avoid erroneous depth measurements could be to not consider the depth information outside the accurate range guaranteed by the device specifications. However, by doing this, the robot could miss some artifacts if it does not get close enough to them.

The camera, by providing more information at near distances with respect to the \gls{lidar}, yields more precise centroid computations because it typically has fewer outliers than the \gls{lidar}. \gls{lidar} outliers can be caused by wrong camera-\gls{lidar} pose calibration and time synchronization, which are essential for these applications, especially when the robot moves fast. In contrast, with \gls{rgbd} cameras, the depth and the \gls{rgb} images are synchronized in time and can be spatially superimposed almost exactly.

It is important to notice that wrong classification errors result from erroneous classifications in the pre-trained instance segmentation neural network, which can be caused by illumination, reflections, or other environmental conditions. They are considered here because the image inference is a module of the proposed pipeline, but such errors can be decreased using more powerful neural networks.
    
\chapter{conclusion} \label{chap:conclusion}
    
The main remarks and demonstrations acquired during the PhD journey are reported in this concluding thesis section. As for the previous sections, the conclusive remarks are distributed in different sections. Section~\ref{sec:re-id-concl} reports the conclusions obtained from the \gls{reid} methods developed, validated, and exposed in this thesis. Environmental understanding conclusions are analyzed and reported in Section~\ref{sec:slam-concl} for the geometric mapping and environment representations and in Section~\ref{sec:semantics-concl} for the semantic understanding that enhances the robotic knowledge, trying to mimic how human beings perceive the environment.

\section{Re-identification for Human-Robot Collaboration} \label{sec:re-id-concl}

The works presented in \gls{reid} all have the same objective: to provide a robust \gls{reid} framework that can be used by a robot to perform \gls{hrc} tasks. Robots can employ this capability to co-exist with different people in an environment, moving beyond their role as passive workers to actively understand the roles of the people around them and act accordingly.

The first work, the FollowMe\,\cite{rollo2023followme}, is a robust framework for following a target person by a mobile robot. It is mainly based on visual \gls{reid} and gesture detection. The experiments confirmed that visual human \gls{reid} is a strong feature to perform person-following tasks and suggested that this ability is crucial for other human-centered applications. Using a simple and non-invasive \gls{rgbd} camera, the robot can efficiently track a selected person, simplifying \gls{hrc}. The FollowMe framework is easily customizable to a target person, avoiding mismatches or ID switches with other persons (\textit{i.e.}, the distractors), even when two persons are similarly dressed.

Despite this, during experiments, some limitations were faced. In specific light conditions, \textit{i.e.}, when a strong light is pointed towards the camera, the detection or \gls{reid} module lacks robustness. This limitation is also connected to the specific camera hardware. Also, \gls{reid} is strictly correlated to online calibration, \textit{i.e.}, if the calibration is not properly finalized, the further recognition of the target could be less robust; the robot should see the person at different distances and in different configurations during calibration, simulating as much as possible the human motion made at inference time.

To solve some of these limitations, a framework for person \gls{reid} that can be adapted to different \gls{hri} scenarios, namely \gls{carpe}\,\cite{rollo2024carpe}, has been developed. This pipeline, similarly to FollowMe, consists of a detection and preliminary person-tracking algorithm, a feature extraction network for appearance representation, and a \gls{reid} structure based on statistical distance. In addition, an adaptable ideal target representation, threshold computation, and a custom version of the \gls{ema} with a damping factor (\gls{dema}) are introduced. This setup enables tracking people even when they are partially or fully occluded or when their appearance changes during tracking (for instance, by wearing a sweatshirt). These are challenging situations that are difficult for traditional \gls{mot} algorithms to handle, as our experiments have shown.

It is worth noting that, even if there are improvements compared to the previous \gls{reid} framework, this work has some limitations. One of them is the issue of catastrophic forgetting, which means that the algorithm tends to forget the appearance of the target as time passes because it continuously adapts to the new information it gathers. This can be faced with an online method that uses continuous learning techniques. The feature extraction network can be trained using people detections extracted during \gls{carpe} execution. This would specialize the feature extractor with the tracked person appearances, forcing the outputted feature to be similar when belonging to the target and different if associated with a distractor.
Proximity to the camera poses another limitation of our algorithm. However, this is not usually a problem in \gls{hrc} scenarios because the human is generally near the robot and the camera.
Lastly, the algorithm's time performance may suffer when more than $20$ people are present in the image. Nevertheless, it is worth noting that such scenarios are rare, especially considering the robot camera's point of view.

To improve \gls{carpe}, the suggested introduction of continual learning techniques is crucial. This technique involves using online self-supervised training to specialize the feature extraction network on the target appearances and to differentiate more effectively from the appearances of other people in the vicinity who may cause distractions.

This can be accomplished with a human \gls{reid} and tracking approach that leverages continual learning and smart image pool generation to fine-tune the feature extractor network specifically for the tracked target's characteristics, thus extending the foundation laid by the previous \gls{carpe} framework. To maintain both accuracy and efficiency (in terms of time), a twin network for feature extraction is adopted, which is trained in parallel with the original network responsible for tracking. Our results demonstrate that this framework outperforms its baseline, effectively tracking the target throughout all videos. To support our findings, saliency maps are provided for enhanced interpretability. These achievements offer valuable insights to the robotics community, promoting a more conscious \gls{hri}. 

However, it is important to note a minor limitation in this work. Given that the Soft Triplet loss function requires both negative and positive samples, a constraint is that it requires at least one distractor in the image for effective training. Conversely, tracking could become trivial if the target is always the only person in the image.
During our evaluation, difficulties arose due to the scarcity of datasets that met our evaluation criteria, particularly those involving changes in target appearance and total occlusions. Consequently, an intriguing avenue for future research could involve curating new datasets or integrating existing ones to address these scenarios better while establishing robust real-world benchmarks for \gls{reid} and tracking performance, especially in \gls{hrc}.

\section{Geometric environmental perception} \label{sec:slam-concl}

The geometric perception of the surroundings is one of the first tasks required by a robot to move freely and safely. To further explore contextual awareness for the robotic platform, three key areas of geometric perception were addressed: \gls{slam}, sensor filtering, and robust loop closure.

To cope with this task, a \gls{lidar} \gls{slam} approach, LEO\raisebox{0.1ex}{-}SLAM\,\cite{rollo2025leoslam}, has been presented. It employs a submap-based keyframe, a multi-level alignment strategy, and a submap-based \gls{sc++} within an adaptive search area. This system differs from the \gls{sota} due to its alignment method, which prioritizes maintaining map consistency between consecutive scans rather than optimizing only the final robot pose, as often occurs in pure \gls{lidar} odometry systems. Our experiments demonstrated that the proposed framework outperforms existing \gls{sota} methods in terms of \gls{ate} and \gls{rpe}.

As with many other \gls{lidar}-based approaches, our \gls{slam} system struggles in low-feature environments, such as long corridors. As shown in our experiments, errors introduced during scan matching can be reduced through the integration of external odometry information or \gls{imu} data. To address this, a multi-sensory approach combined with intelligent multi-odometry source filtering can be used to provide robust final odometry, thereby reducing drift errors.

LEO\raisebox{0.1ex}{-}SLAM provides a stable foundation for future developments aimed at enabling reliable environmental understanding and advancing towards autonomous, conscious robotics.

During the experiments, some environments with reflections and refractions were encountered. In those environments, the maps obtained with the \gls{slam} were not exactly like the original ones. To solve this problem, two different implementations of a ground-aware intensity filtering method have been proposed. They are designed to remove points caused by erroneous reflections from semi-transparent or translucent surfaces, commonly found in indoor environments (\textit{e.g.}, windows). Each implementation offers distinct advantages and drawbacks depending on the environment. However, the simpler solution is preferred due to its faster performance and satisfactory results. Our evaluation of \gls{sota} \gls{lidar} odometry systems confirmed that \gls{lidar}-based systems significantly benefit from our simple yet effective ground-aware intensity filtering in environments with frequent erroneous reflections, enhancing both precision and robustness.

To conclude the geometrical representation study, robust \gls{lcd} has also been improved with the \gls{gsc} algorithm. It is an enhanced variant of \gls{sc++}\,\cite{kim2021scan}, which integrates statistical analysis of point cloud data to improve \gls{lcd} performance. Unlike the original method, which relies solely on the maximal-height point within each context bin, \gls{gsc} models the complete distribution of points using Gaussian statistics. This enables a more robust and descriptive representation, particularly in environments affected by noise and outliers.

Our experimental validation across multiple datasets, including KITTI and \gls{vbr}, demonstrated that \gls{gsc} consistently outperforms the baseline in terms of recall while maintaining perfect precision in most cases. The incorporation of Huber-weighted statistics further enhances robustness by reducing the impact of outlier points.

\section{Semantics for environmental understanding and interaction} \label{sec:semantics-concl}

Once personalized \gls{reid} for \gls{hrc} and geometrical mapping for environmental understanding have been established, the next step is to improve the robot's contextual awareness by introducing other semantics in the environment.

To accomplish this task, the Artifacts Mapping framework\,\cite{rollo2023artifacts} has been presented. It utilizes multi-modal sensor fusion to address the semantic mapping problem, a rare setup in robotics applications. The \gls{lidar} and \gls{rgbd} camera sensor readings are fused to achieve better accuracy for both near and far objects, as opposed to camera-only systems, which lose accuracy for distant objects, or \gls{lidar}-only systems, which lack high-level texture understanding of the environment.

A \gls{ui} application is proposed to interact with the artifacts map obtained during the mapping application. This application is useful for performing autonomous high-level decision-making tasks because it exposes the object's class and location to the robot and the user.

The experiments demonstrated that our application can correctly detect, localize, and map $98\%$ of the objects present in the scene at different distances, providing a small number of detection errors and good localization accuracy. The comparisons with the single-sensor scenario (either camera or \gls{lidar}) proved that sensor fusion is essential for wide areas and high-accuracy applications.


In addition, a semantic loco-manipulation framework for object retrieval has been introduced to prove the usefulness of semantic maps. This application uses high-level semantic scene understanding to enable a robot assistant to search for and bring objects that are not nearby and whose locations may be unknown to the user. It is a foundational platform for enhancing robotic assistance within industrial environments, where robots work alongside human operators and actively participate in the workplace community.

Several future developments for this framework include the generalization of the pose estimation to handle more complex situations and objects, making experiments on a real robot, and improving the reactivity of the \gls{bt}, \textit{e.g.}, if the robot loses the grasp of the object, the \gls{bt} should restart autonomously from the \textit{pose\_estimation} behavior.

\section{Final remarks on robotic contextual awareness and future directions}

In conclusion, the works presented in this thesis all move in the direction of full robotic autonomy. Robots require complete contextual awareness to live and work in a human-populated world. They need to have capabilities similar to those of a person to interact with them, and at the same time be compliant with the environment. Contextual awareness is strictly connected to how a person perceives and interacts with the environment, and this leads to the development of how a robot should do it. This should not be strictly followed because robots can also surpass human capabilities, mimicking other animals or living beings' capabilities, but we can always take into account that robots will interact with humans, and the world we built and are building is mainly shaped for our objectives and capabilities.

To accomplish these tasks, future work in the robotic contextual awareness direction is needed. Improvements in the surrounding understanding are crucial for the robot control. Robots should be able to deeply comprehend the semantics in the environment, their relations, and how they can employ that information to accomplish the tasks that are required. The more the robots can understand their surroundings, the more tasks they can perform. Researchers are quickly moving in this direction as robotic perception and understanding lay the foundations for robotic autonomy.

\bibliographystyle{ieeetr}
\bibliography{biblio.bib}

\appendix
    \chapter{Pseudo Algorithms}

\begin{algorithm}
\caption{\gls{carpe} framework. Reprinted, with permission, from\,\cite{rollo2024carpe}, \textsuperscript{\textcopyright} 2024 \gls{ieee}.} \label{alg:carpe-id}
\begin{algorithmic}[1]
\Statex $\textbf{Input:}\ tracking\_id$ 
\State $\boldsymbol{\mu}, \boldsymbol{\sigma}, \mu_d, \sigma_d, \lambda_d \gets initialize\_variables()$
\While{True}
\State $\boldsymbol{I_{RGB}} \gets cam.getRGB()$
\State $dets \gets mot.infer(\boldsymbol{I_{RGB}})$
\State $feats \gets reid.infer(dets)$
\State $min\_dist \gets MAX\_FLOAT\_NUM$
\State $tracking\_feature \gets Null$
\State $\textcolor{blue}{\texttt{\small // Re-identification}}$ 
\For{$feature \in feats$}
    \If{$tracking\_id = feat.id$}
        \State $tracking\_feature \gets feature$
        \State \textbf{break}
    \Else 
        \State $d_{\boldsymbol{\mu},\boldsymbol{\sigma}} \gets get\_distance(feature,  \boldsymbol{\mu}, \boldsymbol{\sigma}), $
        \If{$d_{\boldsymbol{\mu},\boldsymbol{\sigma}} < min\_dist \And d_{\boldsymbol{\mu},\boldsymbol{\sigma}} < \lambda_d$}
            \State $min\_dis \gets d_{\boldsymbol{\mu},\boldsymbol{\sigma}}$
            \State $tracking\_feature \gets feature$
            \State $tracking\_id \gets feature.id$
        \EndIf
    \EndIf
\EndFor
\State $\textcolor{blue}{\texttt{\small // Target model update}}$
\If{$tracking\_feature \neq Null$} 
    \State $var \gets compute\_var(\boldsymbol{\mu}, tracking\_feature)$
    \State $\boldsymbol{\mu} \gets DEMA(\boldsymbol{\mu}, tracking\_feature, \Delta_{f})$
    \State $\boldsymbol{\sigma} \gets DEMA(\boldsymbol{\sigma}, var, \Delta_{f})$
    \State $var_d \gets compute\_var(\mu_d, min\_dist)$
    \State $\mu_d \gets DEMA(\mu_d, min\_dist, \Delta_{\lambda_d})$
    \State $\sigma_d \gets DEMA(\sigma_d, var_d, \Delta_{\lambda_d})$
    \State $\lambda_d \gets \mu_d + 2 \sigma_d$
\EndIf
\EndWhile
\end{algorithmic}
\end{algorithm}

\begin{algorithm}
\caption{LEO\raisebox{0.1ex}{-}SLAM algorithm. Reprinted, with permission, from\,\cite{rollo2023artifacts}, \textsuperscript{\textcopyright} 2023 \gls{ieee}.}\label{alg:leo-slam}
\setstretch{1.05}
    \begin{algorithmic}[1]
    \Statex $\textbf{Input:}\ \mathcal{P},\ \mathbf{^\mathrm{o}H_\mathrm{b}}\ =\ \mathcal{I}_{4x4}\  (optional)$
    \Statex $\textbf{Output:}\ ^\mathrm{m}\mathbf{H}_{\mathrm{o}}$
    \State $\textcolor{blue}{\texttt{\small// Filtering}}$
    \State $\mathcal{P}_p \gets preprocess(\mathcal{P})$
    \State $\mathcal{P}_f \gets filter(\mathcal{P}_p)$ \textcolor{blue}{\texttt{\small// Used for map construction}}
    \State $^\mathrm{m}\mathbf{H}_{\mathrm{b}}\ \gets\ ^\mathrm{m}\mathbf{H}_{\mathrm{o}}\ *\ ^\mathrm{o}\mathbf{H}_\mathrm{b}$ 
    \State $\textcolor{blue}{\texttt{\small// Scan alignment}}$
    \State $\mathbf{H}^{ICP}_{\mathrm{s}2\mathrm{s}}\ \gets\ GICP(\mathcal{P}_p,\ \mathcal{P}_{prev},\  ^\mathrm{m}\mathbf{H}_{\mathrm{b}})$
    \State $\mathbf{H}^{ICP}_{\mathrm{s}2\mathbb{S}}\ \gets\ GICP(\mathcal{P}_p,\ \mathbb{S},\  \mathbf{H}^{ICP}_{\mathrm{s}2\mathrm{s}})$
    \State $^\mathrm{m}\mathbf{H}_{\mathrm{b}}\ \gets\ \mathbf{H}^{ICP}_{\mathrm{s}2\mathbb{S}} *\ ^\mathrm{m}\mathbf{H}_{\mathrm{b}}$
    \State $d_m,\ d_\theta\ \gets\ computeDistance(^\mathrm{m}\mathbf{H}_{\mathrm{n_{c-1}}},\ ^\mathrm{m}\mathbf{H}_{\mathrm{b}})$
    \State $\mathcal{S}\ \gets\ addPcd(\mathcal{S},\ \mathcal{P}_p)$
    \State $\textcolor{blue}{\texttt{\small// Add new node}}$
    \If{$d_m\ \geq\ \Delta_m\ or\ d_\theta\ \geq\ \Delta_\theta$}
        \State $\textcolor{blue}{\texttt{\small// Submap alignment}}$
        \State $\mathbf{H}^{ICP}_{\mathcal{S}2\mathbb{S}}\ \gets\ GICP(\mathcal{S},\ \mathbb{S},\  ^\mathrm{m}\mathbf{H}_{\mathrm{b}})$
        \State $^\mathrm{m}\mathbf{H}_{\mathrm{b}}\ \gets\ \mathbf{H}^{ICP}_{\mathcal{S}2\mathbb{S}}\ *\ ^\mathrm{m}\mathbf{H}_{\mathrm{b}}$
        \State $\mathrm{n_c}\ \gets\ graph.addNode(\mathcal{S},\ ^\mathrm{m}\mathbf{H}_{\mathrm{b}})$
        \State $\textcolor{blue}{\texttt{\small// Loop Closure Detection}}$
        \State $SC\ =\ computeScanContext(\mathcal{S})$
        \State $LC_{found},\ LC_{ID}\ \gets\ LCD(SC)$
        \If{$LC_{found}$}
            \State $\mathrm{n_l}\ \gets\ graph.getNode(LC_{ID})$
            \State $^\mathrm{n_{c}}\mathbf{t}_\mathrm{n_l}\ \gets\ ^\mathrm{m}\mathbf{t}_{\mathrm{n_c}}\ -\ ^\mathrm{m}\mathbf{t}_{\mathrm{n_l}}$
            \State $\mathbb{S}_c,\ \mathbb{S}_l\ \gets\ computeSubmapsNeighbours()$
            \State $\mathbf{H}^{ICP}_{lc}\ \gets\ GICP(\mathbb{S}_c,\ \mathbb{S}_{l},\  ^\mathrm{n_c}\mathbf{t}_\mathrm{n_l})$
            \State $\textcolor{blue}{\texttt{\small// Pose graph optimization}}$
            \If{$GICP.hasConverged()$}
                \State $^\mathrm{n_\mathrm{c}}\mathbf{H}_{n_l}\ \gets\ (\ \mathbf{H}^{ICP}_{lc}\ *\ ^\mathrm{m}\mathbf{H}_{\mathrm{n_c}})^{-1}\ *\ ^\mathrm{m}\mathbf{H}_{\mathrm{n_{l}}}$
                \State $graph.addLoop(\mathrm{n_c},\ \mathrm{n_{l}},\ ^\mathrm{n_\mathrm{c}}\mathbf{H}_{n_l})$
                \State $graph.isam2optimization()$
                \State $graph.update() $ 
                \State $^\mathrm{m}\mathbf{H}_{\mathrm{b}}\ \gets\ graph.getLastNodePose()$
            \EndIf
        \EndIf
        \State $\mathbb{S}\ \gets getPrevSubmaps()$
    \EndIf
    \State $\mathcal{P}_{prev}\ \gets\ transformPcd(\mathcal{P}_p,\ ^\mathrm{m}\mathbf{H}_{\mathrm{b}})$
    \State $^\mathrm{m}\mathbf{H}_{\mathrm{o}}\ =\ ^\mathrm{m}\mathbf{H}_{\mathrm{b}}\ *\  ^\mathrm{o}\mathbf{H}_\mathrm{b}^{-1}\ $
    \Statex \Return $^\mathrm{m}\mathbf{H}_{\mathrm{o}}$
    \end{algorithmic}
\end{algorithm}

\end{document}